\documentclass[journal]{IEEEtran}
\usepackage{amssymb}
\usepackage{graphics}
\usepackage{graphicx}
\usepackage{subfigure}
\usepackage{epsfig}
\usepackage{amsmath}
\usepackage{bm}
\usepackage{lineno}
\usepackage{url}
\usepackage{float}

\usepackage[normalem]{ulem}
\usepackage[color,final]{showkeys}
\usepackage{multirow}

\ifCLASSINFOpdf
\else
\fi

\begin{document}
%
\title{ Support Driven Wavelet Frame-based Image Deblurring }
%
%
%

\author{    Liangtian He,~\IEEEmembership{Member,~IEEE}, Yilun~Wang*,~\IEEEmembership{Member,~IEEE}, Zhaoyin Xiang
\thanks{  Liangtian He and Zhaoyin Xiang are both with the School of Mathematical Sciences, University of Electronic Science and Technology of China, Chengdu,
Sichuan, 611731 China. Yilun Wang is  with School of Mathematical Sciences and Center for Information in Biomedicine, University of Electronic Science and Technology of China, Chengdu, Sichuan, 611731 P. R. China. (e-mail: yilun.wang@rice.edu). He is also with Center for Applied Mathematics, Cornell University, Ithaca, NY 14853 USA.  *Corresponding author: Yilun Wang}   }


%
%

\markboth{IEEE TRANSACTIONS ON IMAGE PROCESSING}%
{Shell \MakeLowercase{\textit{et al.}}: Bare Demo of IEEEtran.cls for Journals}
%



\maketitle

\begin{abstract}
The wavelet frame systems have been playing an active role in image restoration and many other image processing fields over the past decades, owing to the good capability of sparsely approximating piece-wise smooth functions such as images.
In this paper, we propose a novel wavelet frame based sparse recovery model called \textit{Support Driven Sparse Regularization} (SDSR) for image deblurring, where the partial support information of frame coefficients is attained via a self-learning strategy and exploited via the proposed truncated $\ell_0$ regularization. Moreover, the state-of-the-art image restoration methods can be naturally incorporated into our proposed wavelet frame based sparse recovery framework. In particular, in order to achieve reliable support estimation of the frame coefficients, we make use of the state-of-the-art image restoration result such as that from the IDD-BM3D method as the initial reference image for support estimation.   Our extensive experimental results have shown convincing improvements over existing state-of-the-art deblurring methods.

\end{abstract}


%
\IEEEpeerreviewmaketitle

\section{Introduction}

\IEEEPARstart{I}{mage} restoration is one of the most important research topics in many areas of image processing and computer vision. Its major purpose is to enhance the quality of an observed image (e.g., noisy and blurred) that is corrupted in various ways during the process of imaging, acquisition and communication, and enable us to observe the crucial but subtle objects that reside in the images. Image restoration tasks can often be formulated as an ill-posed linear inverse problem:
\begin{equation}\label{image restoration}
f = Au + \epsilon
\end{equation}
where $u$ and $f$ is the unknown true image and observed degraded image, respectively. $\epsilon$ denotes the additive white Gaussian noise with variance $\sigma^2$. Different image restoration problem corresponds to a different type of linear operator $A$, e.g., an identity operator for image denoising, a projection operator for inpainting, and a convolution operator for deblurring, etc. Most image recovery tasks are ill-posed inverse linear problems. A naive inversion of $A$, such as pseudo-inversion, may result in a restored image with amplified noise and smeared-out edges. Therefore, to obtain a reasonably approximated solution, the regularization methods which try to incorporate both the observation model and the prior information of the underlying solution into a variational formulation, have been widely studied.
Among them, variational approaches and wavelet frame based  methods are extensively studied and adopted [1-16].

In recent years, the sparsity-based prior based on wavelet frame has been playing a very important role in the development of effective image recovery models. The key idea behind
the wavelet frame based image restoration models is that the interested image is compressible in this transform domain. Therefore, the regularized process can be chosen by minimizing the functional that promotes the sparsity of the underlying solution in the transform domain.
The connection of wavelet frame based methods with variational and PDE based approaches is studied in \cite{Cai2011}, \cite{Cai2015}. Such connections explain the reason why wavelet frame based approaches are often superior to some of the variational based models. Generally speaking, the multiresolution structure and redundancy property of wavelet frames allow to adaptively select proper differential operators according to the order of the singularity of the underlying solutions for different regions of a given image.

For regularization methods, exploiting and modeling the appropriate prior knowledge of natural images is one of the most important topics. In other words, the final recovery performance largely depends on the design of the regularization term from the viewpoint of Bayesian statistics. Most existing related works focus more on choices of the classical $\ell_1$ norm, $\ell_p$ ($0<p<1$) or $\ell_0$ quasi-norm as an appropriate sparsity term in their specific problems.  The sparsity-based prior regularization has become so widespread and crowded that it raises the question  whether there still room for further improvement and what is the right direction to head into. One interesting direction is to consider to exploit other important image priors to further improve the recovery performance besides the classical sparsity prior. Recently, Cai et.al \cite{Cai2015} and Ji et.al \cite{Ji2015} proposed the piecewise-smooth image restoration model and added additional regularizations on the locations of image discontinuties, which can be viewed as the variants of the $\ell_1$-norm and Tikhonov regularization.


\begin{figure}[h]
  \centering
   \includegraphics[width=0.48\textwidth]{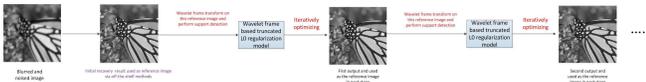} \\

\caption{\small  Overview of the proposed method. Given the noisy and blurry image, we start from obtaining an initial recovered image via any existing image restoration methods, e.g., IDD-BM3D method. Then we perform support detection of the frame coefficients on this recovered result, and develop a truncated $\ell_0$ regularization model. We solve this resulted optimization model, and obtain a new recovered image, and so on. Note that our method is an alternating optimization procedure, which repeatedly applies the support detection and image recovery.   }
\label{fig:flowchart}
\end{figure}

In this paper, we would like to move forward  and aim to further exploit more priors, such as the locations of the nonzero frame coefficients, besides these widely used  classical sparsity priors. Correspondingly, we propose a novel wavelet frame based  \textit{Support  Driven Sparse Regularization} (SDSR) model for image deblurring.
This model makes use of the proposed truncated $\ell_0$ regularization to naturally incorporate the detected partial support information of frame coefficients.
Once we have partial support information of frame coefficients based on the initial reference image, this support information  will be used to produce  a  wavelet frame based truncated $\ell_0$ regularized model. The solution of this new model will be used as the new reference image for the support estimation at second stage.  Then the newly updated partial support information will lead to a new truncated $\ell_0$ regularized model, and so on, resulting into an alternative iterative procedure. Figure 1 illustrates the framework of our method, which is a multi-stage procedure. In figure 2, we provide a first glance of the recovery results via our proposed truncated $\ell_0$ regularization model while the detailed definition and analysis of it are available in Section IV, where the oracle \footnote{The oracle case means the support detection is performed on the original true image. It is infeasible to obtain in practice, since we do not know the true image. However, we use it to illustrate  the potential advantages of making use of support information.} support information of frame coefficients is exploited. The impressive performance indicates the great potential of incorporating the support information into existing sparsity regularized model.

\begin{figure}[h]
  \centering
   \includegraphics[width=0.15\textwidth]{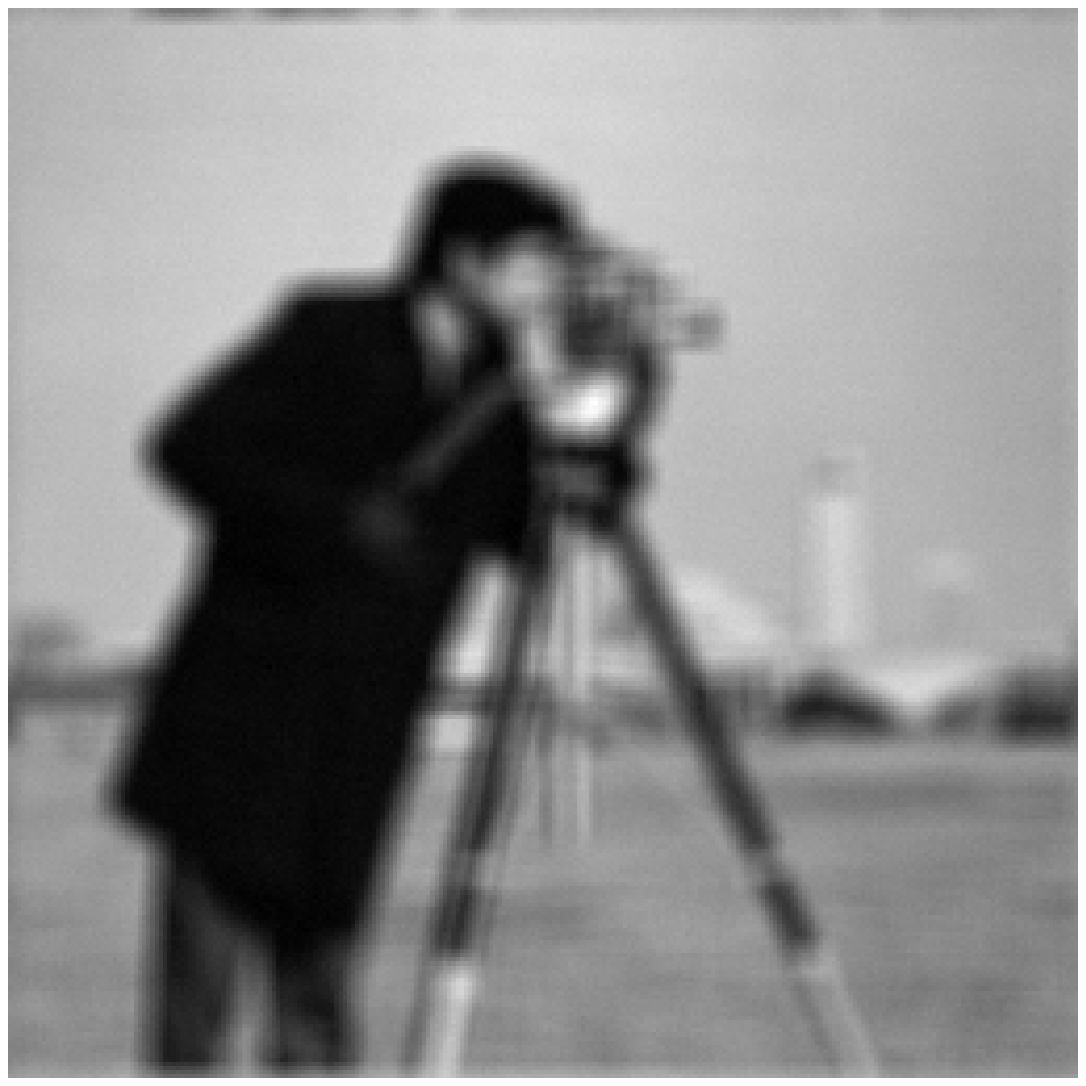}
   \includegraphics[width=0.15\textwidth]{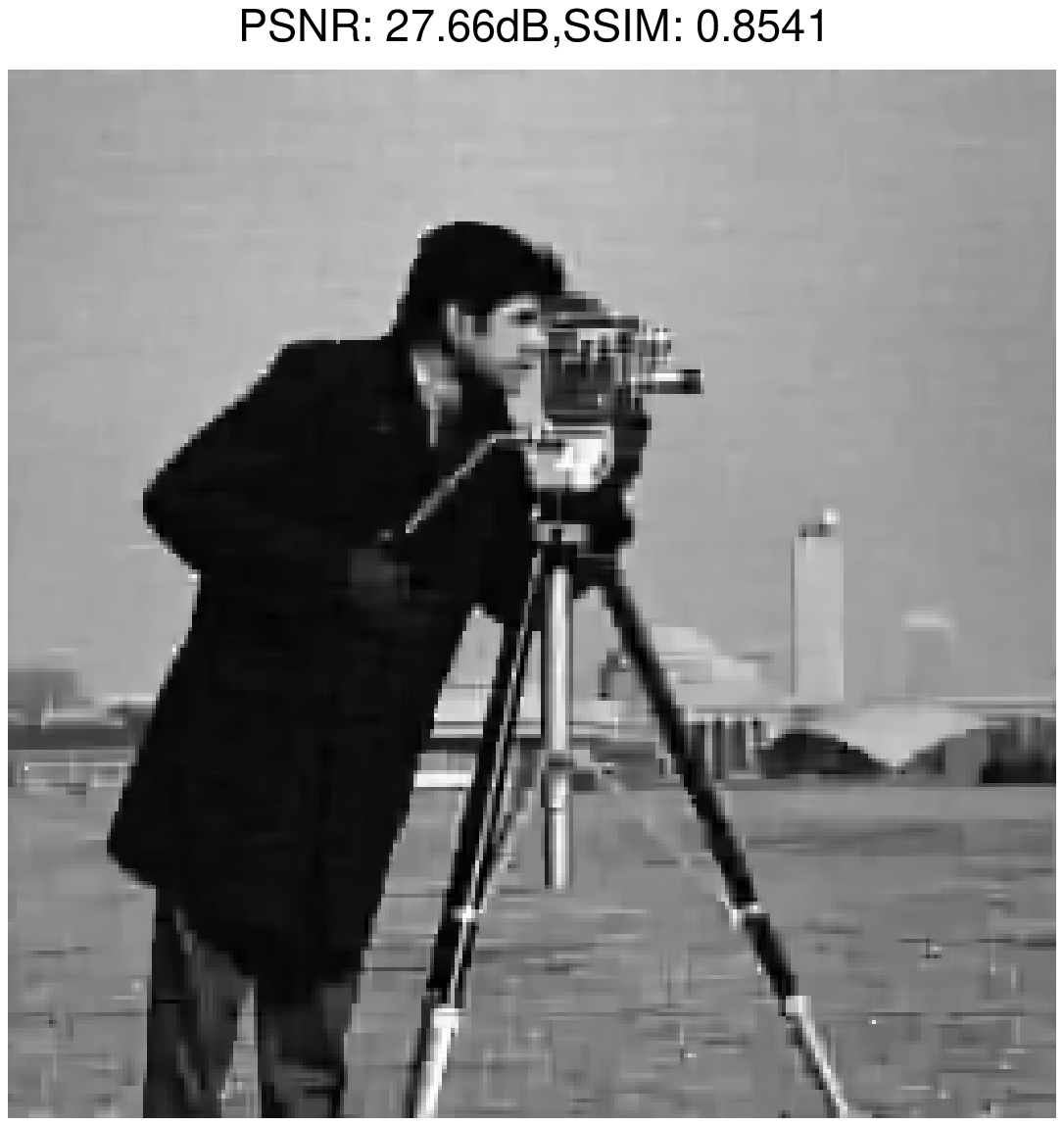}
   \includegraphics[width=0.15\textwidth]{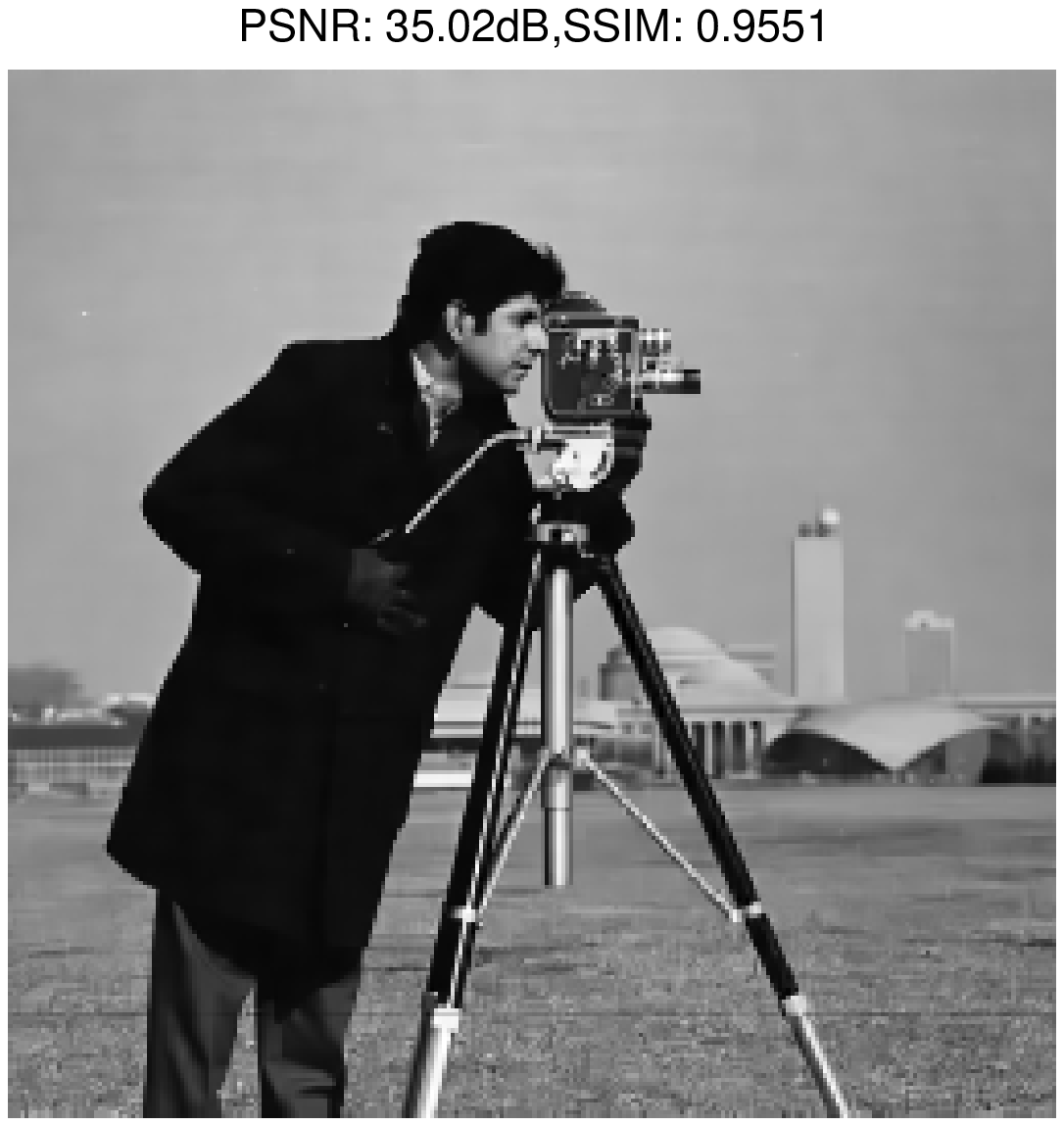}  \\
      \includegraphics[width=0.15\textwidth]{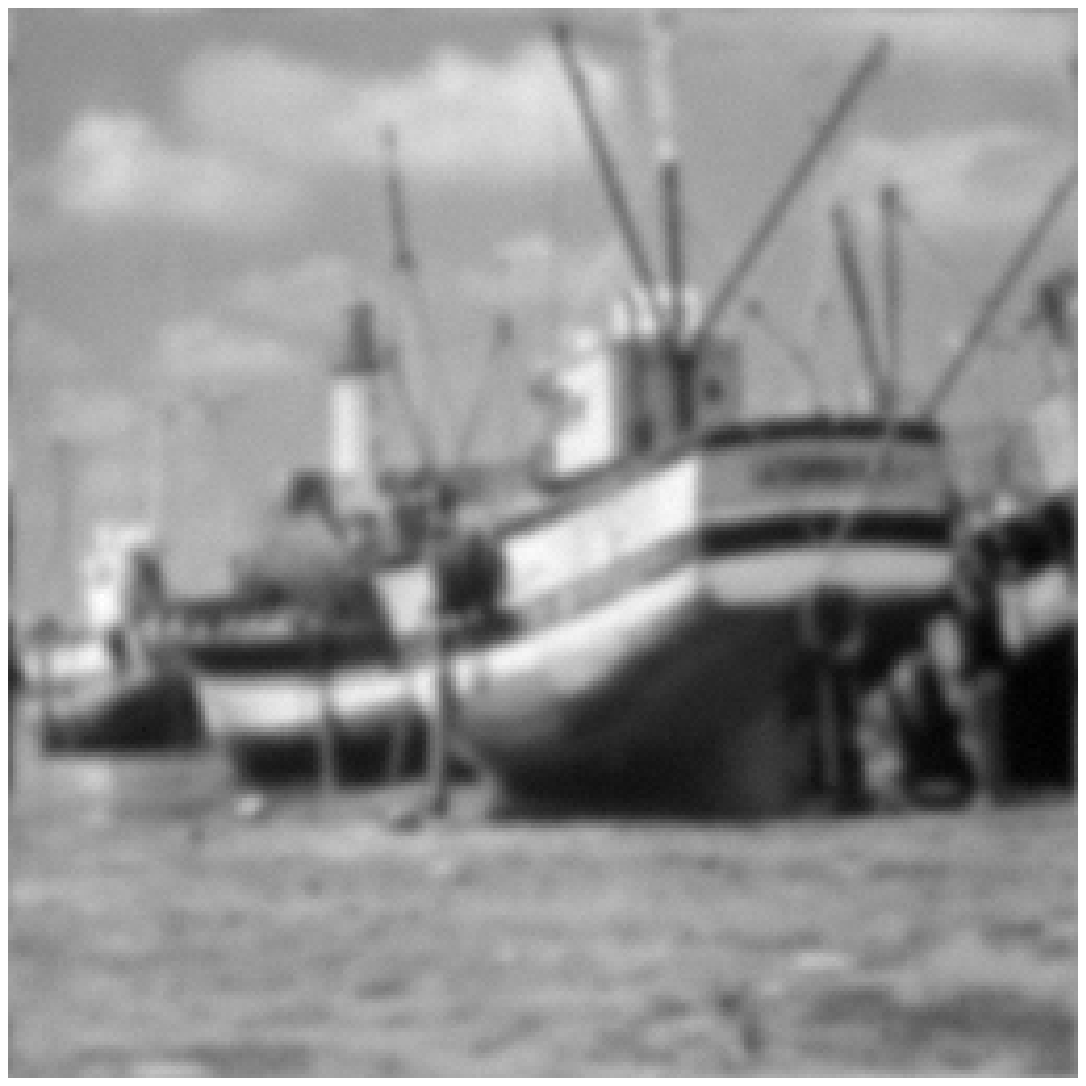}
   \includegraphics[width=0.15\textwidth]{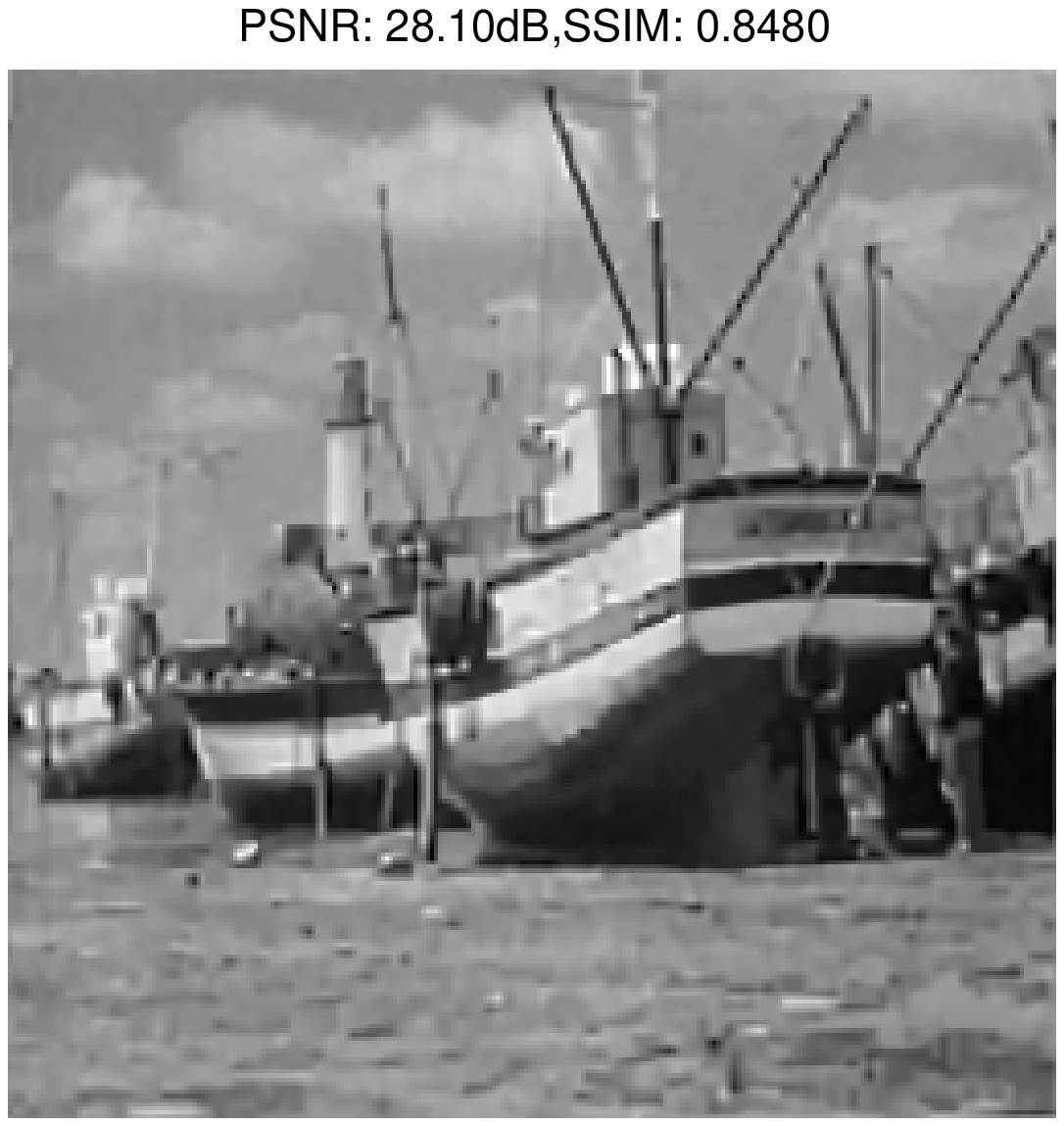}
   \includegraphics[width=0.15\textwidth]{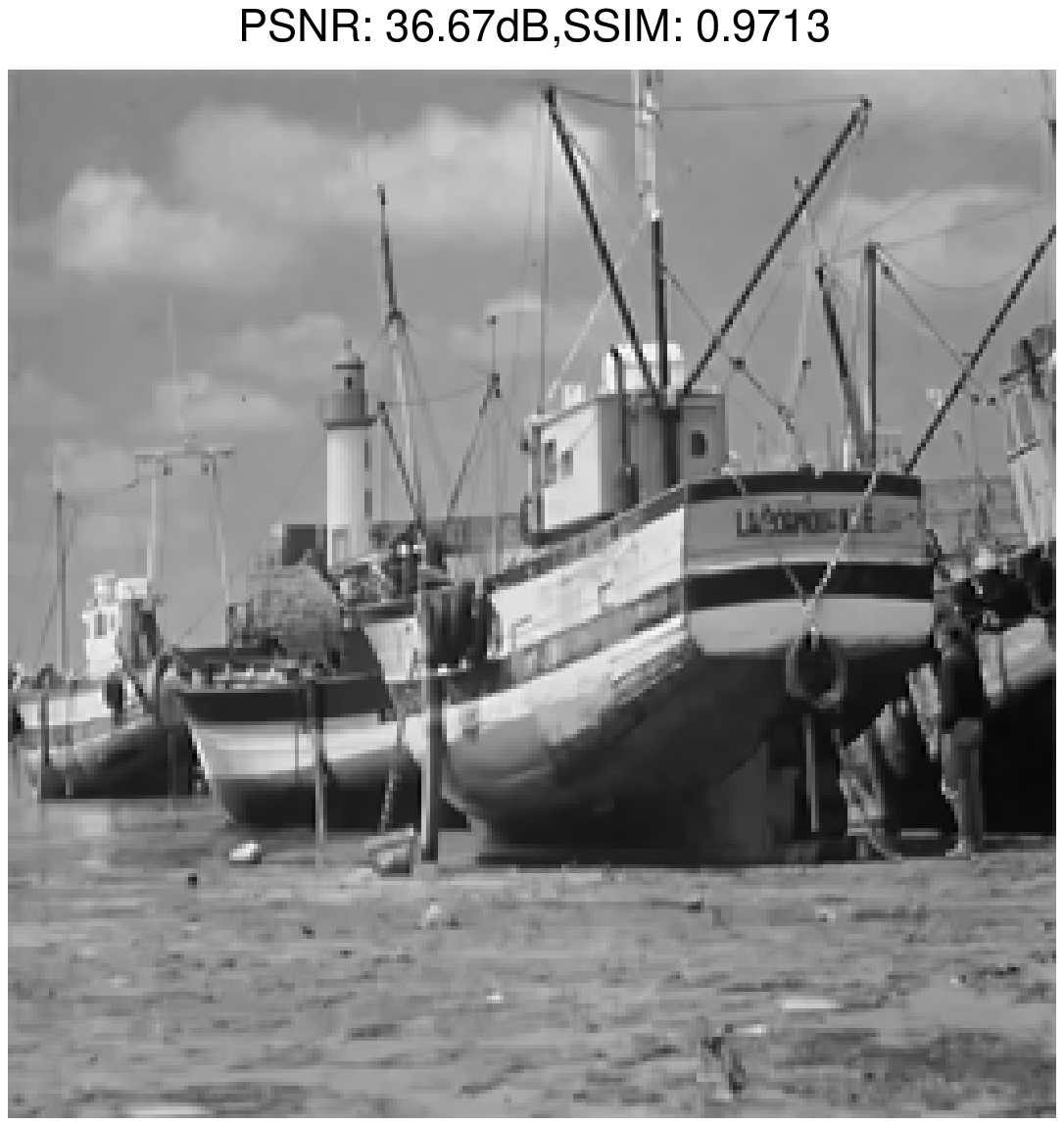}  \\
  \caption{\small  
  From left to right: degraded image, recovered image via classical $\ell_0$ regularization model where only sparsity prior is applied, recovered image via proposed truncated $\ell_0$ regularization model where both the sparsity prior and support information of frame coefficients are exploited.     }
\label{fig:benefit example}
\end{figure}

We would like to emphasize that the key component of our method is the support detection, and the final recovery performance largely depends on the accuracy of the detected support information of wavelet frame coefficients. In order for faithful image restoration, it is expected that the support estimation of frame coefficients should be as close as possible to those of the unknown original image under the given wavelet frame.
Intuitively, we need to have a relatively high quality reference image on which the support estimation is performed. For this purpose, at the initial stage, the support estimation can be performed on the recovery results of exisitng state-of-the-art image recovery methods, e.g., the IDD-BM3D method \cite{Danielyan2012} etc.  In other words, the proposed framework allows us to ``stand on the shoulders of giants".  Ones can pick up any existing image recovery algorithm and use its recovery result as the initial reference image and perform support estimation on it.  As we have known, once we are given reliable partial support information, proper exploitation of it can help improve recovery quality \cite{Saleh2015}, \cite{Wang2010}, \cite{Yu2013support}, \cite{Ince2013CS}. In short, we propose a truncated $\ell_0$ regularization to make use of the partial support information  in the restoration model.
%
%
%
%
The main contributions of this paper are summarized as below:

\begin{itemize}
\item Most existing wavelet frame based $\ell_1$ or $\ell_0$ minimization image processing models only make use of the sparsity prior. On the contrary,  the partial support information of the frame coefficients is learned as a prior and exploited in our work. It is the first time that a truncated $\ell_0$ regularization model based on self-learning of partial support information is proposed and the wavelet frame-based image deblurring  is a specific example in this paper.

\item While there have existed some works on exploiting partial support information to improve sparse recovery performance, they mostly assume that this partial support information is available beforehand \cite{Ince2013CS}. In addition, they are often discussed in the context of compressive sensing. Our method is a multi-stage self-learning procedure and applied to a different field---image deblurring.

\item The proposed algorithmic framework is able to  seamlessly incorporate the existing state-of-the-art image restoration methods by taking their results   as the initial reference image to perform support detection of frame coefficients. More precisely, our method is a self-contained iterative framework with open interface to the available existing image restoration methods. This makes the algorithm able to often achieve state-of-the-art performance. 

\item Moreover, this paper is expected to provide  new insights to other sparsity-based prior regularized image restoration methods. It might chalks out a path for us to explore:  learning (detecting) and exploiting support information is a general idea and can be readily incorporated into existing sparsity-driven methods.

\end{itemize}

The rest of this paper is organized as follows. In the next section, we first  briefly introduce some notations and preliminaries of
the wavelet tight frames. In Section III, the most related wavelet frame based and nonlocal patch based image restoration methods are revisited.
In Section IV, we introduce the proposed SDSR model and summarize the algorithmic framework. In Section V, extensive experiments are conducted to demonstrate the performance of the SDSR model. Section VI is devoted  to the conclusions of this paper and some discussions on possible future work.

\section{Notations and preliminaries}
In this section, we briefly introduce some preliminaries of wavelet tight frames.
Tight wavelet frame are widely applied in image processing. One wavelet frame for $L_2(\mathbb{R})$ is a system generated by the shifts and dilations of a finite set of generators $\Psi = \{ \Psi_1, \Psi_1, \ldots, \Psi_n \} \subset L_2(\mathbb{R})$:
\[
X(\Psi) = \{ \Psi_{l,j,k}, 1\leq l, j\in \mathbb{Z}, k\in \mathbb{Z} \}
\]
where $\Psi_{l,j,k} = 2^{j/2}\Psi_l(2^j \cdot -k)$. Such set $X(\Psi)$ is called tight frame of $L_2(\mathbb{R})$ if
\[
f = \sum_{\psi\in \Psi}<f,\psi>\psi,   \forall f \in L_2(\mathbb{R}).
\]
The construction of framelets can be obtained according to the unitary extension principle (UEP). Following the common experiment implementations, the linear B-spline framelet is used by considering the balance of the quality and time. The linear B-spline framelet has two generators and the associated masks $\{ h_0, h_1, h_2 \}$ are
\[
h_0 = \frac{1}{4}[1,2,1]; h_1=\frac{\sqrt{2}}{4}[1,0,-1]; h_2=\frac{1}{4}[-1,2,-1].
\]
Given the 1D tight wavelet frame, the framelets for $L_2(\mathbb{R}^2)$ can be easily constructed by using tensors products of 1D framelets.

In the discrete setting, we will use $ W \in \mathbb{R}^{m \times n} $ with $ m \geq n $ to denote the transform matrix of framelet decomposition and use $ W^T$ to denote the fast reconstruction. Then according to the unitary extension principle we have $ W^TW = I $. The matrix $W$ is called the analysis (decomposition) operator, and its transpose $W^T$ is called the synthesis (reconstruction) operator. The $L$-level framelet decomposition of $u$ will be further denoted as:
\[
Wu = (\ldots,W_{l,j}u,\ldots) \qquad  \mathrm{for} \quad 0\leq l \leq L-1,j \in \mathcal{I}
\]
where $ \mathcal{I} $ denotes the index set of the framelet bands and $ W_{l,j}u \in \mathbb{R}^n $ is the wavelet frame coefficients of $u$ in bands $j$ at level $l$. The frame coefficients $W_{l,j}u$ can be constructed from the masks associated with the framelets. We consider the $L$-Level undecimal wavelet tight frame system without the down-sampling and up-sampling operators as an example here. Let $h_0$ denote the mask associated with the scaling function and $\{h_1, h_2, \ldots, h_n\}$ denote the masks associated with other framelets. Denote
\begin{equation} \label{eq: frame construction}
h_j^{(l)} = \underbrace{h_0*h_0*\cdots h_0}_{l-1}*h_j
\end{equation}
where $*$ denotes the discrete convolution operator. Then $W_{l,j}$ corresponds to the Toeplitz-plus-Hankel matrix that represents the convolution operator $h_j^{(l)}$ under Neumann boundary condition. We refer the readers to \cite{Cai2012}, \cite{Dong2010} for further detailed introduction of wavelet frame and its applications.

\section{Related work}
The proposed algorithm is based on a truncated $\ell_0$ regularized model, where the truncation depends on the detected support information. The truncated $\ell_0$ regularized model can be considered as a variant of $\ell_0$ regularized model. Therefore, we briefly revisit $\ell_0$ regularized wavelet frame based image recovery model. As a common counterpart, we also review the classical $\ell_1$ norm regularized wavelet frame based image recovery model. The nonlocal patch based methods such as IDD-BM3D \cite{Danielyan2012} will also be reviewed, since they have achieved state-of-the-art image deblurring results, which will be used as the initial reference images for the support estimation.

\subsection{$\ell_1$ norm regularized wavelet frame-based methods}
Due to the redundancy of the wavelet frame systems ($WW^T \neq I$), there are several different wavelet frame based models, including the synthesis model, the analysis model, and the balanced model. However, what these models share in common is that they mostly penalize the $\ell_1$ norm of the wavelet frame coefficients for sparsity constraint in different ways. Detailed description of these different models can be referred in \cite{Shen2011}. Numerical experiments in \cite{Shen2011} have shown that the quality of the recovery images by these models is approximately comparable. Therefore, we only consider the analysis based approach here:
\begin{equation} \label{eq:single l1}
\min_u \frac{1}{2}||Au-f||_2^2 + ||\lambda \cdot Wu||_{1,p}
\end{equation}
where $p = 1$ or $p = 2$ corresponds to anisotropic $ {\ell}_1 $ norm and isotropic $ {\ell}_1 $ norm, respectively.
Here, the generalized $ {\ell}_1 $-norm is defined as
\begin{equation}
||\lambda \cdot Wu||_{1,p} = ||\sum_{l=0}^{L-1}\left(\sum_{j \in \mathcal{I}}\lambda_{l,j}|W_{l,j}u|^p\right)^{1/p}||_1
\end{equation}
where $| \cdot |^p $ and $(\cdot)^{\frac{1}{p}} $ are entrywise operations. We introduce $ \alpha = Wu $ and substitute it into (\ref{eq:single l1}), then we can obtain the rewritten form of (\ref{eq:single l1}) as follows
\begin{equation} \label{eq:constraint single l1}
\min_{u,\alpha} \frac{1}{2}||Au-f||_2^2 + ||\lambda \cdot \alpha||_{1,p} \qquad  s.t.\quad \alpha = Wu.
\end{equation}
Note that the convex optimization problem (\ref{eq:single l1}) or (\ref{eq:constraint single l1}) can be solved via many existing efficient algorithms, e.g., split bregman or alternating direction method \cite{Boyd2010},  \cite{Goldstein2009}.

\subsection{ $\ell_0$ quasi-norm regularized wavelet frame-based methods}

It is well known that the $\ell_1$ norm based approaches are capable of obtaining sparsest solution if the operator $A$ satisfies certain conditions according to compressed sensing theories developed by Candes and Donoho \cite{Candes2008}. For image restoration tasks, unfortunately, the conditions are not necessarily satisfied. Therefore, the $\ell_1$ norm based models often achieve suboptimal performance.

Recently,  $\ell_p$ quasi-norm ($0\leq p < 1$) regularization was further investigated to recover the images with better preserving of sharp edges.  The authors in \cite{ZhangDongBin2013} proposed to use the $\ell_0$ quasi-norm instead of the $\ell_1$ norm in the analysis model:
\begin{equation} \label{eq:single l0}
\min_u \frac{1}{2}||Au-f||_2^2 + \lambda || Wu ||_0
\end{equation}
The $\ell_0$ quasi-norm of $||\alpha||_0$ is defined to be the number of the non-zero elements of $\alpha$. Note that its proximity operator can be easily computed by the hard-thresholding operator. An algorithm called PD method was proposed to solve this $\ell_0$ minimization problem in \cite{ZhangDongBin2013}. Recently, a more efficient algorithm called MDAL method was developed for solving the same problem \cite{DongBin2013}.

\subsection{The nonlocal patch based methods}
The nonlocal approaches are built on the observation that image structures of small regions tend to repeat themselves in spatial domain, which is suitable for exploiting the redundancy information in natural images. It is started with the nonlocal means method proposed by Buades et.al \cite{Buades2005} for image denoising and has been extended to solve other inverse problems in image processing tasks; see e.g., \cite{Zhangxiaoqun2010}, \cite{Dabov2007}, \cite{Danielyan2012}. Very recently, the nonlocal idea is combined with the patch dictionary methods and generate the current state-of-the-art methods in the quality of restored images \cite{Aharon2006}, \cite{Elad2006}, \cite{Mairal2009}, \cite{Dong2011CVPR}, \cite{Dong2011},  \cite{Zhang2014}.

\section{Wavelet frame-based support driven sparse regularization (SDSR) model} \label{Sec:SDSR}

In this section, we propose a wavelet frame-based support driven sparse regularization (SDSR) model based on a truncated $\ell_0$ regularized term.
The model is formulated as follows.


\begin{equation} \label{eq:new truncatedl0l2}
\min_u \frac{1}{2}||Au-f||_2^2 + \lambda|| (Wu)_T ||_0
\end{equation}
where $(Wu)_T$ is the truncated version of $Wu$ and this truncation is the main difference with the $\ell_0$ regularized model. The index set of frame coefficients $T$ denotes the complementary set of the detected support set $I$, i.e., $T = I^{C}$. $I$ is  unknown beforehand.  
 In the common $\ell_0$ model, $I$ is an empty set.

In order to avoid confusion to the readers, we give a toy example to illustrate these notations here.
Assuming that we want to recover a underlying sparse vector $\bar{a} = (0, 10, 0, 25, 20, 0)$ via this model, the true support index set of $\bar{a}$ is $\bar{I} = \{ 2, 4, 5\}$. The detected partial support set might be $I = \{ 2, 4 \}$, and $T = \{ 1, 3,  5, 6\}$. The components corresponding to $T$ will remain in the truncated $\ell_0$ quasi-norm while those corresponding to $I$ will be truncated out.  

The truncated $\ell_0$ regularization comes from the simple intuition that a frame coefficient should be not forced to move closer to 0 and needs to be moved out of the regularizer term, if this coefficient is believed to be a nonzero component. In other words, once the locations of some nonzero components (especially those of large magnitude) are identified, this kind of truncation aims to make them not  be shrunk. The resulted benefit is the better preserving of sharp edges.

Note that the index set of $T$ or $I$ is unknown beforehand as the underlying true image is not available.
Thus, the key question is to perform the support detection to determine $I$, i.e.,  which frame coefficients should be truncated out of the $\ell_0$ quasi-norm, and this is determined via a self-learning strategy.  After the acquiring of the partial support information, we will have a truncated $\ell_0$ regularized  model (\ref{eq:new truncatedl0l2}) with $T$ known.

In summary, SDSR is a multi-stage alternative optimization procedure, which repeatedly applies the following two steps when applied to model (\ref{eq:new truncatedl0l2}):

 $\bullet$ Component 1: we perform support detection on a reference image and determine the $I$.

 $\bullet$ Component 2: we solve the truncated $\ell_0$ regularized  model (\ref{eq:new truncatedl0l2}) with $T$ known.  The result will acts as a reference image for support detection in Component  $1$ of the next stage. \\


\subsection{Component 1: Support detection to determine $T$}
Given a reference image, support detection is performed on it to estimate some partial support information of the underlying true image.
 For the first stage, the initial reference image comes from the results of other state-of-the-art image deblurring methods such as IDD-BM3D method. 

In this work, we adopt a heuristic but effective support detection method, which is similar to the strategy proposed in \cite{Wang2010}. The support detection is based on the thresholding strategy, where we retain indexes of frame coefficients whose magnitude larger than the threshold value as the support set. For $s$-th stage, note that we have the intermediate recovery result $u^{(s)}$ at hand, and the support index set of frame coefficients is obtained as follows:
\begin{equation} \label{eq: support detection}
I^{(s+1)}:= \{ i: |(Wu^{(s)})_i| > \epsilon^{(s)} \}
\end{equation}
Correspondingly, the index set of frame coefficients kept in the $\ell_0$ quasi-norm is
$T^{(s+1)}=(I^{(s+1)})^C.$
We set the threshold value:
\begin{equation}\label{eq:threshold-value setting}
\epsilon^{(s)} := ||Wu^{(s)}||_{\infty}/\rho.
\end{equation}
with $\rho > 0$.
Empirically, the performance of our method is not very sensitive to the choice of $\rho$, and a small percentage of wrong support detection will not degrade the performance of the proposed method. 

\subsection{Component 2: Solving the truncated $\ell_0$ model with $T$ known}
Given $T$, (\ref{eq:new truncatedl0l2}) becomes a non-convex problem in terms of $u$ and most existing algorithms for the common $\ell_0$ (without truncation) regularized model can be slightly modified and applied. Here the mean doubly augmented Lagrangian (MDAL) method \cite{DongBin2013} is adopted.

We introduce $ \alpha = Wu $, and the equivalent constraint optimization problem of (\ref{eq:new truncatedl0l2}) is:
\begin{equation} \label{eq: Truncated l0l2 constraint}
\min_{u, \alpha} \frac{1}{2}||Au-f||_2^2 + \lambda || \alpha_T ||_0 , \quad s.t. \quad \alpha = Wu
\end{equation}
The MDAL method applied to (\ref{eq: Truncated l0l2 constraint}) is formulated as:
\begin{equation}\label{eq: DAL truncated l0l2}
\left\{\begin{array}{lll}
u^{k+1} = \mathrm{arg} \min_u \frac{1}{2}||Au - f||_2^2 + \frac{\mu}{2}||Wu - \alpha^{k} + b^{k}||_2^2 \\
\ \ \ \ \ \ \ \ \ \ \ \ \ \ \ \ \ \ \ \ \ \ \ \ \ \ \ \ \ \  \ \ \ \ \ \ \  + \frac{\gamma}{2}||u - u^{k}||_2^2 \\
\alpha^{k+1} = \mathrm{arg} \min_{\alpha} \lambda||\alpha_T||_0  + \frac{\mu}{2}||\alpha - (Wu^{k+1} + b^{k})||_2^2 \\
\ \ \ \ \ \ \ \ \ \ \ \ \ \ \ \ \ \ \ \ \ \ \ \ \ \ \ \ \ \  \ \ \ \ \ \ \  + \frac{\gamma}{2}||\alpha - \alpha^{k}||_2^2 \\
b^{k+1} = b^{k} + Wu^{k+1} - \alpha^{k+1}
\end{array}\right.
\end{equation}
Compared with  the common (without truncation) $\ell_0$ quasi-norm regularized model, the main modification of MDAL here lies in the subproblem $\alpha_{k+1}$. Specifically,
\begin{equation}\label{eq:subproblem alpha}
\alpha^{k+1} =  \mathcal{H}_{T, \lambda, \mu, \gamma}( Wu^{k+1}+b^k, \alpha^k)
\end{equation}
where the operator $\mathcal{H}$ is a generalized component-wisely \textit{selective} (defined by $T$) hard-threhsolding operator defined as follows:
\begin{equation} \label{eq:selective operator}
(\mathcal{H}_{T, \lambda, \mu, \gamma}(x,y))_i = \left\{ \begin{array}{ll}
0,  \mathrm{if} \quad  i \in T  \quad \mathrm{and} \quad | \frac{ \mu x_i + \gamma y_i}{\mu + \gamma} | < \sqrt{ \frac{2\lambda}{\mu + \gamma} }       \\
\frac{ \mu x_i + \gamma y_i}{\mu + \gamma}, \quad \mathrm{otherwise} \end{array} \right.
\end{equation}
 It is well known that the edges of an image should correspond to the large nonzero frame coefficients. Note that only the components of $\alpha$ indexed in $T$ perform the hard-threhsolding, while the nonzero components belonging to the index set $I$ are not shrunk. Thus this selective hard-threhsolding operator expects to reduce the wrong shrinkage, leading to better edge preserving performance of the recovered images.

 As the original MDAL, the minimization with respect to $u$ remains to be  a least square problem with the normal equation
\begin{equation}\label{eq:subproblem u}
(A^TA + (\mu + \gamma) I)u^{k+1} = A^Tf + \gamma u^k + \mu W^T(\alpha^{k} - v^k)
\end{equation}
Under the periodic boundary conditions for $u$, the entire left-hand side matrix in (\ref{eq:subproblem u}) can be diagonalized
by the discrete Fourier transform, and thus it is simple and fast to solve.

Following the implementation of \cite{DongBin2013}, we use the arithmetic means of the solution sequence, denoted by
\begin{equation} \label{eq:mean step}
\bar{u}^{k} = \frac{1}{k+1}\sum_{j=0}^{k} u^j; \quad  \bar{\alpha}^{k} = \frac{1}{k+1}\sum_{j=0}^{k} \alpha^j.
\end{equation}
as the final output instead of the sequence $(u^k, \alpha^k)$ itself.

\subsection{Summary of the algorithm}
From the analysis in previous sections, we can see that the SDSR model in Component 1 and Component 2 work together to gradually detect the support set and improve the recovery performance. Now, we summarize the SDSR deblurring algorithmic framework below.  \\

\begin{tabular}{l}
\hline
\textbf{Algorithm 1} Image deblurring via wavelet frame-based \\
  \qquad   \qquad  \quad     SDSR model  \\
\hline
Given observed image $f$ and convolution operator $A$. \\
1. \textbf{Initialization}:  Compute an initial recovery result via  \\
 any image restoration methods, e.g., IDD-BM3D method, \\
 as the initial reference image for initial support detection.  \\

2. \textbf{Outer loop (stage)}: iteration on $s=$ $1,\ldots,S $ \\

 \quad (a)  Perform  support detection on the reference image. \\

 \quad (b) \textbf{Inner loop} (solving the truncated $\ell_0$ model (\ref{eq:new truncatedl0l2})) \\
 \qquad\textbf{ While} the stopping condition is not satisfied, \\
 \ \ \ \ \ \ \ \ \ \ \ \ \ \ \ \ iterate on $k=1,2,\ldots,K$ \textbf{Do}   \\
 \ \ \ \ \ \ \ \ \ \ \ \ \ \ \ \ (I)   Image estimate $  u^{k+1}  $  via (\ref{eq:subproblem u}).                                \\
\ \ \ \ \ \ \ \ \ \ \ \ \ \ \ \  (II)  Compute $ \alpha^{k+1} $ via (\ref{eq:subproblem alpha}).  \\
\ \ \ \ \ \ \ \ \ \ \ \ \ \ \ \  (III) Update  $   b^{k+1} = b^{k} + Wu^{k+1} - \alpha^{k+1} $.       \\
 \qquad\textbf{ End}   \\
 \qquad Compute the arithmetic means of the solution \\
\qquad  sequence as the final output via (\ref{eq:mean step}). It also acts\\
\qquad  as the reference image for support detection of \\
\qquad the  next stage. \\
\hline
\end{tabular} \\

\subsection{Taking an even closer look at the proposed algorithm}


The discrete wavelet frame coefficients are obtained by applying wavelet frame filters to a given image. Since the wavelet frame filters are designed to be standard difference operators with various orders, the locations of large wavelet frame coefficients indicate the edges of a given image. The locations of small wavelet frame coefficients indicate the region where image is smooth.
A good image restoration method should preserve smooth image components while enhancing sharp image edges. This is a rather challenging
task since smoothing and preservation of edges are often contradictory to each other.

The basic motivation behind the wavelet  frame based sparsity regularized methods is to  promote the sparsity of the wavelet frame coefficients of the recovery images via shrinkage operators so that edges can be well preserved.
It is well known that soft-thresholding operator and hard-thresholding operator are equivalent as the minimization of $\ell_1$-norm and $\ell_0$-quasi-norm based optimization model, respectively.  However,
simply applying either  $\ell_1$-norm or $\ell_0$-quasi-norm penalization may weaken the sharpness of the edges and introduce unwanted artifacts in smooth regions. Tuning the regularization parameter in the model may reduce these artifacts, but  it may smear out edges at the same time, see \cite{DongBin2013} for details.

Instead of just passively using a sparsity promoting function (such as the $\ell_1$-norm and $\ell_0$-quasi-norm) and hoping  the paradox between smoothness and sharpness can be resolved automatically. In this paper, we actively exploiting other helpful information to rectify this shortcoming, i.e., detecting (learning) the location information of large nonzero frame coefficients.
The selective hard shrinkage operator (\ref{eq:selective operator}) is equivalent as the minimization of an truncated $\ell_0$ quasi-norm based optimization model,  which can be viewed as a data-driven adaptive shrinkage operator. 

The key component of our algorithm is the support detection, and the final recovery performance largely depends on this prior. Note that given an original clean image, its support information with the given wavelet frame is unique, however unknown for us in practice. Thus we need to design a method to learn this useful information or at least part of it, for some reference images. 
Intuitively, the higher quality of the reference image where the support detection performed on, more accurate support information should be acquired. For this purpose, we perform the support detection of first stage based on the recovery results of existing state-of-the-art methods, e.g., IDD-BM3D method etc. ``A good beginning is half the battle", and we can acquire even more reliable support set as the iteration of our algorithm proceeds, leading to gradually improved recovery results.

It should be pointed out that our algorithm is not just a post-processing. Since our algorithmic framework has open interface to any available image recovery results as the initial reference images to perform support detection, thus we can ``stand on the
shoulders of the giants", i.e., using the state-of-the-art results as the initial reference images. However, our algorithm itself is an self-contained iterative procedure, alternatively performing support detection on the recent recovery and returning an updated one by solving the resulted truncated $\ell_0$ model. From the viewpoint of non-convex optimization,  we admit the importance of picking an appropriate initial point, for example, the result of state-of-the-art image debulrring algorithms can be used here, we would like to emphasize the importance of a well-design searching method, which corresponds to the self-contained iterative procedure based on the support detection and the solving of a truncated $\ell_0$ model.

The proposed Algorithm is simple in implementation and efficient in computation. We emphasize that mostly computational cost of Algorithm 1 is the \textbf{Initialization} process, for example, for a gray-scale image of size $256\times 256$, it takes about 5 minutes for IDD-BM3D method, while the total cost of a single \textbf{Outer loop (Stage)} is merely around 20 seconds.

\section{Numerical experiments}

\begin{figure}[h]
  \centering
   \includegraphics[width=0.08\textwidth]{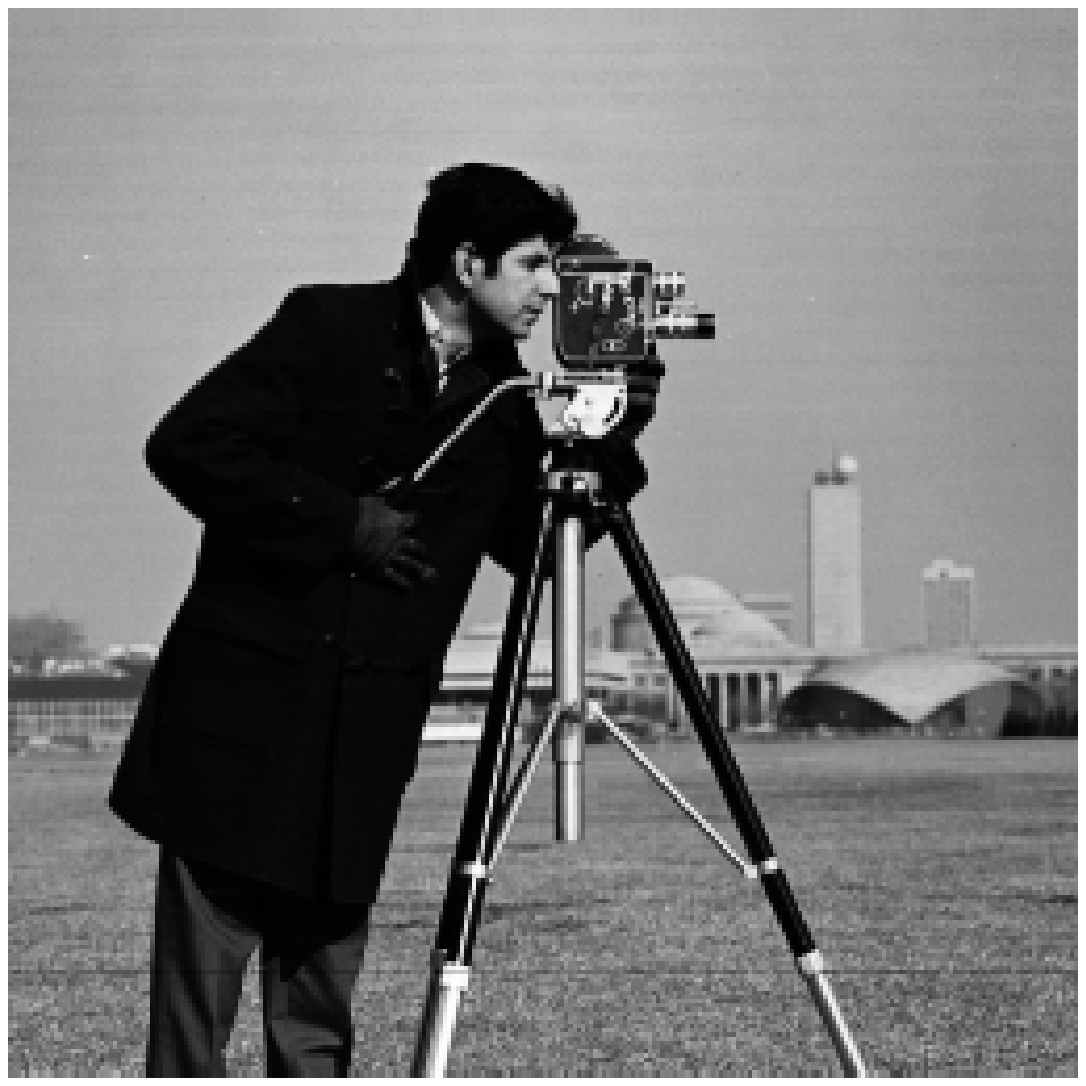}
   \includegraphics[width=0.08\textwidth]{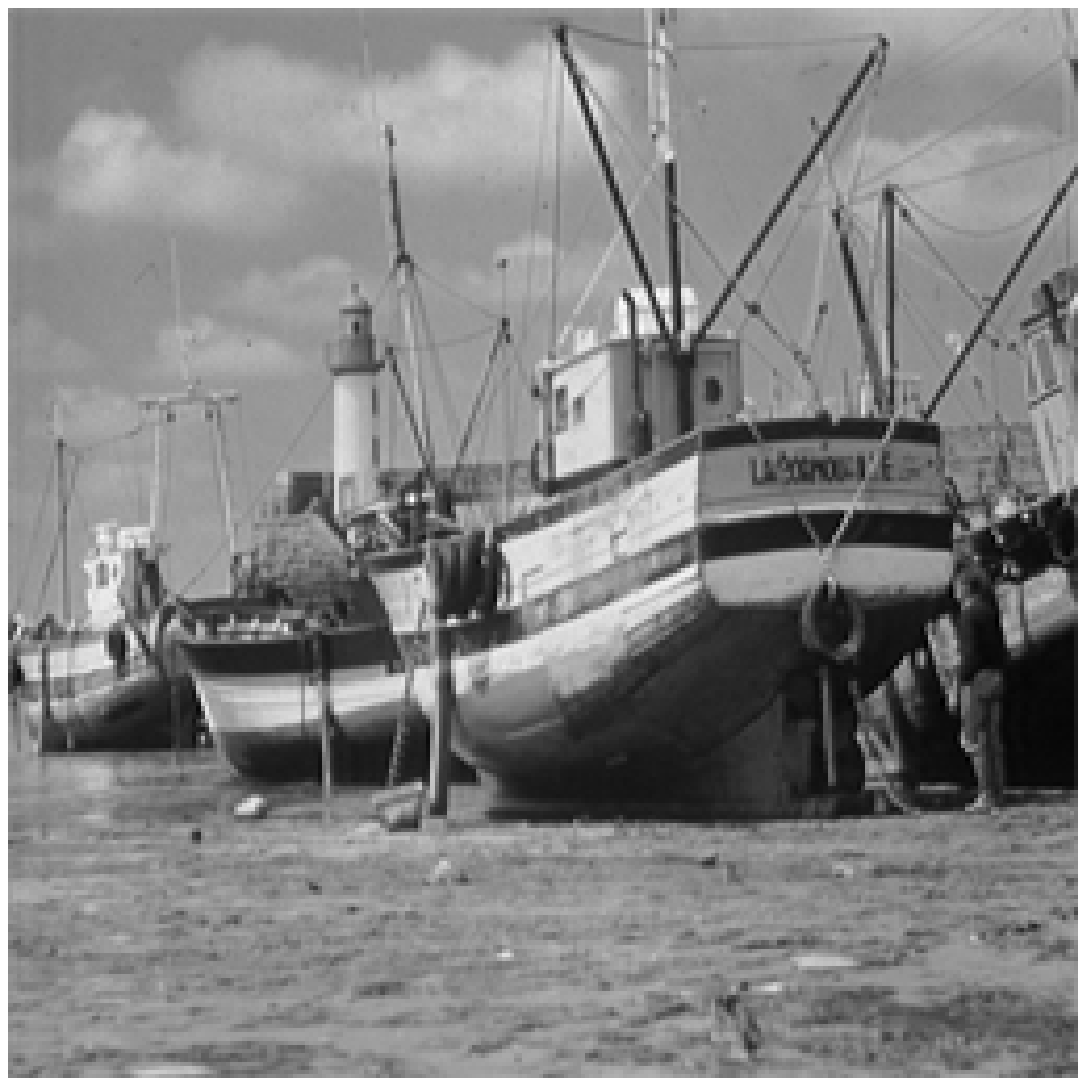}
   \includegraphics[width=0.08\textwidth]{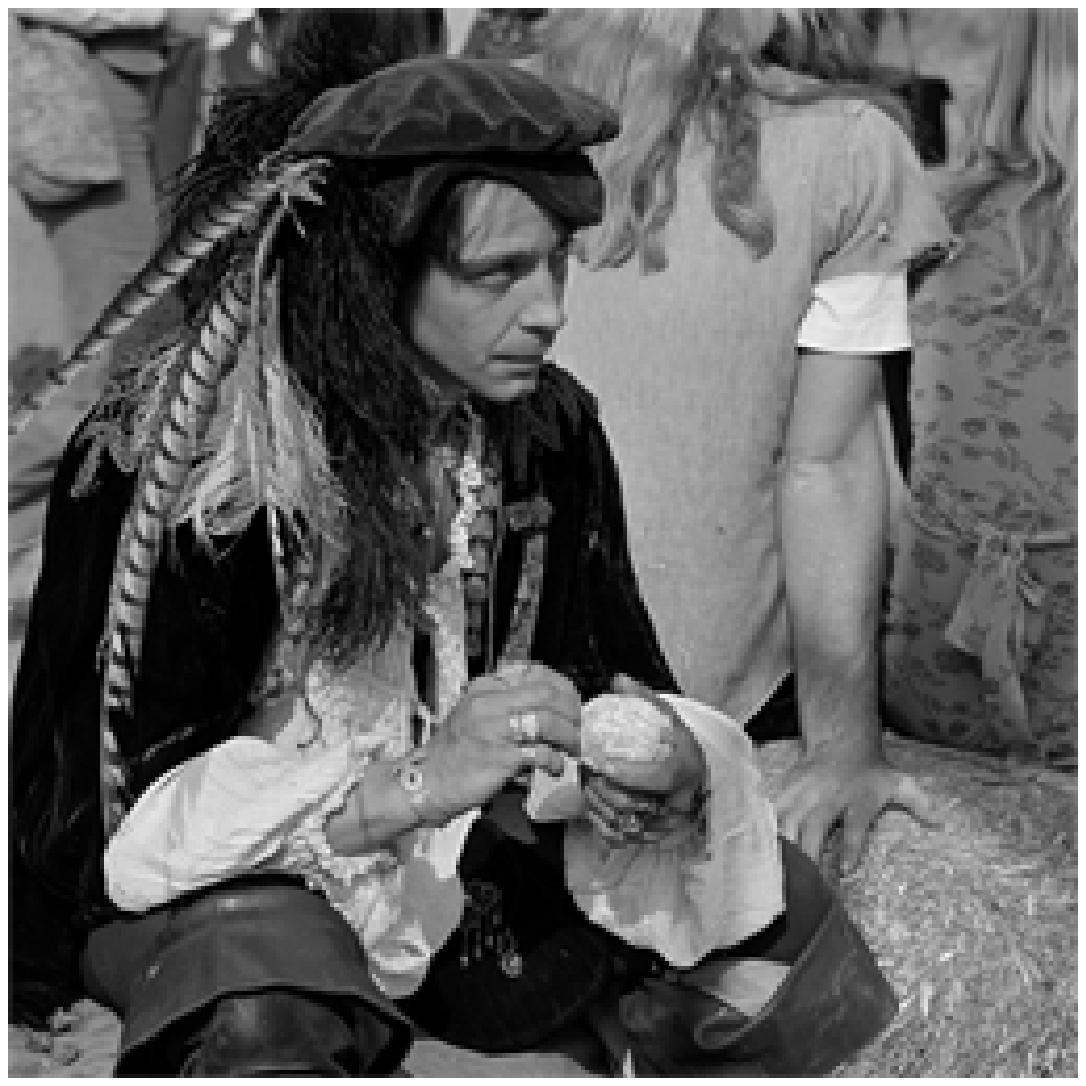}
   \includegraphics[width=0.08\textwidth]{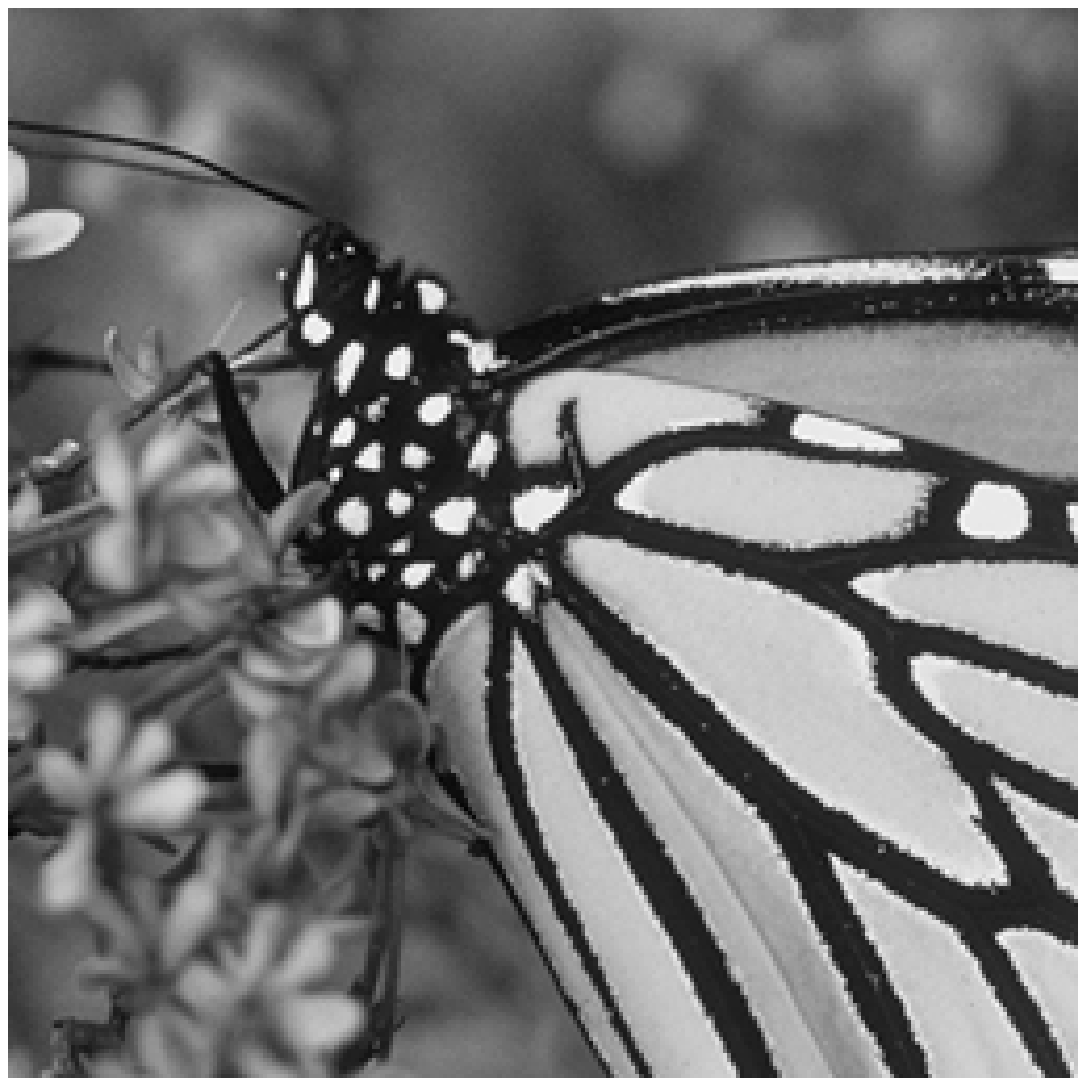}
   \includegraphics[width=0.08\textwidth]{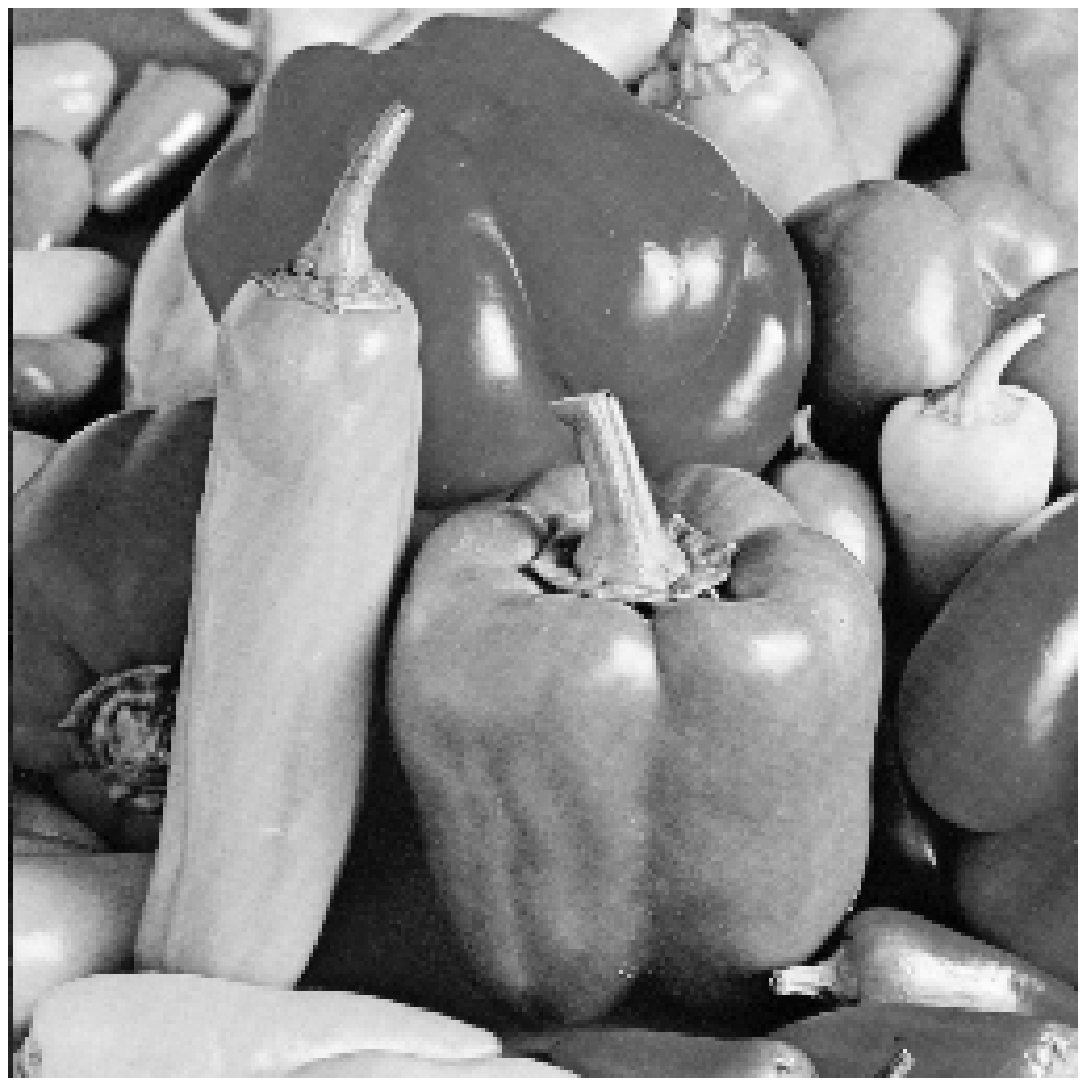} \\
      \includegraphics[width=0.08\textwidth]{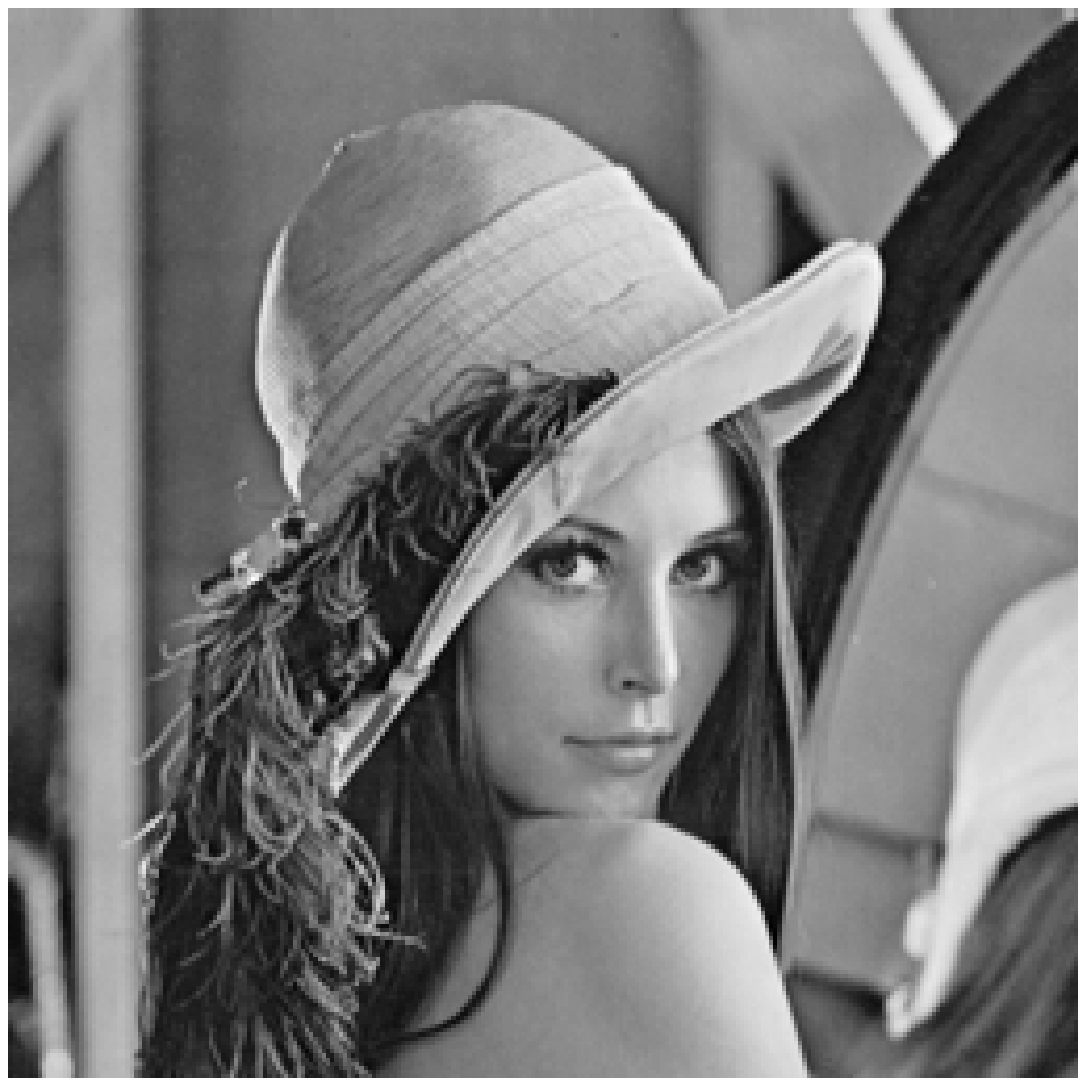}
   \includegraphics[width=0.08\textwidth]{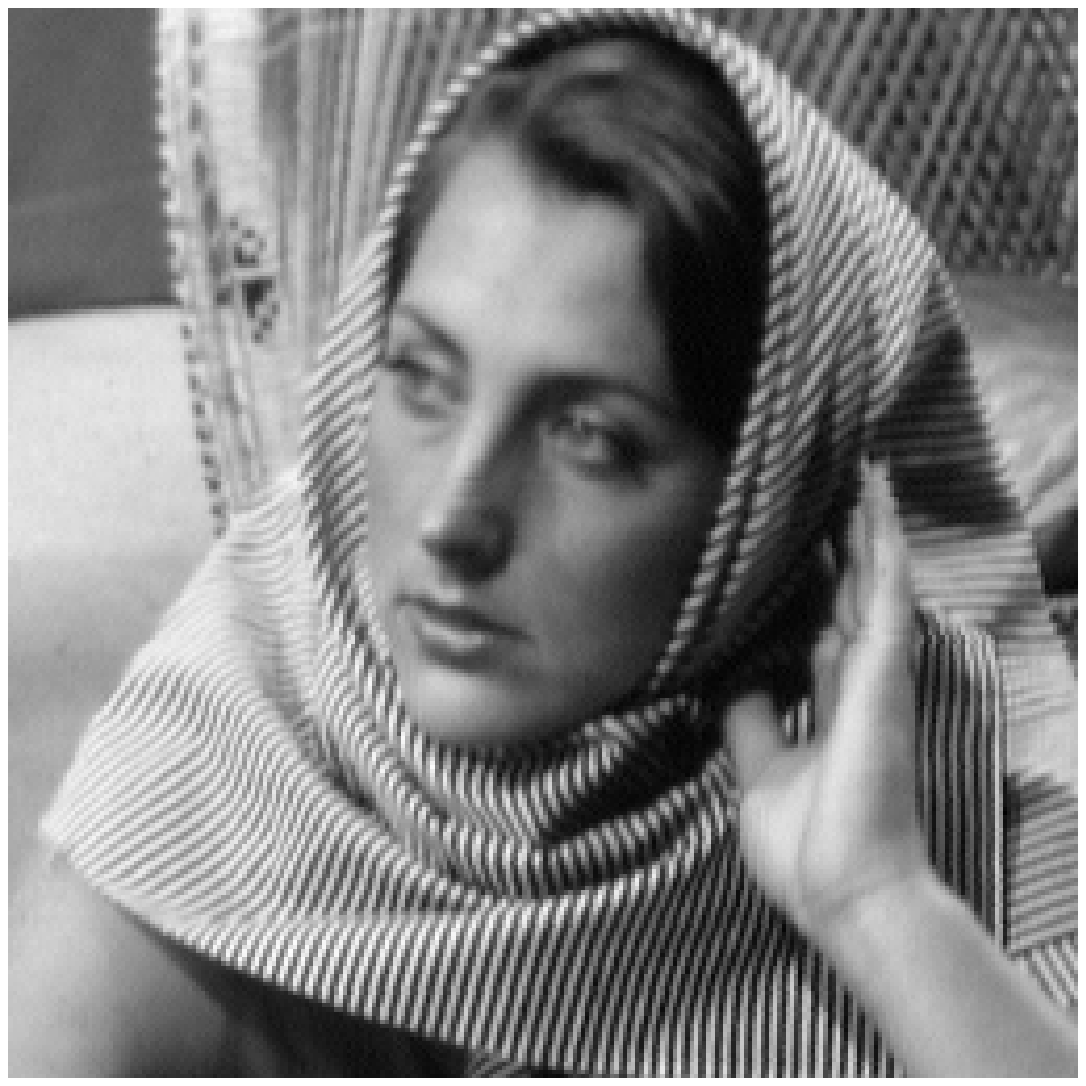}
   \includegraphics[width=0.08\textwidth]{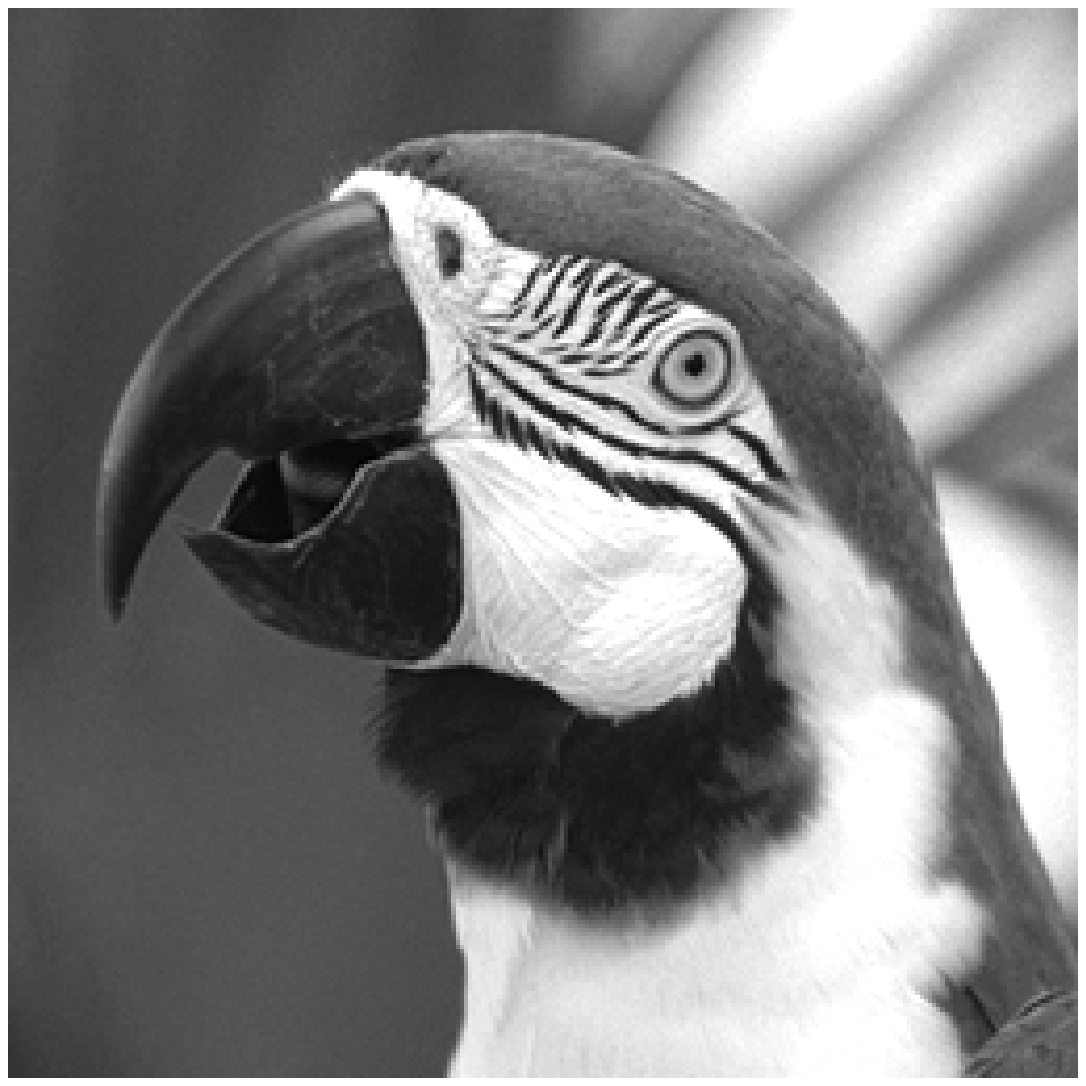}
   \includegraphics[width=0.08\textwidth]{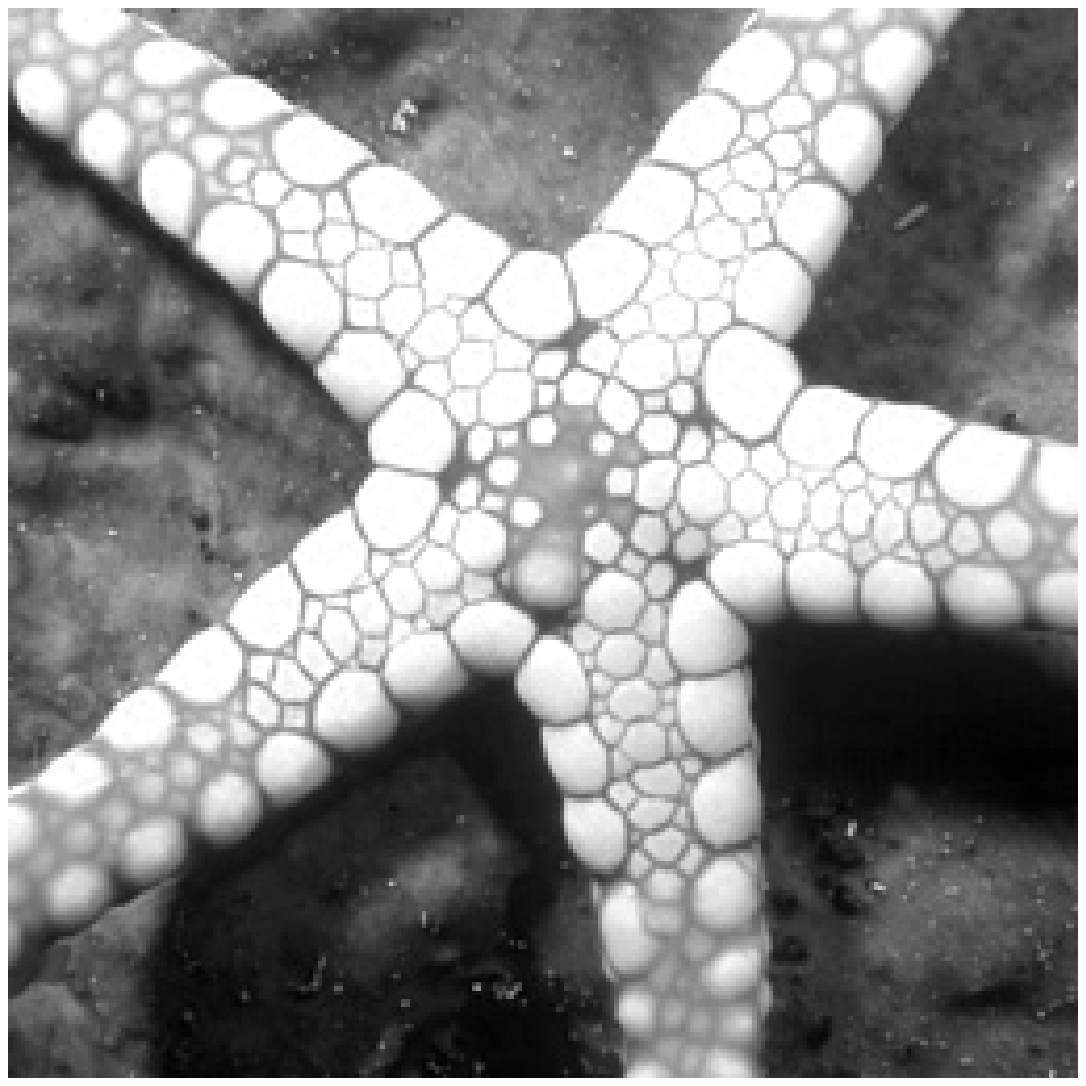}
   \includegraphics[width=0.08\textwidth]{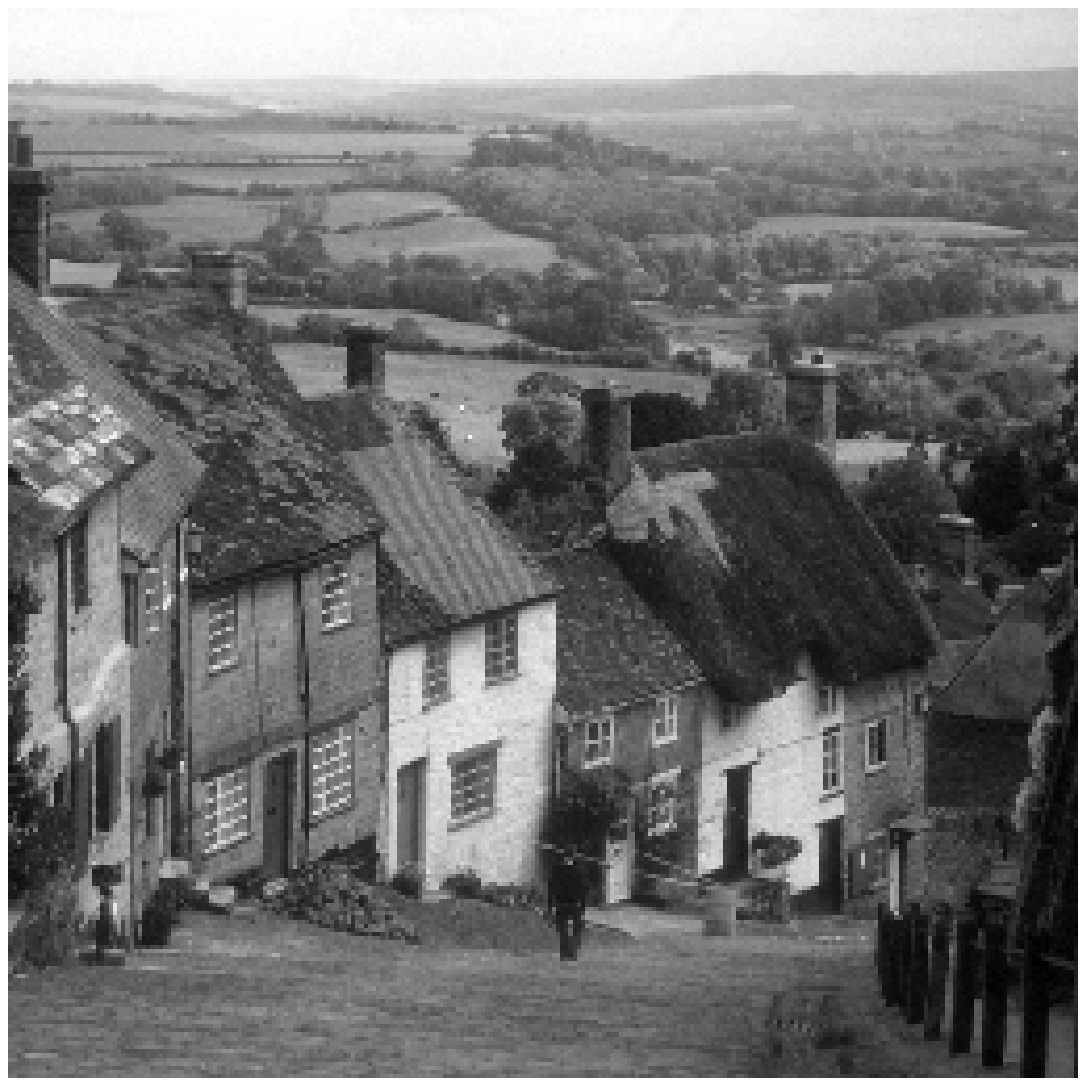} \\
  \caption{\small  All experimental test images. From left to right and top to bottom: \textit{C.man, Boat, Man, Monarch, Peppers, Lena, Barbara, Parrots, Starfish, Goldhill}, respectively.}
\label{fig:testimages}
\end{figure}

\subsection{Experimental settings}
 In this section,  extensive experiments are conducted to demonstrate the performance of our proposed SDSR model for image deblurring. The intensity of a pixel of these test images ranges from 0 to 255. To simulate a blur image, the original images are blurred by a blur kernel and then additive Gaussian noise with standard deviations  $\sigma = \sqrt{2}$ and $\sigma = 2$ are added, respectively. Four blur kernels are used for simulation. The whole experimental settings (degraded scenarios) are summarized in Table 1. The $10$ test images are showed in Figure \ref{fig:testimages}.

\begin{table} \centering { 

\begin{tabular}{|c|c|c|}

\hline 

 Scenario & PSF  &  $\sigma$          \\
 \hline

 1    &    $ 1/(z_1^2+z_2^2), z_1, z_2 = -7,\ldots,7 $  & $\sqrt{2}$     \\
  \hline
  2    &    $ 1/(z_1^2+z_2^2), z_1, z_2 = -7,\ldots,7 $  & 2     \\
   \hline

     \hline
 3   &    uniform 9   &  $\sqrt{2}$        \\
  \hline
  4    &    uniform 9   & 2        \\
  \hline

   \hline
 5    &    fspecial(gaussian,25,1.6)   & $\sqrt{2}$         \\
   \hline
  6    &    fspecial(gaussian,25,1.6)  & 2        \\
   \hline

   \hline
  7    &   fspecial(motion,15,30)    & $\sqrt{2}$        \\
    \hline
  8    &    fspecial(motion,15,30)   & 2        \\

\hline

\end{tabular}}  \\
\caption{ \small Eight typical deblurring experiments with various blur PSFs and noise standard variances. }

\end{table}

\subsection{Evaluation measures}
 The quality of the recovered image is evaluated by the peak signal to noise ratio (PSNR) value defined as:
 \[
\textrm{ PSNR}(u,\bar{u}):= 10\textrm{log}10 \frac{255^2}{\frac{1}{MN}\sum_{i=1}^{M}\sum_{j=1}^{N}(u(i,j)-\bar{u}(i,j))^2}
 \]
 where $M$ and $N$ are the dimensions of the image, and $u(i,j),\bar{u}(i,j)$ are the pixel values of the input evaluated image and original true image at the pixel location $(i,j)$.
 In addition to PSNR, which is used to evaluate the objective image quality, we use another image quality assessment: Structural SIMilarity (SSIM) \cite{Wang2004SSIM}, which aims to be more consistent with human eye perception. The higher SSIM value means better visual quality. We refer the readers to \cite{Wang2004SSIM} for details.

\subsection{Comparison methods}

%
%
%
%
%

The comparison methods include: wavelet frame based SB \cite{Cai2009a} \footnote{http://www.math.ust.hk/\url{~}jfcai/}, MDAL \cite{DongBin2013} \footnote{http://bicmr.pku.edu.cn/\url{~}dongbin/Publications.html}, nonlocal patch based IDD-BM3D \cite{Danielyan2012} \footnote{http://www.cs.tut.fi/\url{~}foi/GCF\url{-}BM3D/index.html\url{#}ref\url{_}software}, CSR \cite{Dong2011} \footnote{http://see.xidian.edu.cn/faculty/wsdong/wsdong\url{_}Publication.htm}, GSR \cite{Zhang2014} \footnote{http://124.207.250.90/staff/zhangjian/\url{#}Publications}. As far as we know, IDD-BM3D, CSR and GSR provide the current state-of-the-art image deblurring results in the literature. All parameters involved in the competing algorithms were optimally assigned or automatically chosen as described in the reference papers.

For the proposed SDSR method, at the first stage, the partial support information  is obtained based on the initial reference image, which is the  result of  various nonlocal patch based methods including IDD-BM3D, CSR and GSR. In a nutshell, our SDSR algorithm can be viewed as the hybrid of wavelet frame based sparse regularization method and the state-of-the-art nonlocal patch based image restoration methods. Therefore, we name them as IDD-BM3D+SDSR, CSR+SDSR, GSR+SDSR, respectively. In addition, in what follows, we also give the ORACLE recovered results of our proposed method, i.e., the support detection is based on the original true image. Clearly, we usually do not know the original true image in practice. Here, we just use it as an ideal golden upper bound of the performance for our proposed method. It aims to  demonstrate the advantage by exploiting the detected support knowledge of frame coefficients and show  the probably largest room of  further improvement.

\subsection{Implementation details}
The linear B-spline framelet and  two decomposition levels are adopted  for the wavelet frames used in  Algorithm 1, i.e., $L=1$ and $L=4$, respectively. For all the cases, we fix the parameter $\mu=0.01$, $\gamma = 0.003$. The parameters $\rho$ and $\lambda$ control the overall performance.
Specifically, the setting of regularization parameter $\lambda$ in (\ref{eq:new truncatedl0l2}) is the same as that in literature \cite{DongBin2013}, in order  for the optimal PSNR and SSIM values.

Empirically, the parameter $\rho$ is not very sensitive to the type of images, blurs
and noise levels. In our tests, we have found that $\rho=200$ for $L=1$, and $\rho=250$ for $L=4$ consistently yield good performance. Optimal adjustments of the parameter $\rho$ may improve the results over what are presented here. However, it will also reduce the practicality of the algorithms since
more parameters need to be adjusted by users. Therefore, we choose to fix this parameter.
The stopping criterion of the inner loop in Algorithm 1 is:
\begin{equation}
\mathrm{min}\left\{\frac{||u^k-u^{k-1}||_2}{||u^k||_2},\frac{||Au^k-f||_2}{||f||_2}\right\}<5\times10^{-4}
\end{equation}
Empirically, we have found that our algorithm  performs well even when the outer loop only executes one iteration (larger $S$ values do not always lead to significantly noticeable
PSNR and SSIM improvement). Hence, we  set  $S=2$  and  save much computational complexity of the proposed algorithm.
All the experiments were performed under Windows 7 and MATLAB v7.10.0 (R2010a) running on a desktop with an Intel(R) Core(TM) i7-4790 CPU (3.60GHz) and 32GB of memory.

\subsection{Results and discussions}

Table \ref{Table: all PSNR SSIM comparison} is the PSNR and SSIM results of the 10 test photographic images on 8 degraded scenarios. Our method is compared with IDD-BM3D since the initial support estimation is performed  on the recovered results of IDD-BM3D method.  We observe that the proposed IDD-BM3D+SDSR method have overall significant improvements compared to the IDD-BM3D method in terms of both PSNR and SSIM values. In average, IDD-BM3D+SDSR(L=1) and IDD-BM3D+SDSR(L=4) outperform IDD-BM3D by (0.41 dB, 0.0114) and (0.48 dB, 0.0140), respectively. In addition, we emphasize that the proposed Algorithm with $L=4$ slightly better than $L=1$ in most cases, but the corresponding computational time is also longer.


\begin{table*} { \tiny
\centering
\begin{tabular}{|c|c|c|c|c|c|c|c|c|c|c|c|c|}
\hline 
\textbf{Scenario} & \textbf{Method} & \textit{C.man} & \textit{Boat} & \textit{Man} & \textit{Monarch} & \textit{Peppers} & \textit{Lena} & \textit{Barbara} & \textit{Parrots} & \textit{Starfish} & \textit{Goldhill} & \textbf{Average} \\
 \hline
  \multirow{7}{*}{1} & SB \cite{Cai2009a} & 29.55/0.8830 & 29.87/0.8716	& 29.07/0.8688  & 31.76/0.9408  &  30.58/0.8845 & 31.98/0.9133 & 28.72/0.8568	& 32.47/0.9201  & 30.65/0.8991 & 28.84/0.8062 & 30.35/0.8844 \\
    \cline{2-13}
    & MDAL \cite{DongBin2013} & 30.15/0.8927 & 30.06/0.8767	&  29.03/0.8716 &  31.28/0.9431  &  31.43/0.8860 & 32.02/0.9184 & 29.10/0.8765	&  32.71/0.9252 &  30.99/0.9074 & 29.10/0.8124  & 30.59/0.8910  \\
  \cline{2-13}
    & IDD-BM3D \cite{Danielyan2012}& 31.08/0.8916 & 30.96/0.8911	& 29.65/0.8810 & 32.47/0.9452  & 31.98/0.8893   & 33.27/0.9241  & 32.77/0.9296	& 33.90/0.9233  & 31.95/0.9129  &  29.51/0.8223 & 31.75/0.9010  \\
   \cline{2-13}
    & \textbf{IDD-BM3D+SDSR (L=1)} & 31.51/0.9033& 31.67/0.9058  & 30.28/0.8945  & 32.97/0.9554 & 32.39/0.8975 & 33.61/0.9339& 32.36/0.9328 & 34.21/0.9328  & 32.61/0.9248  & 29.99/0.8389  &  32.16/0.9120 \\
  \cline{2-13}
      & \textbf{IDD-BM3D+SDSR (L=4)} & \textbf{31.71}/\textbf{0.9057}& \textbf{31.82}/\textbf{0.9095}  & \textbf{30.38}/\textbf{0.8967}  & \textbf{33.08}/\textbf{0.9573} & \textbf{32.46}/\textbf{0.8982} & \textbf{33.76}/\textbf{0.9368}& \textbf{32.82}/\textbf{0.9361} & \textbf{34.44}/\textbf{0.9353}  & \textbf{32.74}/\textbf{0.9269}  & \textbf{30.08}/\textbf{0.8416}  &  \textbf{32.33/0.9144}  \\
  \cline{2-13}
    & \textbf{ ORACLE (L=1)} &36.27/0.9604  & 36.28/0.9695	& 34.90/0.9682  & 37.10/0.9778  & 36.65/0.9548  &37.35/0.9713  & 34.10/0.9654	&37.62/0.9641 &37.04/0.9728 &35.61/0.9572 & 36.29/0.9662 \\
  \cline{2-13}
      & \textbf{ ORACLE (L=4)} &36.74/0.9639  & 36.93/0.9750	& 35.85/0.9746  & 37.41/0.9820  & 37.36/0.9597  &37.70/0.9762  & 34.83/0.9716	&37.89/0.9702 &37.74/0.9784 &36.54/0.9642 & 36.90/0.9716 \\
  \cline{2-13}

\hline
\hline

  \multirow{7}{*}{2} & SB \cite{Cai2009a}& 28.47/0.8654 & 28.80/0.8421	&  28.08/0.8398 & 30.62/0.9269  & 29.71/0.8694 & 30.98/0.8952 & 27.35/0.8142 & 31.40/0.9090  & 29.59/0.8749  & 28.10/0.7739  &  29.31/0.8611      \\
  \cline{2-13}
    & MDAL \cite{DongBin2013}& 29.10/0.8742 & 29.06/0.8530	& 28.04/0.8407  & 30.19/0.9295  & 30.67/0.8736  & 31.04/0.9003 & 27.82/0.8395	& 31.64/0.9142  & 29.99/0.8820 & 28.31/0.7823 & 29.59/0.8689   \\
  \cline{2-13}
    & IDD-BM3D \cite{Danielyan2012}&  30.01/0.8760 & 29.79/0.8664& 28.56/0.8556  & 31.23/0.9329  & 31.08/0.8745  & 32.13/0.9083 & \textbf{31.31}/0.9087	& 32.47/0.9113  & 30.77/0.8935  & 28.77/0.7947 &  30.61/0.8822      \\
   \cline{2-13}
    &\textbf{IDD-BM3D+SDSR (L=1)}& 30.42/0.8879 & 30.26/0.8810& 28.96/0.8686  & 31.67/0.9441  & 31.49/0.8842  &32.35/0.9176  & 30.63/0.9108 & 32.69/0.9206  & 31.29/0.9059  & 29.10/0.8085 &   30.89/0.8929       \\
  \cline{2-13}
  &\textbf{IDD-BM3D+SDSR (L=4)}& \textbf{30.56/0.8915} & \textbf{30.31/0.8849}& \textbf{29.02/0.8705}  & \textbf{31.73/0.9479}  & \textbf{31.60/0.8860}  &\textbf{32.46/0.9225}  & 31.24/\textbf{0.9144} & \textbf{32.92/0.9240}  & \textbf{31.37/0.9074}  & \textbf{29.21/0.8121} &   \textbf{31.04/0.8961}       \\
  \cline{2-13}
    &\textbf{ ORACLE (L=1)} &34.25/0.9500  & 34.26/0.9576	&32.70/0.9547   &35.05/0.9705   &34.79/0.9427   &35.14/0.9605  & 31.53/0.9489	&35.51/0.9545   &35.05/0.9620 &33.75/0.9425  &   34.20/0.9544      \\
  \cline{2-13}
  &\textbf{ ORACLE (L=4)} &34.95/0.9583  & 35.13/0.9681	& 33.89/0.9672   &35.55/0.9781   &35.73/0.9534   &35.68/0.9699  & 32.56/0.9610	&35.83/0.9657   &35.95/0.9721 &35.01/0.9559  & 35.03/0.9650 \\
  \cline{2-13}

\hline
\hline

  \multirow{7}{*}{3} & SB \cite{Cai2009a}& 26.74/0.8335 &	26.95/0.7854& 25.90/0.7556  & 27.47/0.8752   & 28.57/0.8274 & 28.59/0.8438 & 26.35/0.7678	& 27.39/0.8708  &  27.09/0.7934 & 27.47/0.7357 &  27.25/0.8089     \\
  \cline{2-13}
    & MDAL \cite{DongBin2013}& 27.64/0.8545 & 27.56/0.8177	&  26.15/0.7690 & 27.95/0.8921  & 29.17/0.8395 & 28.93/0.8589 & 26.59/0.7842	& 28.64/0.8868  & 27.81/0.8232  & 27.72/0.7496 &  27.82/0.8276     \\
 \cline{2-13}
    & IDD-BM3D \cite{Danielyan2012}& 28.54/0.8586 & 28.06/0.8219& 26.55/0.7799  & 29.04/0.9034  & 29.62/0.8427  & 29.71/0.8658 & 27.99/0.8227& 29.98/0.8914  & 28.35/0.8321 &27.92/0.7526 &  28.58/0.8371  \\
  \cline{2-13}
    &\textbf{ IDD-BM3D+SDSR (L=1)} & 29.04/0.8726 &28.54/0.8402 & \textbf{26.98/0.7998}  & 29.59/0.9138  & 30.02/0.8544 & 30.05/0.8752  & 28.10/0.8291	& 30.24/0.8982  & \textbf{28.84/0.8492}  & 28.30/0.7712 &  28.97/0.8502       \\
  \cline{2-13}
  &\textbf{IDD-BM3D+SDSR (L=4)}& \textbf{29.07/0.8744} & \textbf{28.58/0.8410}& 26.97/0.7993  & \textbf{29.63/0.9159}  & \textbf{30.06/0.8553}  &\textbf{30.10/0.8771}  & \textbf{28.17}/\textbf{0.8316} & \textbf{30.36/0.9001}  & 28.81/0.8484  & \textbf{28.32/0.7722} &   \textbf{ 29.00/0.8513}     \\
  \cline{2-13}
    &\textbf{ ORACLE (L=1)}& 35.02/0.9551 & 35.05/0.9578 	& 33.62/0.9552  & 36.07/0.9733  & 35.69/0.9440  & 35.77/0.9597 & 31.93/0.9414	& 35.03/0.9563  & 35.67/0.9619  & 34.59/0.9410 &   34.84/0.9456    \\
  \cline{2-13}
  &\textbf{ ORACLE (L=4)}& 36.33/0.9575 & 36.86/0.9671 	& 36.25/0.9699  & 37.80/0.9811  & 37.42/0.9523  & 37.90/0.9711 & 34.92/0.9648	& 37.07/0.9640  & 37.70/0.9729  & 36.18/0.9729 &  36.84/0.9674     \\
  \cline{2-13}

\hline
\hline

  \multirow{7}{*}{4} & SB \cite{Cai2009a}& 26.09/0.8152 & 26.37/0.7585	& 25.37/0.7296  & 26.86/0.8589  & 28.01/0.8121 & 28.09/0.8278 & 25.69/0.7395	& 26.83/0.8601  &  26.51/0.7692 & 26.92/0.7048 &   26.67/0.7876       \\
  \cline{2-13}
    & MDAL \cite{DongBin2013}& 27.01/0.8360 & 26.76/0.7826	& 25.51/0.7436  &  27.10/0.8738 & 28.51/0.8254 & 28.31/0.8429 & 25.86/0.7559	& 27.87/0.8751  & 27.01/0.7951  & 27.18/0.7215 &  27.11/0.8052     \\
  \cline{2-13}
    & IDD-BM3D \cite{Danielyan2012}& 27.69/0.8393 & 27.27/0.7935& 25.94/0.7540  & 28.24/0.8875  & 28.97/0.8263  & 29.03/0.8473 & 27.25/0.7949	&29.20/0.8784 &27.60/0.8066& 27.39/0.7262 &   27.86/0.8154  \\
  \cline{2-13}
    &\textbf{ IDD-BM3D+SDSR (L=1)} &28.13/0.8533 & 27.70/0.8124 & \textbf{26.32/0.7740}  & 28.66/0.8987   & 29.31/0.8398  &29.31/0.8587  & 27.20/0.8012	& 29.47/0.8875  & \textbf{28.00/0.8239}  & 27.70/0.7428 &   28.18/0.8192         \\
  \cline{2-13}
      &\textbf{ IDD-BM3D+SDSR (L=4)} &\textbf{28.20/0.8553} & \textbf{27.76/0.8142}& 26.31/0.7729  & \textbf{28.68/0.9014}   & \textbf{29.33/0.8407}  &\textbf{29.33/0.8608}  & \textbf{27.34/0.8046}	& \textbf{29.58/0.8902}  & 27.99/0.8236  & \textbf{27.74/0.7441} &   \textbf{28.23/0.8308 }      \\
  \cline{2-13}
    &\textbf{ ORACLE (L=1)}  & 33.40/0.9477 & 33.36/0.9467	&31.79/0.9428   &34.28/0.9671   &34.18/0.9351   &34.10/0.9509  & 30.08/0.9268	&33.61/0.9496   &34.14/0.9523   & 33.13/0.9281  &   33.21/0.9447     \\
  \cline{2-13}
   &\textbf{ ORACLE (L=4)}  & 35.07/0.9536 & 35.43/0.9615	&34.63/0.9647   &36.24/0.9784   &36.18/0.9484   &36.31/0.9669  & 33.53/0.9592	&35.93/0.9617   &36.31/0.9683   & 33.13/0.9281  &   35.28/0.9591    \\
  \cline{2-13}

\hline
\hline

  \multirow{7}{*}{5} & SB \cite{Cai2009a}& 27.00/0.8601 &	28.01/0.8396 & 27.46/0.8376  & 30.35/0.9335  & 29.14/0.8781 & 30.71/0.9025 & 25.12/0.7640	&  30.31/0.9163 & 29.08/0.8768  & 27.66/0.7699 &  28.48/0.8578        \\
 \cline{2-13}
    & MDAL \cite{DongBin2013}& 27.34/0.8687 & 28.09/0.8487	& 27.14/0.8349  & 29.39/0.9326  & 29.18/0.8775 & 30.21/0.9038  & 25.61/0.7752	&  29.96/0.9180  & 29.31/0.8923  & 27.20/0.7690 &    28.34/0.8621     \\
  \cline{2-13}
    & IDD-BM3D \cite{Danielyan2012}& 28.10/0.8687 & 28.73/0.8528& 27.83/0.8441  & 30.90/0.9380  &29.97/0.8799&31.41/0.9089  &27.08/0.8205& 31.55/0.9179  &30.36/0.8924&28.18/0.7787    &    29.41/0.8702   \\
 \cline{2-13}
    &\textbf{ IDD-BM3D+SDSR (L=1)}&28.36/0.8781 &29.08/0.8641& 28.12/0.8543&31.34/0.9442& 30.24/0.8862& 31.59/0.9149 & 27.22/0.8245 &31.85/0.9244 &  30.83/0.9019 & 28.31/0.7876  & 29.68/0.8780  \\
  \cline{2-13}
      &\textbf{ IDD-BM3D+SDSR (L=4)}&\textbf{28.42/0.8796} &\textbf{29.12/0.8661}& \textbf{28.14/0.8551}&\textbf{31.37/0.9466}& \textbf{30.27/0.8867}& \textbf{31.67/0.9168} & \textbf{27.26/0.8272} &\textbf{31.93/0.9267} &  \textbf{30.91/0.9039} &\textbf{ 28.34/0.7882 } &  \textbf{29.74/0.8797}    \\
  \cline{2-13}
    &\textbf{ ORACLE (L=1)} &35.56/0.9616  & 36.67/0.9713	&35.21/0.9706   & 37.89/0.9811  & 37.23/0.9579  &38.08/0.9751  &31.97/0.9433 & 36.86/0.9669  & 38.09/0.9765 &35.58/0.9551 &  36.31/0.9659 \\
  \cline{2-13}
      &\textbf{ ORACLE (L=4)} &37.51/0.9658  & 38.81/0.9781	&38.30/0.9798   & 39.48/0.9850  & 38.83/0.9627  & 40.02/0.9805  &35.89/0.9724 & 38.36/0.9717  & 39.73/0.9824 &37.62/0.9659 &  38.46/0.9744  \\
  \cline{2-13}

\hline
\hline

  \multirow{7}{*}{6} & SB \cite{Cai2009a}& 26.73/0.8496 &	27.59/0.8215 & 27.08/0.8206  & 29.84/0.9245  & 28.85/0.8694 & 30.31/0.8914 & 24.69/0.7480	& 29.94/0.9087  & 28.69/0.8631  &  27.44/0.7533  &   28.12/0.8450        \\
  \cline{2-13}
    & MDAL \cite{DongBin2013}& 27.08/0.8574 & 27.68/0.8306	&  26.83/0.8189 & 29.05/0.9249  & 29.33/0.8708 & 29.94/0.8943 & 24.57/0.7467	& 29.56/0.9099  & 28.68/0.8744  & 27.07/0.7563 &   27.98/0.8484     \\
  \cline{2-13}
    & IDD-BM3D \cite{Danielyan2012}&27.63/0.8609  &28.05/0.8311&27.28/0.8201   & 30.24/0.9338  &29.55/0.8725& 30.84/0.9019 & 26.02/0.7853	&30.99/0.9161   &29.69/0.8757& 27.71/0.7549 &  28.80/0.8552    \\
  \cline{2-13}
    &\textbf{ IDD-BM3D+SDSR (L=1)}& 27.85/0.8646 &28.49/0.8450& 27.63/0.8349  &30.57/0.9361&29.85/0.8780 & 30.94/0.9025  &\textbf{26.32/0.7926}& 31.18/0.9157 & 30.12/0.8857 & 27.88/0.7693   &  29.08/0.8624     \\
 \cline{2-13}
     &\textbf{ IDD-BM3D+SDSR (L=4)}& \textbf{27.96/0.8671} &\textbf{28.53/0.8479}& \textbf{27.68/0.8366}  &\textbf{30.73/0.9381}&\textbf{29.91/0.8787} & \textbf{31.04/0.9055}  &26.24/0.7921& \textbf{31.30/0.9191} & \textbf{30.25/0.8887} & \textbf{27.91/0.7701 }  &  \textbf{29.16/0.8644 }      \\
 \cline{2-13}
    &\textbf{ ORACLE (L=1)} &34.48/0.9558  & 35.32/0.9647	& 33.64/0.9626  & 36.36/0.9767  &35.97/0.9510 &36.45/0.9687  &30.83/0.9341 	& 35.87/0.9614  &36.50/0.9699 & 34.56/0.9477  &   35.00/0.9593    \\
 \cline{2-13}
     &\textbf{ ORACLE (L=4)} &36.80/0.9637  & 37.82/0.9755	& 37.25/0.9771  & 38.21/0.9829  &38.03/0.9600 & 38.80/0.9777  &35.20/0.9697 	& 37.66/0.9696  &38.40/0.9791 & 37.04/0.9633  &   37.52/0.9719      \\
 \cline{2-13}

\hline
\hline

  \multirow{7}{*}{7} & SB \cite{Cai2009a}& 28.76/0.8672 &	28.53/0.8383 & 27.86/0.8231  & 29.37/0.9035  & 29.82/0.8614  & 30.37/0.8837 & 27.25/0.8226	&  31.00/0.9065 &  28.96/0.8614 &  28.21/0.7793   &   29.01/0.8547      \\
  \cline{2-13}
    & MDAL \cite{DongBin2013}& 30.31/0.8886 &	29.58/0.8672 & 28.39/0.8425  & 30.70/0.9286  & 31.29/0.8744 & 30.98/0.9053 & 27.89/0.8521	& 32.34/0.9202  & 30.18/0.8895  &  28.86/0.7994   &  30.05/0.8768    \\
  \cline{2-13}
    & IDD-BM3D \cite{Danielyan2012}&30.97/0.8841  & 30.35/0.8776& 28.85/0.8548  &31.65/0.9287   &31.60/0.8746 & 32.24/0.9062 & 31.76/0.9170 & 32.47/0.9081  & 31.01/0.8950  &29.26/0.8114   &   31.02/0.8858      \\
 \cline{2-13}
    &\textbf{ IDD-BM3D+SDSR (L=1)} & 31.71/0.9059 & \textbf{31.17/0.8964}& \textbf{29.66/0.8779}  & \textbf{32.66/0.9467}  &\textbf{32.41/0.8904} &\textbf{32.91/0.9276}  & 31.42/0.9234 &33.50/0.9278   &\textbf{31.96/0.9153 }&  29.88/0.8336   &  31.73/\textbf{0.9045}      \\
  \cline{2-13}
      &\textbf{ IDD-BM3D+SDSR (L=4)} & \textbf{31.77/0.9063} & 31.15/0.8954 & 29.66/0.8768  & 32.49/0.9459  &32.38/0.8894 & 32.85/0.9269  & \textbf{31.95}/\textbf{0.9255} &\textbf{33.54/0.9284 }  & 31.88/0.9133 & \textbf{29.93/0.8340}   &  \textbf{31.76}/0.9042    \\
  \cline{2-13}
    &\textbf{ ORACLE (L=1)} &36.53/0.9606  &36.25/0.9694 & 35.16/0.9680  & 37.19/0.9773  & 36.64/0.9573  &37.24/0.9713  &34.14/0.9651 	& 37.62/0.9646  & 36.86/0.9725  & 35.78/0.9610  &  36.34/0.9667     \\
  \cline{2-13}
      &\textbf{ ORACLE (L=4)} &37.09/0.9645  &37.09/0.9750 & 36.33/0.9754  & 37.73/0.9817  & 37.55/0.9631  &37.83/0.9767  &35.44/0.9735 	& 38.24/0.9700  & 37.78/0.9782  & 36.91/0.9679  &   37.20/0.9726     \\
  \cline{2-13}

\hline
\hline

  \multirow{7}{*}{8} & SB \cite{Cai2009a}& 27.57/0.8447 &	27.57/0.8042&  26.86/0.7899 & 28.16/0.8857  & 28.77/0.8420  & 29.31/0.8610 & 26.21/0.7853	& 29.87/0.8957  &  27.99/0.8333 & 27.37/0.7425   &  27.97/0.8284      \\
  \cline{2-13}
    & MDAL \cite{DongBin2013}& 29.12/0.8661 & 28.49/0.8368	&  27.27/0.8038 & 29.39/0.9105  & 30.21/0.8548 & 29.92/0.8836 & 26.67/0.8092	&  31.10/0.9069 & 28.90/0.8585  & 28.05/0.7643   &  28.91/0.8495       \\
  \cline{2-13}
    & IDD-BM3D \cite{Danielyan2012}& 29.65/0.8675 & 29.06/0.8510	&27.58/0.8215   & 30.30/0.9144  & 30.47/0.8564  & 31.01/0.8880 & \textbf{30.31}/0.8940	&30.93/0.8974 & 29.67/0.8698  &28.34/0.7784  &   29.73/0.8638        \\
  \cline{2-13}
    &\textbf{ IDD-BM3D+SDSR (L=1)}& 30.37/0.8879  & \textbf{29.81/0.8707}&\textbf{28.27/0.8466}   & \textbf{31.19/0.9322}  & \textbf{31.24/0.8736 } &\textbf{31.51}/0.9084  &29.62/0.8961 & 32.07/0.9151  & \textbf{30.48/0.8906}   &\textbf{28.85/0.7992 } &    30.34/\textbf{0.8820}           \\
  \cline{2-13}
      &\textbf{ IDD-BM3D+SDSR (L=4)}&\textbf{30.39/0.8887 } & 29.68/0.8685 &   28.20/0.8438   & 31.02/0.9318  & 31.19/0.8727  &31.49/\textbf{0.9095}  &30.26/\textbf{0.8982} & \textbf{32.10/0.9168}  & 30.35/0.8881   &28.82/0.7981  &   \textbf{30.35}/0.8816           \\
  \cline{2-13}
    &\textbf{ ORACLE (L=1)} & 34.53/0.9509 & 34.14/0.9569	& 33.05/0.9553  &35.20/0.9701 & 34.71/0.9442  &35.12/0.9608  &31.66/0.9484 	& 35.61/0.9552  & 34.90/0.9620  & 33.86/0.9461 &   34.28/0.9550  \\
 \cline{2-13}
     &\textbf{ ORACLE (L=4)} & 35.22/0.9575 & 35.23/0.9672	& 34.37/0.9676  &35.88/0.9772 & 35.83/0.9547  &35.80/0.9695  &33.24/0.9632 	& 36.31/0.9641  & 35.97/0.9714  & 35.18/0.9575 & 35.30/0.9650    \\
 \cline{2-13}
\hline
\end{tabular}}
\caption{\small The comparison of PSNR (dB) and SSIM results of our proposed IDD-BM3D+SDSR together with different alternative methods. Bold values denote the highest PSNR or SSIM values excluding the ORACLE cases.}
\label{Table: all PSNR SSIM comparison}
\end{table*}

Besides the IDD-BM3D  method, two other nonlocal patch based methods CSR and GSR are also considered. Due to the space limit, in what follows,  we just present the results of scenario 3 and scenario 5 in Table I, since the other cases have the same conclusions. The PSNR and SSIM results of  IDD-BM3D, CSR, GSR and corresponding IDD-BM3D+SDSR, CSR+SDSR and GSR+SDSR  are reported in Table \ref{Table: part PSNR SSIM comparison}.  It can be observed that the final recovered results of the proposed IDD-BM3D+SDSR, CSR+SDSR and GSR+SDSR are slightly different due to the different initial reference images,  but overall comparable to each other. We can also observe that our proposed methods have overall performance improvements compared to the initial recoveries by IDD-BM3D, CSR and GSR, respectively. This observation demonstrates that SDSR is very promising.   It is not surprising that the ORACLE (support detection based on the true image) situation of the proposed method achieves the best recovery performance in all the cases. The above observations demonstrate the power of making use of the support prior of frame coefficients. Ones can achieve significant performance gain as long as the support estimation is reliable.

\begin{table*}{ 
\centering
\begin{tabular}{|c|c|c|c|c|c|c|c|c|c|c|c|c|c}
  \cline{1-13} 
\textbf{Scenario} & \textbf{Method} & \textit{C.man} & \textit{Boat} & \textit{Man} & \textit{Monarch} & \textit{Peppers} & \textit{Lena} & \textit{Barbara} & \textit{Parrots} & \textit{Starfish} & \textit{Goldhill} & \textbf{Average} \\
   \cline{1-13}
  \multirow{14}{*}{3} & \multirow{2}{*}{IDD-BM3D  \cite{Danielyan2012}}& 28.54 & 28.06 & 26.55  & 29.04  & 29.62  & 29.71 & 27.99 & 29.98  & 28.35 & 27.92 &  28.58  \\
  & & 0.8586 & 0.8219& 0.7799 & 0.9034& 0.8427  & 0.8658   & 0.8227  & 0.8914 & 0.8321& 0.7526  & 0.8371  \\
  \cline{2-13}
  & \multirow{2}{*}{CSR \cite{Dong2011}}& 28.53  & 28.40& 26.90  & 29.05  & 29.63  & 29.91 & 27.96& 30.55  & 28.83 & 27.88 & 28.76   \\
  & & 0.8563  & 0.8297& 0.7924 &0.8970 & 0.8403  &0.8655   &0.8214   & 0.8883 &0.8469 & 0.7607  &  0.8399  \\
  \cline{2-13}
  & \multirow{2}{*}{GSR \cite{Zhang2014}}& 28.28 & 28.27 & 26.66  & 28.99  & 29.66  & 30.10 & 28.95  & 30.40  & 28.56 & 27.96   &  28.78    \\
  &  & 0.8538  & 0.8316 & 0.7887  & 0.9074  & 0.8484  & 0.8772  & 0.8488 & 0.8923   & 0.8407  & 0.7602 &  0.8449   \\
  \cline{2-13}
  & \multirow{2}{*}{\textbf{ IDD-BM3D+SDSR (L=1)}}&29.04  & 28.54& 26.98  &  29.59 & 30.02  & 30.05 & 28.10   &  30.24 & 28.84 &  28.30 &  28.97   \\
  & & 0.8726  & 0.8402 & 0.7998 & 0.9138 & 0.8544  & 0.8752   & 0.8291  & 0.8982 &  0.8492&  0.7712 &  0.8502  \\
  \cline{2-13}
    & \multirow{2}{*}{\textbf{ IDD-BM3D+SDSR (L=4)}}& 29.07 & 28.58 & 26.97  & 29.63  & 30.06  & 30.10 & 28.17 & 30.36  & 28.81 & 28.32 &  29.00  \\
  &  & 0.8744 & 0.8410 &  0.7993 & 0.9159   & 0.8553  & 0.8771  &  0.8316 & 0.9001 &  0.8484 & 0.7722 & 0.8513  \\
  \cline{2-13}
  & \multirow{2}{*}{\textbf{ CSR+SDSR (L=1)}}& 29.02 &  28.74 & \textbf{27.17}  & 29.63  &  30.13 & 30.29   &  28.11 & 30.81  & 29.26 & 28.33 &  29.15 \\
  &  & 0.8751 & \textbf{0.8477} & \textbf{0.8067} & 0.9143  & 0.8565  & 0.8802  & 0.8309 & 0.9019 &  0.8611 & 0.7773  & 0.8551   \\
  \cline{2-13}
    & \multirow{2}{*}{\textbf{ CSR+SDSR (L=4)}}& \textbf{29.12} & 28.50 & 27.00  & \textbf{29.65}  &  \textbf{30.16} & 30.35 & 28.17 & \textbf{30.95}  & \textbf{29.28} & \textbf{28.35} &  29.15   \\
  &  & \textbf{0.8769} & 0.8409 & 0.8019 & 0.9161  & 0.8576  & 0.8823   & 0.8331 & \textbf{0.9051} &  \textbf{0.8612} & \textbf{0.7778} &  0.8552 \\
  \cline{2-13}
  & \multirow{2}{*}{\textbf{ GSR+SDSR (L=1)}}& 28.83 & 28.73   &  27.07 & 29.42  & 30.01  & 30.32  &  28.83  & 30.61  & 29.01 & 28.32  &  29.11  \\
  &  & 0.8728   & 0.8464    & 0.8046 & 0.9161  & 0.8569  & 0.8825  & 0.8493 & 0.9015  & 0.8549  & 0.7751 &  0.8558   \\
  \cline{2-13}
    & \multirow{2}{*}{\textbf{ GSR+SDSR (L=4)}}& 28.91 & \textbf{28.75} & 27.07  & 29.49  &  30.07 & \textbf{30.39} & \textbf{28.97} & 30.77  & 29.00 & \textbf{28.35} &  \textbf{29.17}  \\
  &  & 0.8741  & 0.8473  &  0.8042 & \textbf{0.9184}  & \textbf{0.8583}  & \textbf{0.8848}  & \textbf{0.8518} & 0.9040 &  0.8542 & 0.7754 &  \textbf{0.8571}  \\
  \cline{2-13}

\hline
\hline

  \multirow{14}{*}{5} & \multirow{2}{*}{IDD-BM3D  \cite{Danielyan2012}}& 28.10 & 28.73 &  27.83 &  30.90 & 29.97  & 31.41 & 27.08 & 31.55  & 30.36 & 28.18 &  29.41   \\
  &  & 0.8687 & 0.8528 &  0.8441 & 0.9380  & 0.8799  & 0.9089  & 0.8205 & 0.9179 & 0.8924  & 0.7787 & 0.8702  \\
  \cline{2-13}
  & \multirow{2}{*}{CSR \cite{Dong2011}}& 28.27 & 29.07 & 27.98  &  30.36 &  30.17 & 31.23 & 27.80 & 31.76  & 30.97 & 27.97 & 29.56   \\
  &  &  0.8554 & 0.8605  & 0.8501 &  0.9226  & 0.8670  &  0.8970 &  0.8257 & 0.9054 & 0.8980  & 0.7760  & 0.8658  \\
  \cline{2-13}
  & \multirow{2}{*}{GSR \cite{Zhang2014}}& 27.77 & 28.64  & 27.58  &  30.29 & 30.20  & 31.47 & 28.26  & 31.40  & 30.19 & 28.06 &  29.39   \\
  &  & 0.8666  & 0.8557 & 0.8427  & 0.9357  & 0.8793  & 0.9135  & 0.8436 & 0.9179  & 0.8900  & 0.7793 &  0.8724 \\
  \cline{2-13}
  & \multirow{2}{*}{\textbf{ IDD-BM3D+SDSR (L=1)}}&  28.36 & 29.08 & 28.12  & 31.34  & 30.24  & 31.59 &  27.22 & 31.85  & 30.83 & 28.31 &  29.68   \\
  & & 0.8781  & 0.8641 & 0.8543 & 0.9442 &  0.8862 & 0.9149  & 0.8245  & 0.9244 & 0.9019  &  0.7876 &  0.8780  \\
  \cline{2-13}
    & \multirow{2}{*}{\textbf{ IDD-BM3D+SDSR (L=4)}}& 28.42 & 29.12 & 28.14  & \textbf{31.37}  & 30.27  & 31.67 & 27.26 & 31.93  & 30.91 & \textbf{28.34} &  29.74 \\
  &  & 0.8796 & 0.8661 & 0.8551 & \textbf{0.9466}  & 0.8867  &  0.9168 & 0.8272 &  0.9267 &  0.9039 & 0.7882 & 0.8797  \\
  \cline{2-13}
  & \multirow{2}{*}{\textbf{ CSR+SDSR (L=1)}}& 28.60 & \textbf{29.27} &  \textbf{28.15} & 30.92  & 30.73  & 31.57 & 27.92 & 32.18  & 31.56 &  28.25 &  29.92 \\
  &  & 0.8810 & \textbf{0.8697} &  \textbf{0.8567} & 0.9442  &  0.8868 &  0.9157 & 0.8363 & 0.9249 & 0.9126  & 0.7893 &  0.8817 \\
  \cline{2-13}
    & \multirow{2}{*}{\textbf{ CSR+SDSR (L=4)}}& \textbf{28.63} & 29.20 & 28.03  & 31.01  & \textbf{30.75}  & 31.71 & 27.97 & \textbf{32.27}  & \textbf{31.62} & 28.27 & \textbf{29.95}   \\
  &  & \textbf{0.8828} & 0.8696 & 0.8551 &  0.9437 & \textbf{0.8873}  &  0.9179 & 0.8392 &  \textbf{0.9280} & \textbf{0.9140}  & 0.\textbf{7899} & \textbf{0.8827}  \\
  \cline{2-13}
  & \multirow{2}{*}{\textbf{ GSR+SDSR (L=1)}}& 28.08 & 29.00 &  27.92 & 30.67  & 30.54  & 31.60 & 28.23  & 31.68  & 30.77 &  28.23 & 29.67 \\
  &  & 0.8756  & 0.8652 & 0.8527  & 0.9431  & 0.8862  & 0.9180  & 0.8441 & 0.9242  &  0.9019  & 0.7869 & 0.8798  \\
  \cline{2-13}
    & \multirow{2}{*}{\textbf{ GSR+SDSR (L=4)}}& 28.17  & 29.03 & 27.96  & 30.81  & 30.59  & \textbf{31.73} & \textbf{28.30} & 31.79  & 30.85 &  28.26 &   29.75 \\
  &  & 0.8776 & 0.8676 &  0.8544 & 0.9447  &  0.8872 & \textbf{0.9202}  & \textbf{0.8471} &  0.9271 &  0.9038 & 0.7879 &  0.8818 \\
  \cline{2-13}

\end{tabular}}

\caption{\small The comparison of PSNR (dB) and SSIM results by our proposed IDD-BM3D+SDSR, CSR+SDSR, and GSR+SDSR together with different alternative methods. Bold values denote the highest PSNR or SSIM values.}
\label{Table: part PSNR SSIM comparison}
\end{table*}

\begin{figure*}[h]
  \centering
  \includegraphics[width=0.19\textwidth]{cameraman.eps}
   \includegraphics[width=0.19\textwidth]{cameraman_blurred_uniform9sigma1.eps}
   \includegraphics[width=0.19\textwidth]{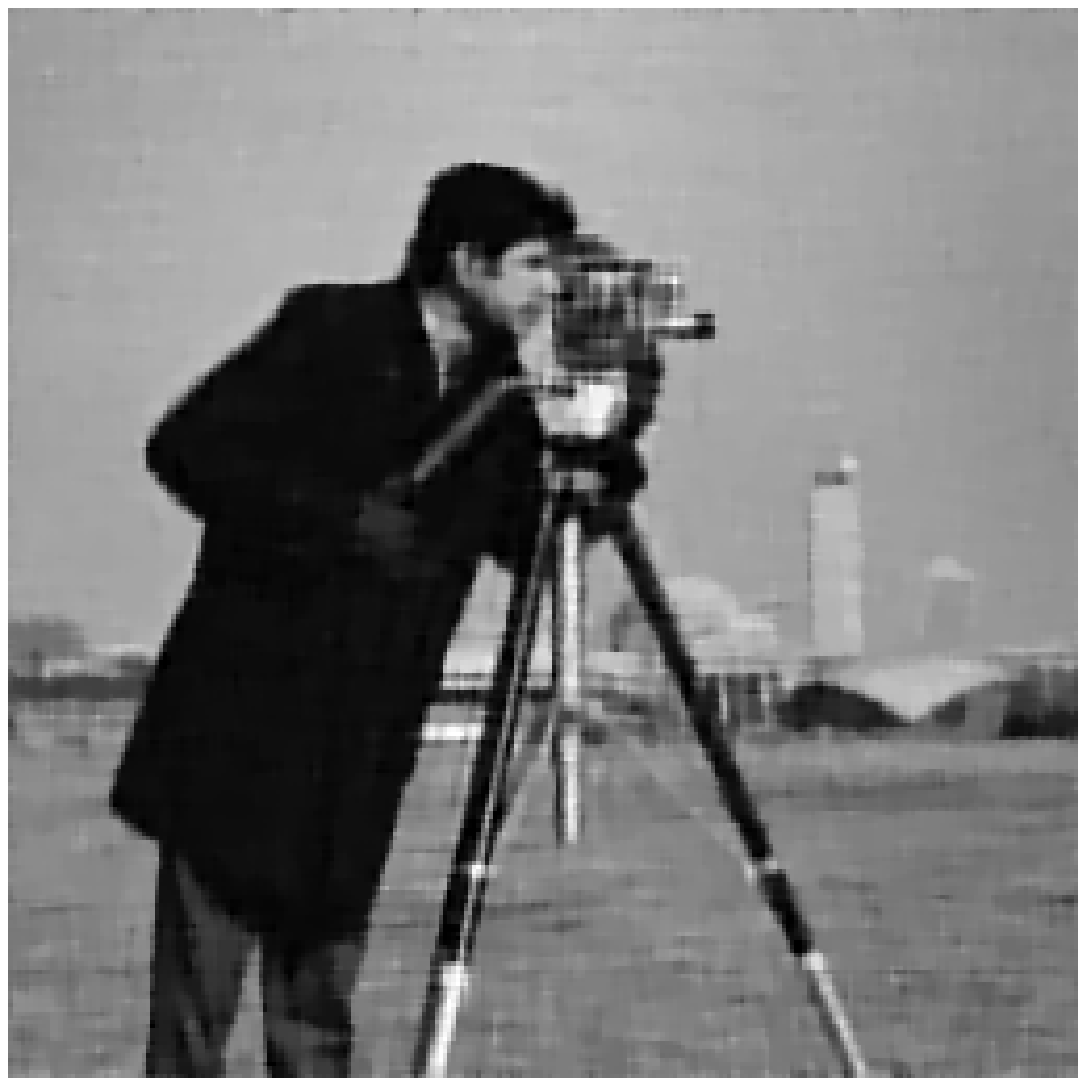}
   \includegraphics[width=0.19\textwidth]{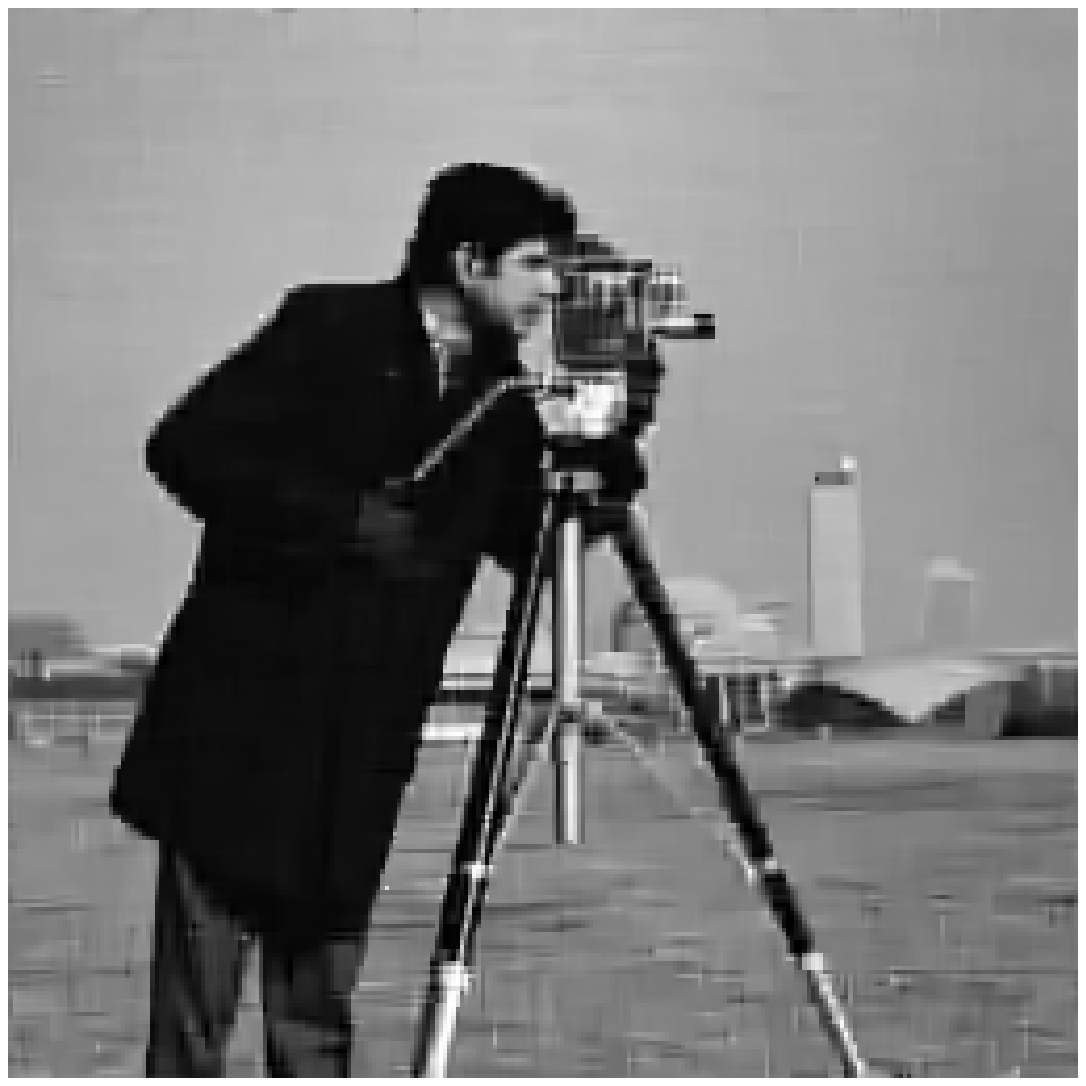}
   \includegraphics[width=0.19\textwidth]{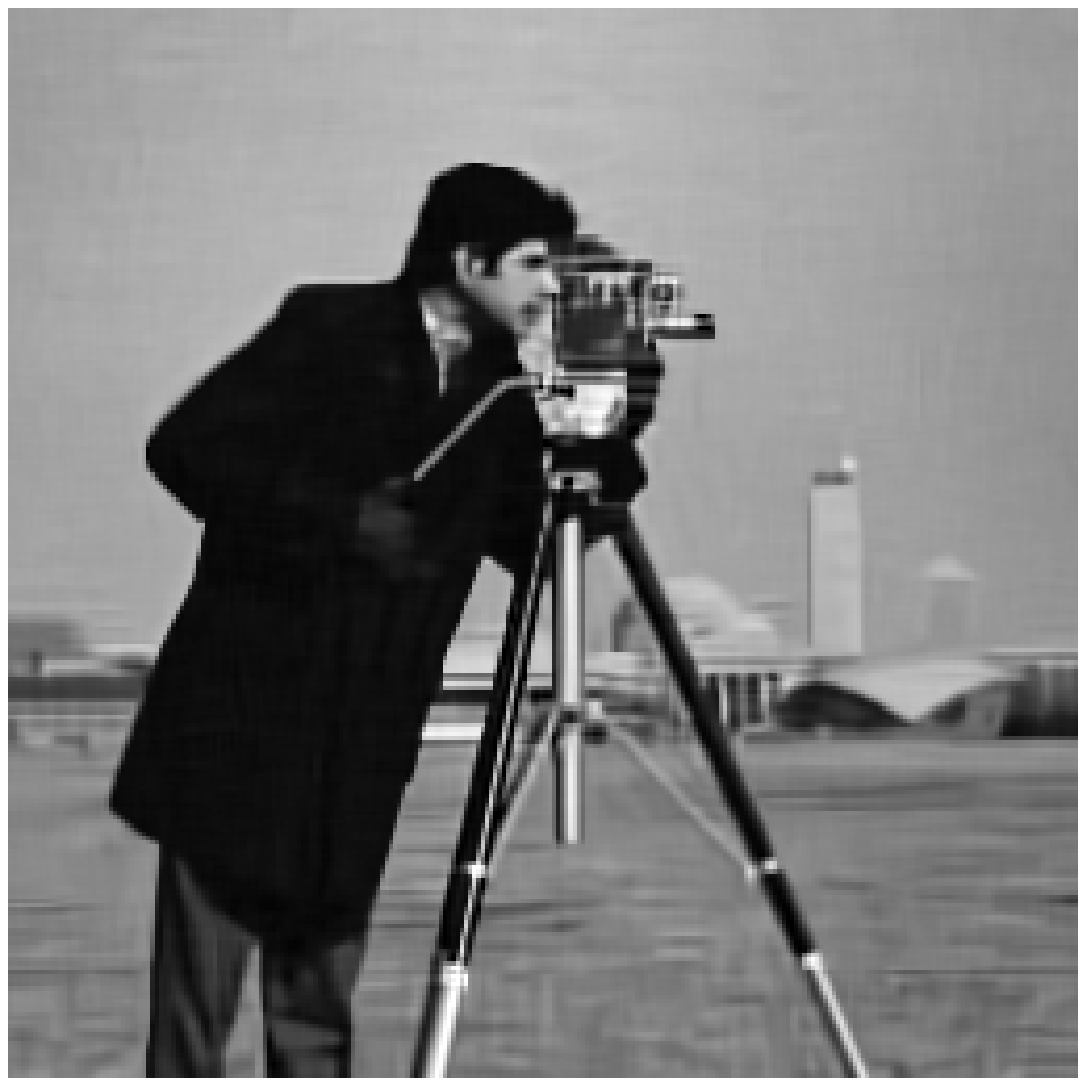} \\
   \includegraphics[width=0.19\textwidth]{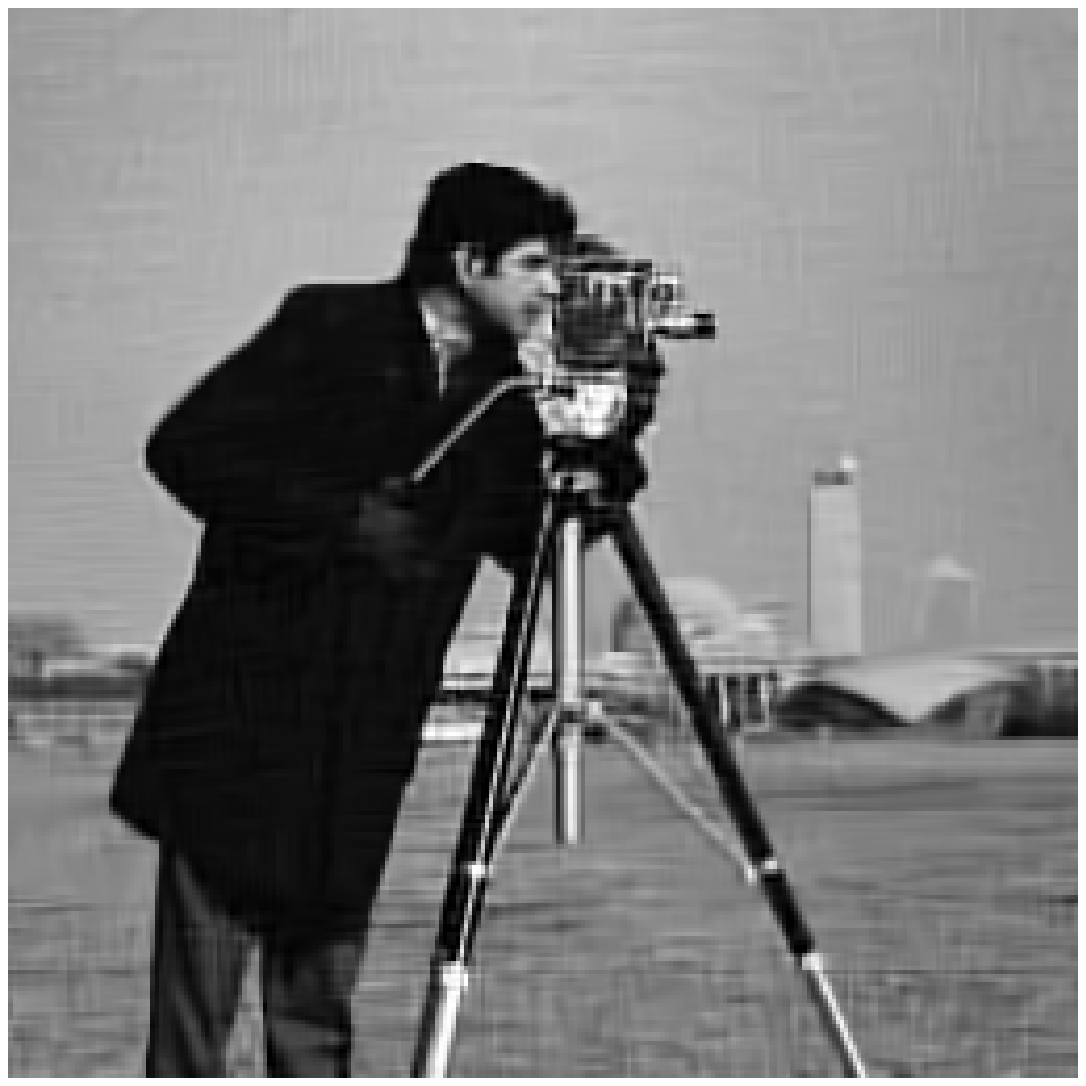}
   \includegraphics[width=0.19\textwidth]{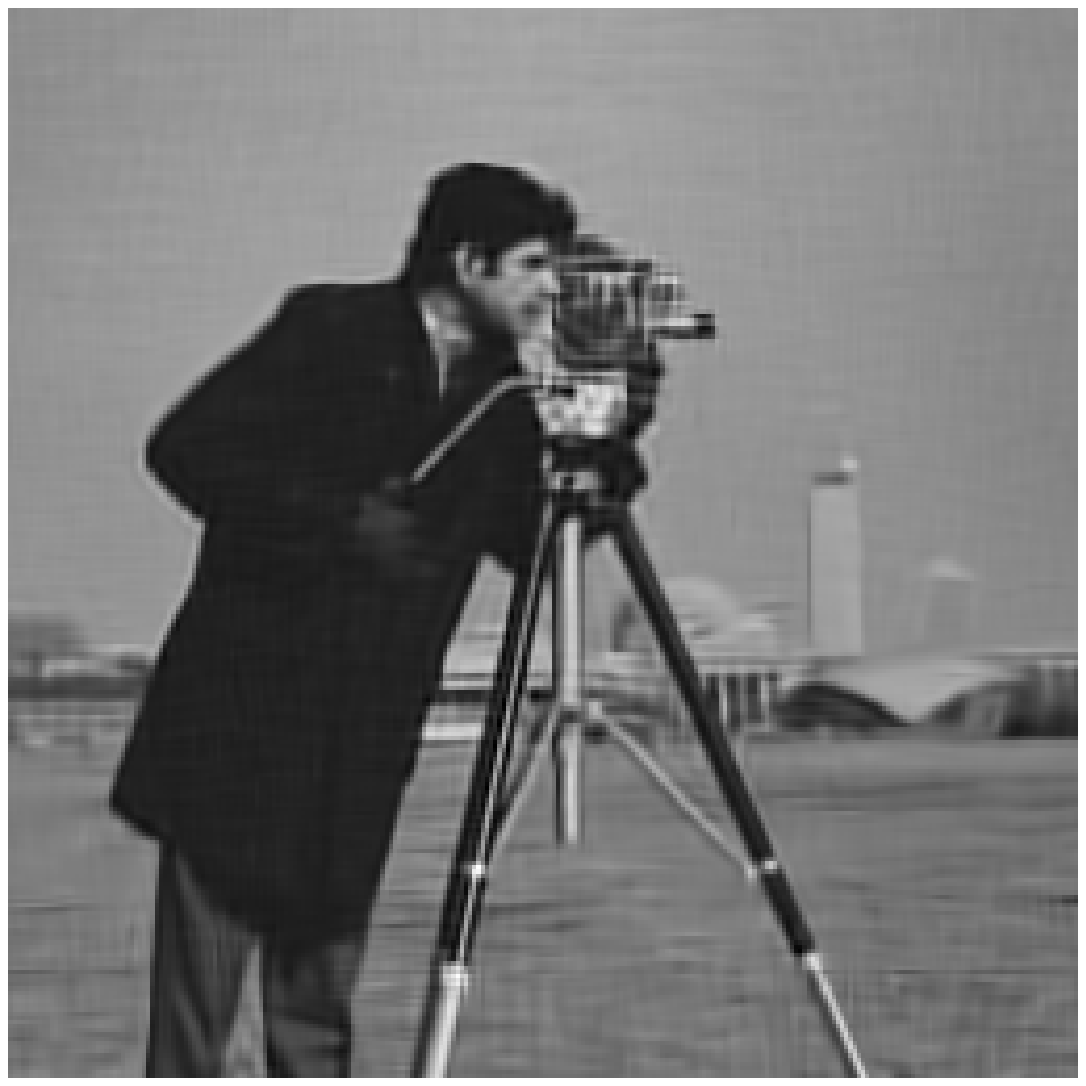}
   \includegraphics[width=0.19\textwidth]{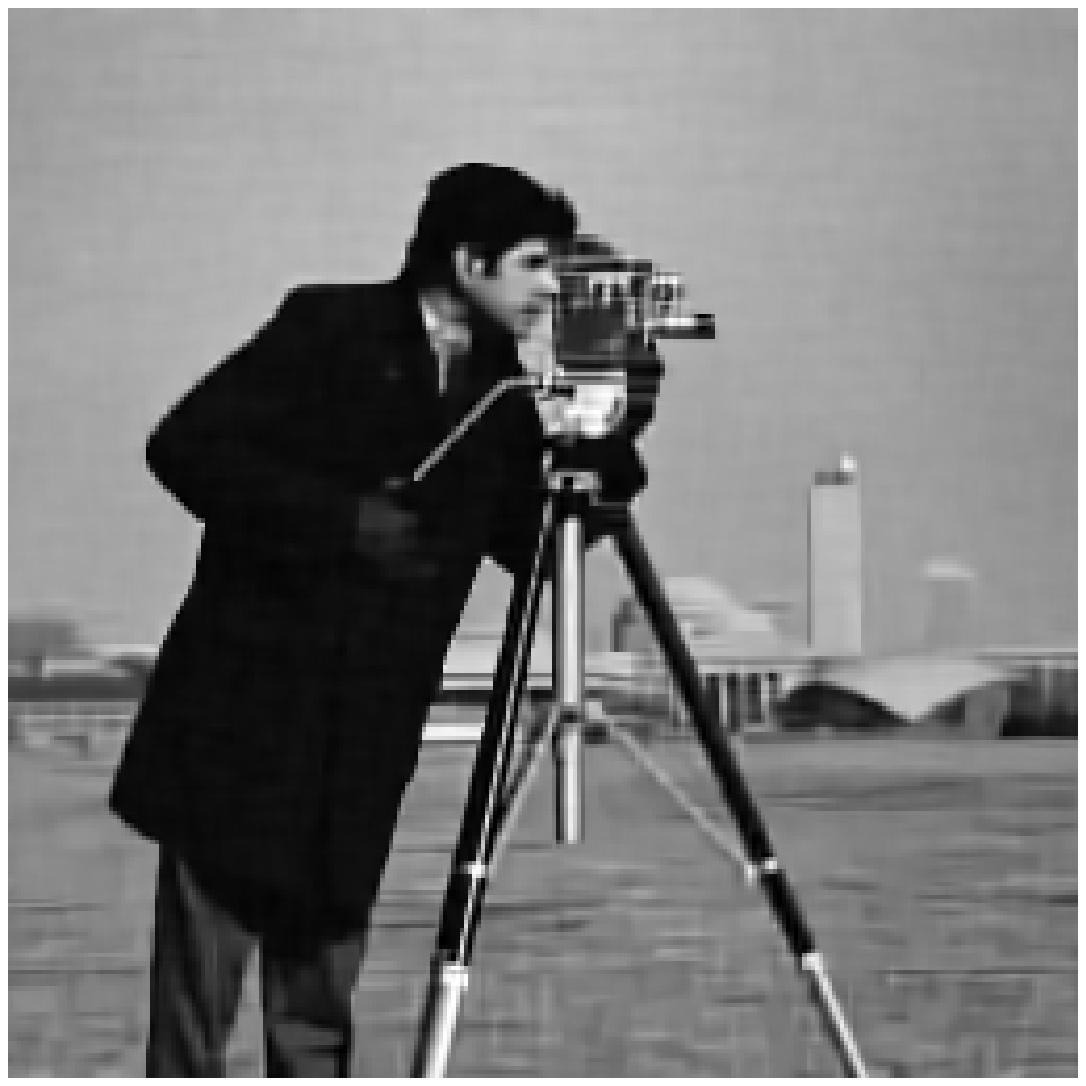}
    \includegraphics[width=0.19\textwidth]{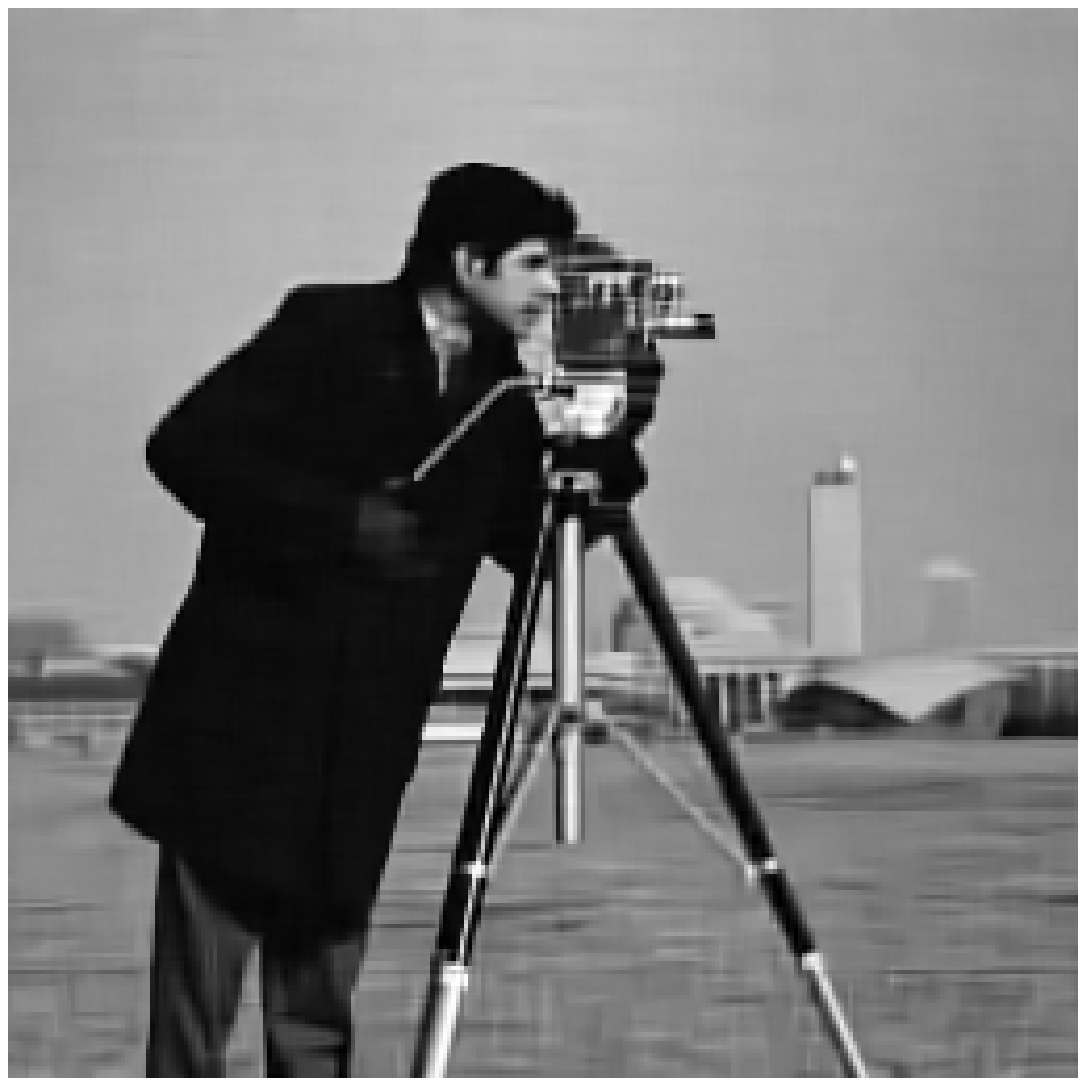}
   \includegraphics[width=0.19\textwidth]{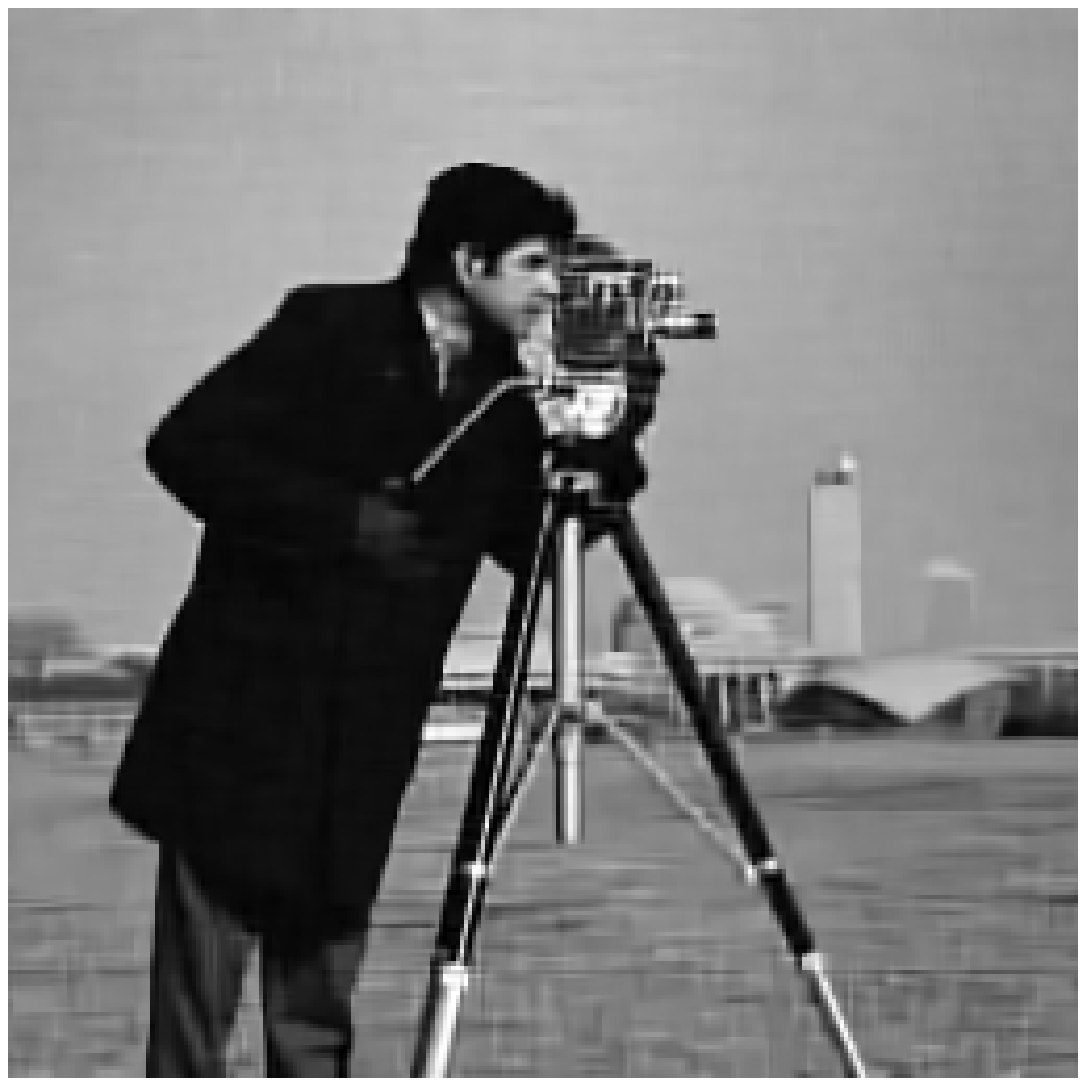} \\
   \includegraphics[width=0.19\textwidth]{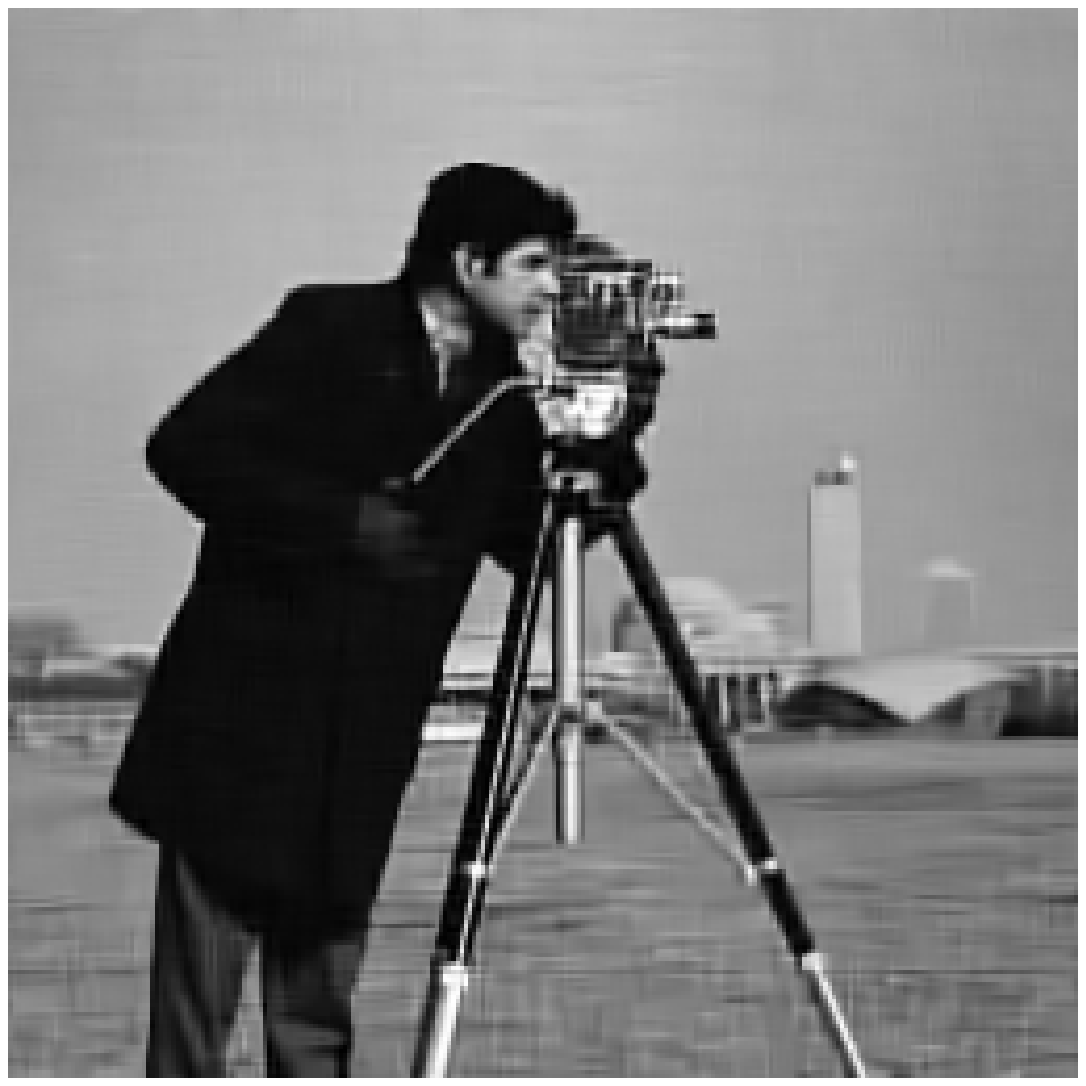}
   \includegraphics[width=0.19\textwidth]{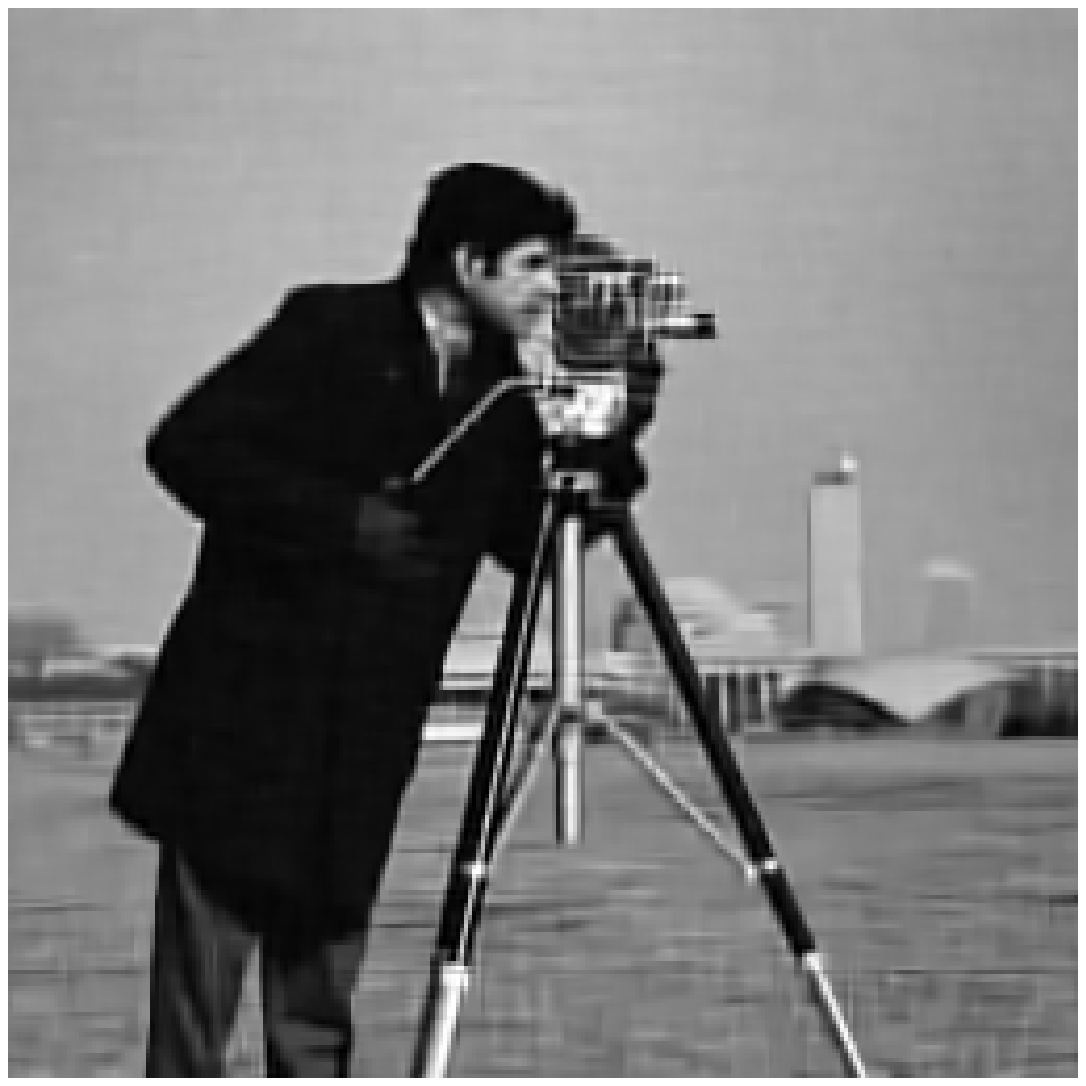}
   \includegraphics[width=0.19\textwidth]{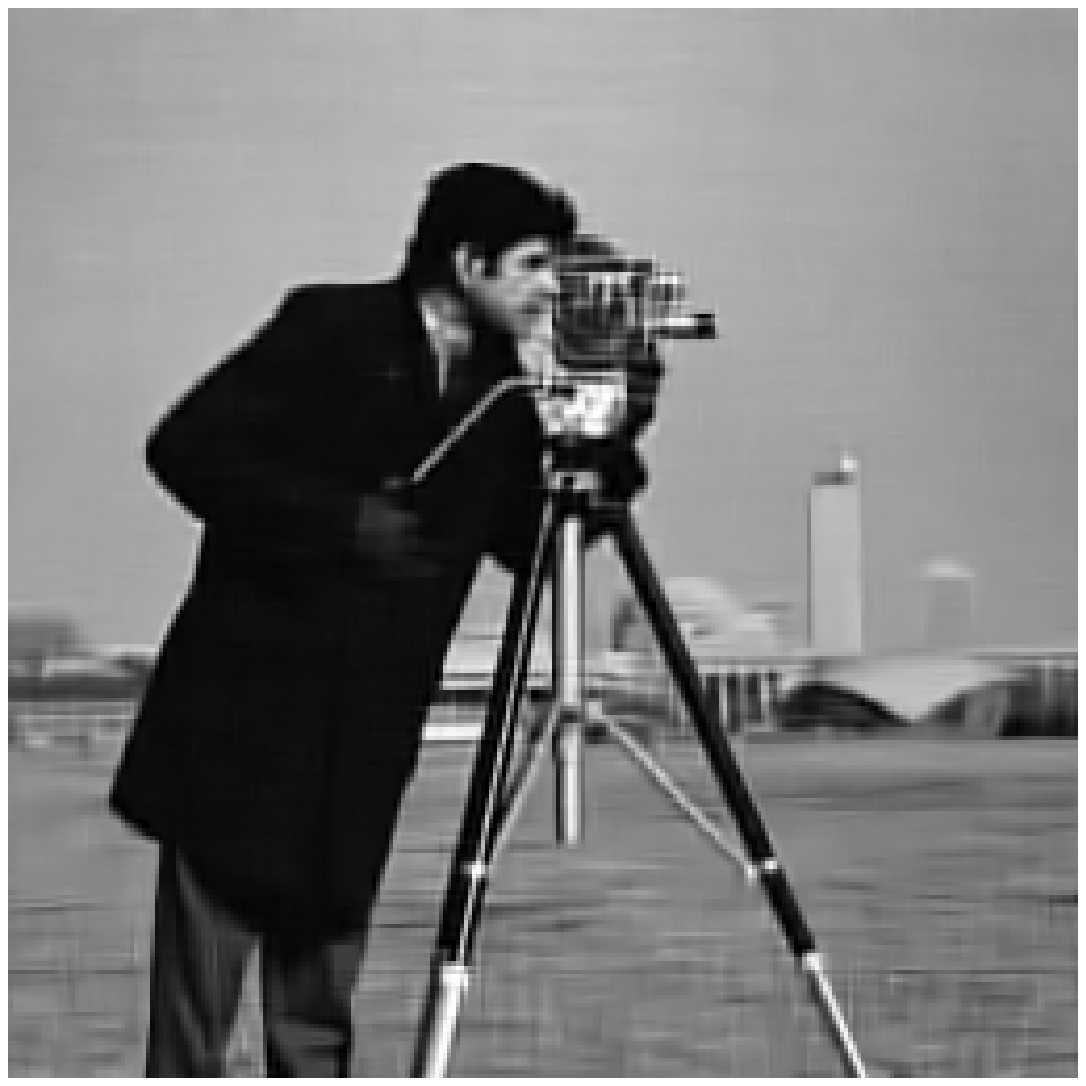}
   \includegraphics[width=0.19\textwidth]{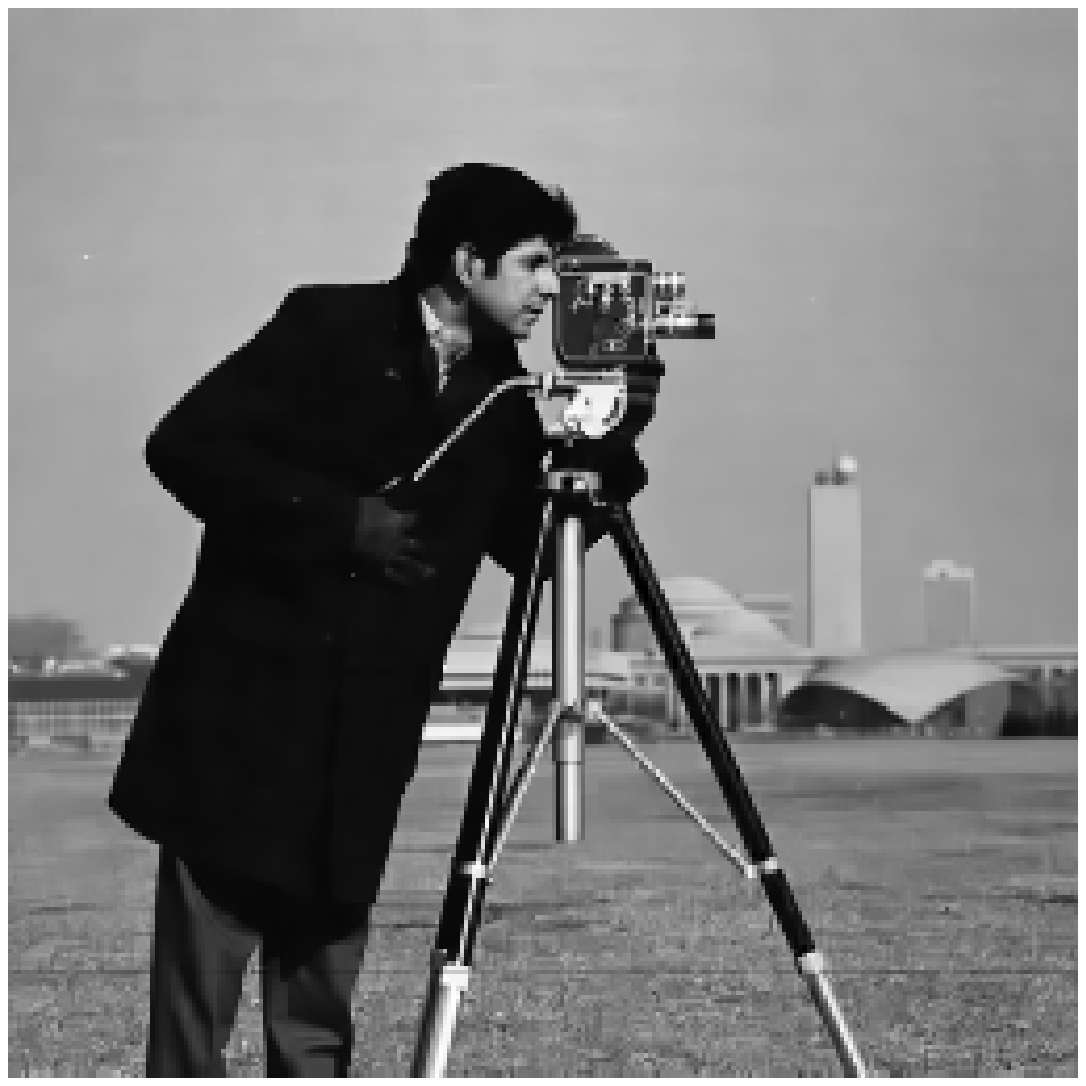}
   \includegraphics[width=0.19\textwidth]{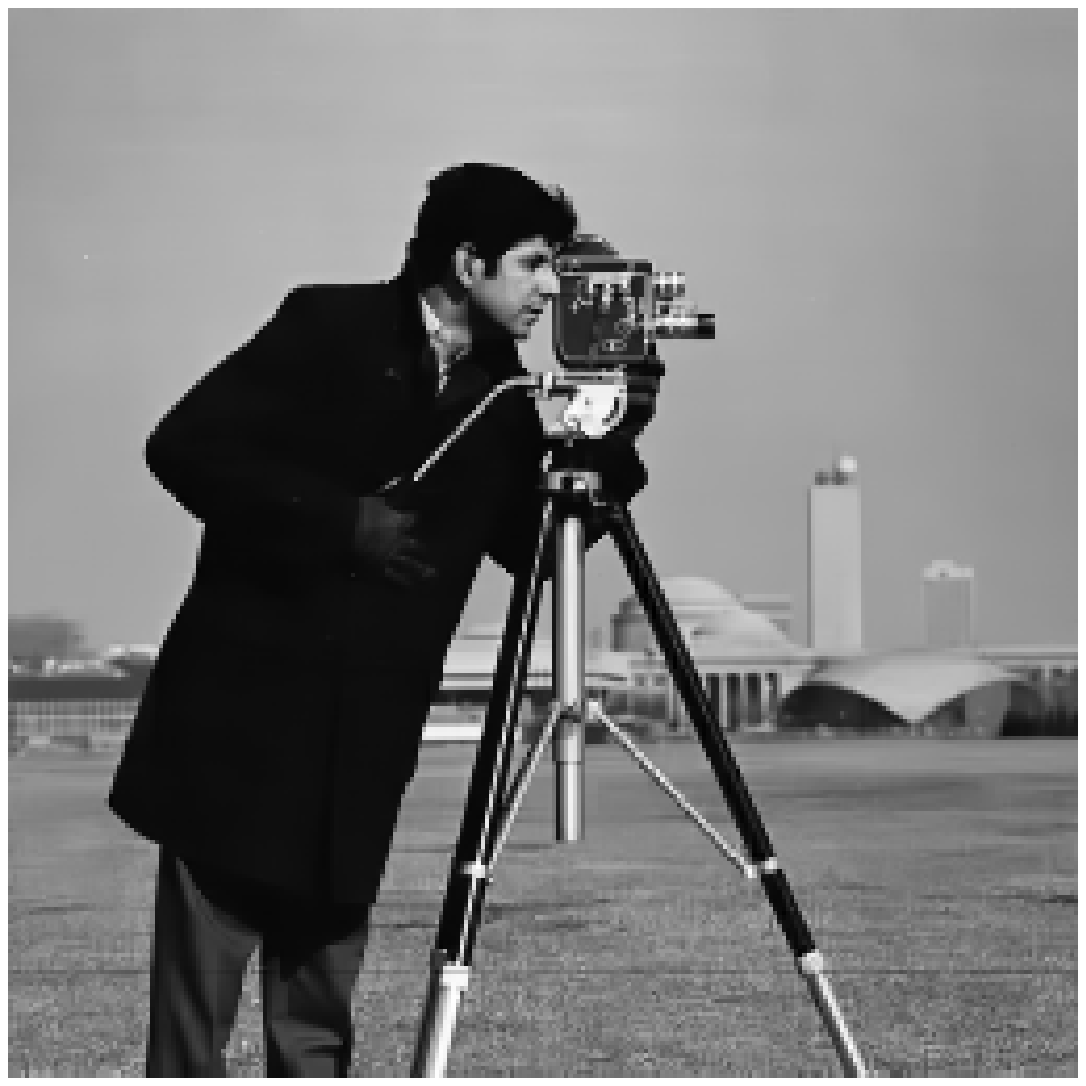} \\
  \caption{\small Visual quality comparison of image deburring results on image \textit{Cameraman} ($256\times 256$). From left to right and top to bottom: original image, degraded image (Scenario 3), the recovered image by SB \cite{Cai2009a} (PSNR=26.74; SSIM=0.8335), MDAL \cite{DongBin2013} (PSNR=27.64; SSIM=0.8545), IDD-BM3D \cite{Danielyan2012} (PSNR=28.54;SSIM=0.8586), CSR \cite{Dong2011} (PSNR=28.53;SSIM=0.8563), GSR \cite{Zhang2014} (PSNR=28.28;SSIM=0.8538), Our proposed IDD-BM3D+SDSR(L=1)  (PSNR=29.04;SSIM=0.8726), IDD-BM3D+SDSR(L=4)  (PSNR=29.07;SSIM=0.8744), CSR+SDSR(L=1) (PSNR=29.02;SSIM=0.8751), CSR+SDSR(L=4) (PSNR=\textbf{29.12};SSIM=\textbf{0.8769}), GSR+SDSR(L=1) (PSNR=28.83;SSIM=0.8728), GSR+SDSR(L=4) (PSNR=28.91;SSIM=0.8741), ORACLE(L=1) (PSNR=35.02;SSIM=0.9551), ORACLE(L=4) (PSNR=36.33;SSIM=0.9575). Bold values denote the highest PSNR or SSIM values excluding the ORACLE cases.}
\label{fig:cameraman comparison}
\end{figure*}

\begin{figure*}[h]
  \centering
  \includegraphics[width=0.19\textwidth]{parrots.eps}
   \includegraphics[width=0.19\textwidth]{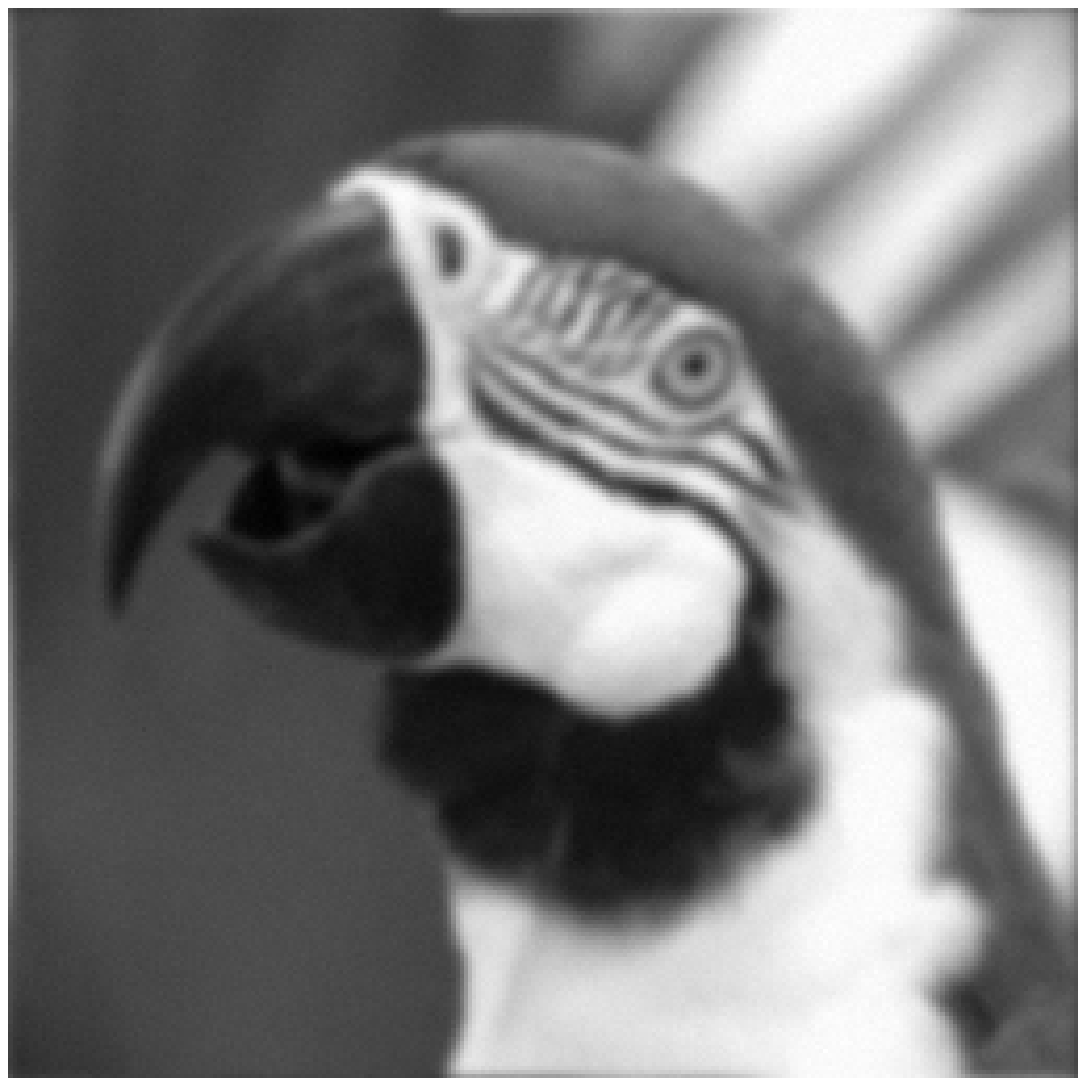}
   \includegraphics[width=0.19\textwidth]{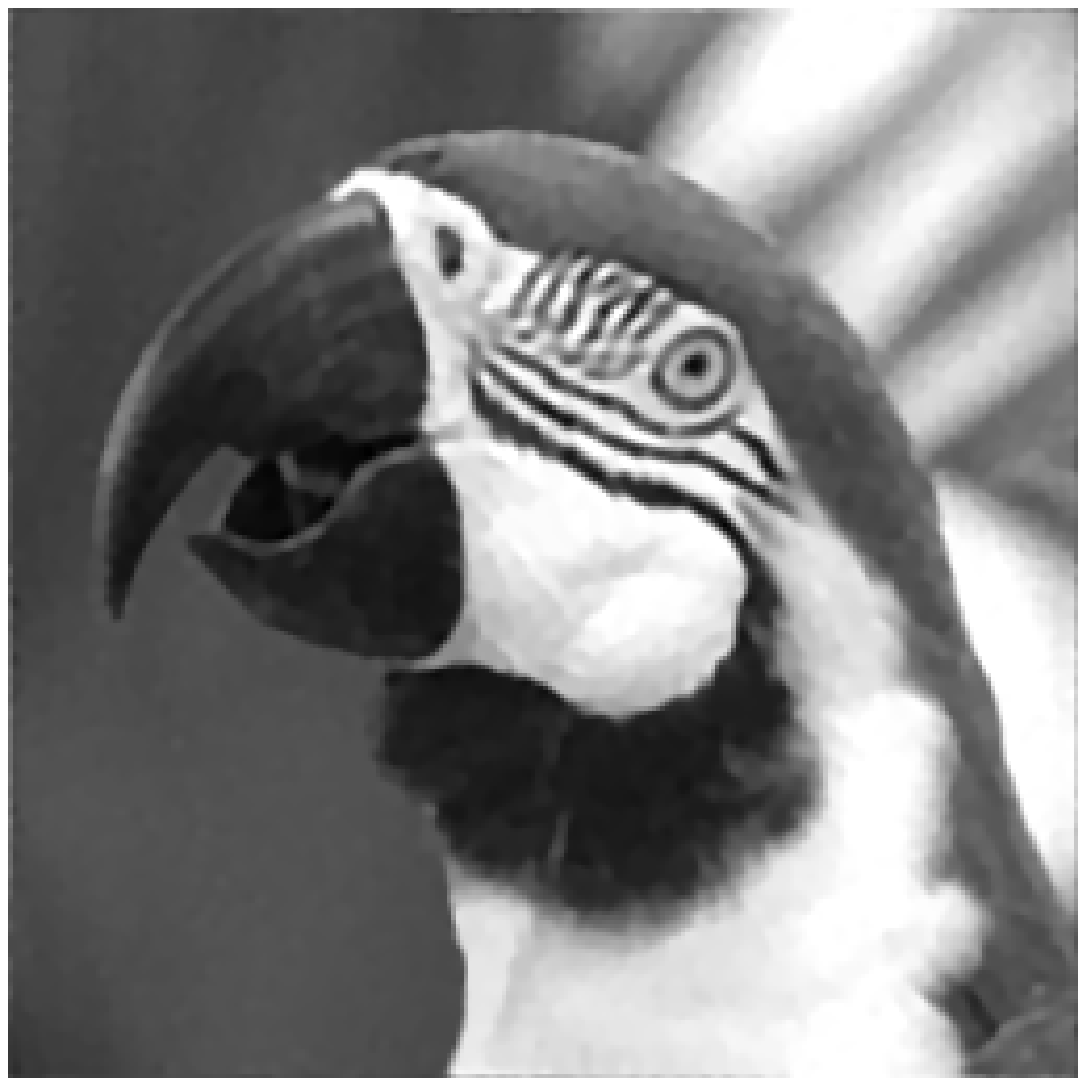}
   \includegraphics[width=0.19\textwidth]{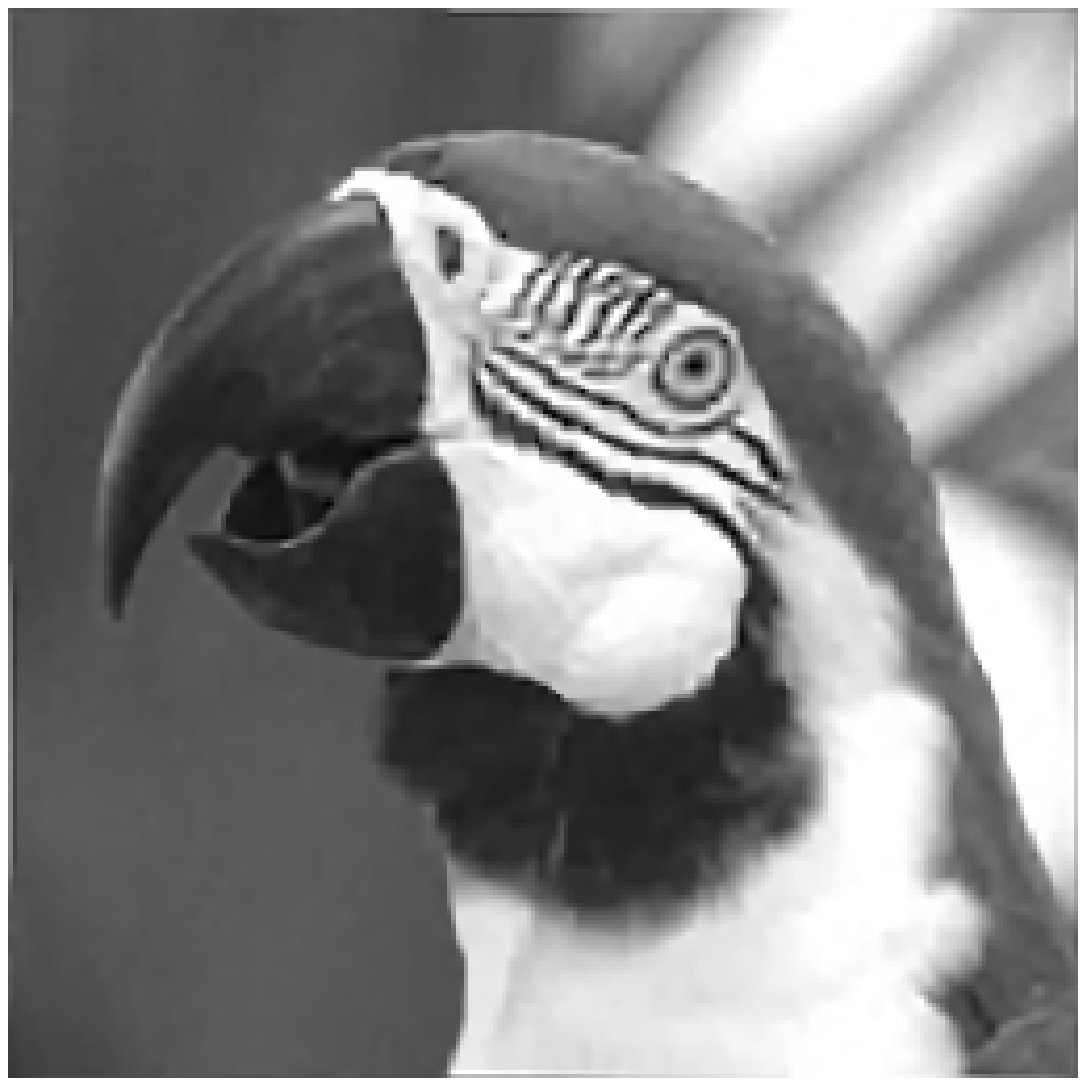}
   \includegraphics[width=0.19\textwidth]{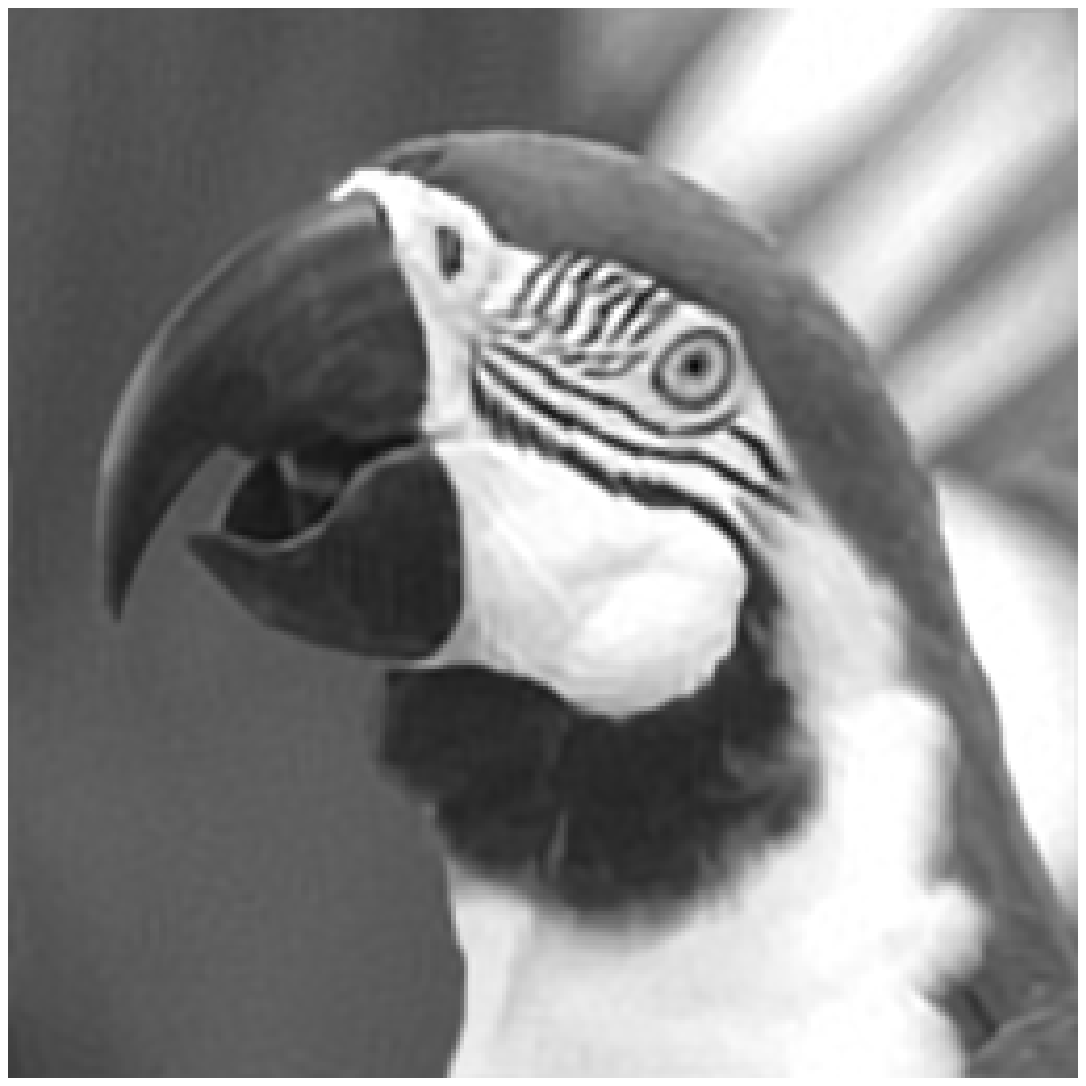} \\
   \includegraphics[width=0.19\textwidth]{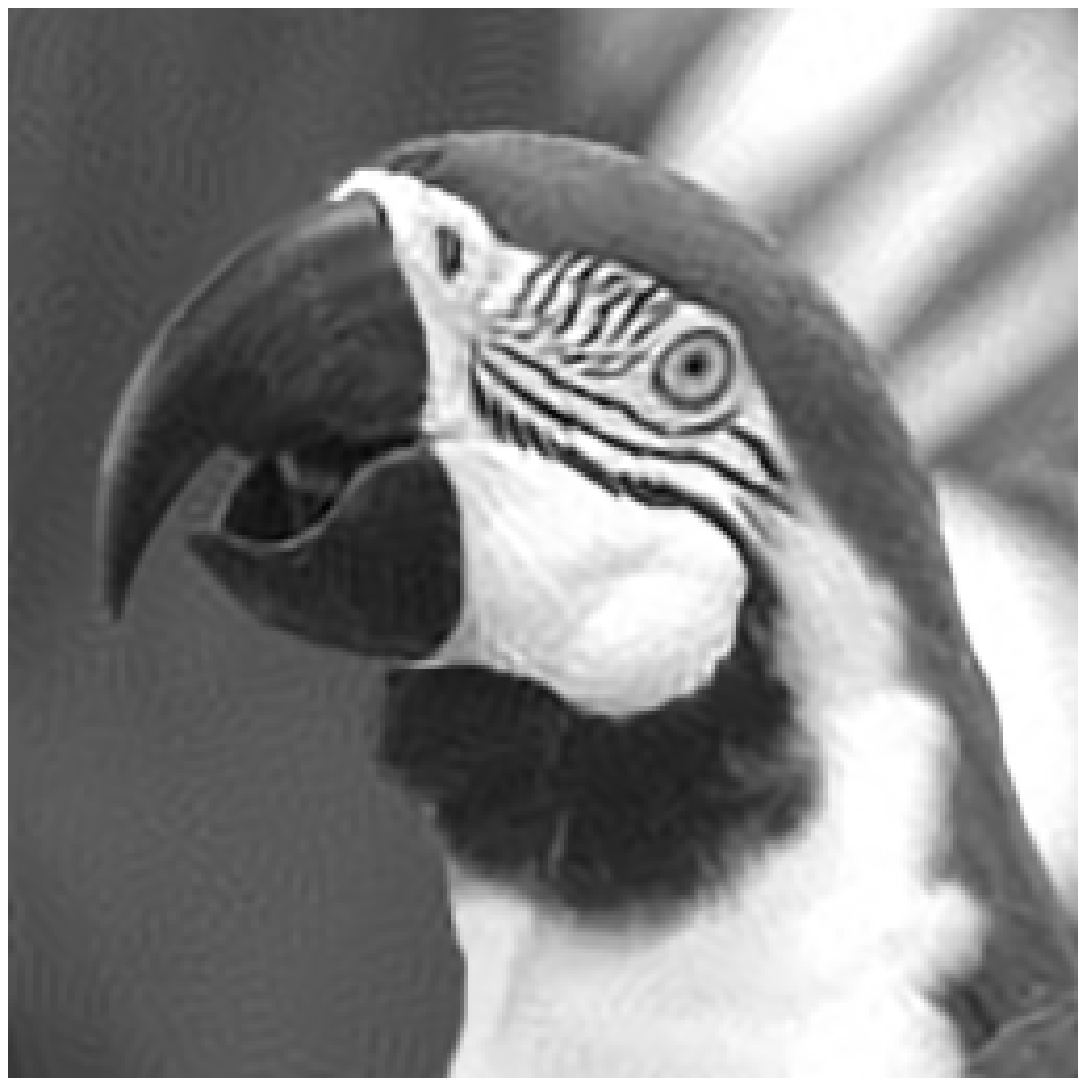}
   \includegraphics[width=0.19\textwidth]{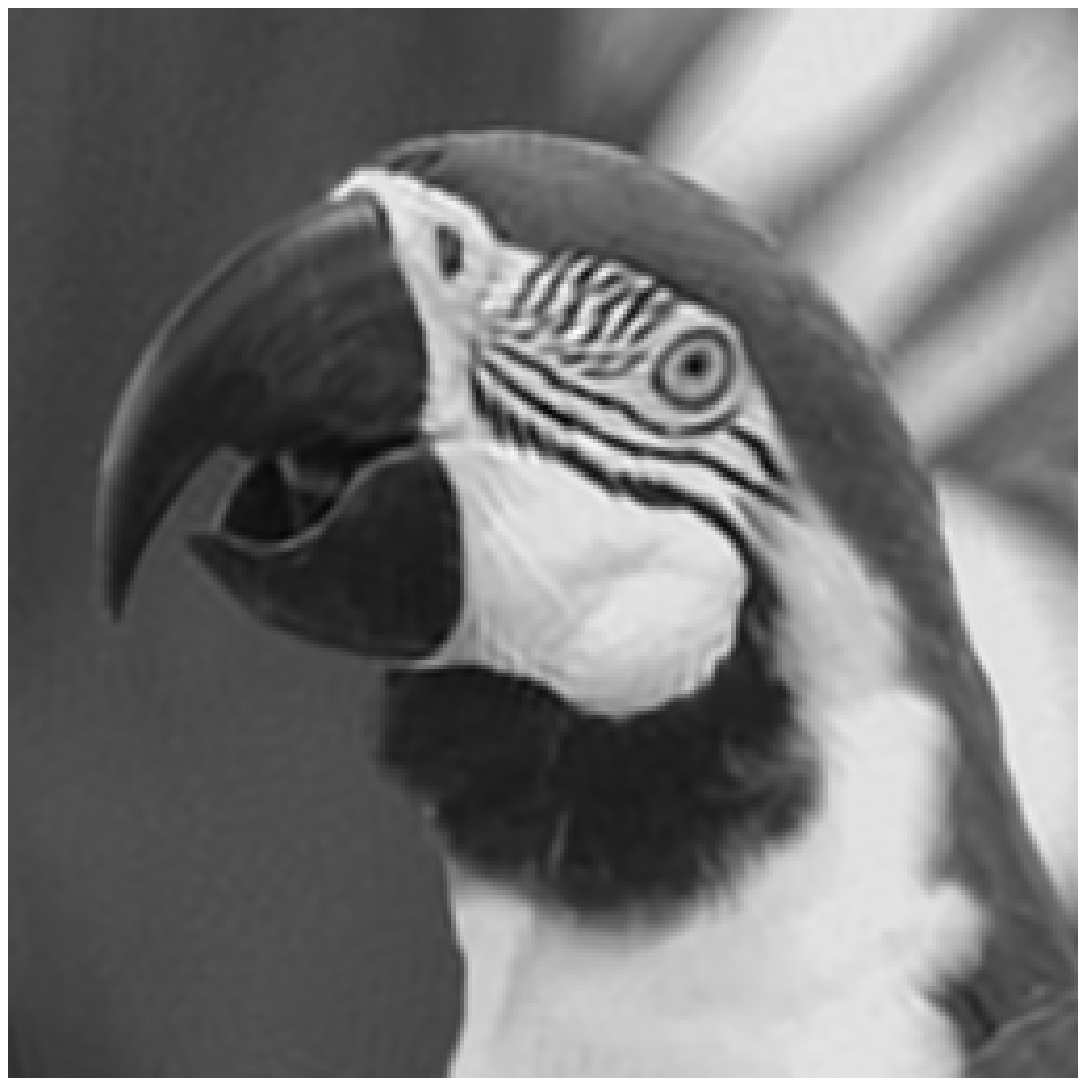}
   \includegraphics[width=0.19\textwidth]{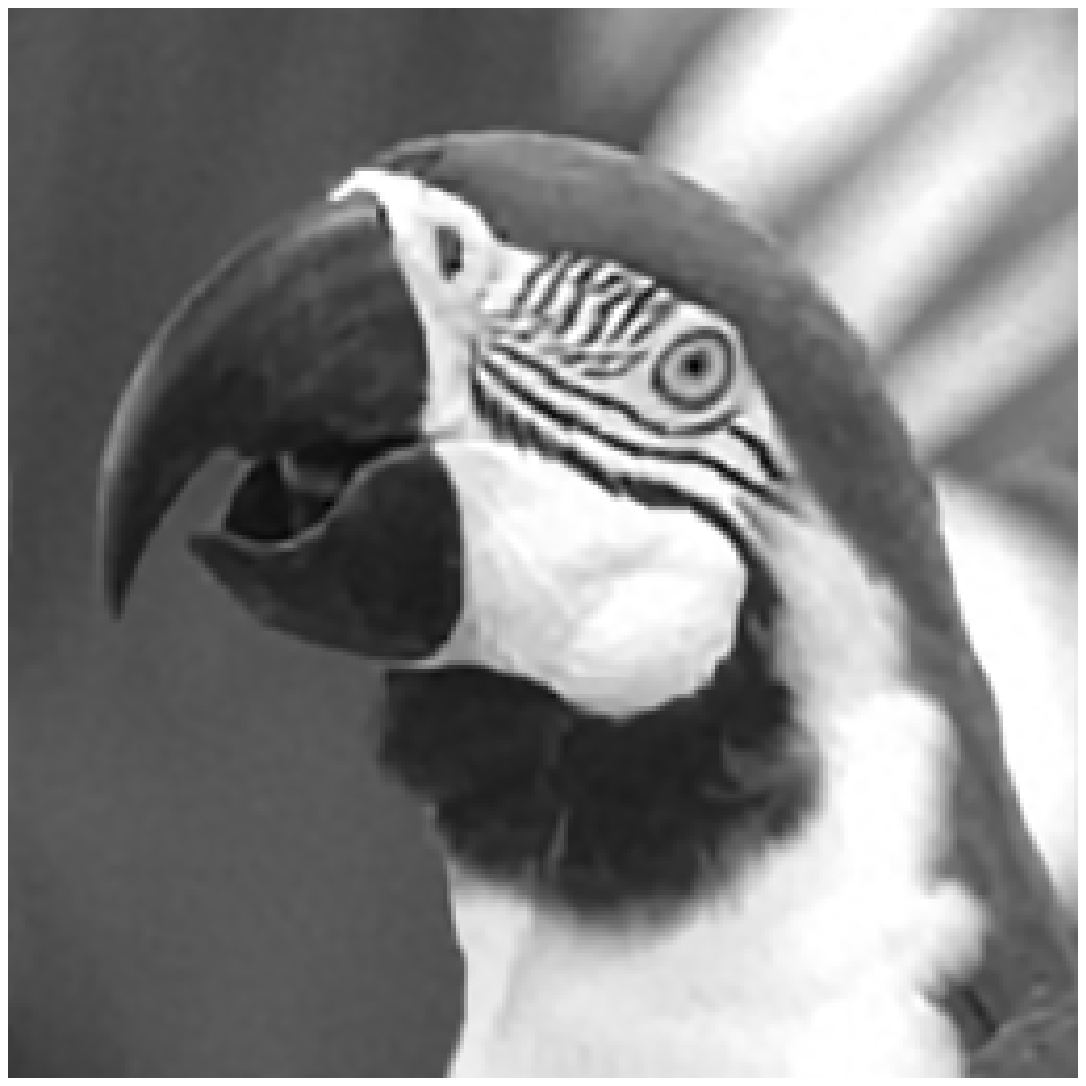}
    \includegraphics[width=0.19\textwidth]{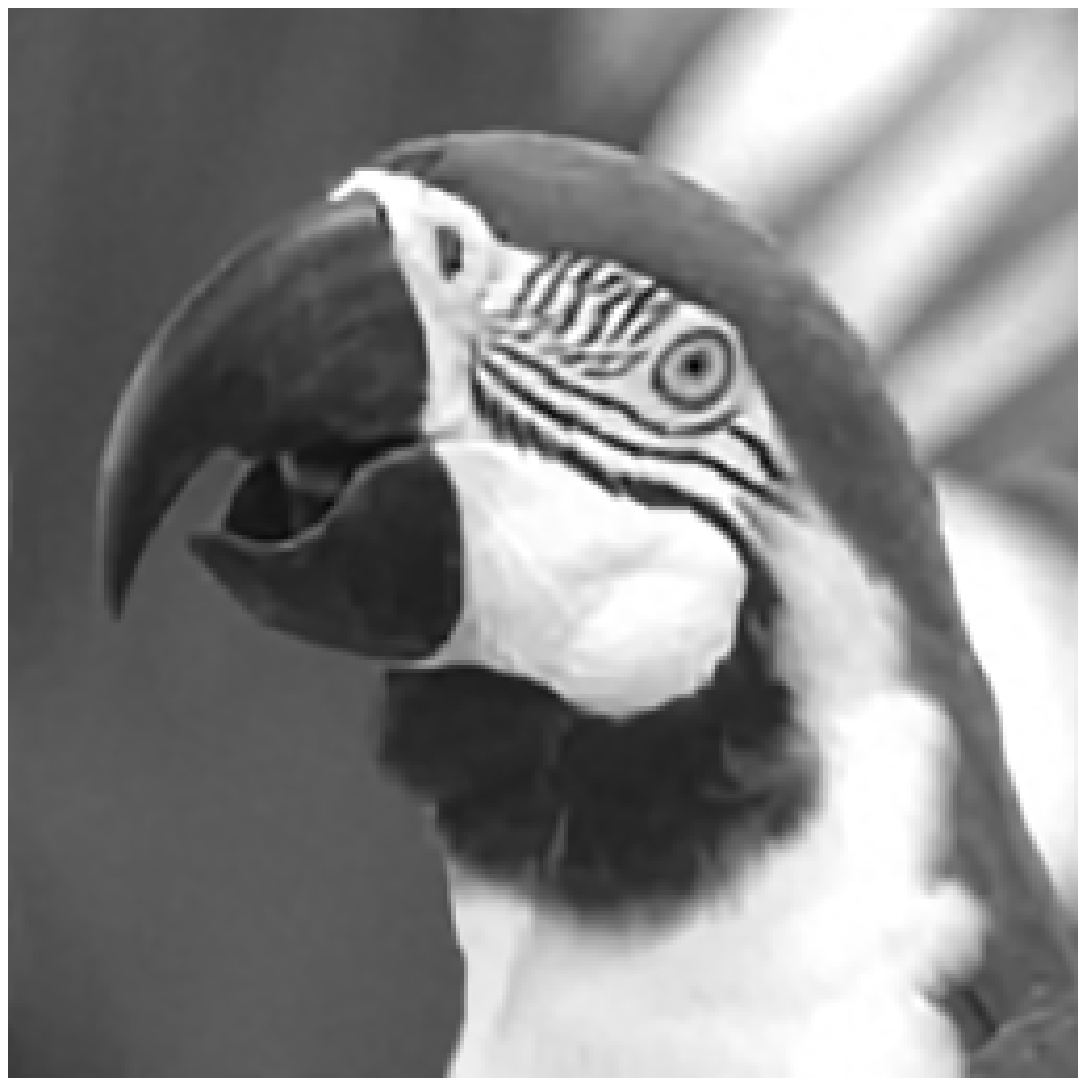}
   \includegraphics[width=0.19\textwidth]{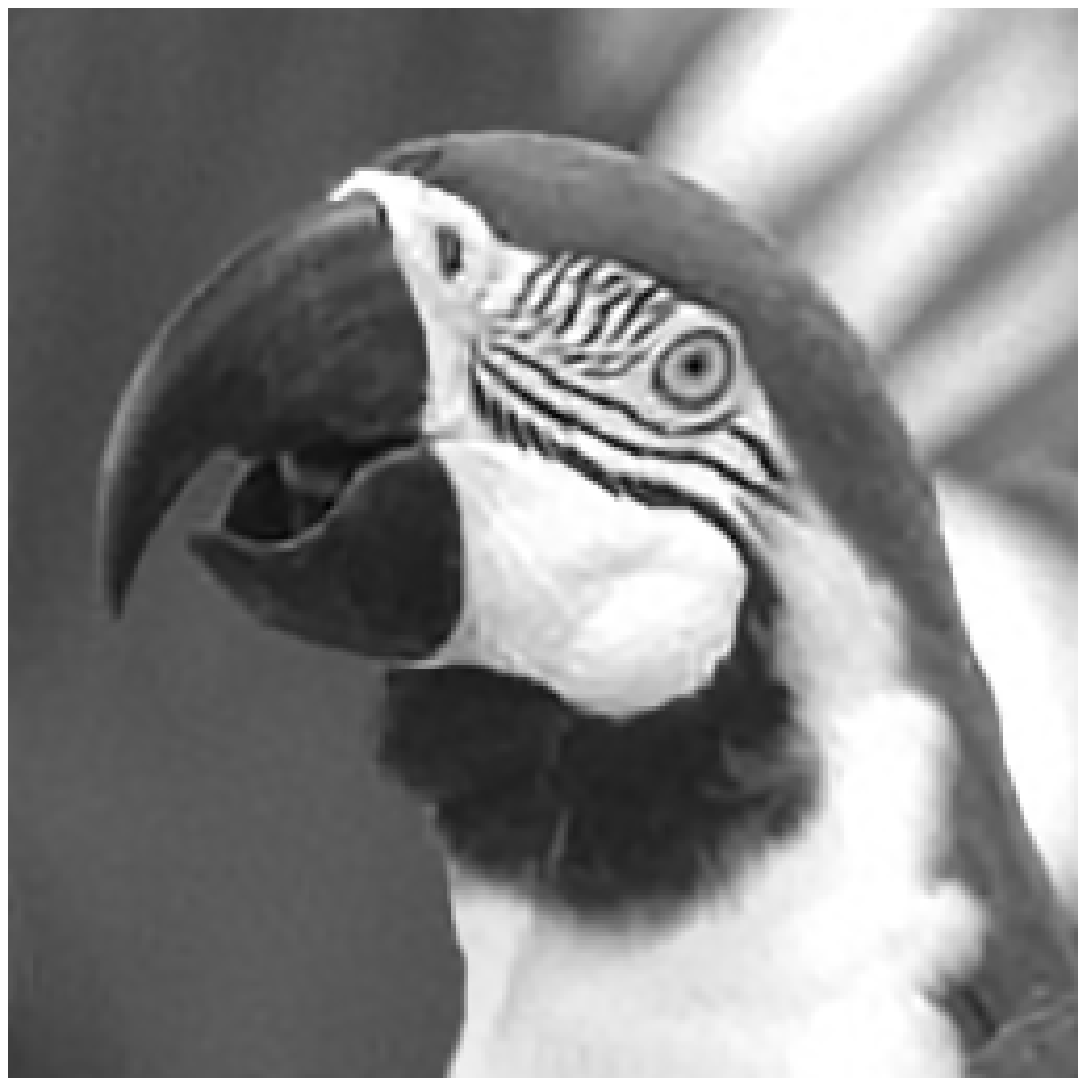} \\
   \includegraphics[width=0.19\textwidth]{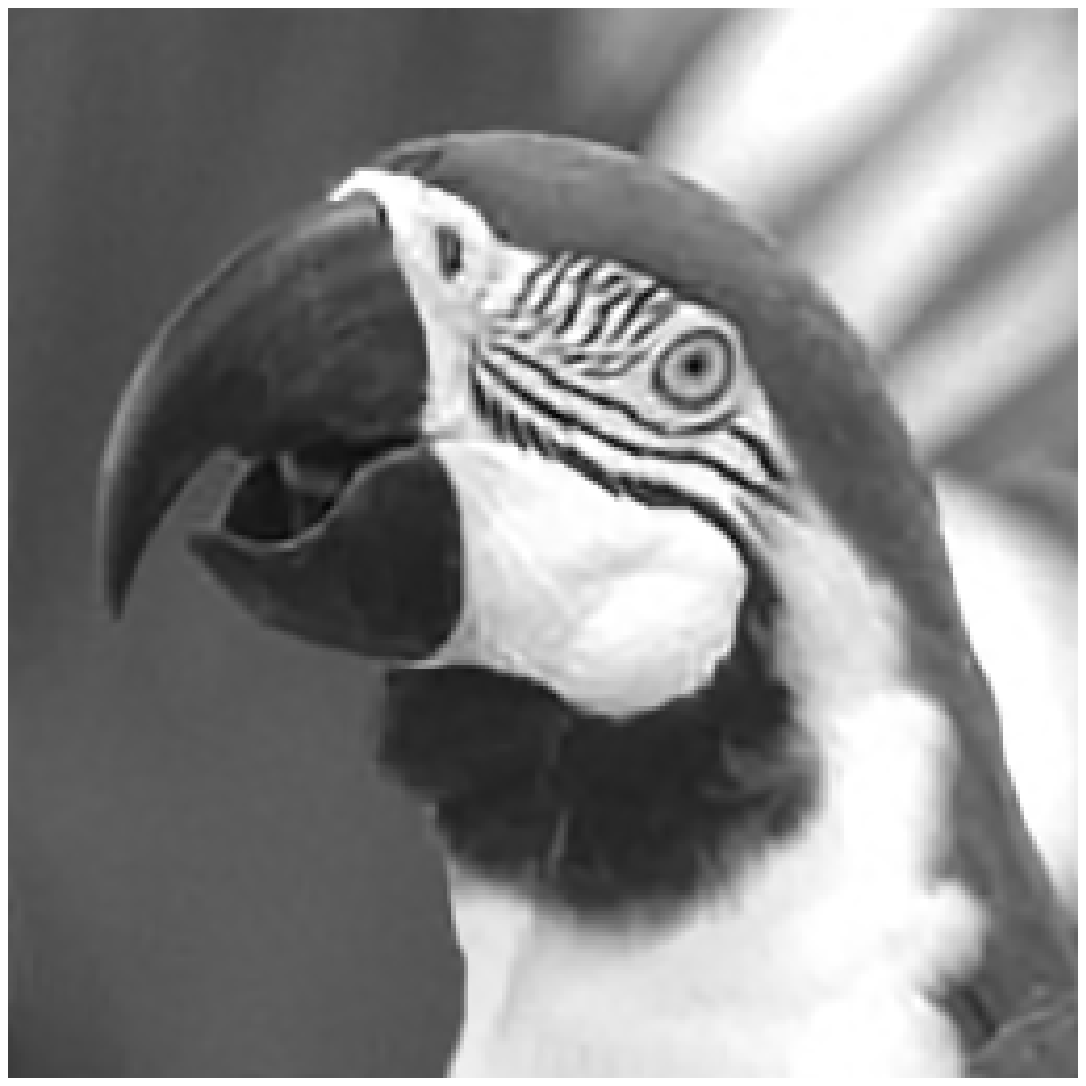}
   \includegraphics[width=0.19\textwidth]{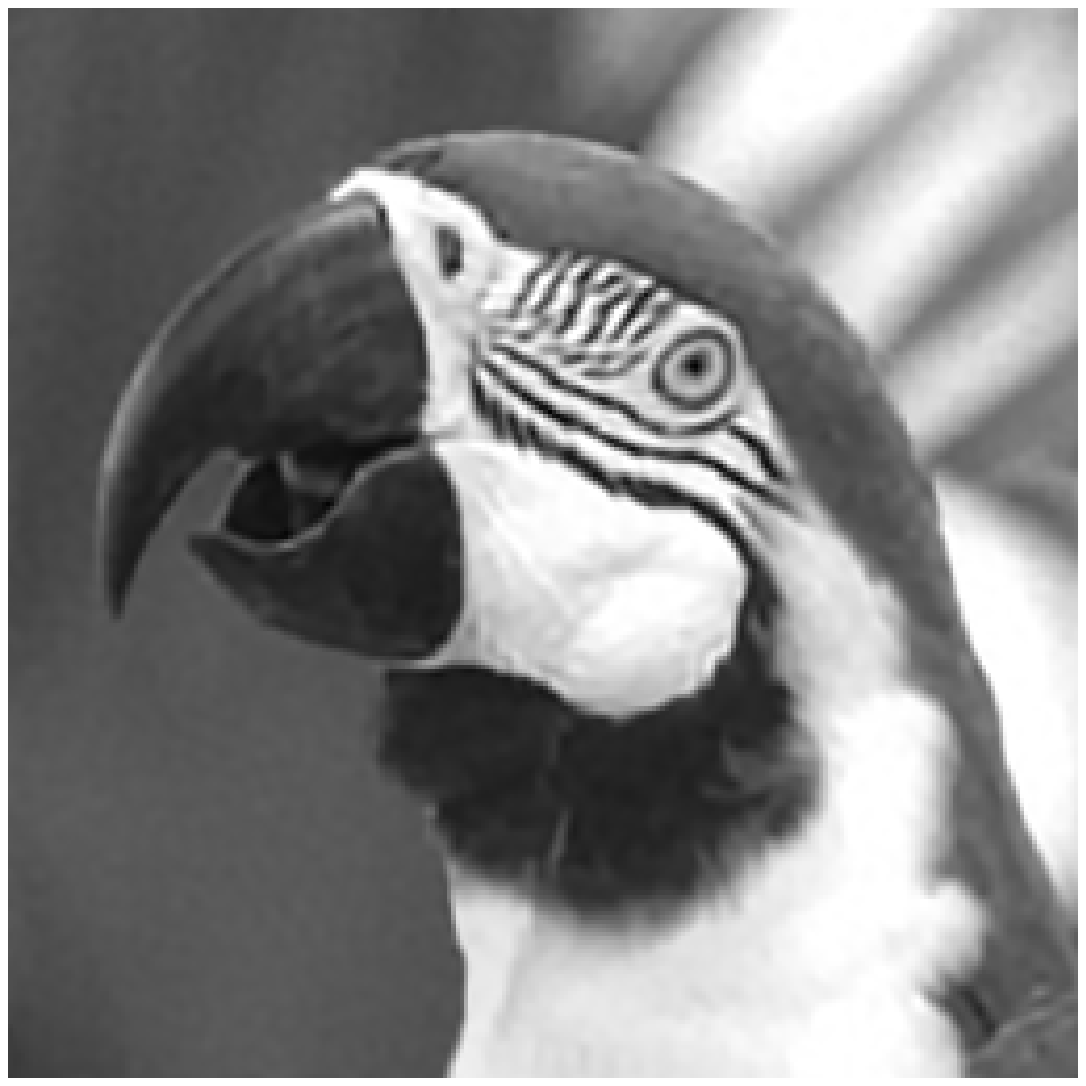}
   \includegraphics[width=0.19\textwidth]{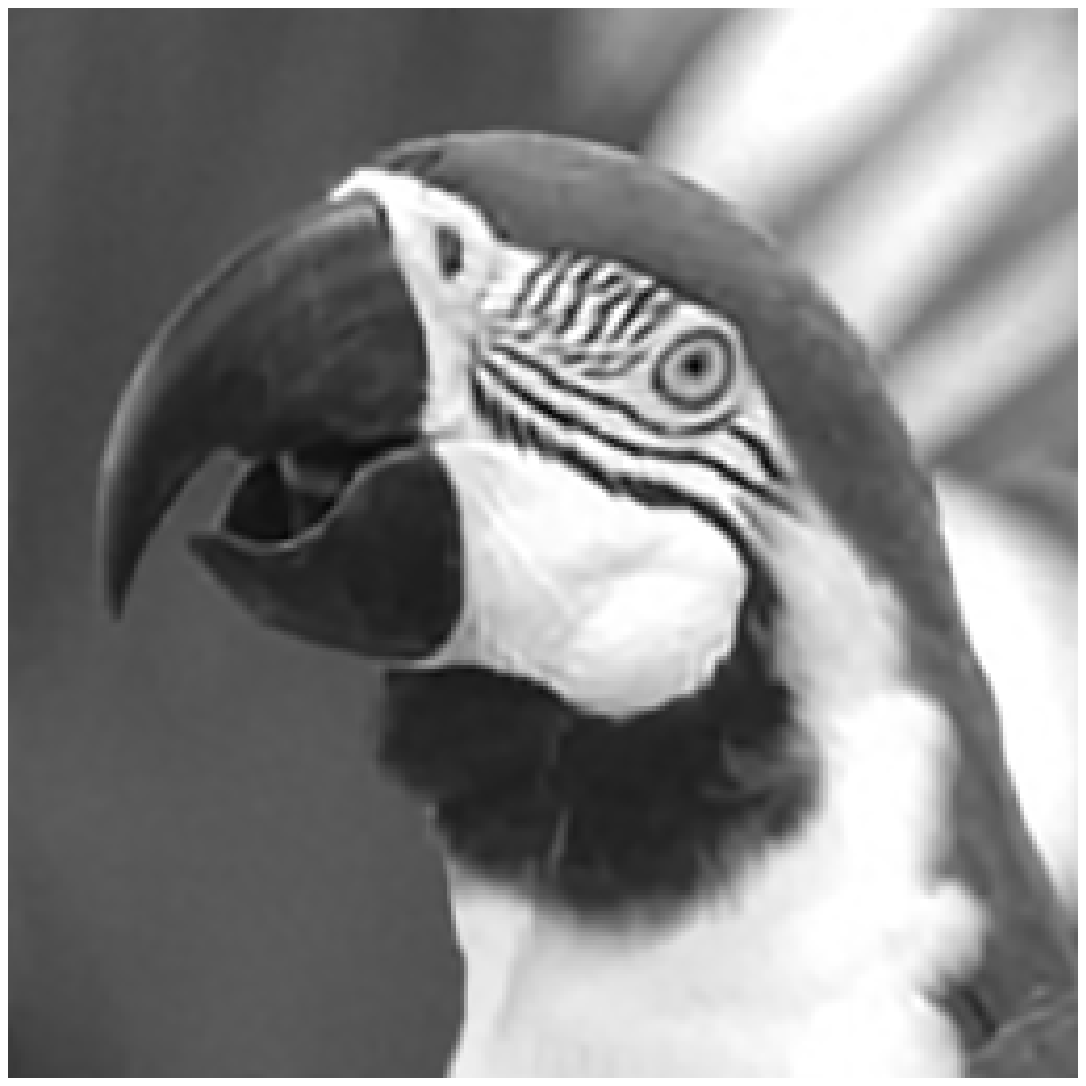}
   \includegraphics[width=0.19\textwidth]{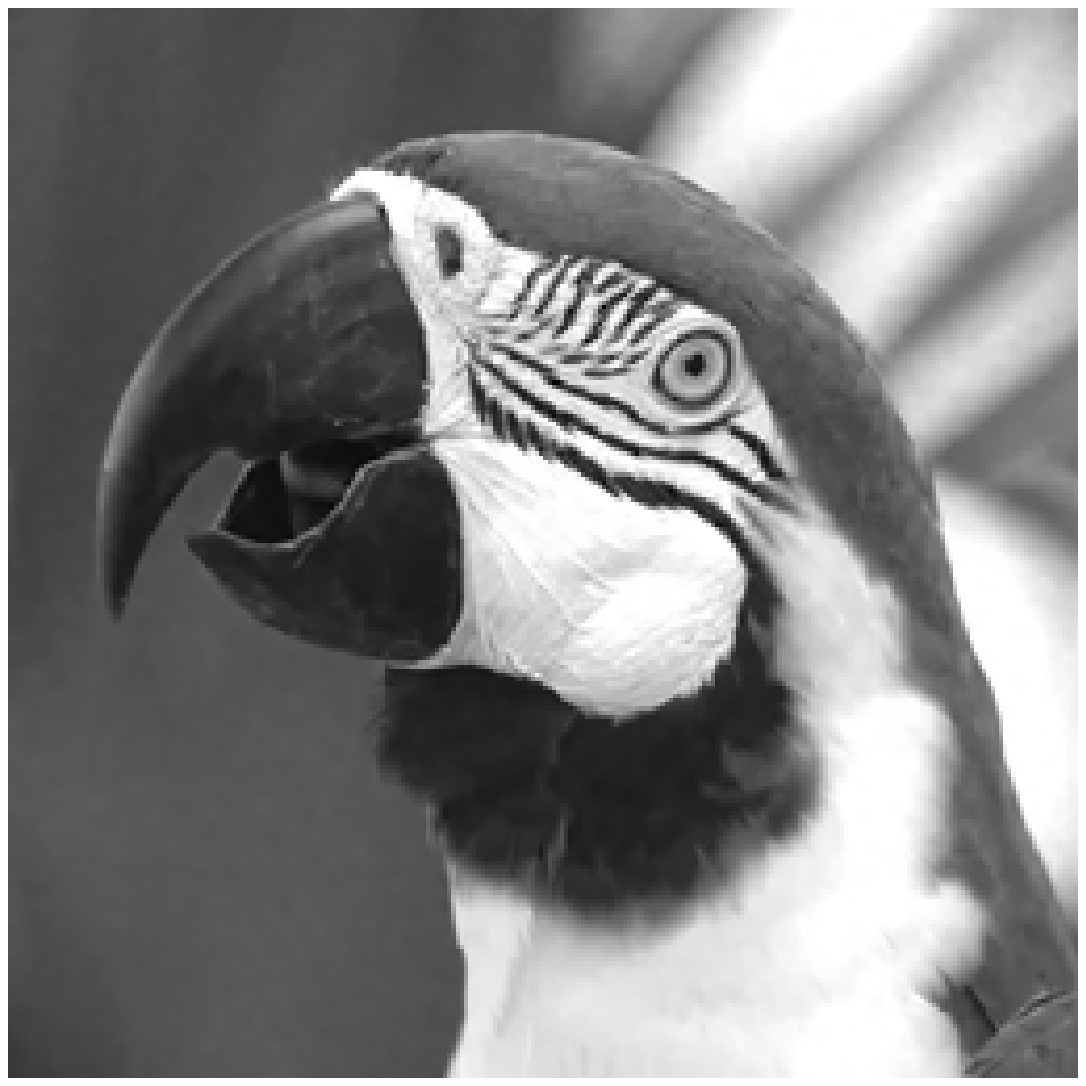}
   \includegraphics[width=0.19\textwidth]{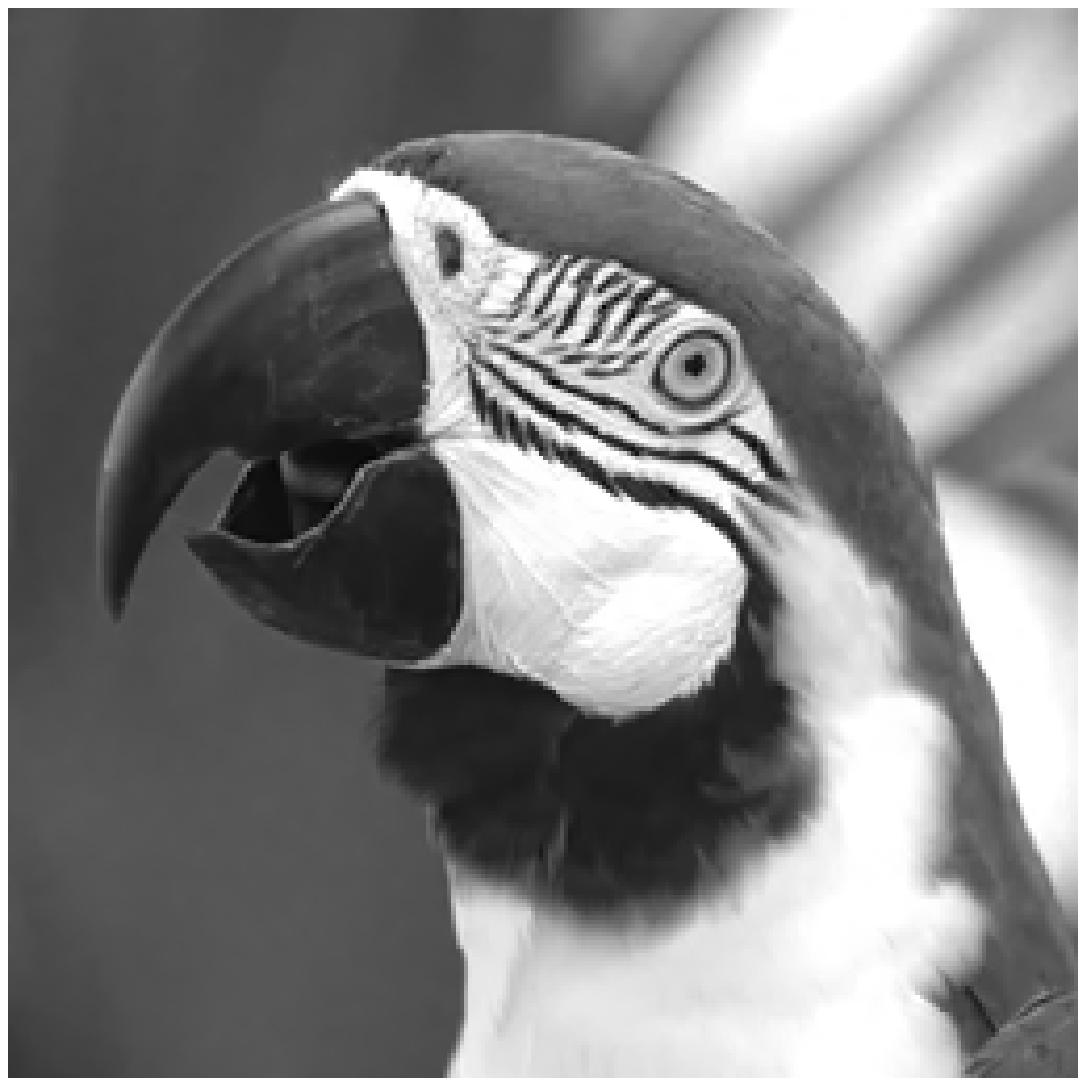} \\
  \caption{\small Visual quality comparison of image deburring results on image \textit{Parrots} ($256\times 256$). From left to right and top to bottom: original image, degraded image (Scenario 3), the recovered image by SB \cite{Cai2009a} (PSNR=30.31; SSIM=0.9163), MDAL \cite{DongBin2013} (PSNR=29.96; SSIM=0.9180), IDD-BM3D \cite{Danielyan2012} (PSNR=31.55; SSIM=0.9179), CSR \cite{Dong2011} (PSNR=31.76; SSIM=0.9054), GSR \cite{Zhang2014} (PSNR=31.40;SSIM=0.9179), Our proposed IDD-BM3D+SDSR(L=1)  (PSNR=31.85;SSIM=0.9244), IDD-BM3D+SDSR(L=4)  (PSNR=31.93;SSIM=0.9267), CSR+SDSR(L=1) (PSNR=32.18;SSIM=0.9249), CSR+SDSR(L=4) (PSNR=\textbf{32.27};SSIM=\textbf{0.9280}), GSR+SDSR(L=1) (PSNR=31.68;SSIM=0.9242), GSR+SDSR(L=4) (PSNR=31.79;SSIM=0.9271), ORACLE(L=1) (PSNR=36.86;SSIM=0.9669), ORACLE(L=4) (PSNR=38.36;SSIM=0.9717). Bold values denote the highest PSNR or SSIM values excluding the ORACLE cases.}
\label{fig:parrots comparison}
\end{figure*}

\begin{figure*}[h]
  \centering
   \includegraphics[width=0.32\textwidth]{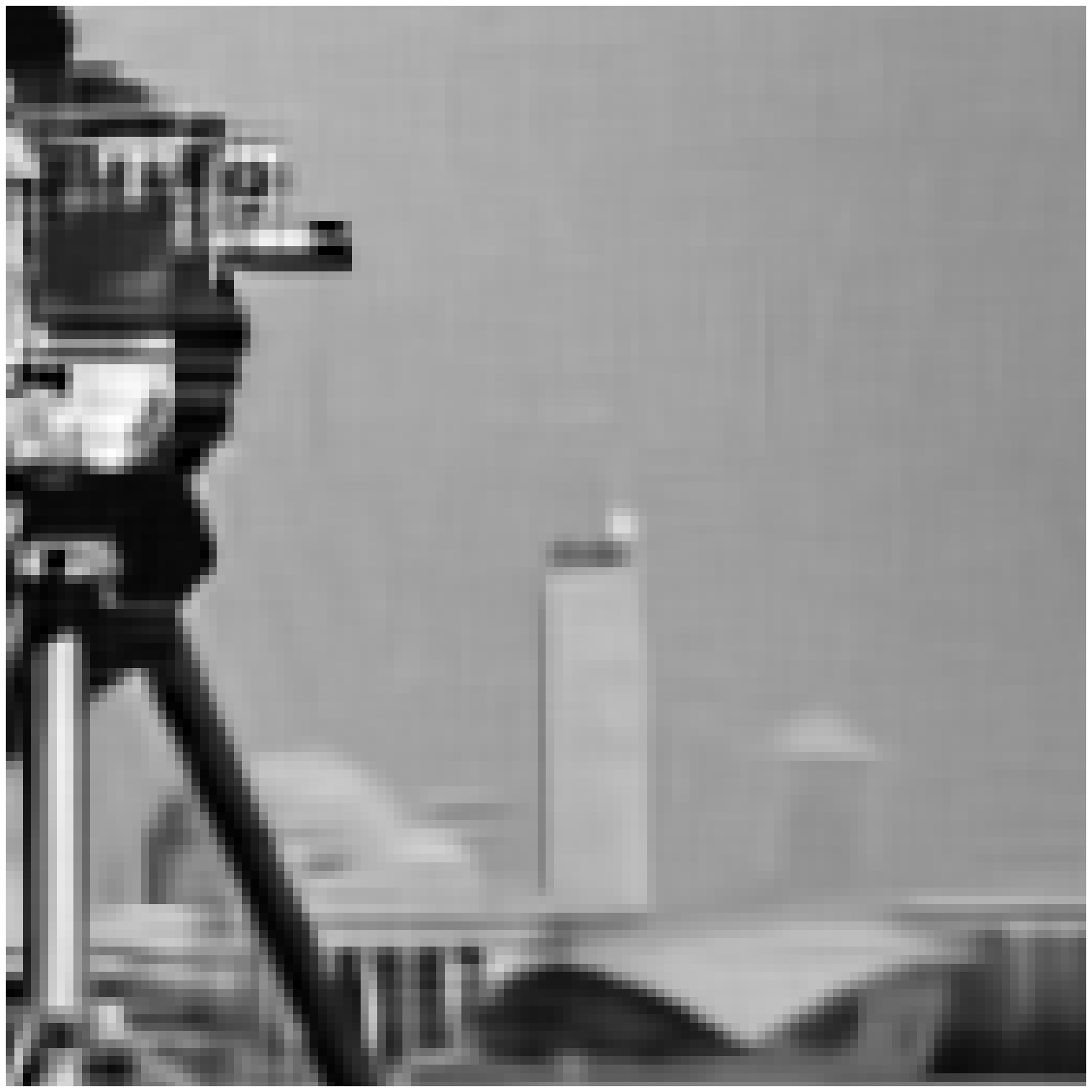}
   \includegraphics[width=0.32\textwidth]{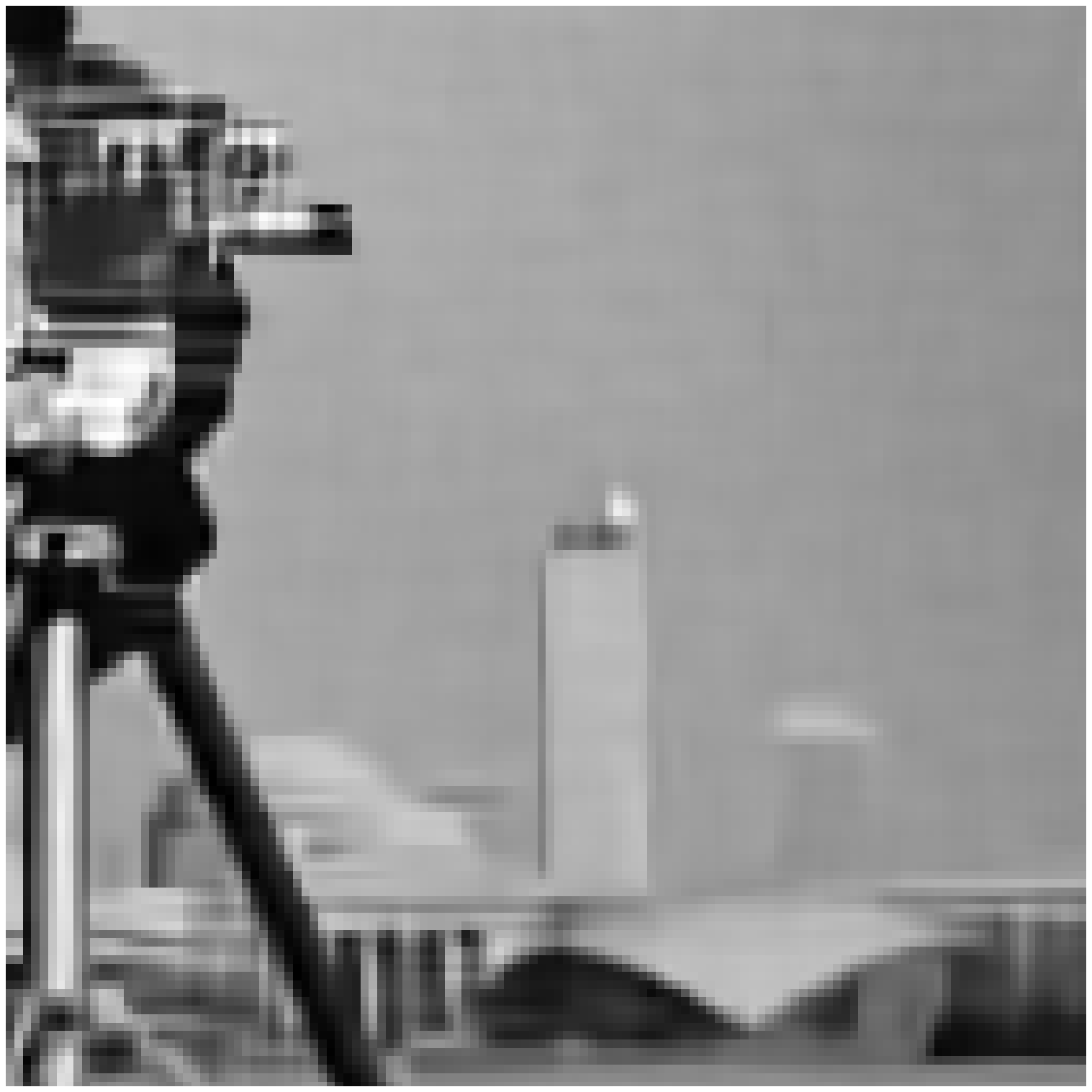}
   \includegraphics[width=0.32\textwidth]{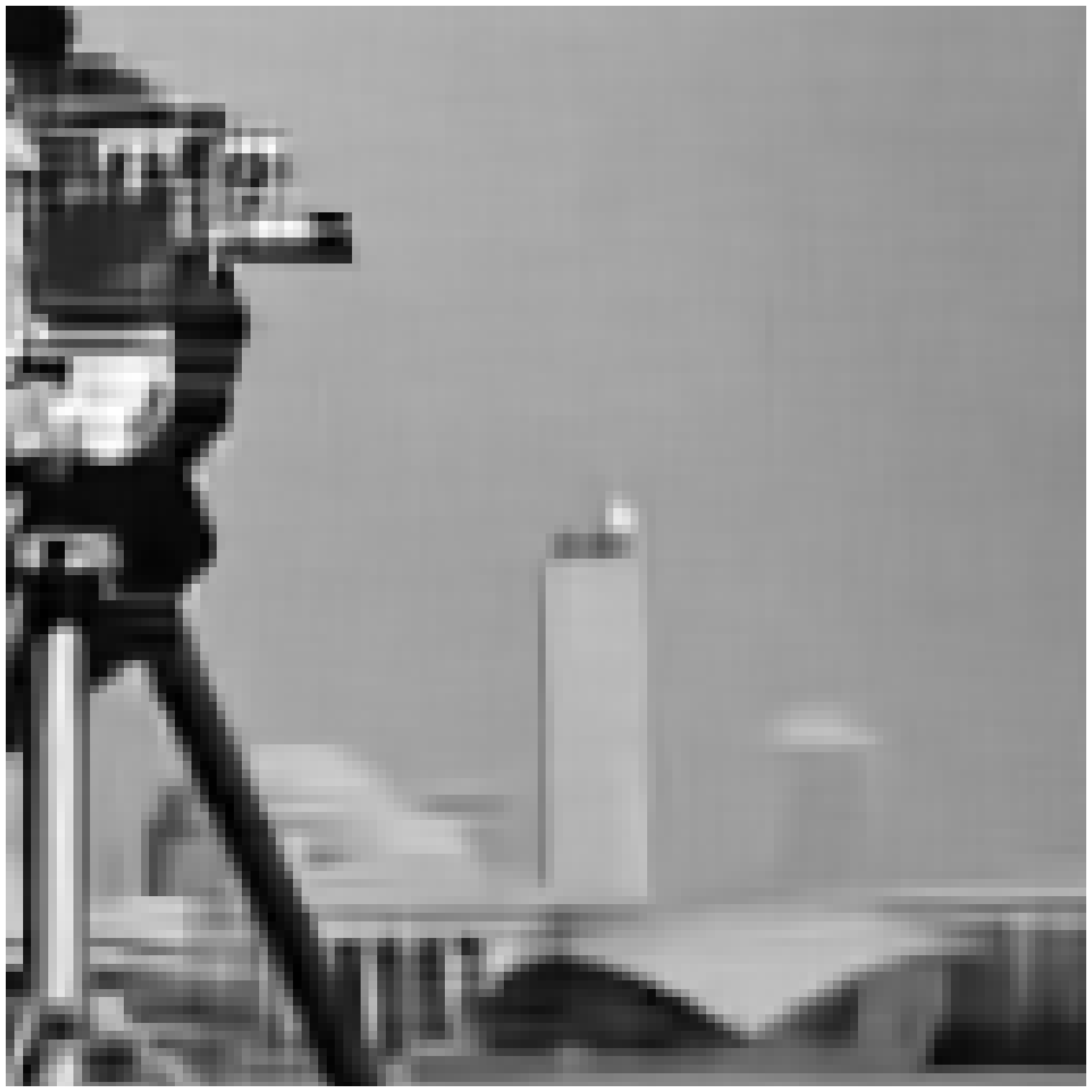} \\
      \includegraphics[width=0.32\textwidth]{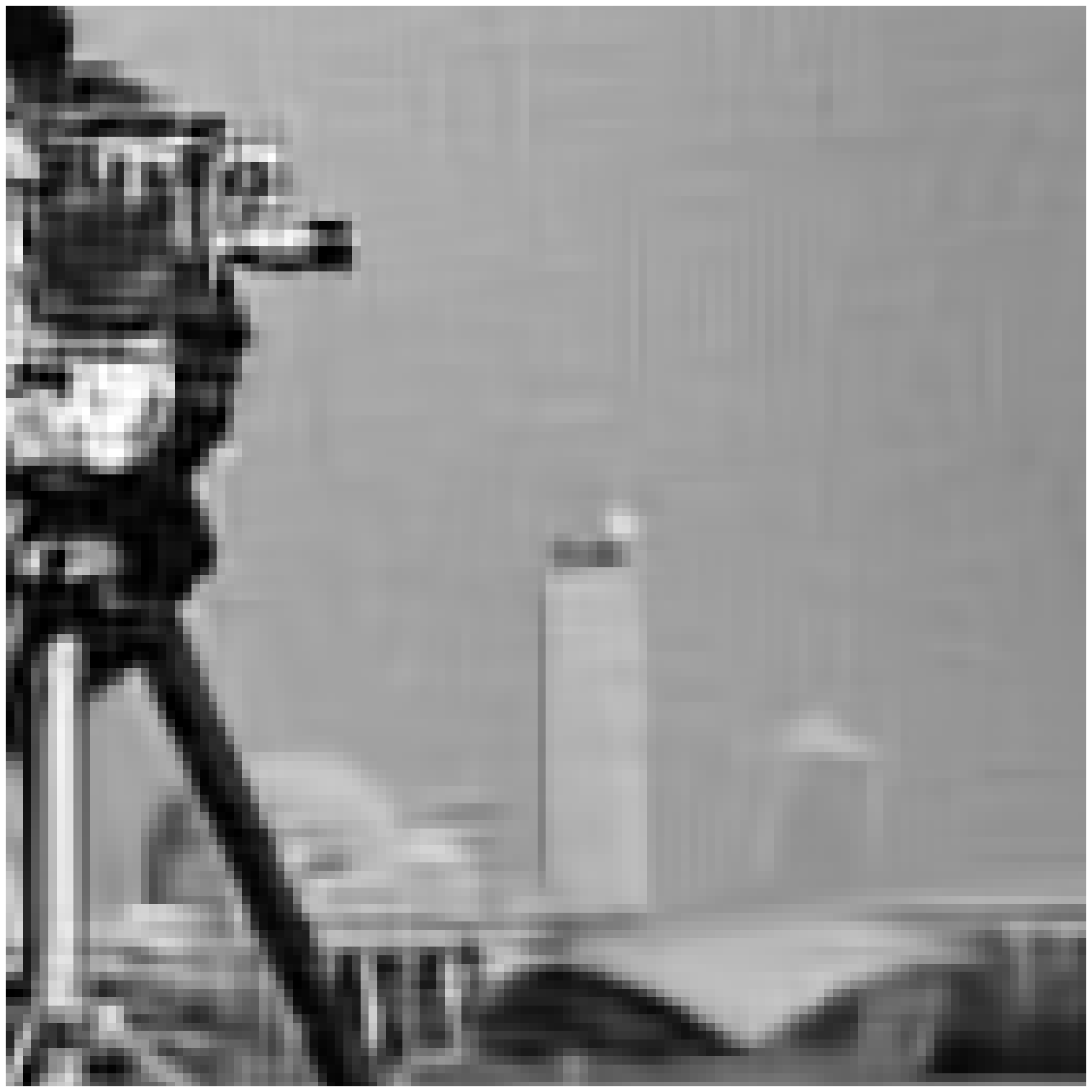}
   \includegraphics[width=0.32\textwidth]{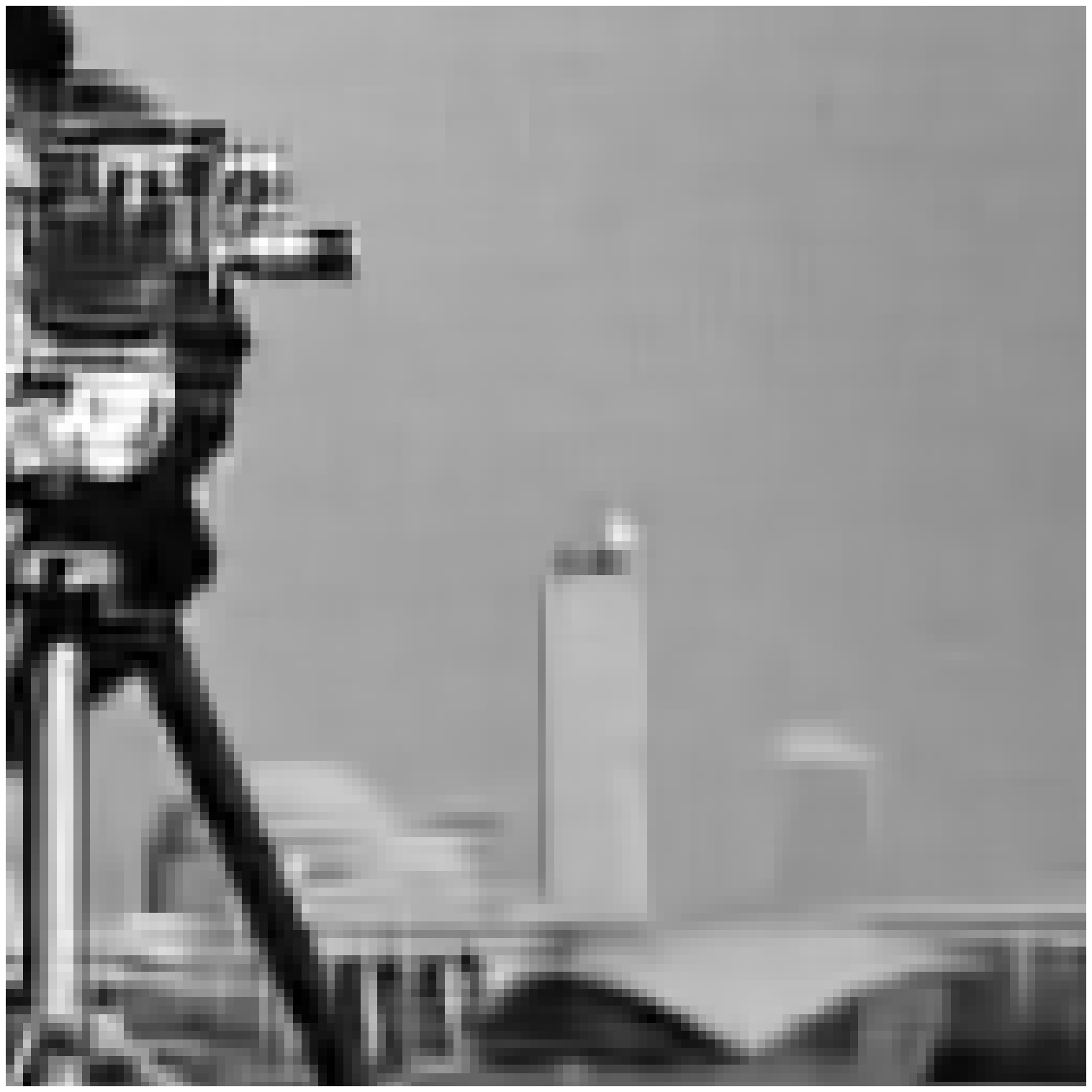}
   \includegraphics[width=0.32\textwidth]{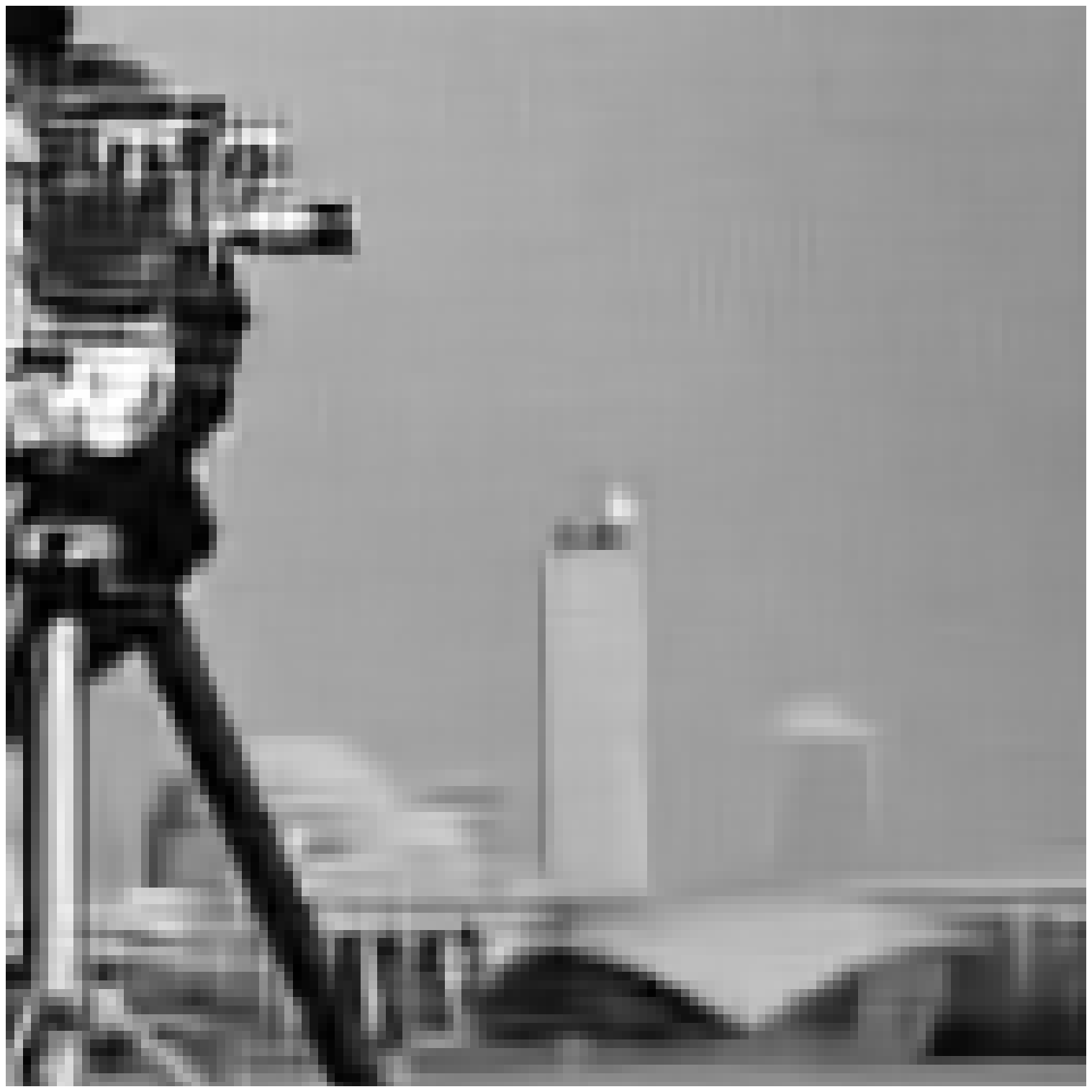} \\
         \includegraphics[width=0.32\textwidth]{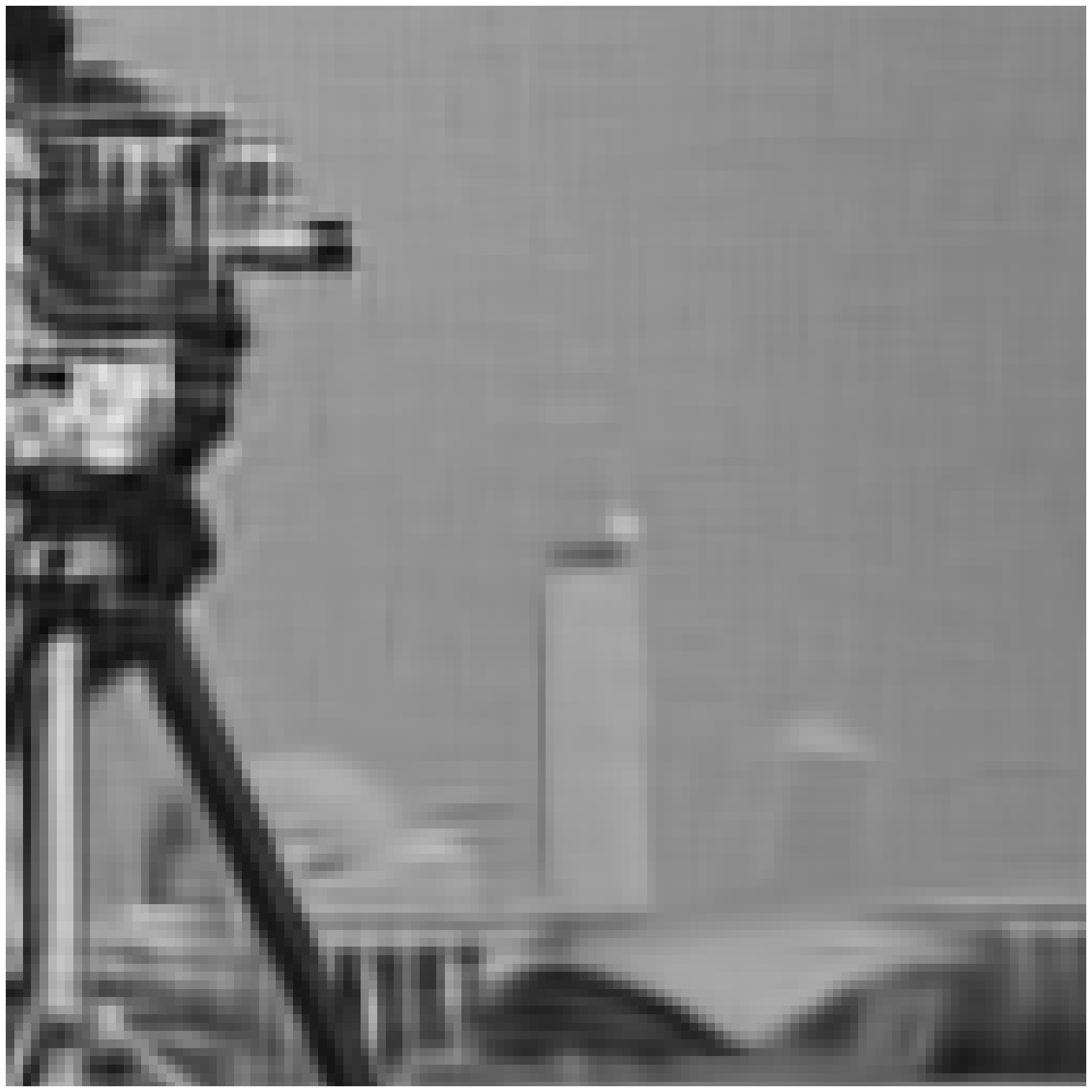}
   \includegraphics[width=0.32\textwidth]{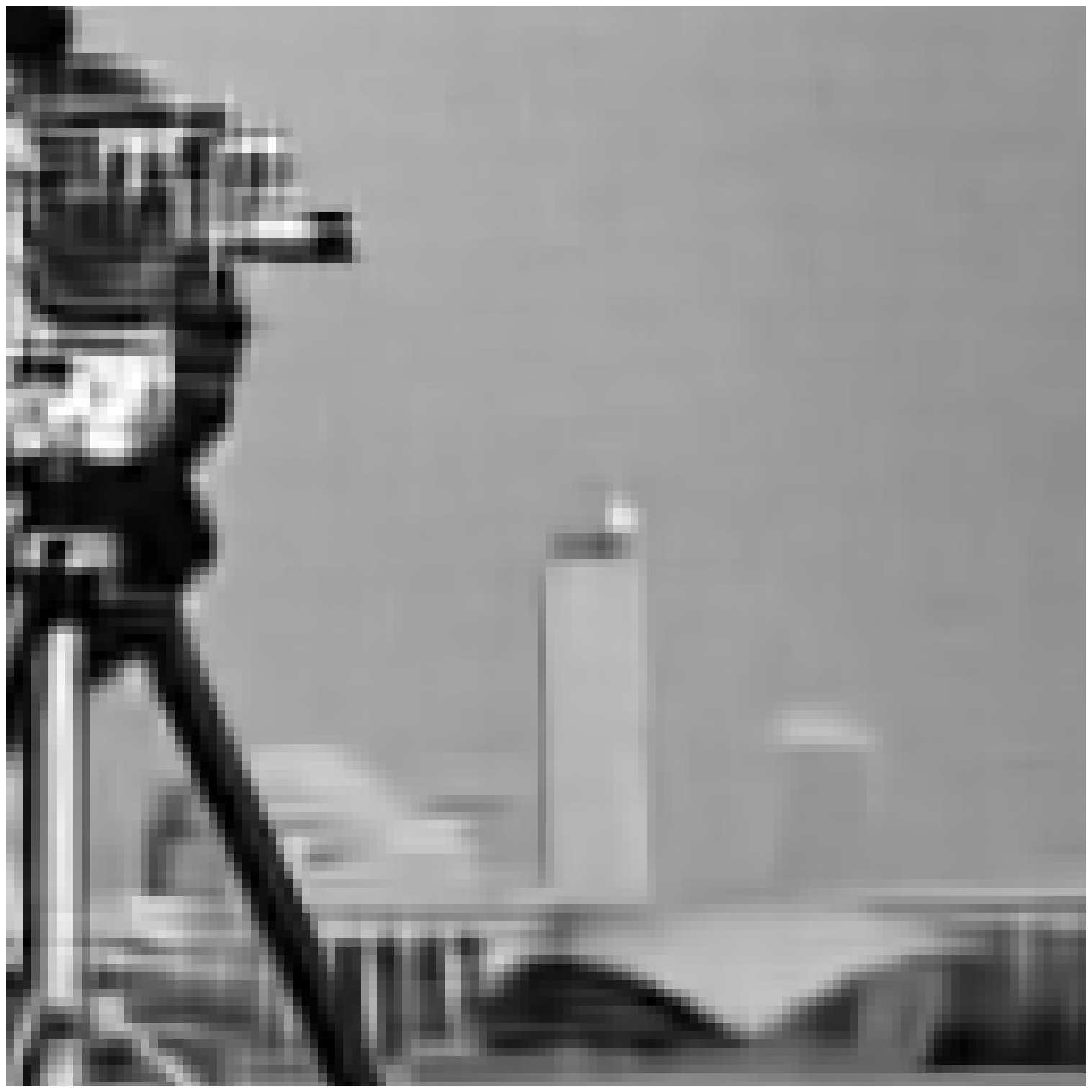}
   \includegraphics[width=0.32\textwidth]{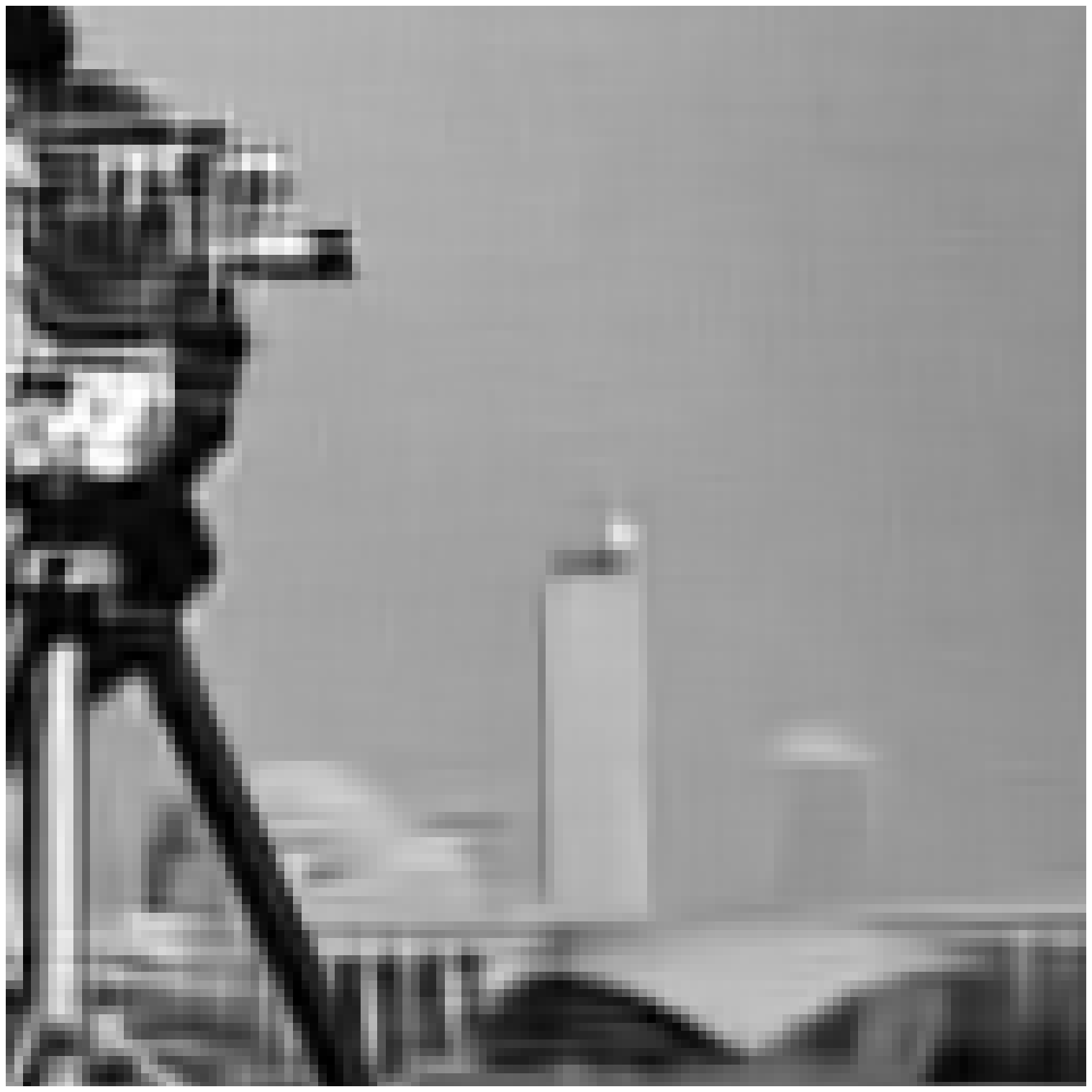} \\

  \caption{\small The zoom in visual comparisons corresponding to Fig \ref{fig:cameraman comparison}. From left to right and top to bottom: IDD-BM3D, Proposed IDD-BM3D+SDSR(L=1), IDD-BM3D+SDSR(L=4), CSR, Proposed CSR+SDSR(L=1), CSR+SDSR(L=4), GSR, Proposed GSR+SDSR(L=1), GSR+SDSR(L=4). }
\label{fig:cameraman zoomin}
\end{figure*}

\begin{figure*}[h]
  \centering
   \includegraphics[width=0.32\textwidth]{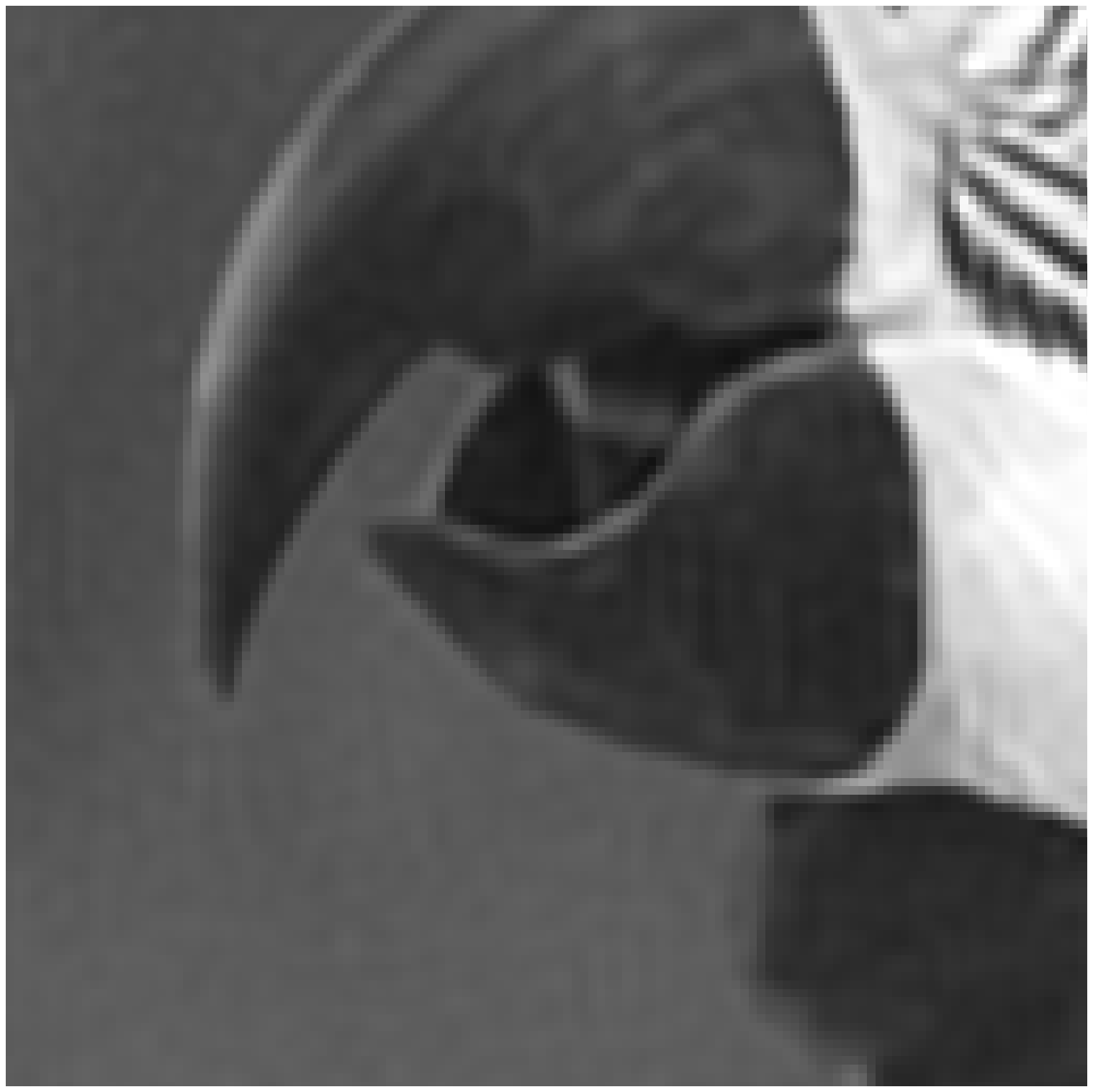}
   \includegraphics[width=0.32\textwidth]{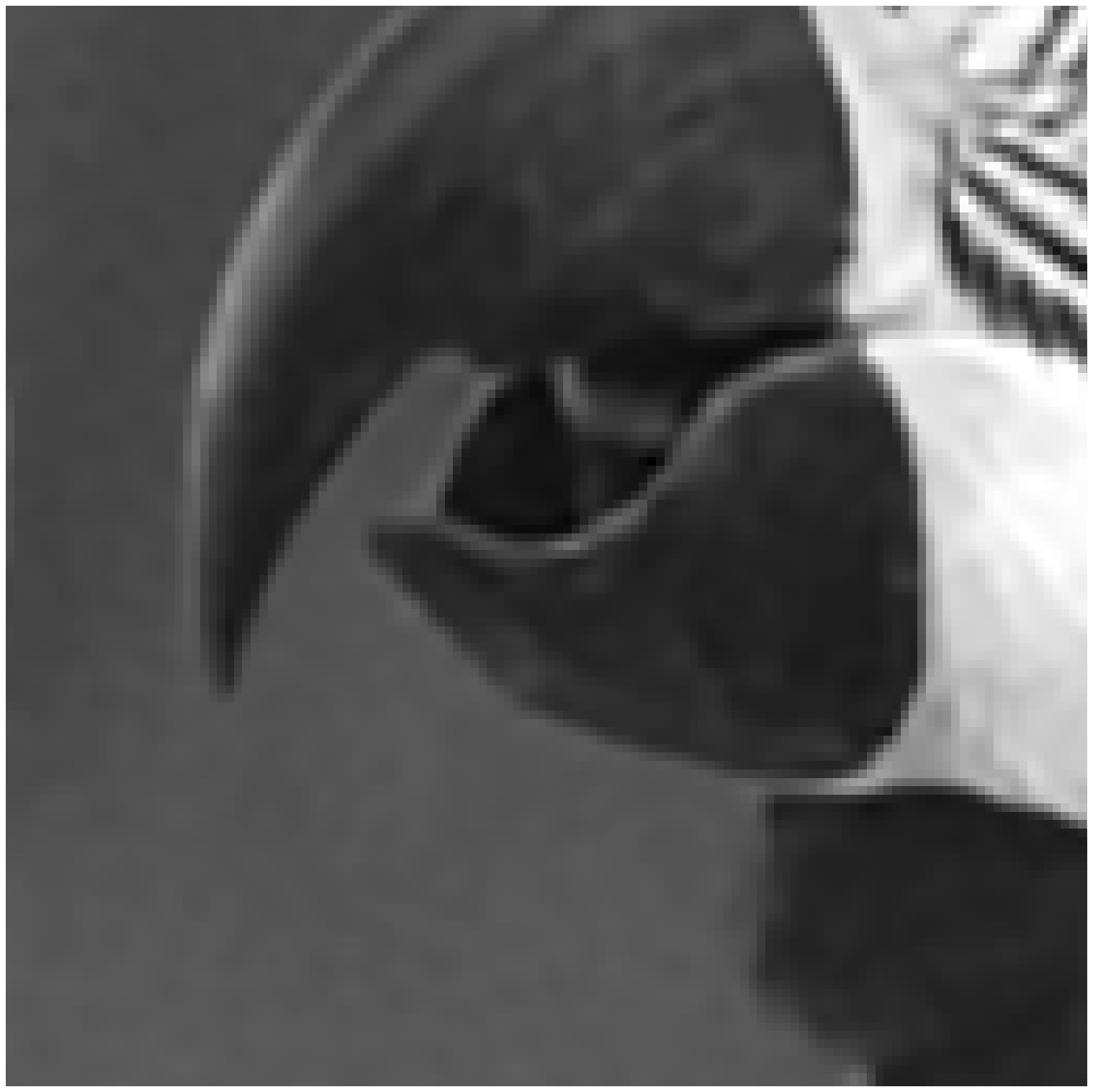}
   \includegraphics[width=0.32\textwidth]{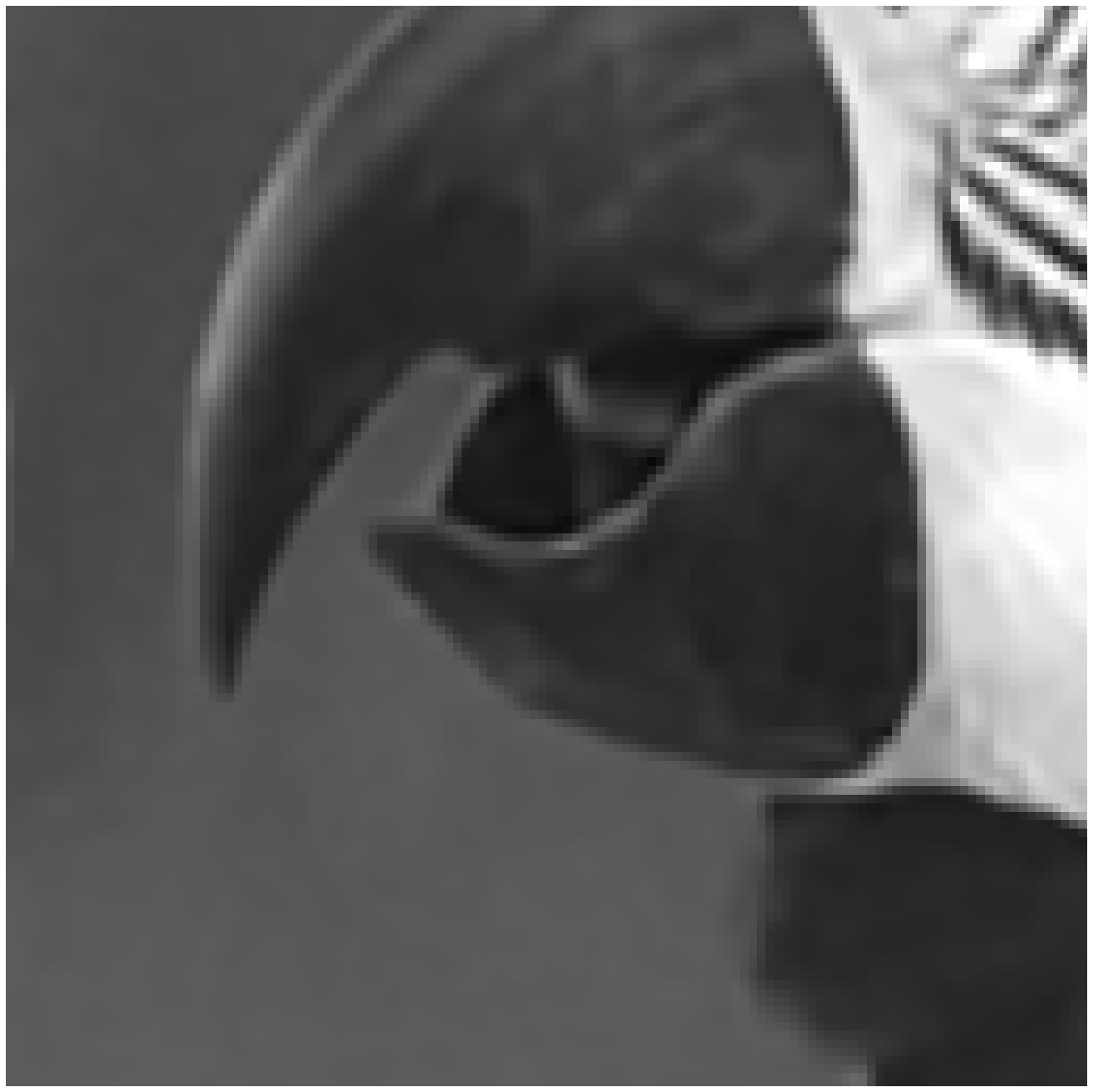} \\
   \includegraphics[width=0.32\textwidth]{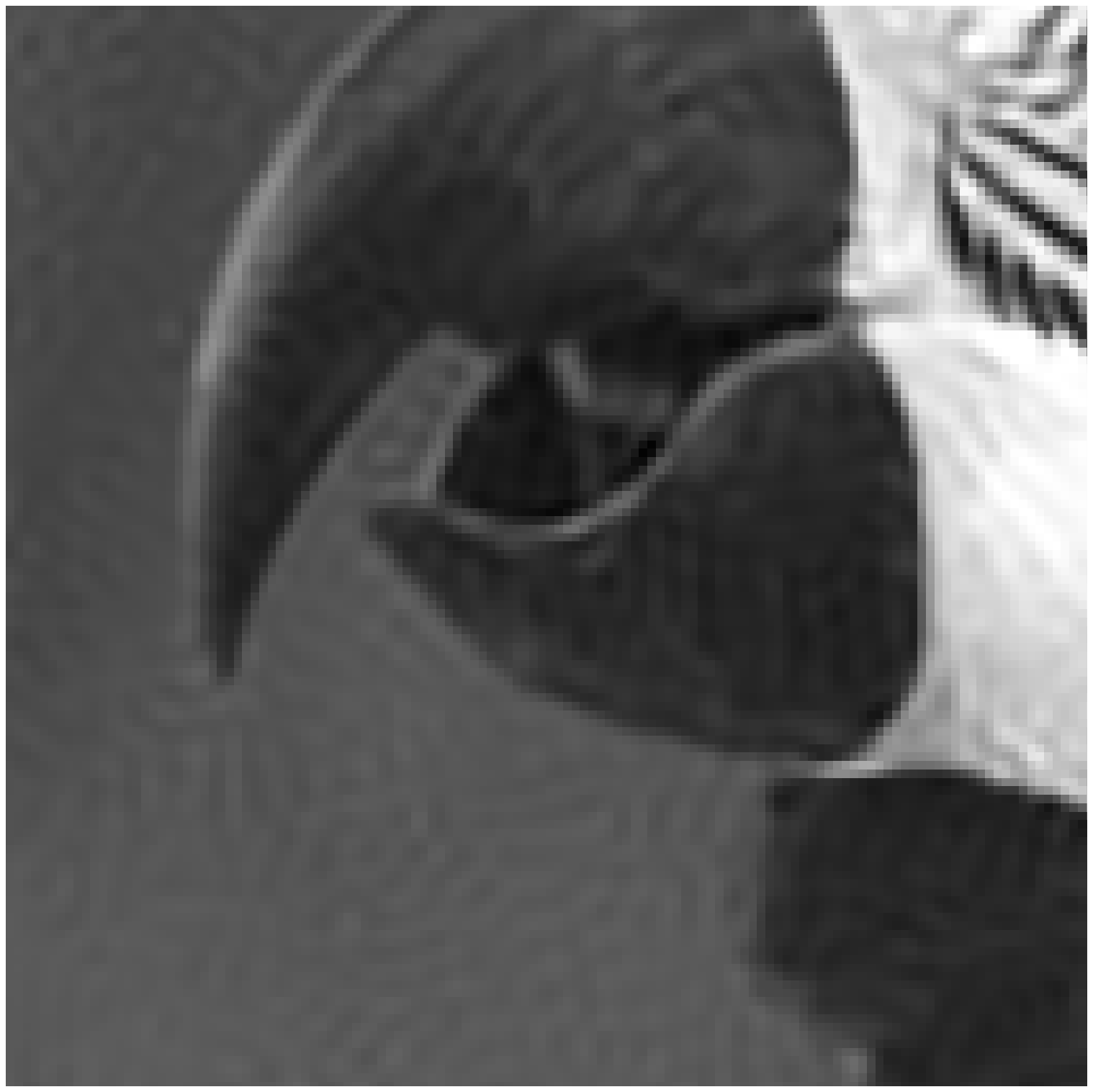}
   \includegraphics[width=0.32\textwidth]{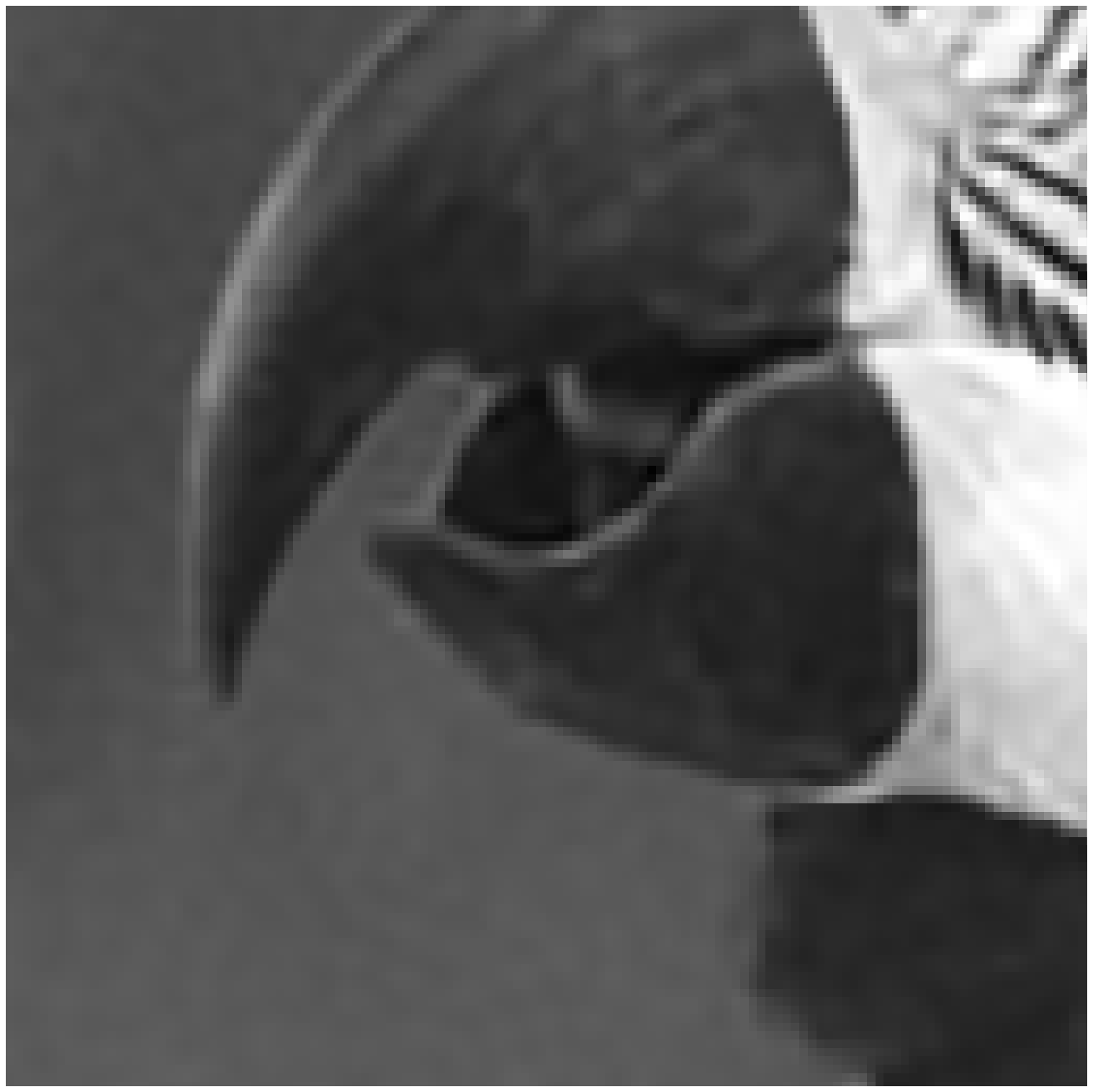}
   \includegraphics[width=0.32\textwidth]{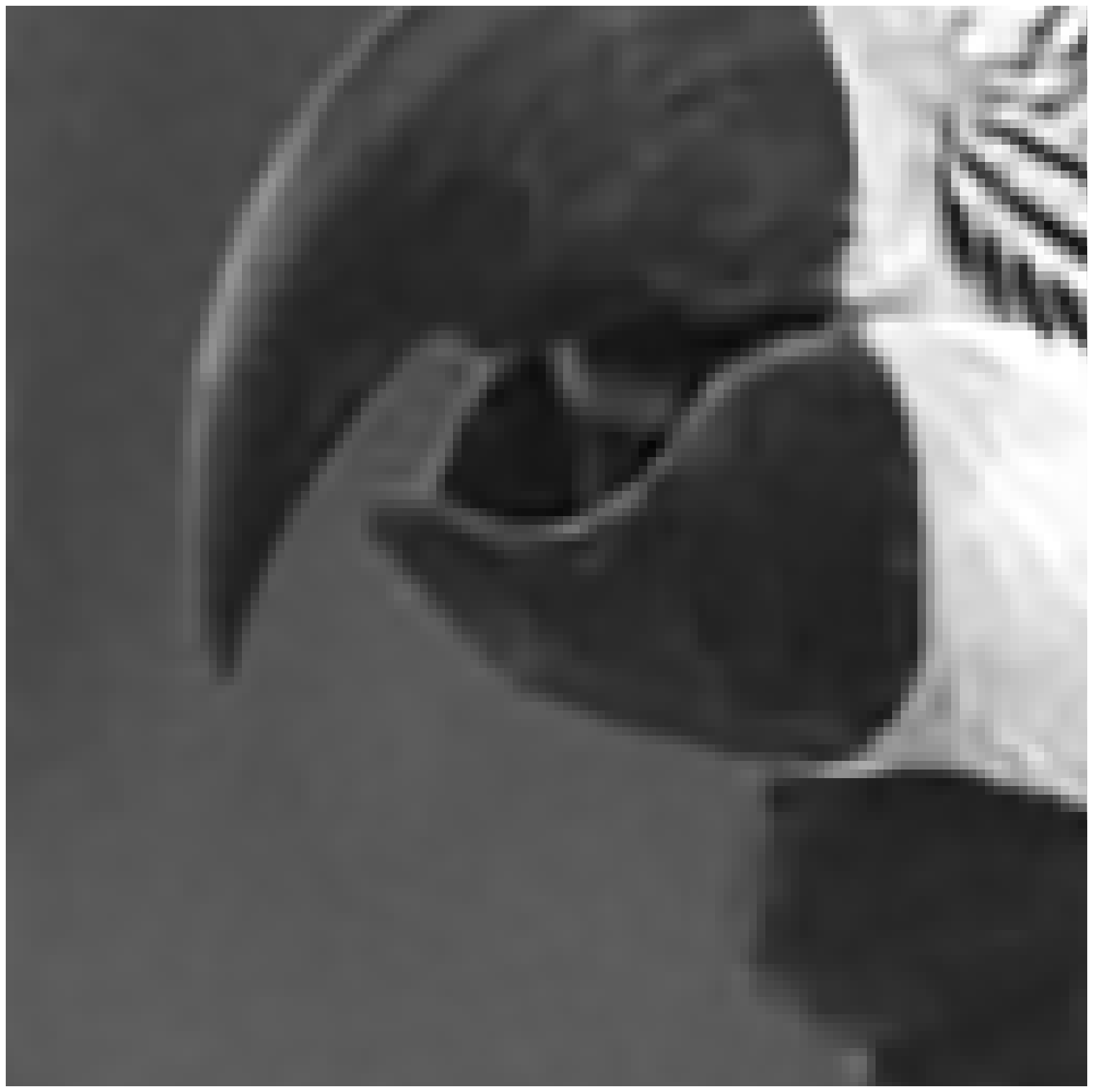} \\
   \includegraphics[width=0.32\textwidth]{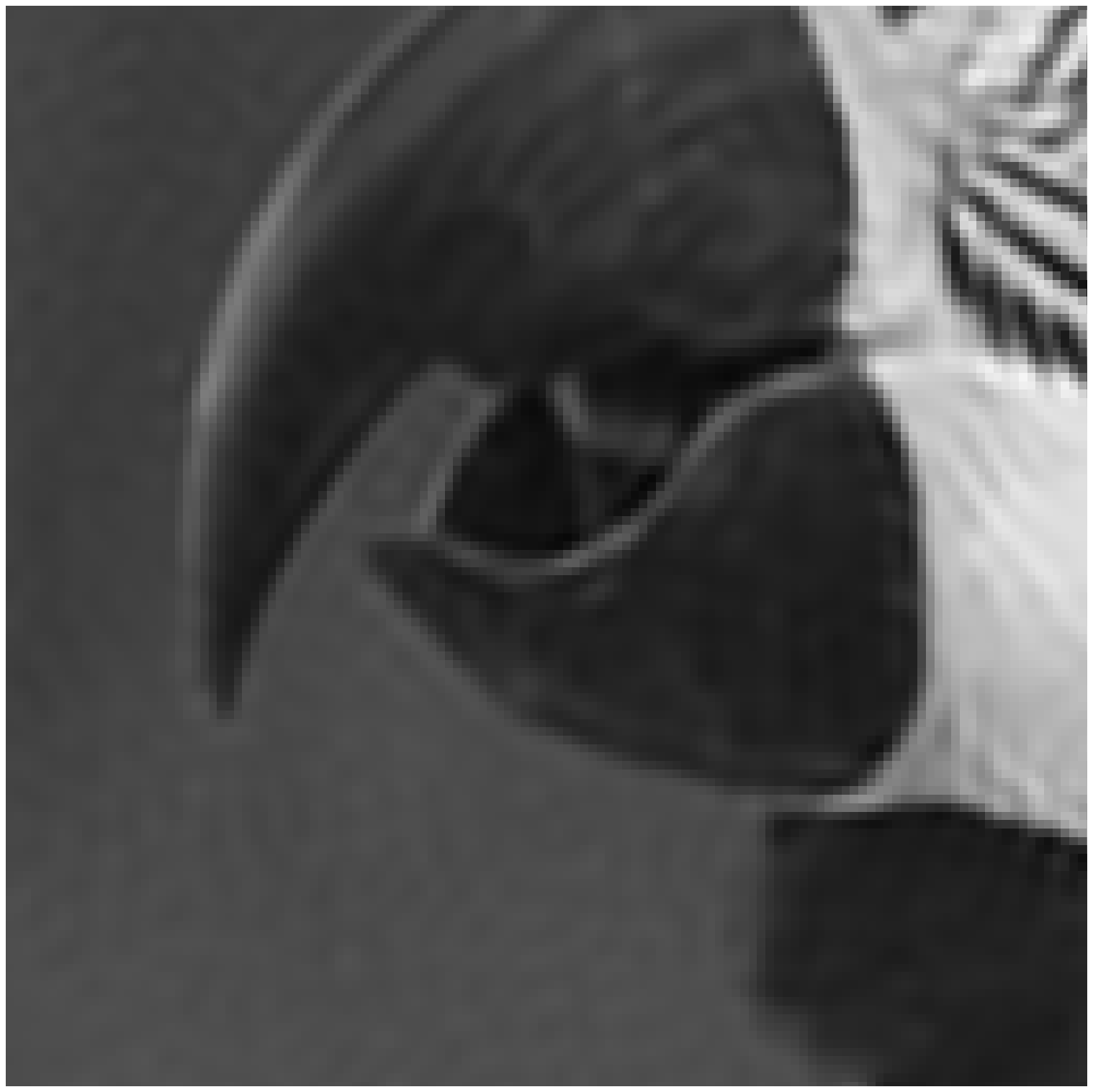}
   \includegraphics[width=0.32\textwidth]{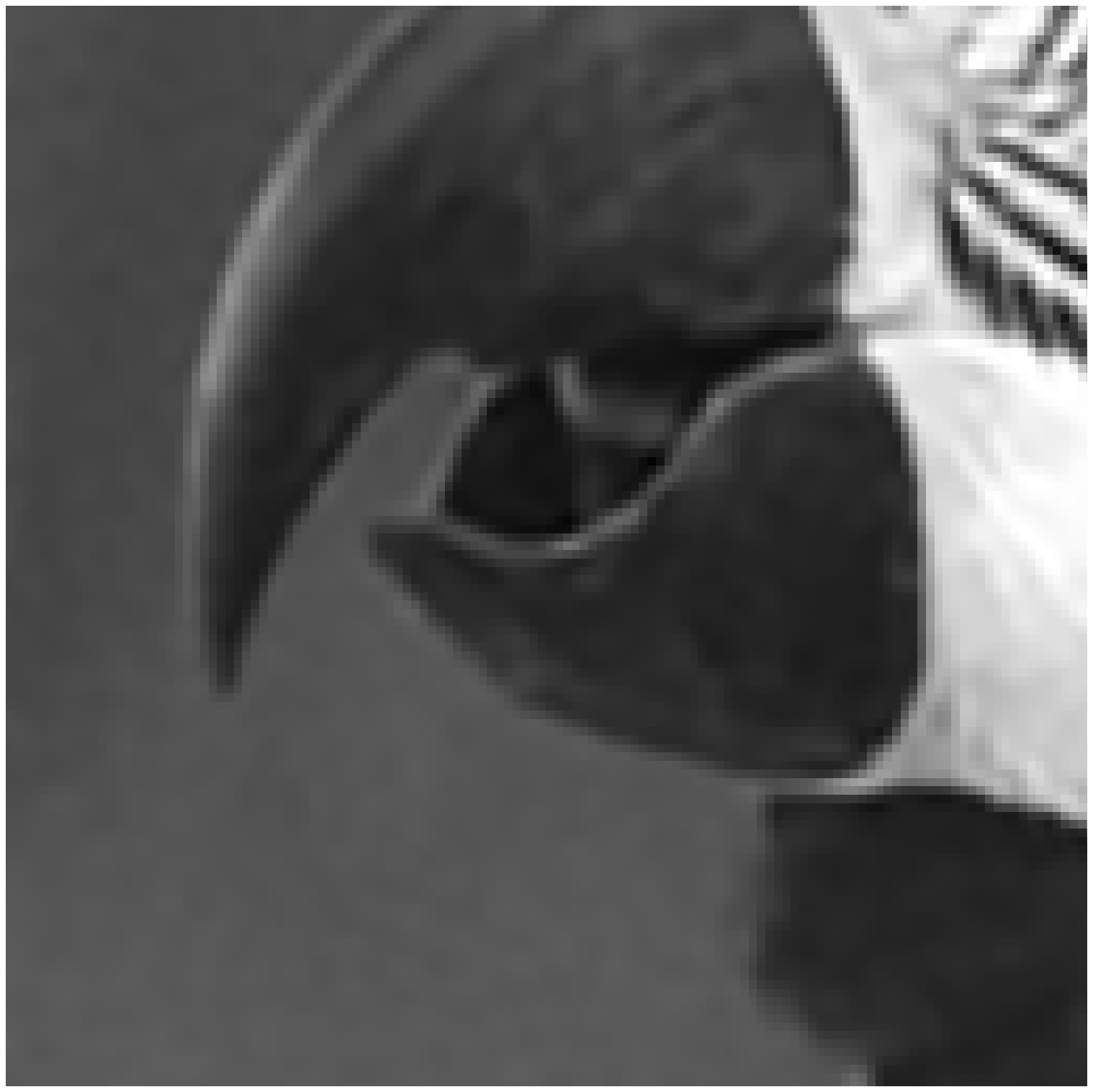}
   \includegraphics[width=0.32\textwidth]{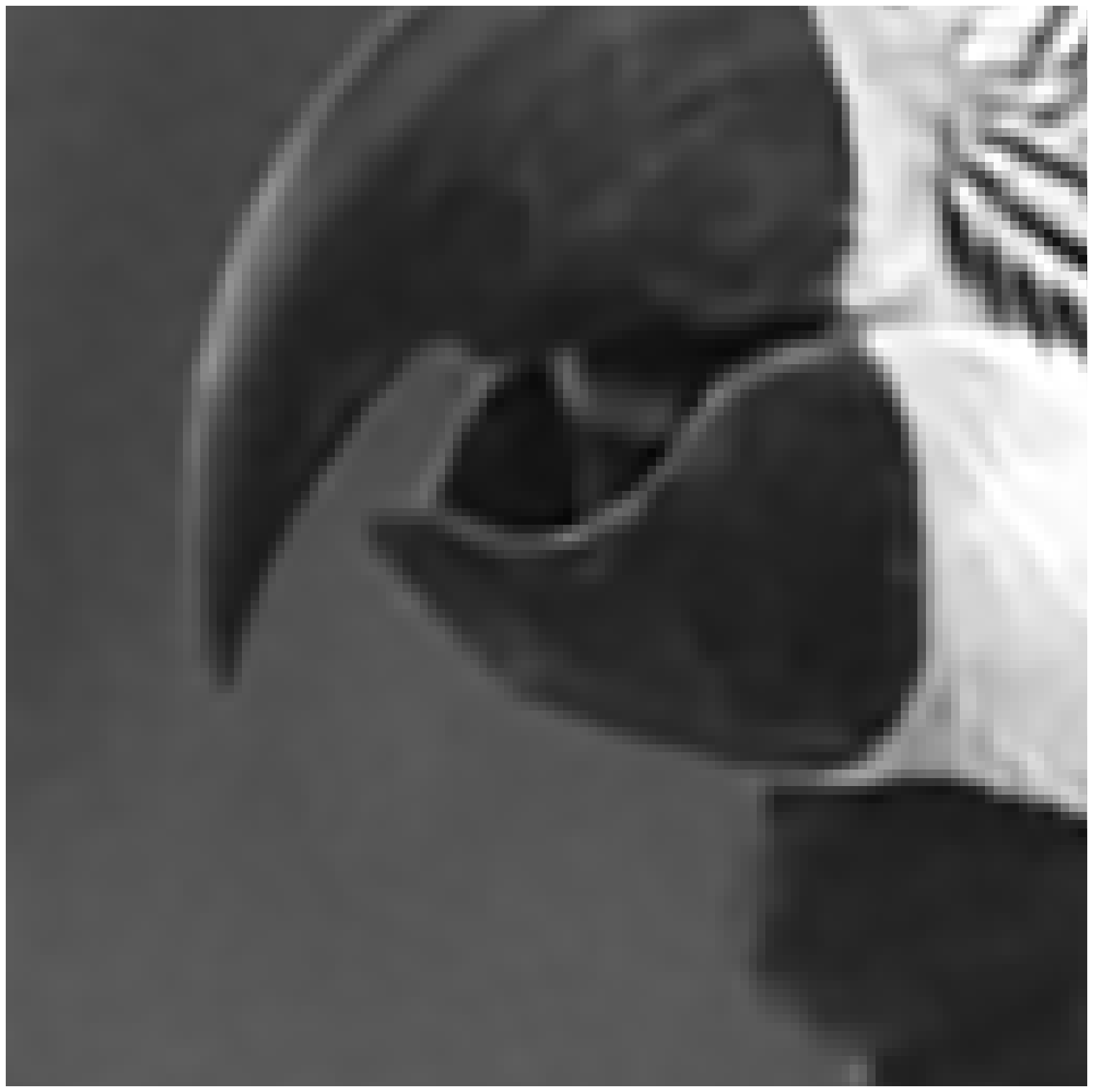} \\
  \caption{\small  The zoom in visual comparisons corresponding to Fig \ref{fig:parrots comparison}. From left to right and top to bottom: IDD-BM3D, Proposed IDD-BM3D+SDSR(L=1), IDD-BM3D+SDSR(L=4), CSR, Proposed CSR+SDSR(L=1), CSR+SDSR(L=4), GSR, Proposed GSR+SDSR(L=1), GSR+SDSR(L=4).  }
\label{fig:parrots zoomin}
\end{figure*}

\begin{figure}[h]
  \centering
   \includegraphics[width=0.11\textwidth]{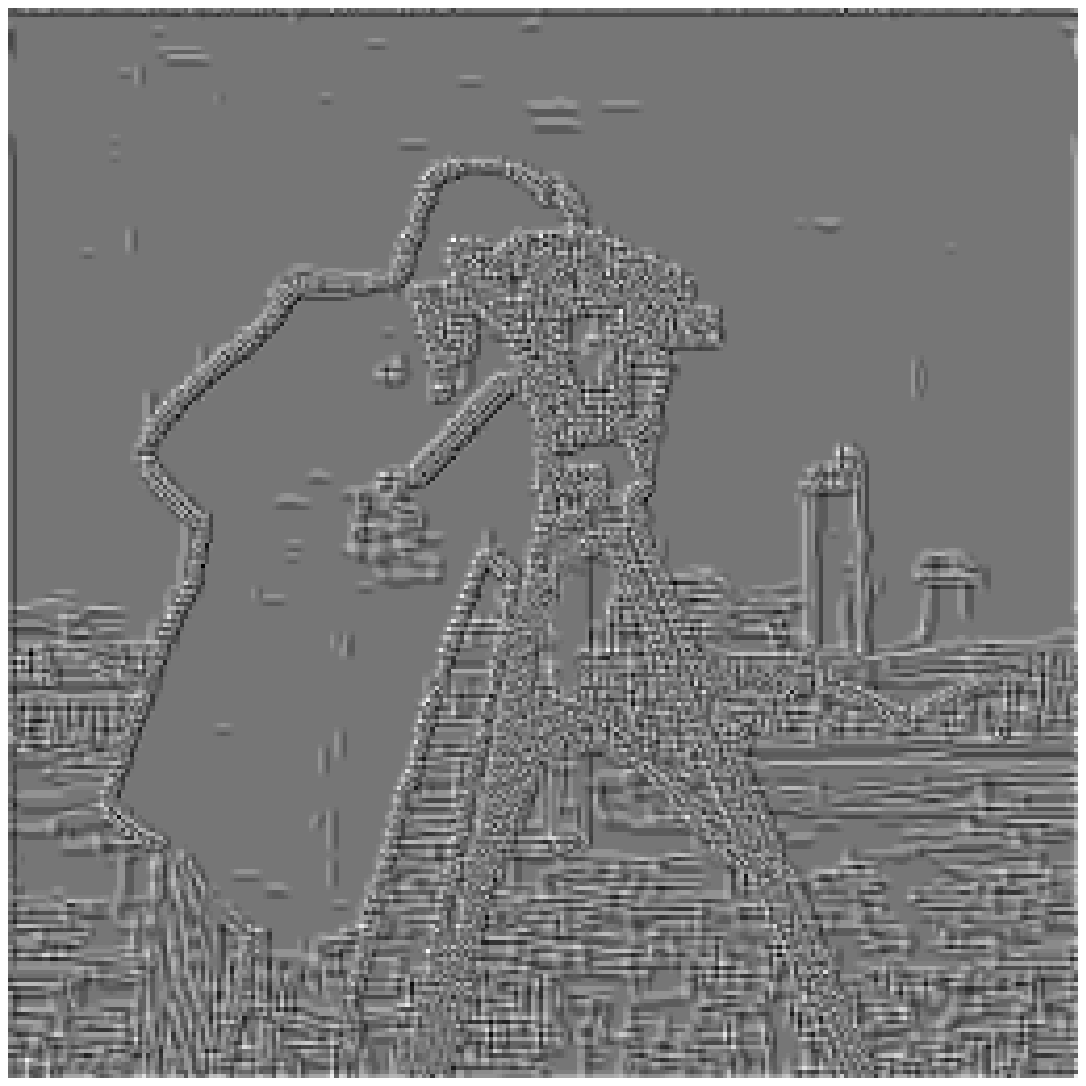}
   \includegraphics[width=0.11\textwidth]{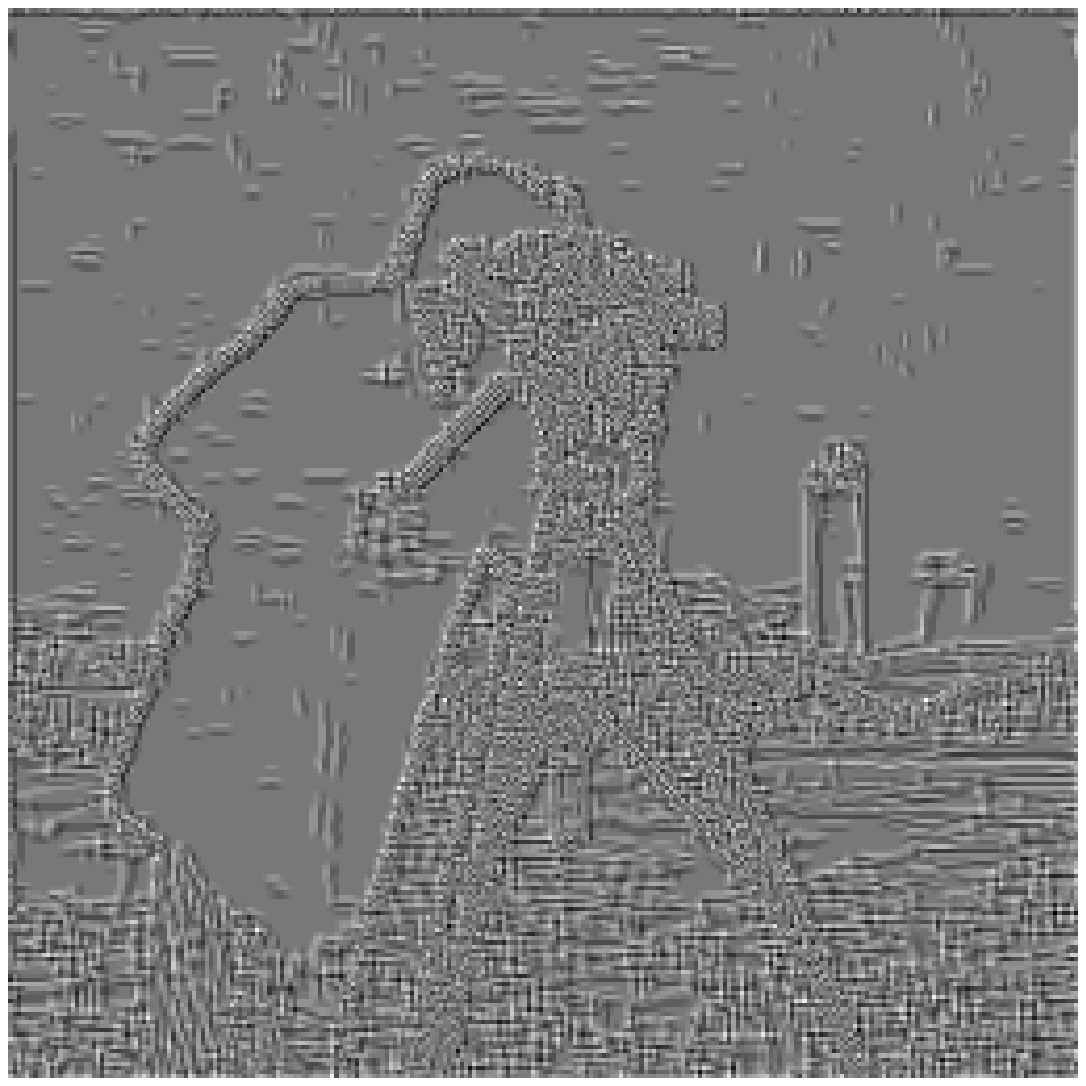}
   \includegraphics[width=0.11\textwidth]{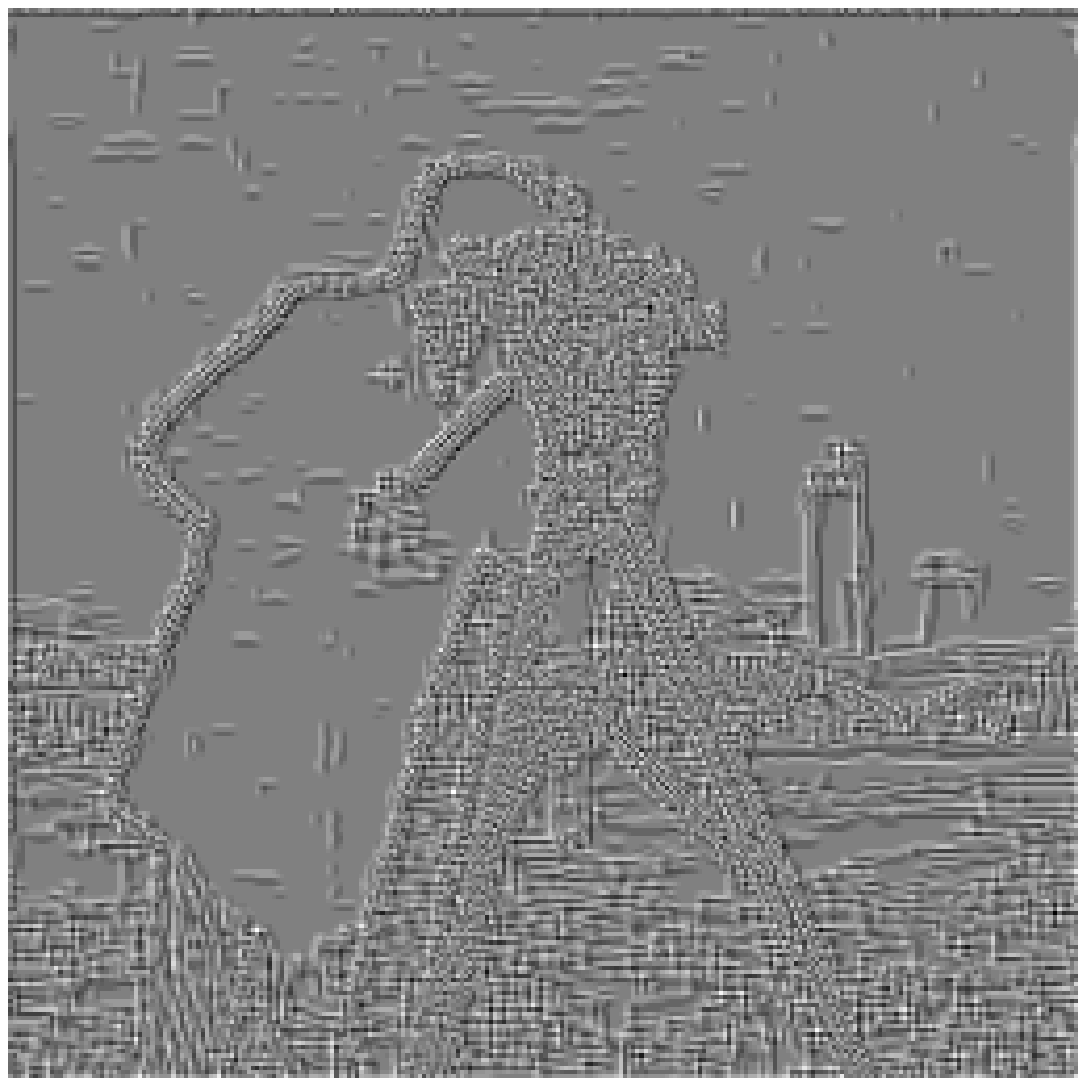}
   \includegraphics[width=0.11\textwidth]{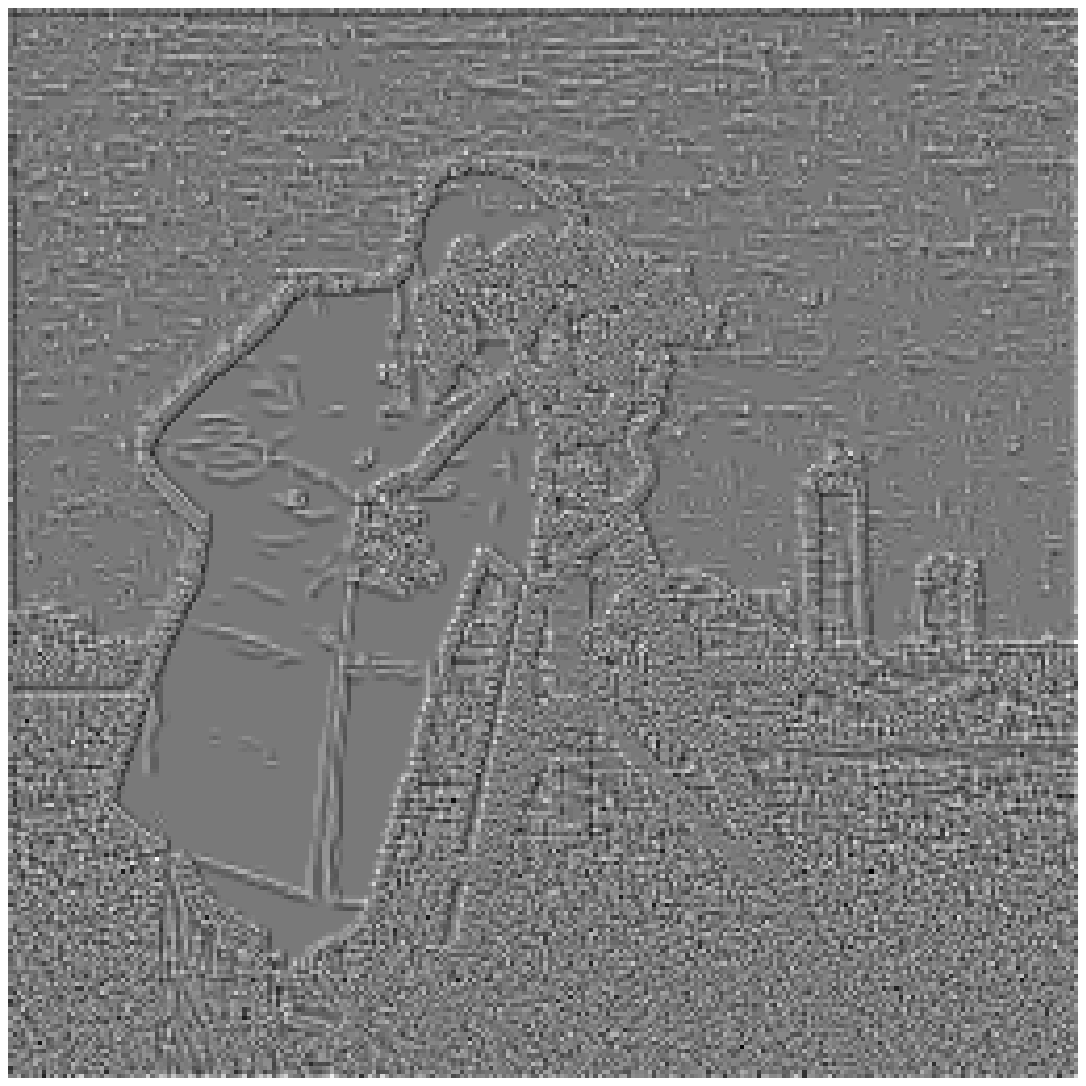}  \\
  \includegraphics[width=0.11\textwidth]{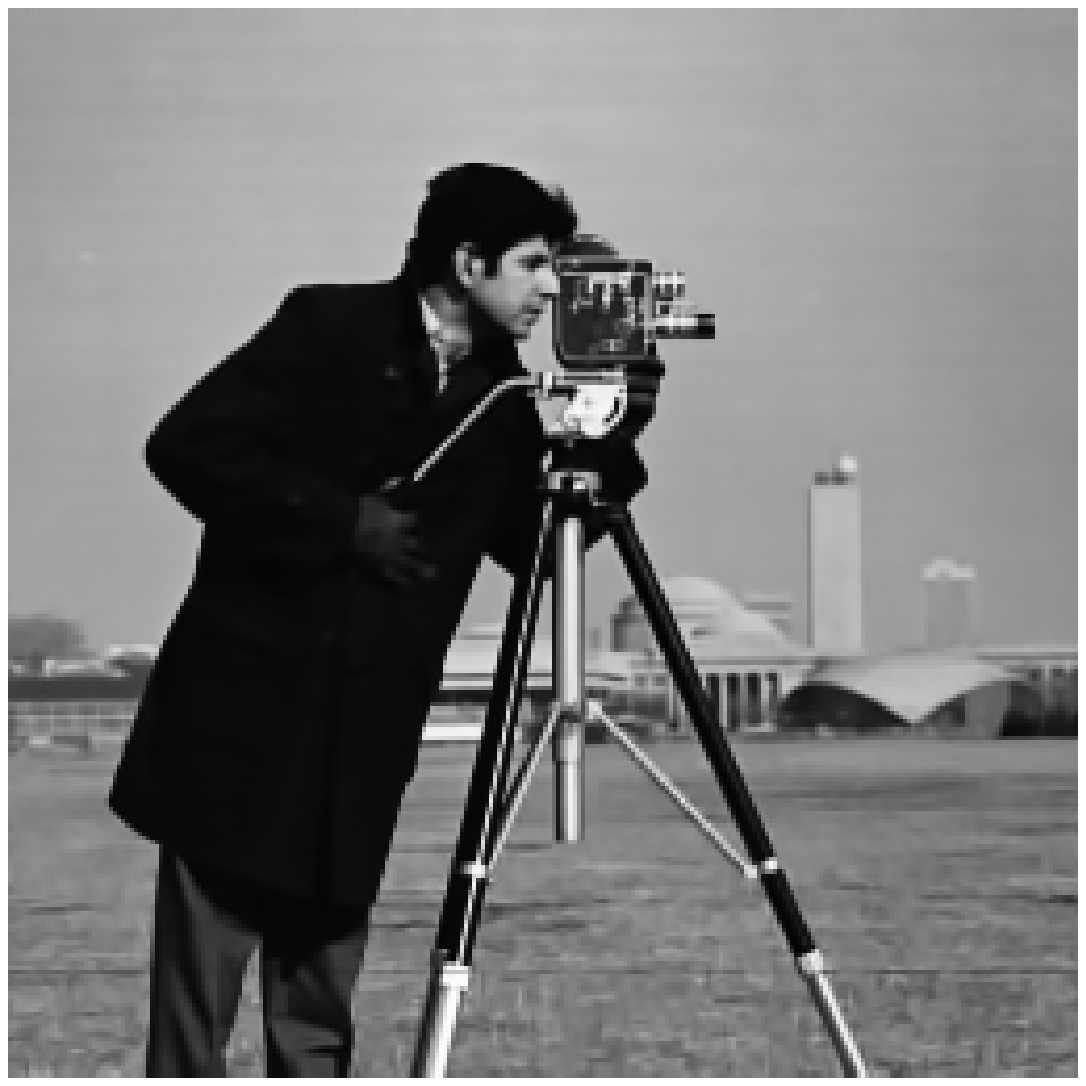}
   \includegraphics[width=0.11\textwidth]{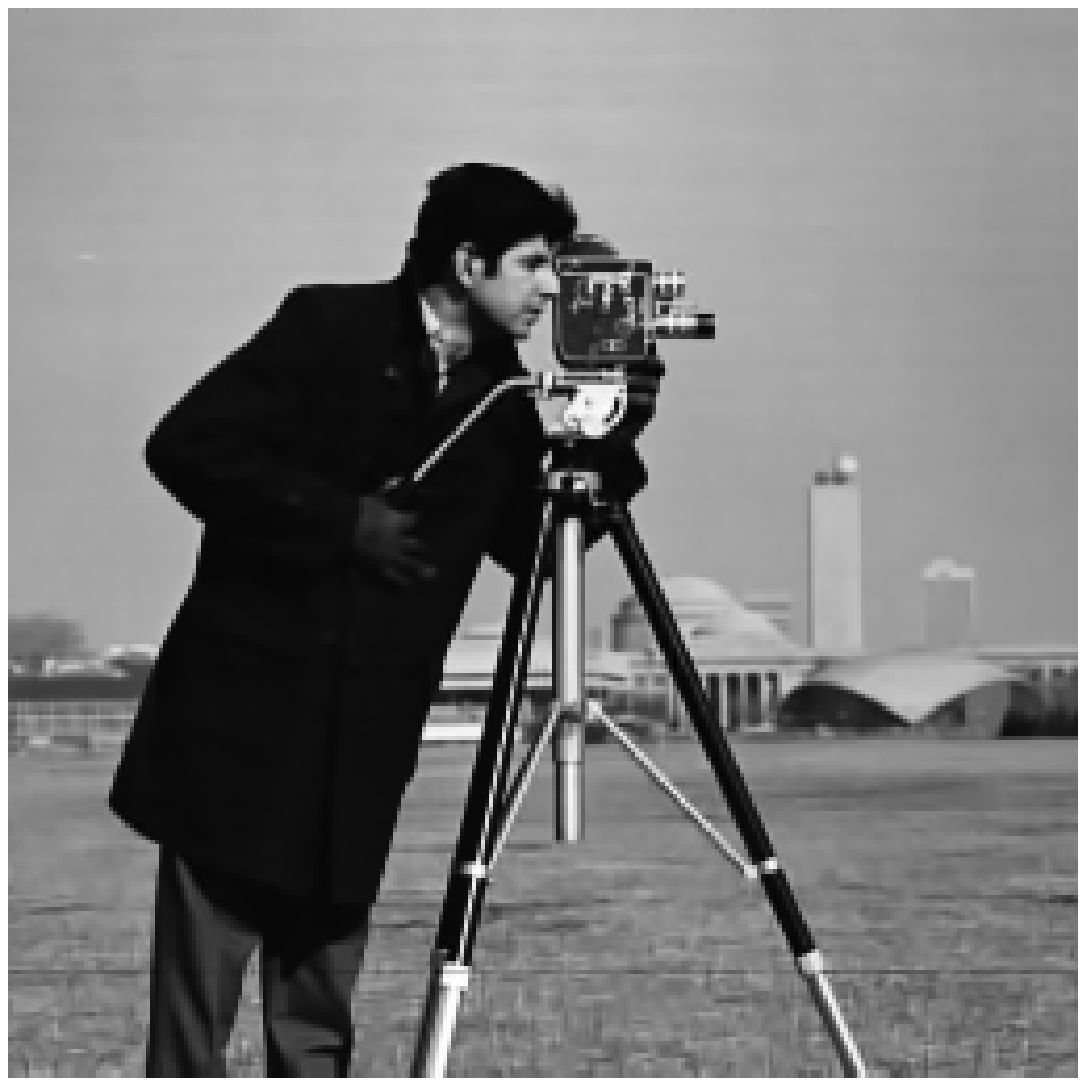}
   \includegraphics[width=0.11\textwidth]{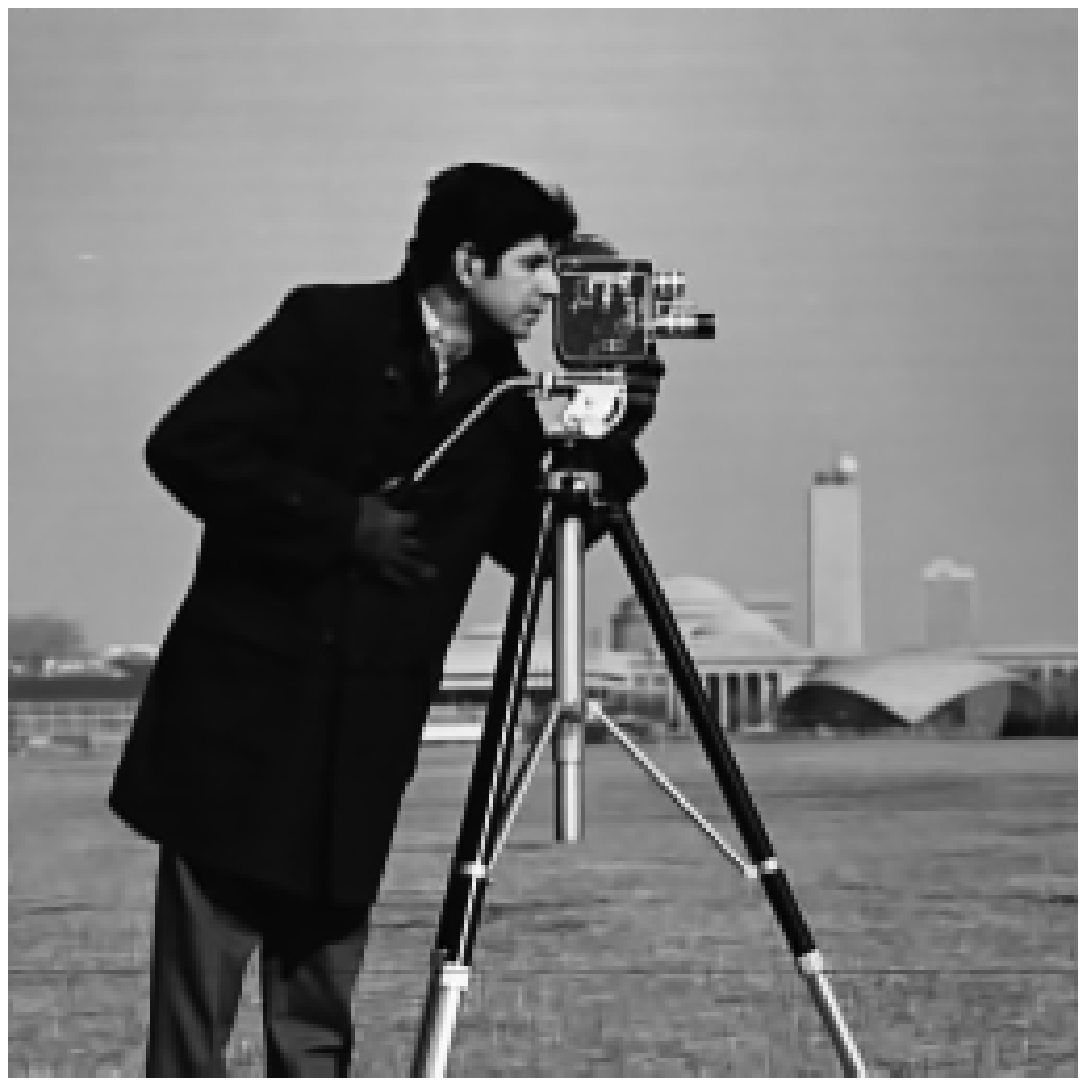}
   \includegraphics[width=0.11\textwidth]{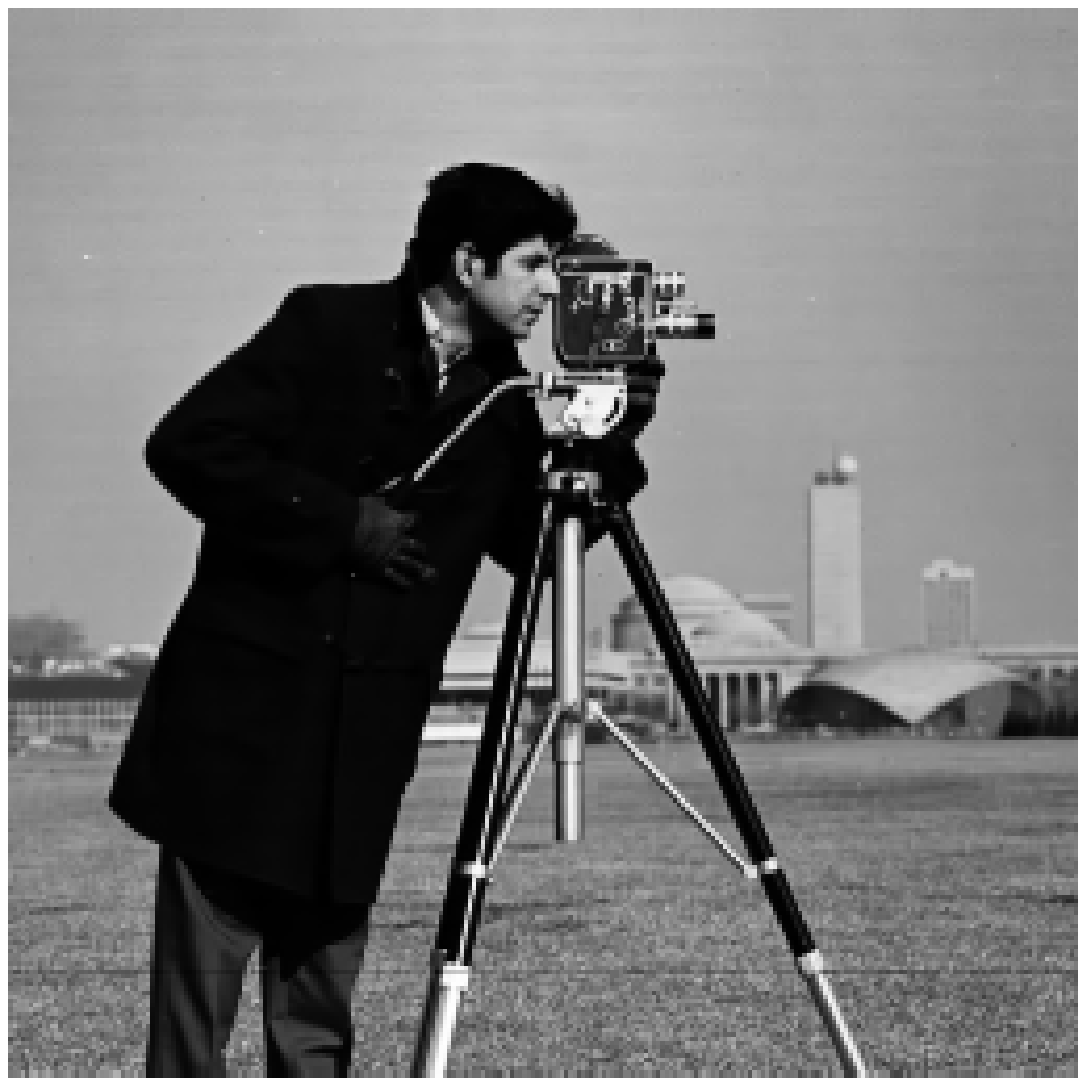}  \\
   \includegraphics[width=0.11\textwidth]{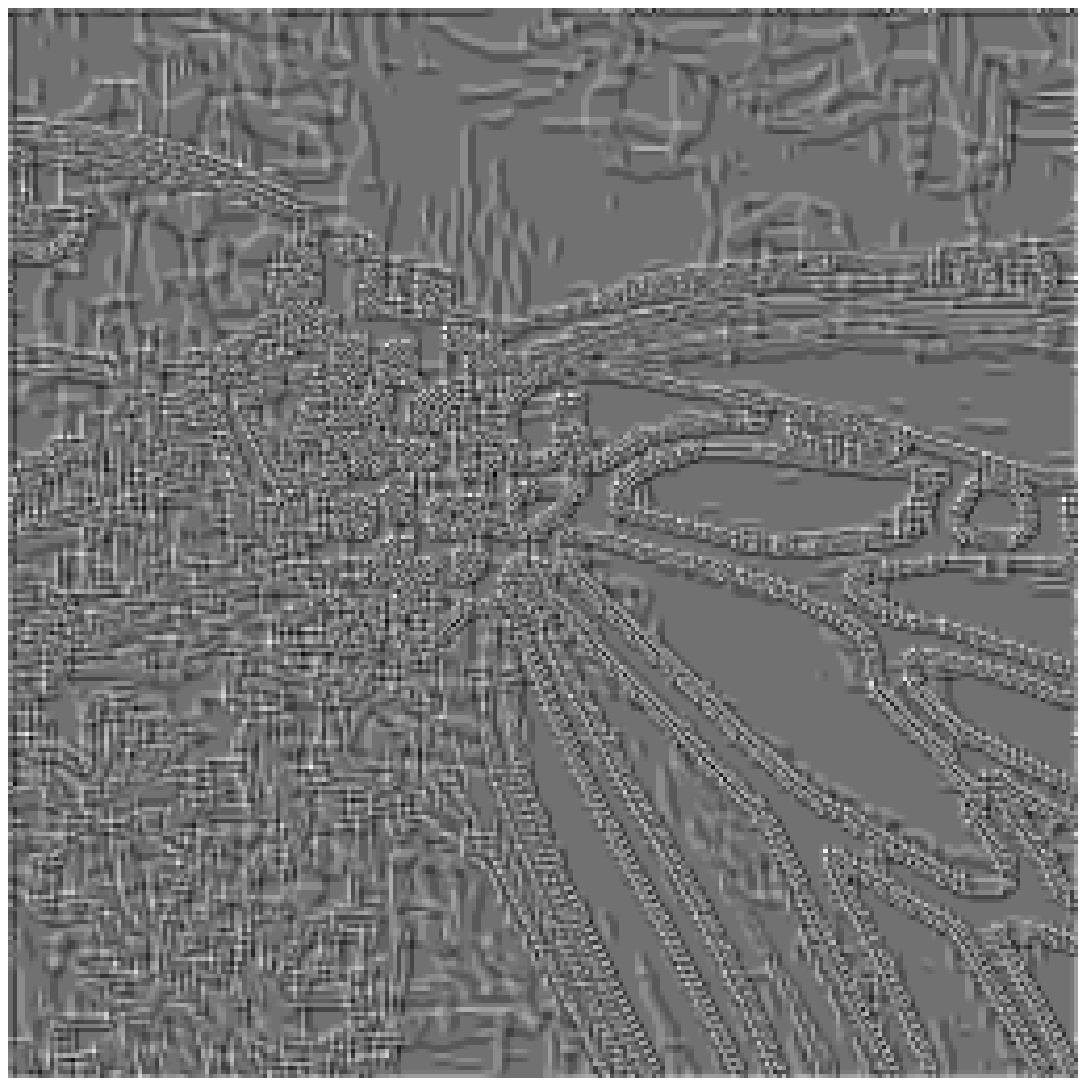}
   \includegraphics[width=0.11\textwidth]{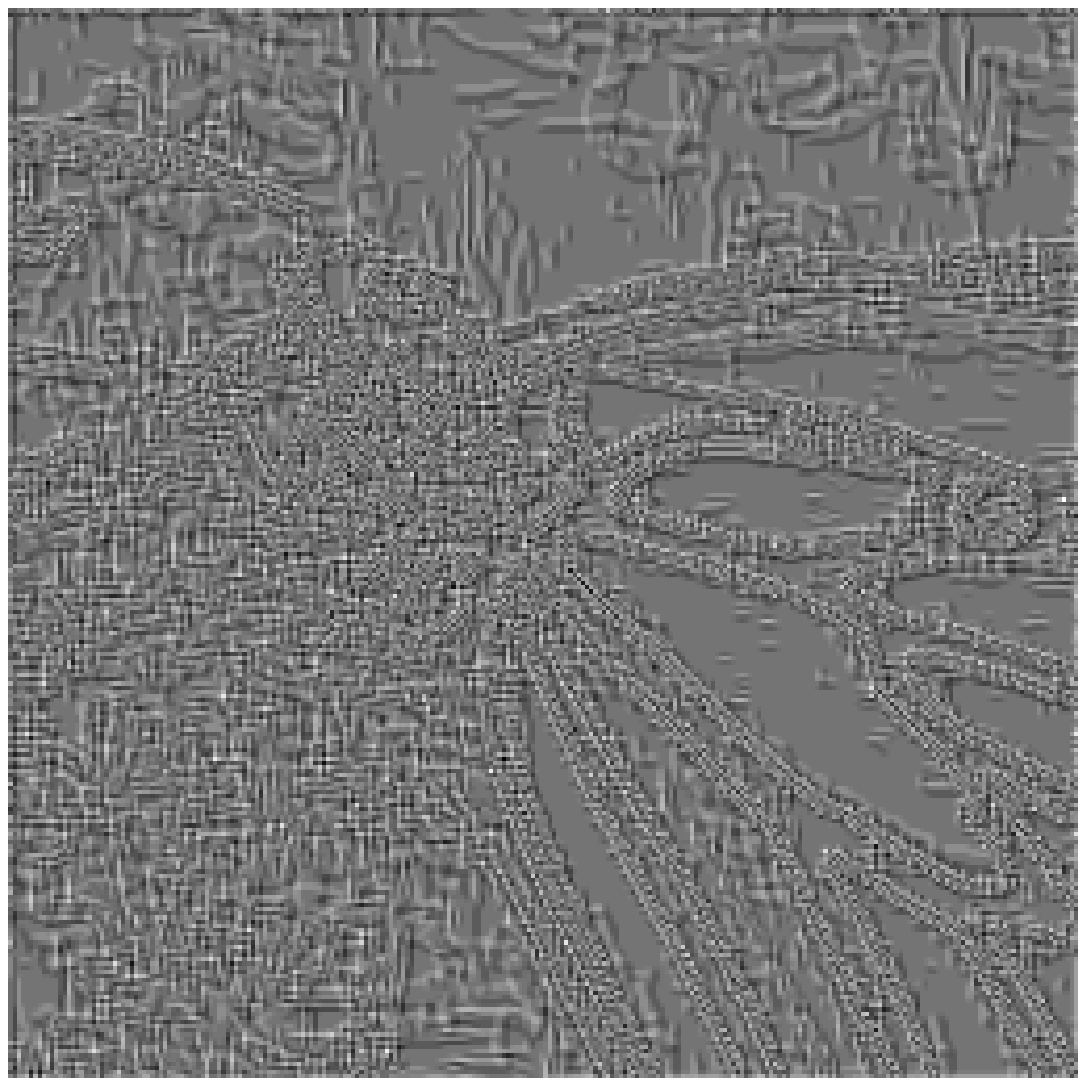}
   \includegraphics[width=0.11\textwidth]{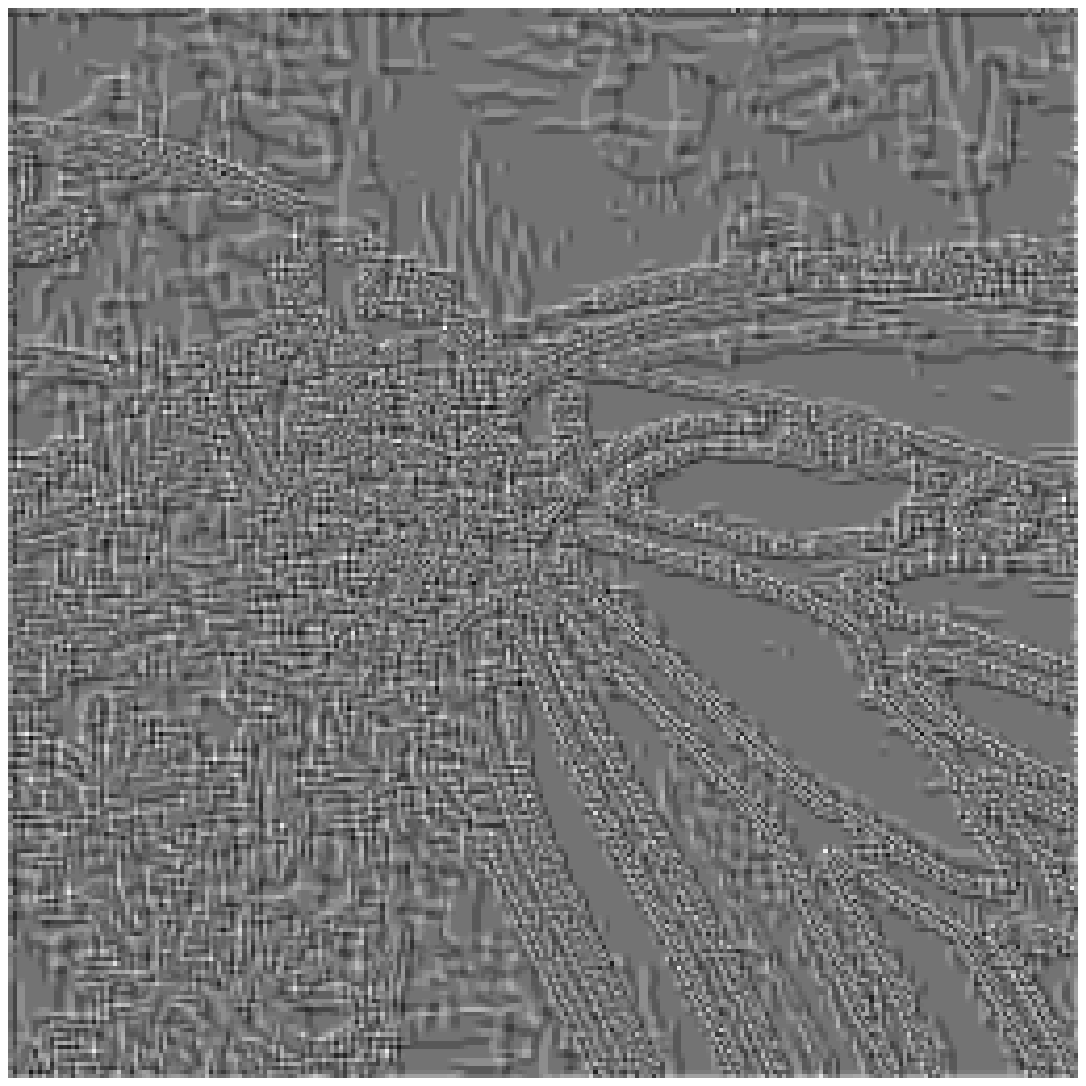}
   \includegraphics[width=0.11\textwidth]{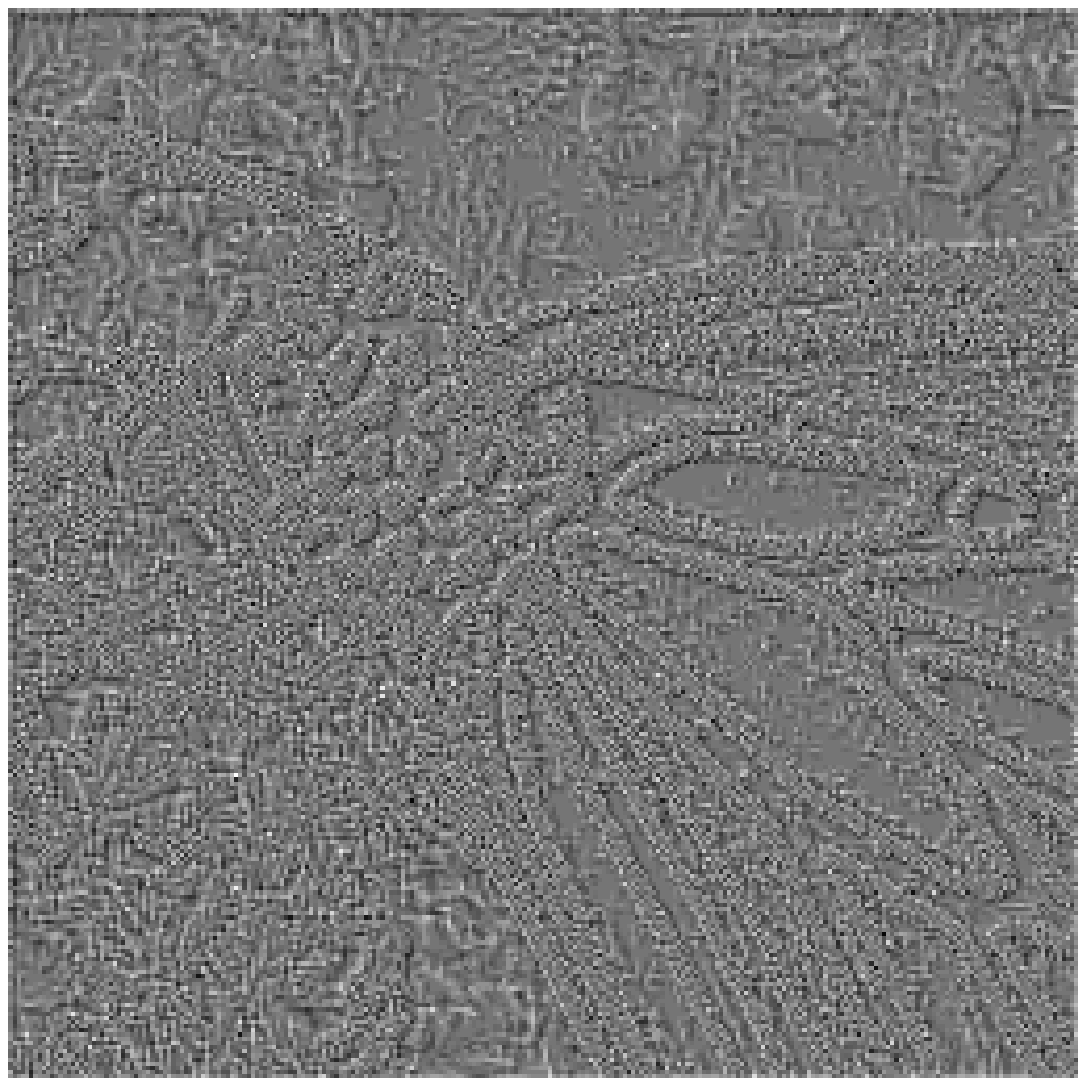}  \\
      \includegraphics[width=0.11\textwidth]{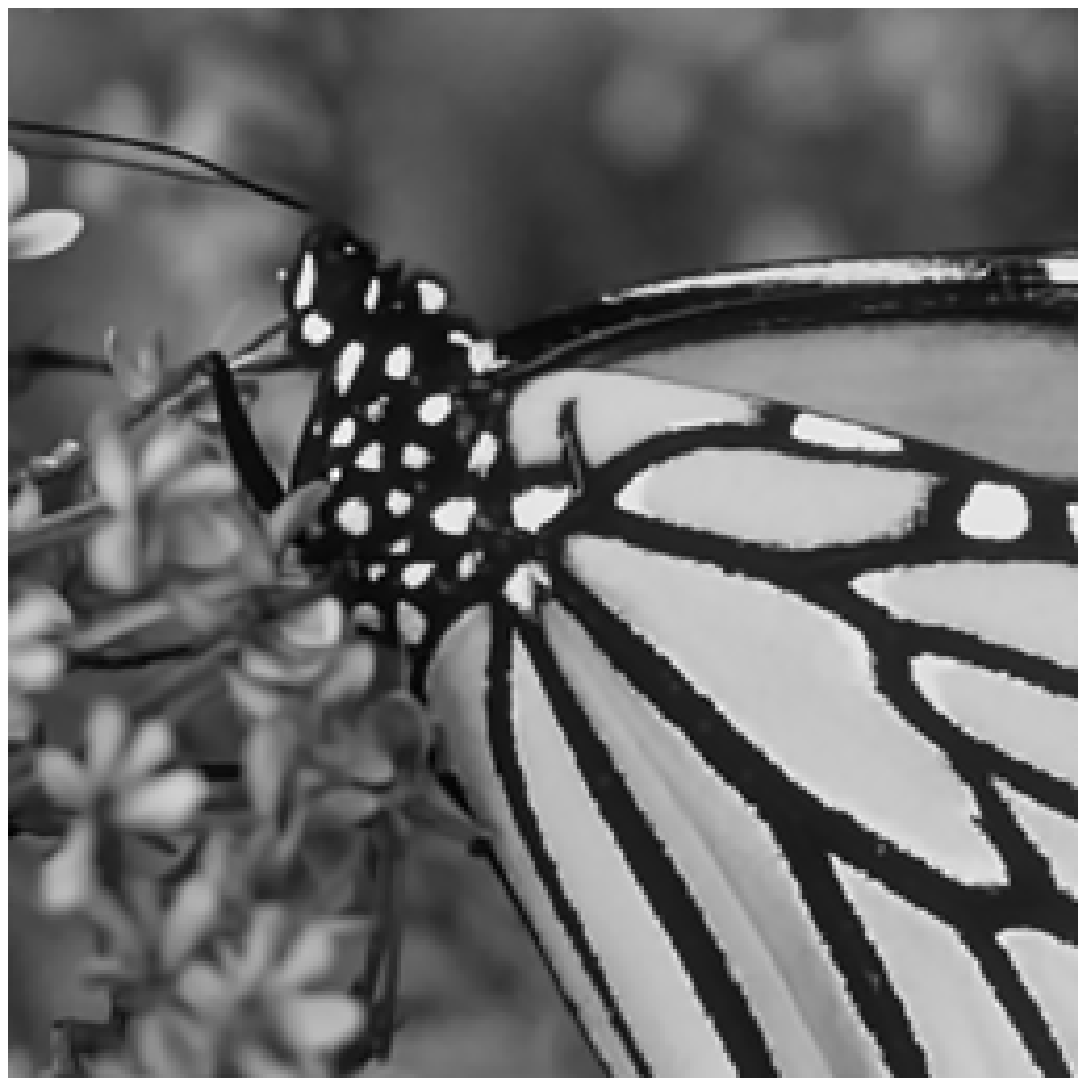}
   \includegraphics[width=0.11\textwidth]{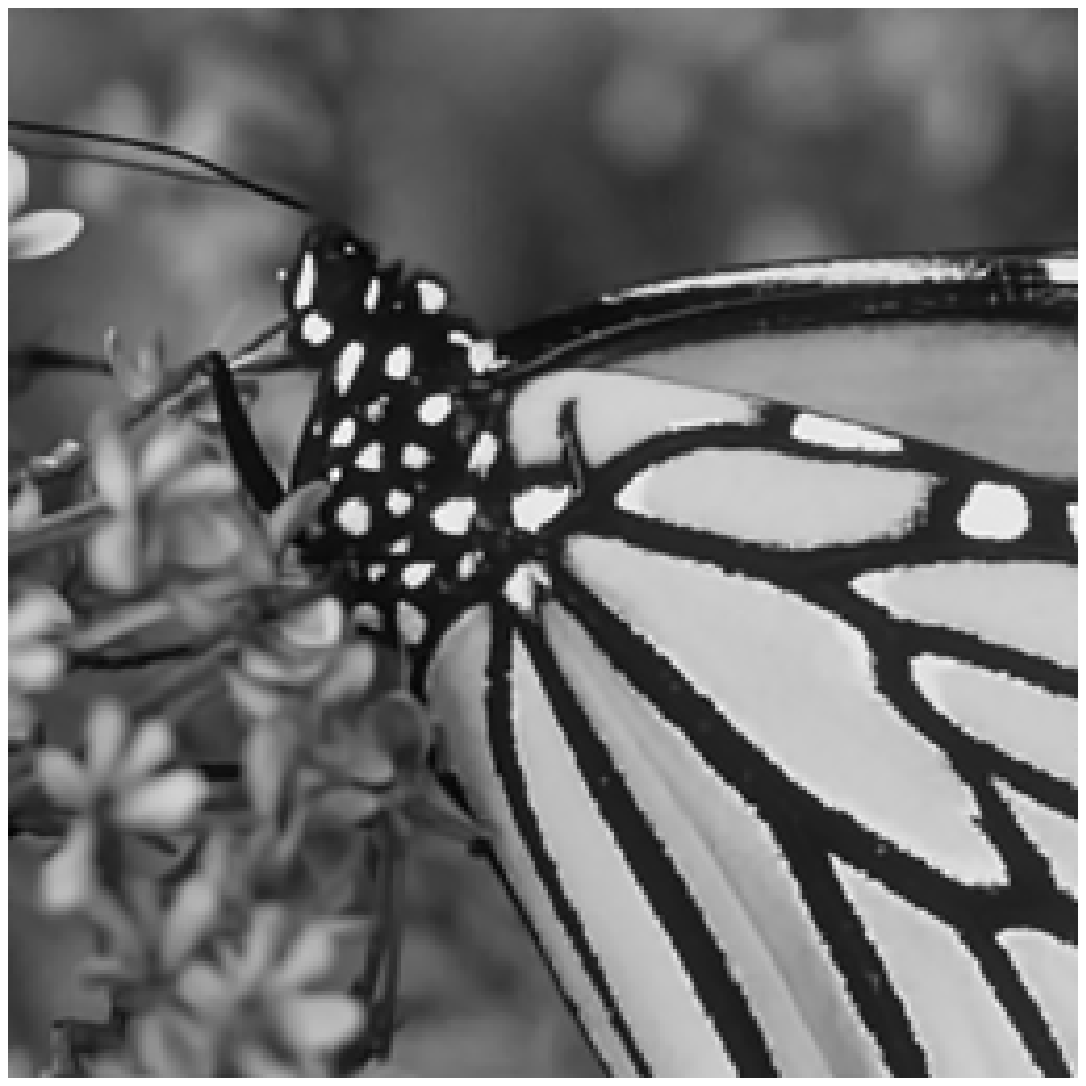}
   \includegraphics[width=0.11\textwidth]{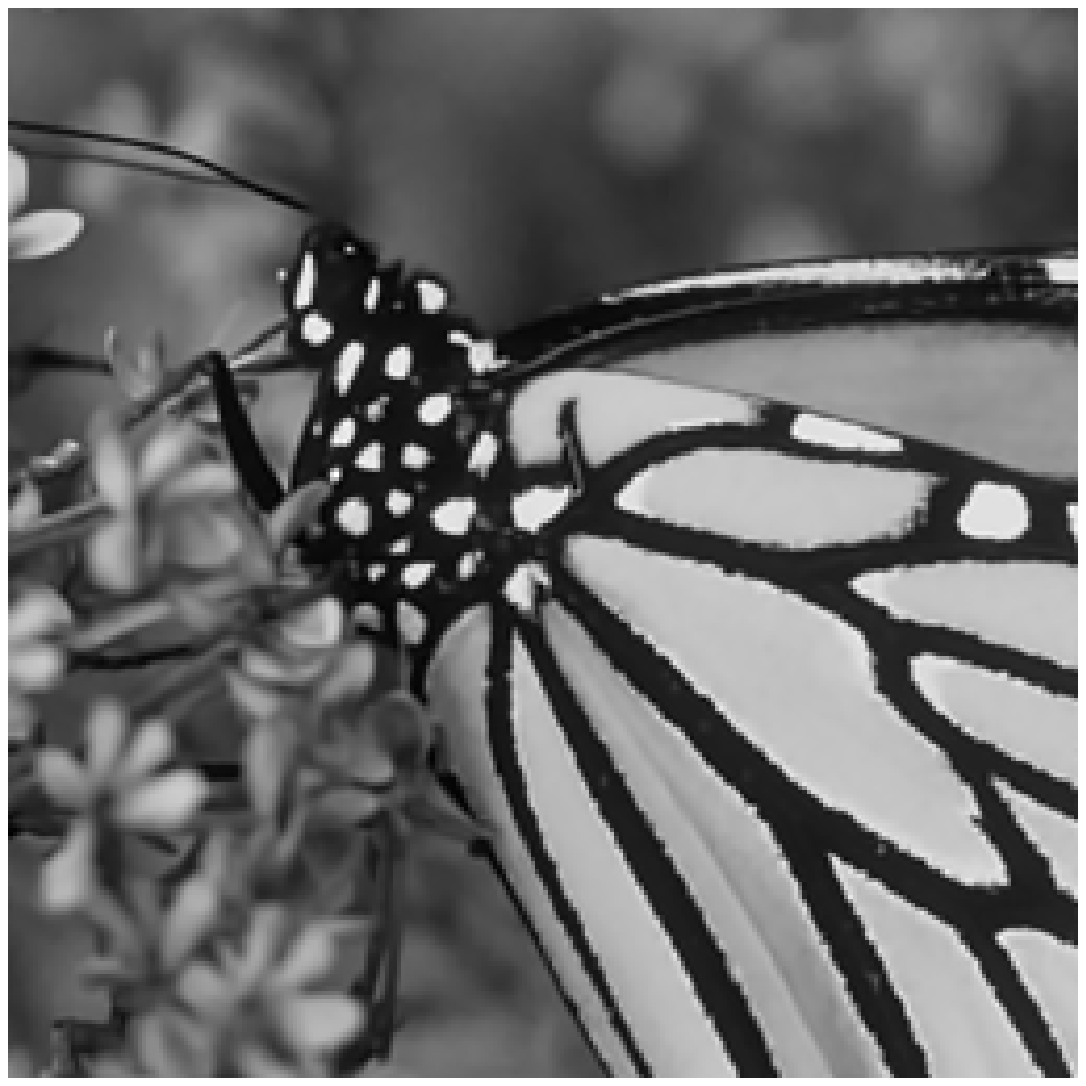}
   \includegraphics[width=0.11\textwidth]{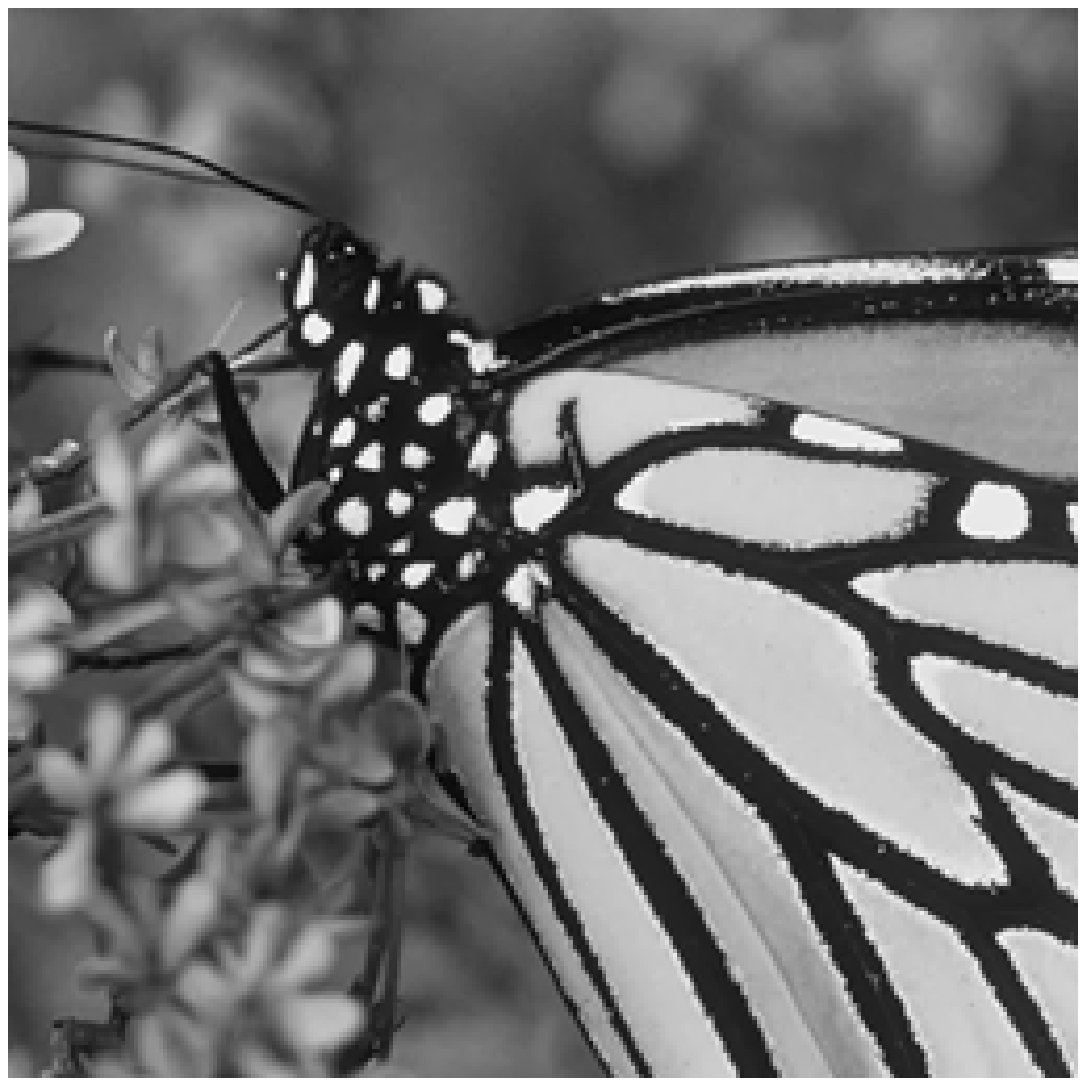}  \\

  \caption{\small The comparison of support maps (the first row and third row, which are obtained by directly inverse wavelet frame transform to the support detection) and back projection results (the second row and fourth row, which are obtained by only reserving the large wavelet frame coefficients to the original true image at the support locations) via different initial methods; IDD-BM3D method (first column), CSR method (second column), GSR method (third column) and ORACLE true imgae (fourth column). Scenario: 3. The decomposition level $L=1$. }
\label{fig:cameraman monrch support detection}
\end{figure}

\begin{figure}[h]
  \centering
   \includegraphics[width=0.11\textwidth]{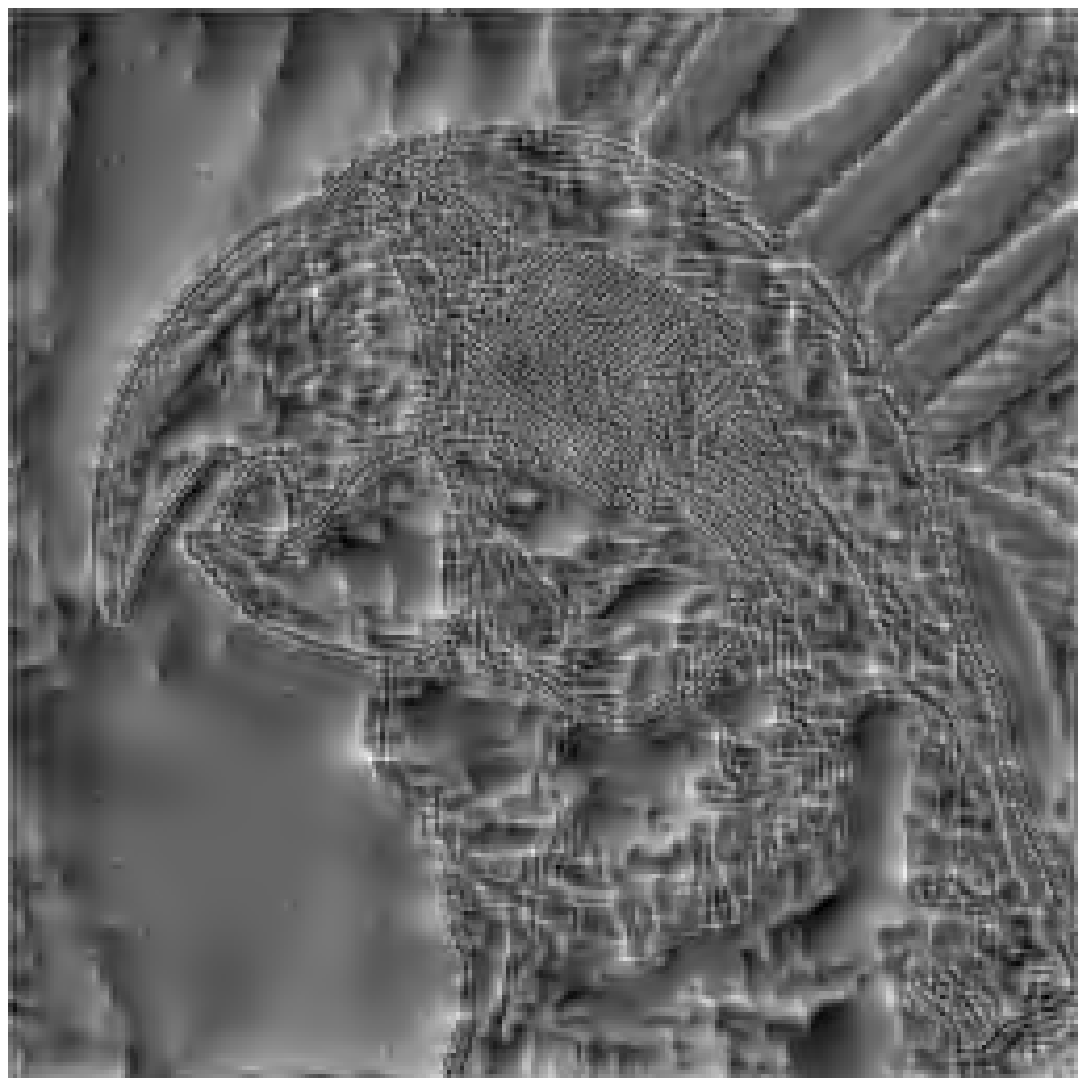}
   \includegraphics[width=0.11\textwidth]{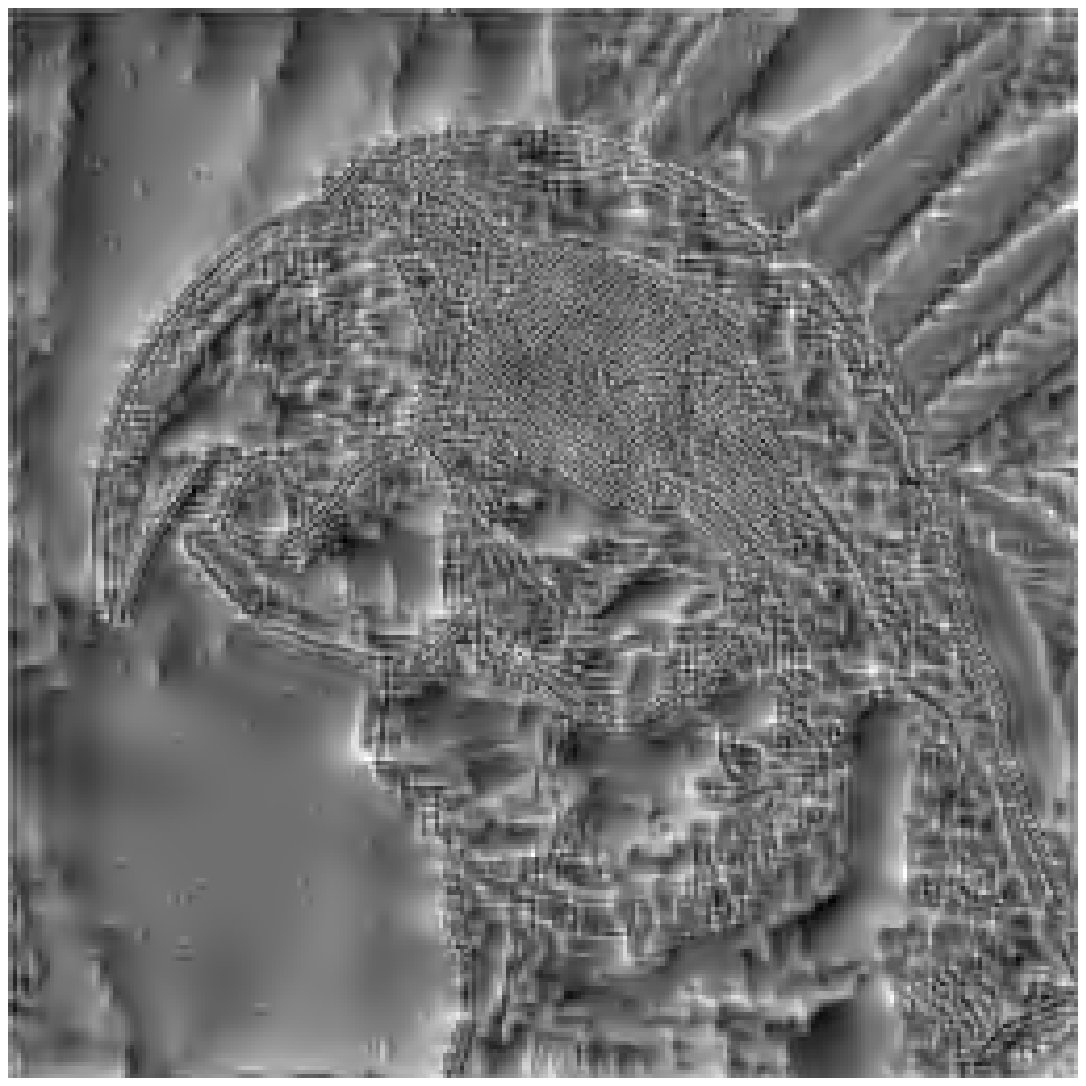}
   \includegraphics[width=0.11\textwidth]{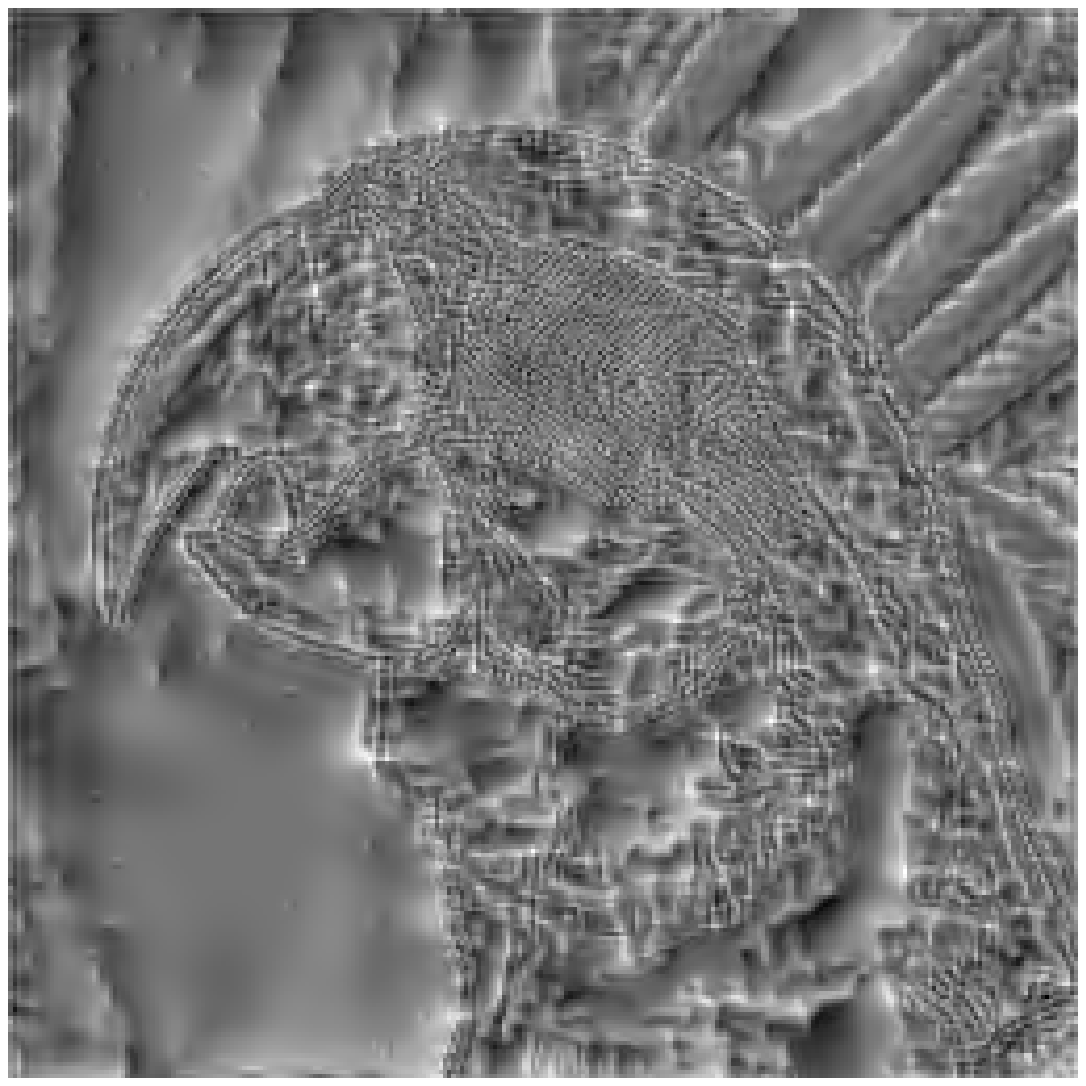}
   \includegraphics[width=0.11\textwidth]{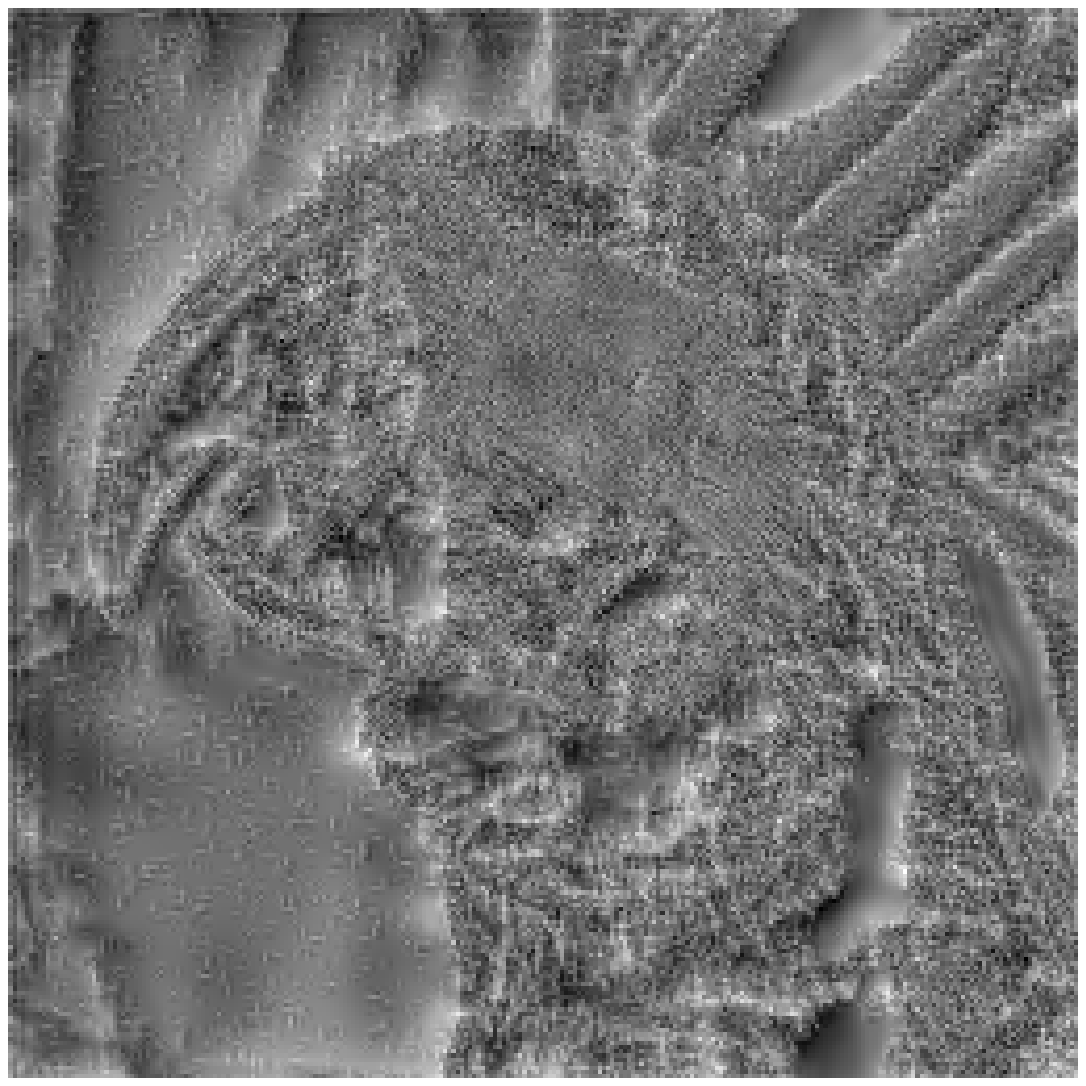}  \\
   \includegraphics[width=0.11\textwidth]{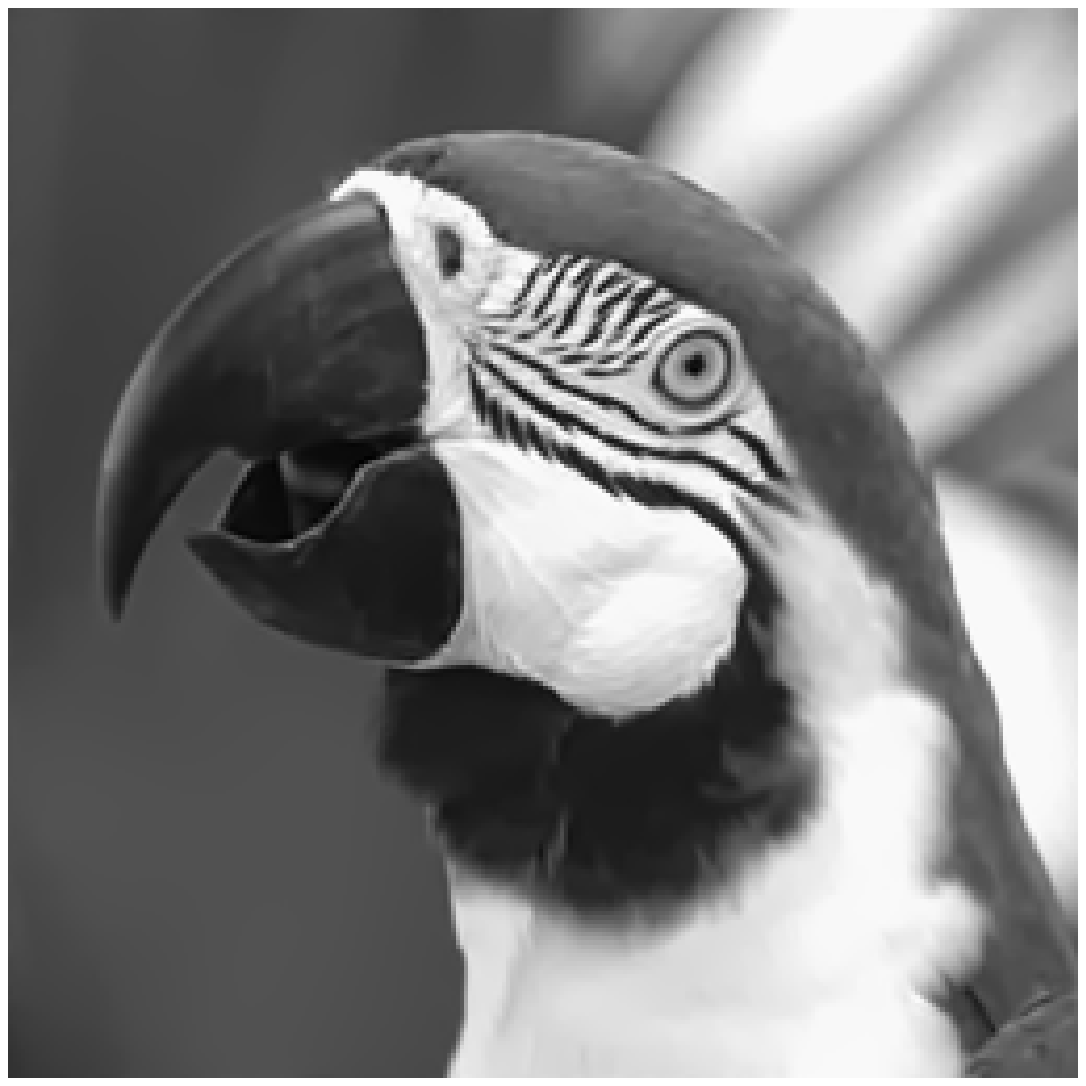}
   \includegraphics[width=0.11\textwidth]{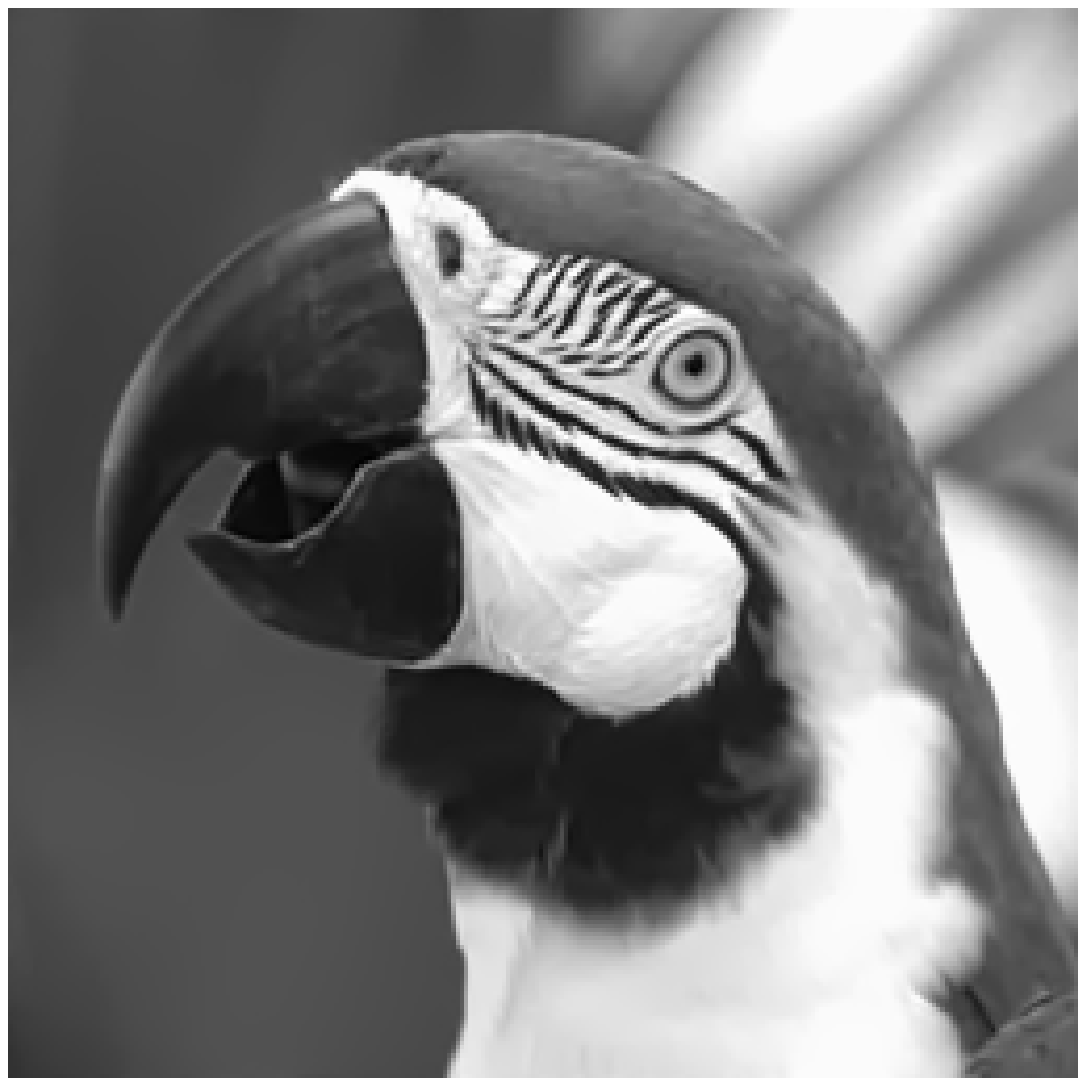}
   \includegraphics[width=0.11\textwidth]{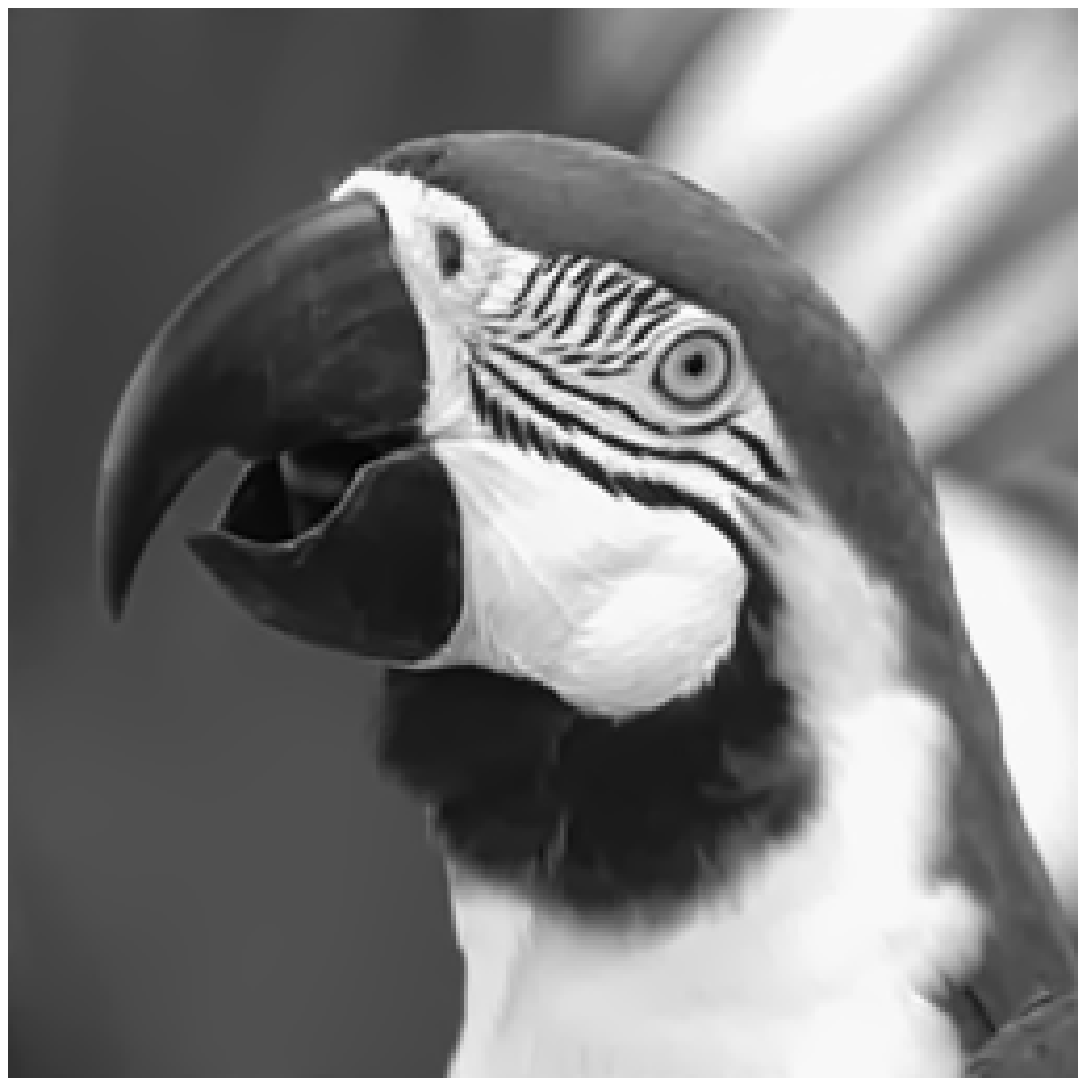}
   \includegraphics[width=0.11\textwidth]{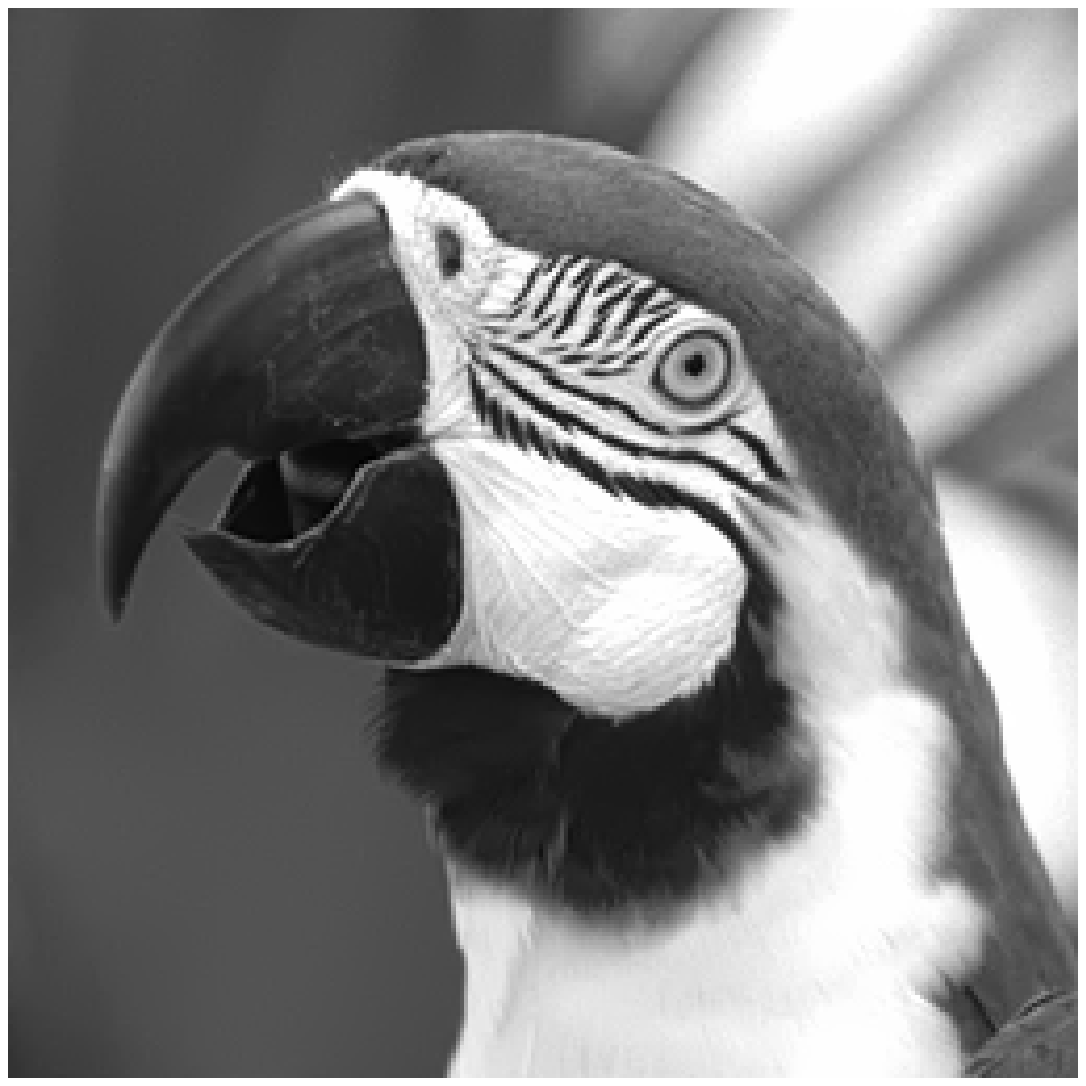}  \\
   \includegraphics[width=0.11\textwidth]{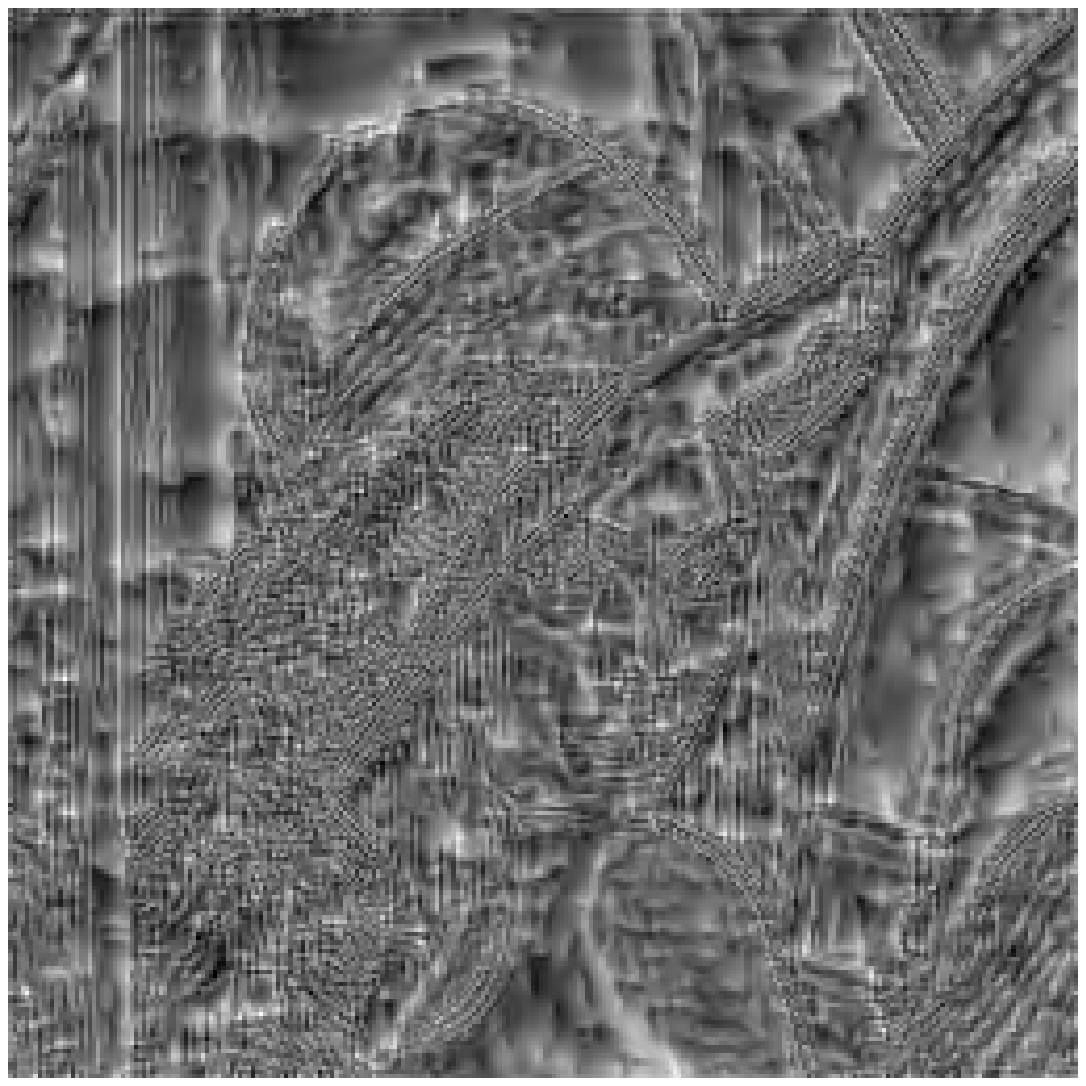}
   \includegraphics[width=0.11\textwidth]{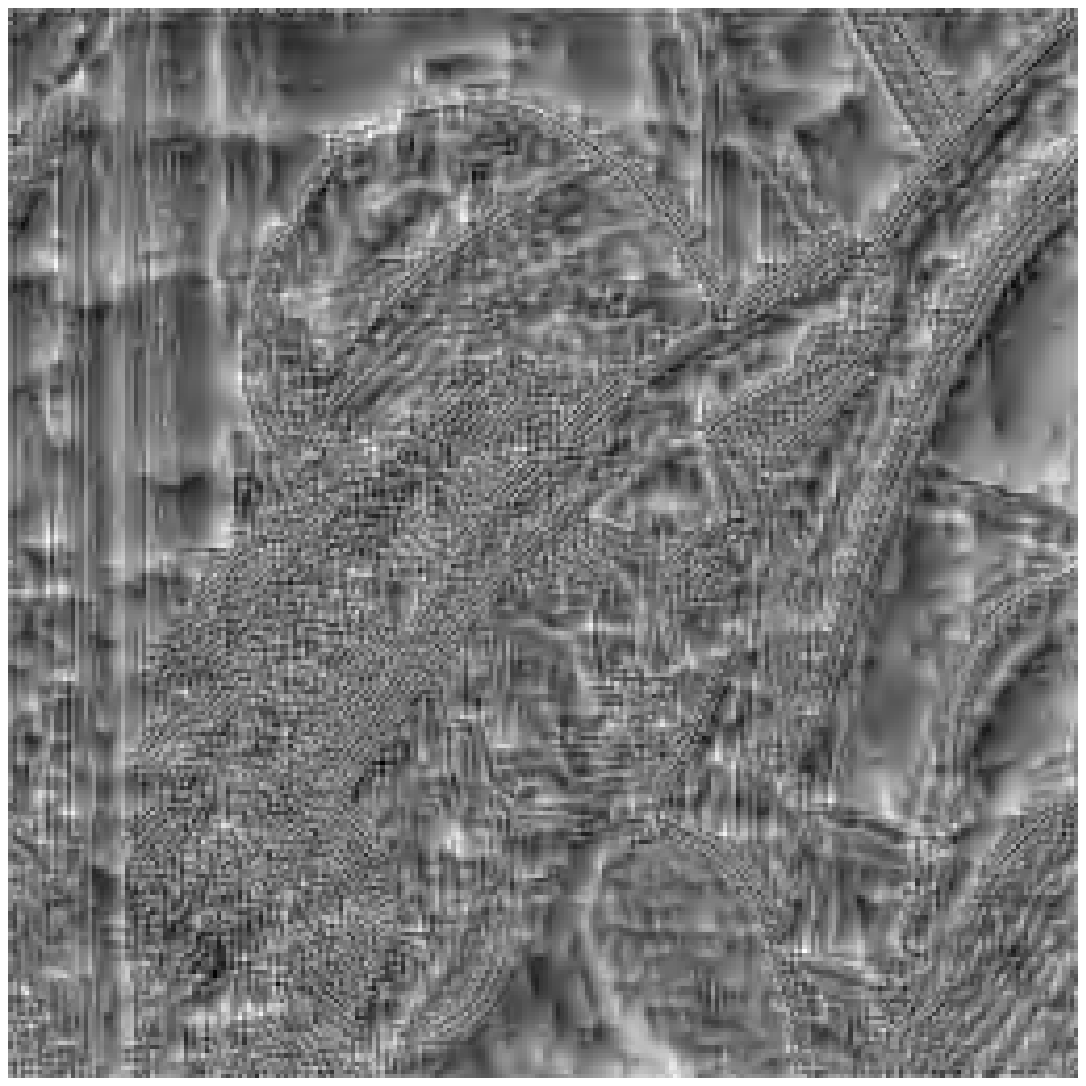}
   \includegraphics[width=0.11\textwidth]{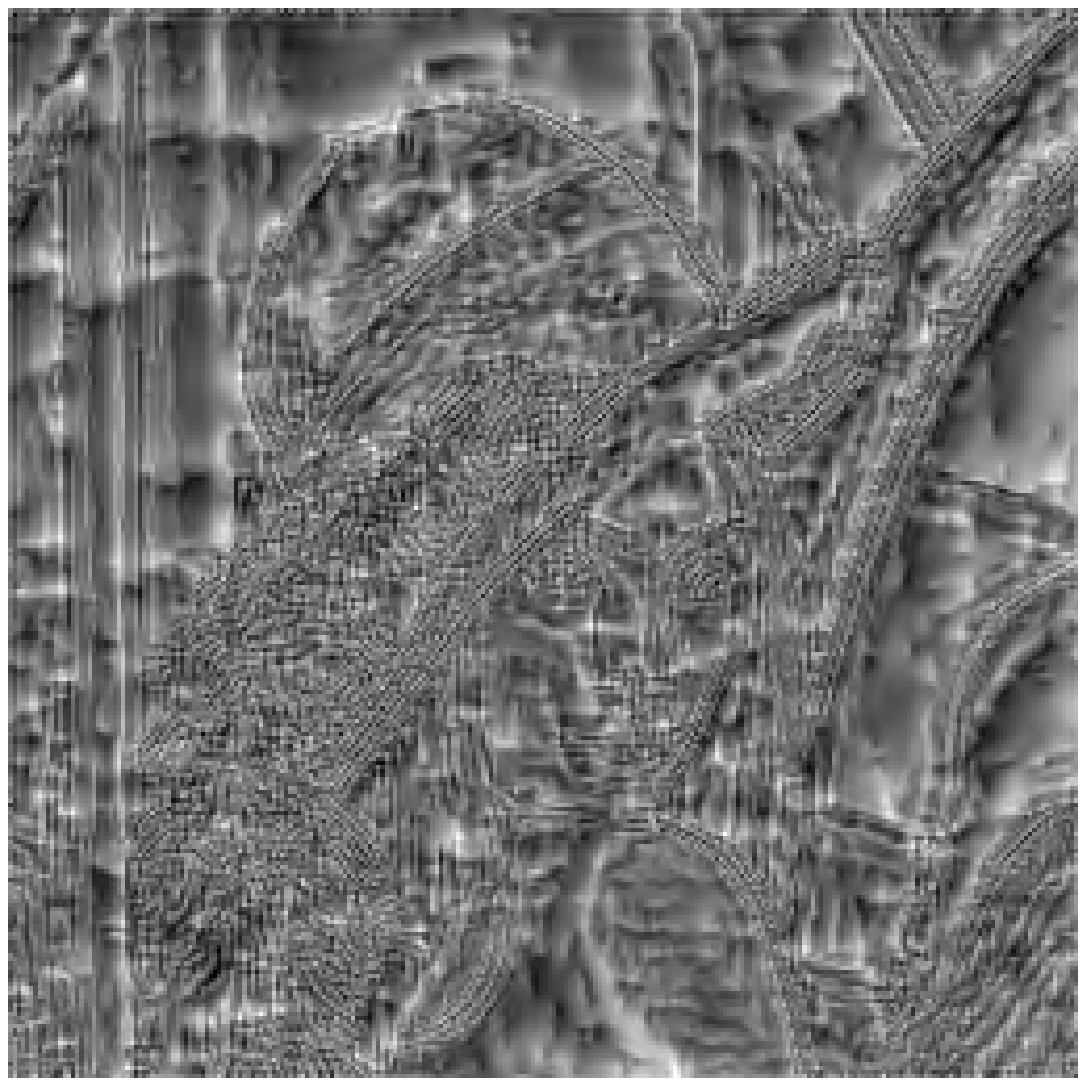}
   \includegraphics[width=0.11\textwidth]{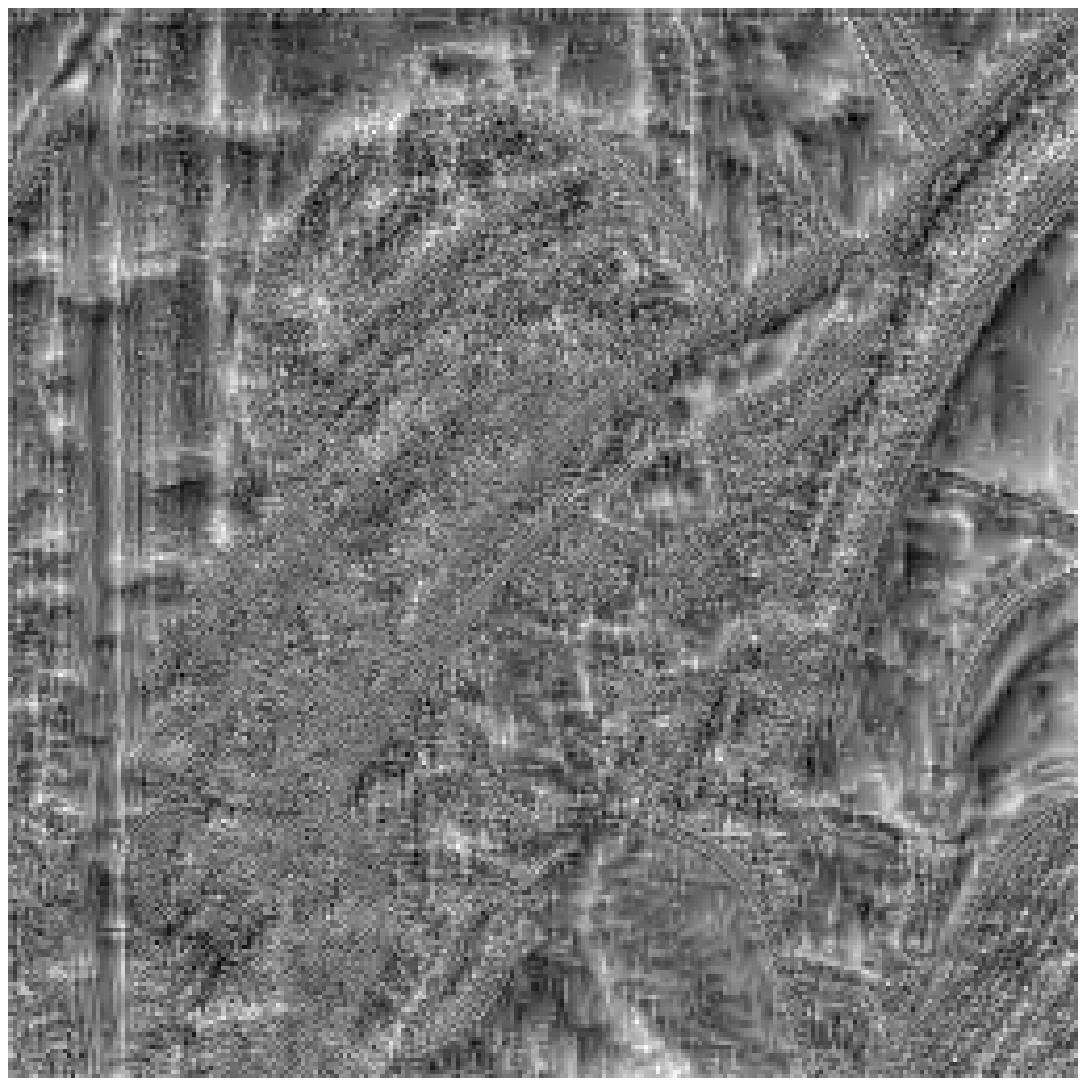}  \\
   \includegraphics[width=0.11\textwidth]{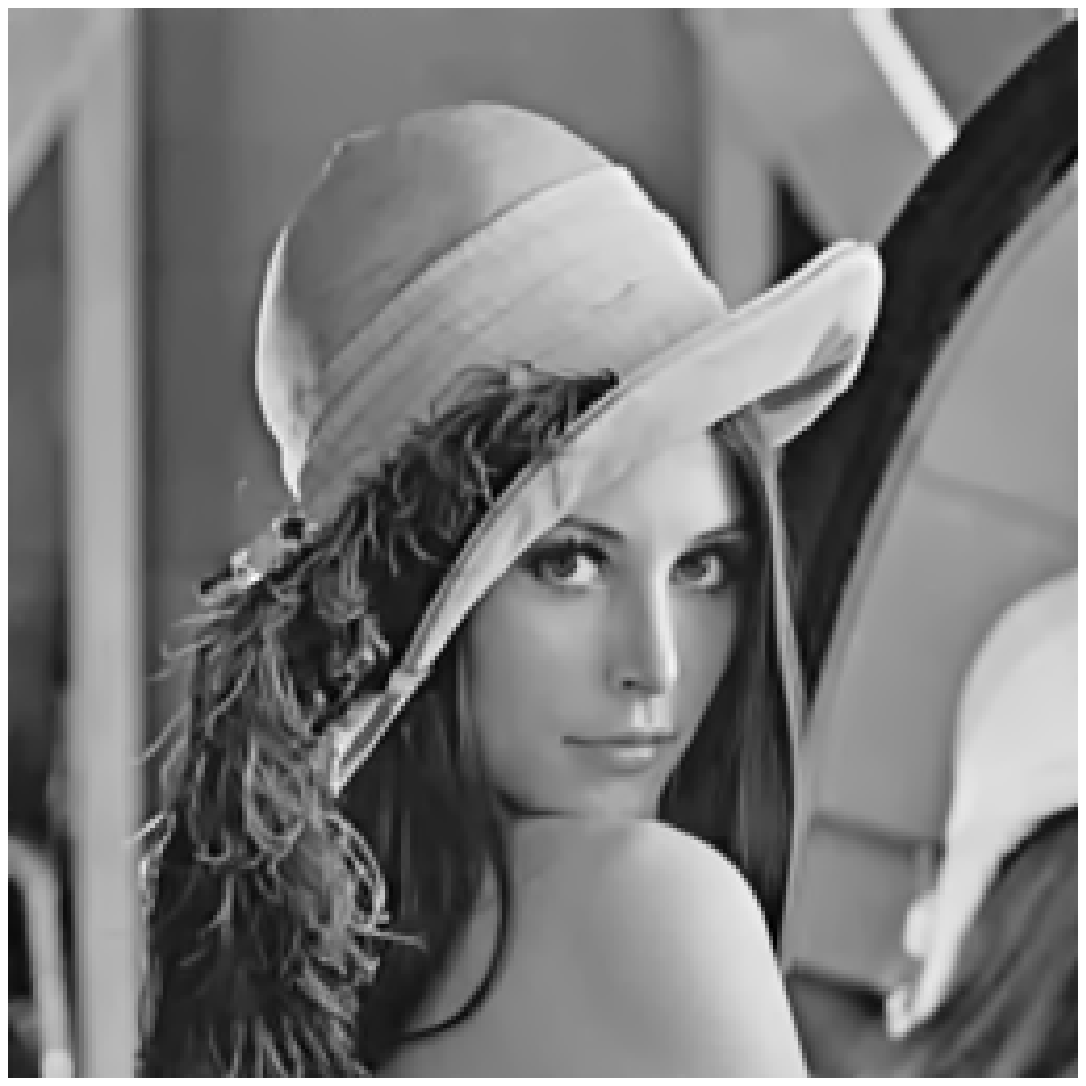}
   \includegraphics[width=0.11\textwidth]{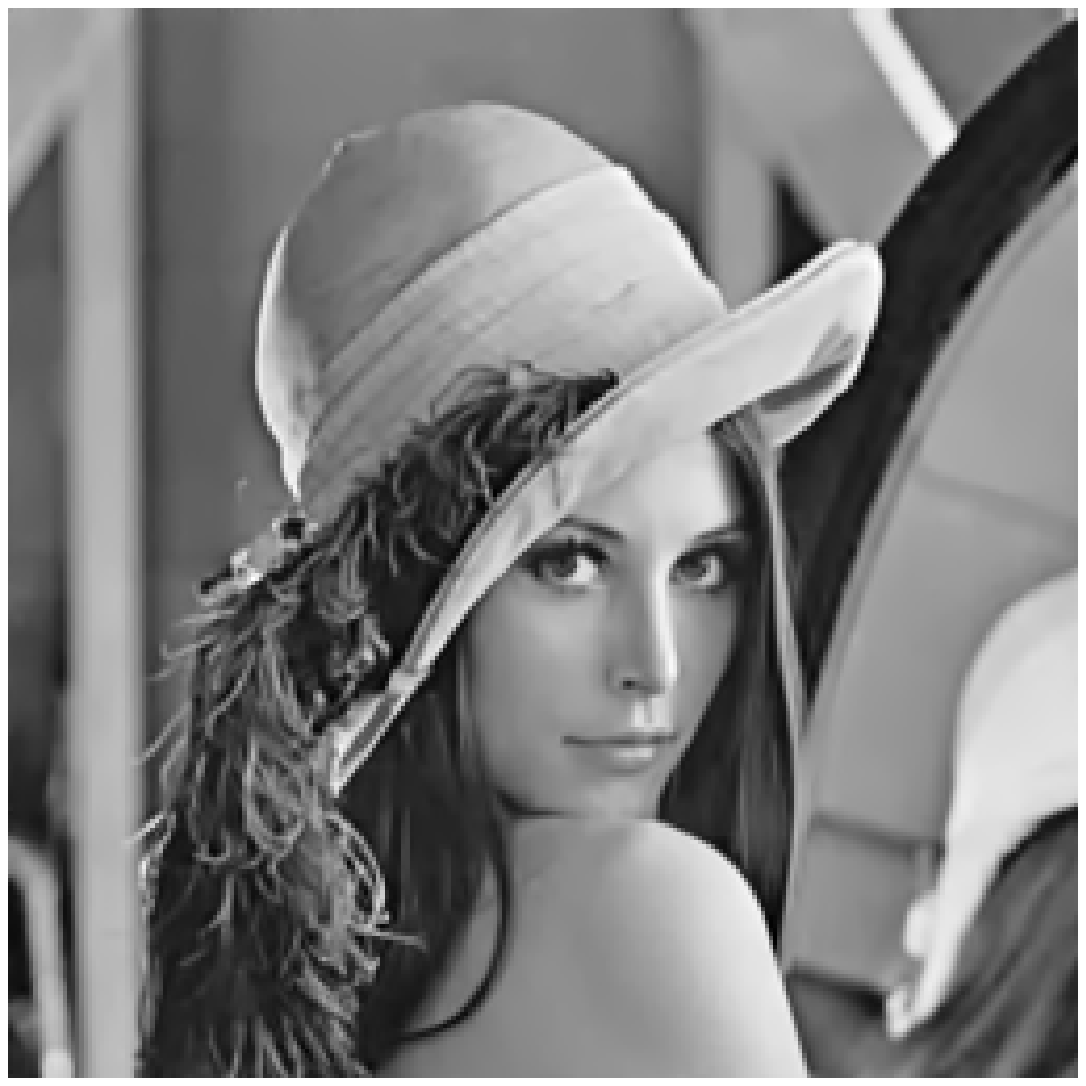}
   \includegraphics[width=0.11\textwidth]{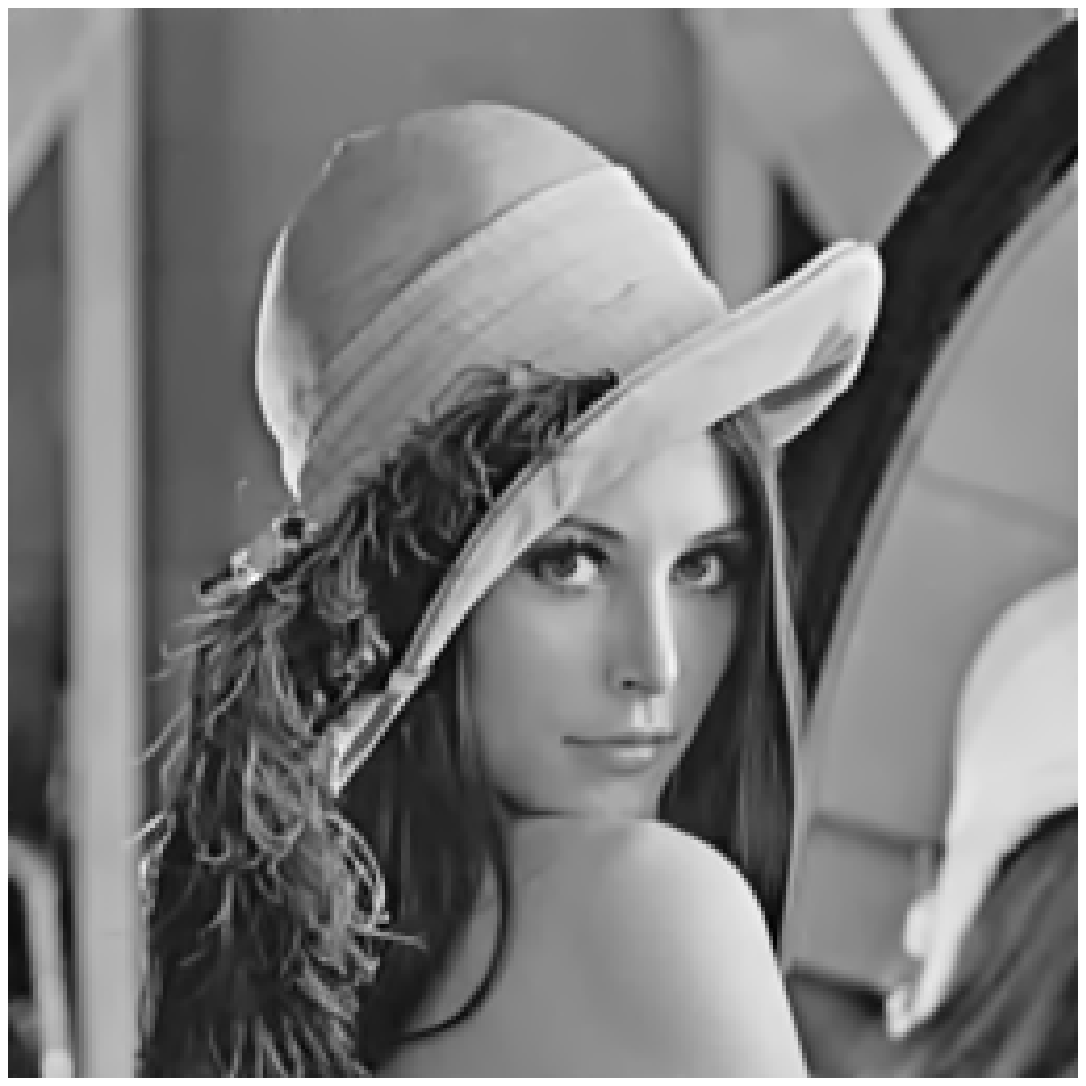}
   \includegraphics[width=0.11\textwidth]{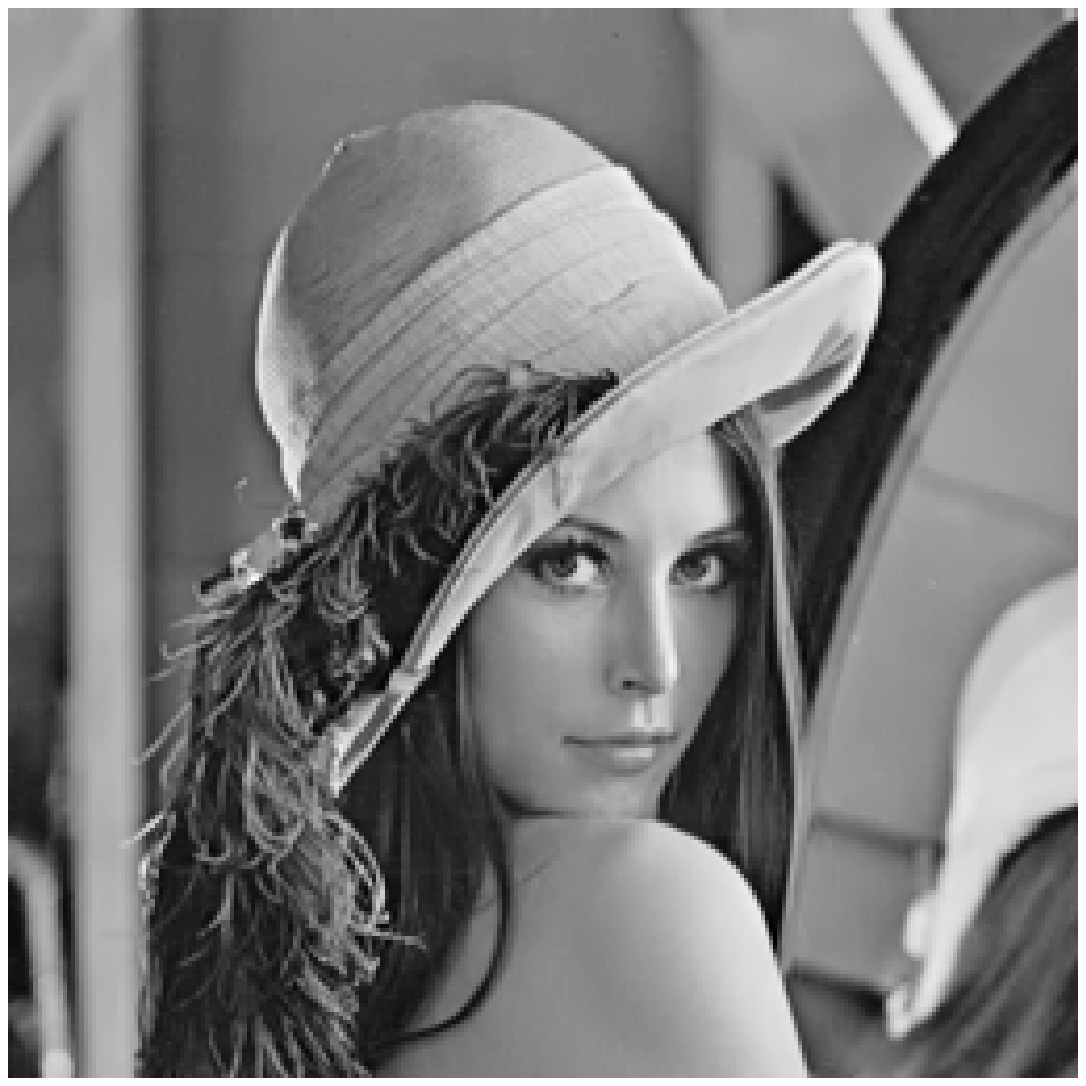}  \\
  \caption{\small The comparison of support maps (the first row and third row, which are obtained by directly inverse wavelet frame transform to the support detection) and back projection results (the second row and fourth row, which are obtained by only reserving the large wavelet frame coefficients to the original true image at the support locations) via different initial methods , IDD-BM3D method (first column), CSR method (second column), GSR method (third column) and ORACLE true imgae (fourth column). Scenario: 5. The decomposition level $L=4$.  }
\label{fig:parrots lena support detection}
\end{figure}

The advantage of our algorithm over other methods in terms of the PSNR and SSIM values is also consistent with the improvement of the visual quality. The subjective visual comparisons of different deblurring methods are shown in Figure \ref{fig:cameraman comparison}, \ref{fig:parrots comparison}. In addition, for better visual comparisons, in Figure \ref{fig:cameraman zoomin} and Figure \ref{fig:parrots zoomin}, we present the close-up views corresponding to the Figure \ref{fig:cameraman comparison} and Figure \ref{fig:parrots comparison}, respectively (These figures are best viewed on screen, rather than in print). We can see that the proposed SDSR algorithm leads to less artifacts, much cleaner and sharper image than other competing methods.
To further study the proposed method, in Figure \ref{fig:cameraman monrch support detection} and \ref{fig:parrots lena support detection}, we explicitly present the support maps, which are obtained by directly inverse wavelet frame transform to the support detection (binary 0-1 coefficients, the coefficients on support locations are 1, the remainder are 0) and back projection results, which are obtained by only reserving the large wavelet frame coefficients to the original true image at the support locations. Due to the space limit, here we just present the results of the first stage.
Moreover, we list the support detection accuracy rate of different initial methods in Table \ref{Table: support accuracy rate L1} and Table \ref{Table: support accuracy rate L4}, here the accuracy rate is defined as follows:
\begin{equation}
\mathrm{AR} = \frac{\# \{I^{detected}\cap I^{true}\} + \# \{T^{detected}\cap T^{true}\}}{\# \{I^{true}\} + \# \{T^{true}\}}
\end{equation}
where $I^{true}$ is the support index detected on the original true image, $I^{detected}$ is the support index detected on the initial reference image (the recovered results via different methods, i.e., IDD-BM3D, CSR and GSR in this work). Note that $T^{detected}$ and $T^{true}$ is the complementary set of $I^{detected}$ and $I^{true}$, respectively. $\# \{ \cdot \}$ denotes the cardinality of a given set. Clearly, the accuracy rate of support detection on the true image is $100\%$. From the above observations, we can conclude that: 1) The accuracy rates of the above three methods are approximately comparable.  We can acquire more reliable support information as the outer stage iteration of proposed algorithm proceeds, and the higher accuracy rate of support detection, it tends to achieve better final recovery result. 2) It  inevitably contains wrong support indexes in the detected support set in practice. However, our proposed SDSR is robust to the detected support information and certain percentage of wrong support information would not degrade the final recovery performance.  To our best  knowledge, this is the first time that an algorithm is able to consistently outperforms the IDD-BM3D, CSR and GSR in terms of image deblurring.

\begin{table} \centering { \tiny
\begin{tabular}{|c|c|c|c|}
\hline 
Image name (Scenario)& Initial method   & Accuracy rate (1st stage) & Accuracy rate (2nd stage)  \\
 \hline
  \multirow{3}{*}{Cameraman (3)} & IDD-BM3D & 81.67\% (29.02/0.8713)  &  81.95\% (29.04/0.8726) \\
    \cline{2-4}
    & CSR & 80.86\% (28.96/0.8744) & 81.99\% (29.02/0.8751)	 \\
    \cline{2-4}
    & GSR & 80.44\%  (28.80/0.8718) & 81.51\% (28.83/0.8728)    \\
    \cline{2-4}
 \hline
  \multirow{3}{*}{Monarch (3)} & IDD-BM3D & 81.22\% (29.54/0.9130)  &  81.54\% (29.59/0.9138)  \\
    \cline{2-4}
    & CSR &  79.94\% (29.55/0.9139) &	81.17\% (29.63/0.9143) \\
    \cline{2-4}
    & GSR & 80.63\% (29.26/0.9150)  &  81.24\% (29.42/0.9161) \\
    \cline{2-4}
 \hline
  \multirow{3}{*}{Parrots (5)} & IDD-BM3D &  85.65\% (31.80/0.9234) &  85.97\% (31.85/0.9244)  \\
    \cline{2-4}
    & CSR &  84.39\% (32.07/0.9236)	&  86.02\% (32.18/0.9249)  \\
    \cline{2-4}
    & GSR &  85.67\% (31.51/0.9218) &  85.95\% (31.68/0.9242)  \\
    \cline{2-4}
 \hline
  \multirow{3}{*}{Lena (5)} & IDD-BM3D & 82.94\% (31.53/0.9146)  &   83.17\% (31.59/0.9149)  \\
    \cline{2-4}
    & CSR &  81.69\% (31.54/0.9147)  &	83.21\% (31.57/0.9157) \\
    \cline{2-4}
    & GSR & 83.30\% (31.48/0.9150) &   83.45\% (31.60/0.9180) \\
    \cline{2-4}

\hline
\end{tabular}}

\caption{ The comparison of support detection accuracy rate, the PSNR and SSIM results of each stage when the initial reference image is produced by IDD-BM3D, CSR and GSR, respectively. The decomposition level $L=1$. }
\label{Table: support accuracy rate L1}
\end{table}

\begin{table} \centering { \tiny
\begin{tabular}{|c|c|c|c|}
\hline 
Image name (Scenario)& Initial method   & Accuracy rate (1st stage) & Accuracy rate (2nd stage)  \\
 \hline
  \multirow{3}{*}{Cameraman (3)} & IDD-BM3D & 78.61\% (29.04/0.8722)  &  79.13\% (29.07/0.8744) \\
    \cline{2-4}
    & CSR & 77.91\% (29.09/0.8758) & 79.28\% (29.12/0.8769)	 \\
    \cline{2-4}
    & GSR & 77.12\% (28.87/0.8728)  &   78.74\% (28.91/0.8741) \\
    \cline{2-4}
 \hline
  \multirow{3}{*}{Monarch (3)} & IDD-BM3D & 78.06\% (29.52/0.9136)  &  78.55\% (29.63/0.9159)  \\
    \cline{2-4}
    & CSR &  77.15\% (29.62/0.9158)& 78.21\% (29.65/0.9161)	 \\
    \cline{2-4}
    & GSR & 77.63\% (29.42/0.9162) & 78.31\% (29.49/0.9184) \\
    \cline{2-4}
 \hline
  \multirow{3}{*}{Parrots (5)} & IDD-BM3D &  82.08\% (31.80/0.9242)  &  82.72\% (31.93/0.9267) \\
    \cline{2-4}
    & CSR &  80.63\% (32.20/0.9267) 	&  82.84\% (32.27/0.9280)  \\
    \cline{2-4}
    & GSR &  82.16\%  (31.71/0.9248) & 82.80\% (31.79/0.9271)  \\
    \cline{2-4}
 \hline
  \multirow{3}{*}{Lena (5)} & IDD-BM3D & 80.09\% (31.62/0.9147)  &  80.49\% (31.67/0.9168) \\
    \cline{2-4}
    & CSR &  78.79\% (31.64/0.9166) &  80.60\% (31.71/0.9179)	 \\
    \cline{2-4}
    & GSR & 80.53\% (31.63/0.9176)  &  80.77\% (31.73/0.9202) \\
    \cline{2-4}

\hline
\end{tabular}}

\caption{ The comparison of support detection accuracy rate, the PSNR and SSIM results of each stage when the initial reference image is produced by IDD-BM3D, CSR and GSR, respectively. The decomposition level $L=4$. }
\label{Table: support accuracy rate L4}
\end{table}

%
%

Finally, we would like  to  show the part of the total performance improvement gained solely by support detection. We tried two cases. The first one is to solve a common $\ell_0$ model and the second one is to solve a truncated $\ell_0$ model. Both are non-convex models and when we solve them, the same initial points are used, i.e., we both use the results of BM3D as the initial points. Specifically, we first choose to  fix $\rho=1$ in order to focus on the contributions due to support detection. When $\rho=1$ in our SDSR algorithm, it is easy to see that the set $T$ in (\ref{eq:new truncatedl0l2}) contains all the frame coefficients and the truncated $\ell_0$ model degrades to plain $\ell_0$ model,  two scenarios of deblurring experiments are conducted with various kernels and noise variances, i.e., scenario 3 and scenario 5 in Table \ref{Table: all PSNR SSIM comparison}.  Only the PSNR and SSIM results of proposed IDD-BM3D+SDSR(L=4) are shown here, since the other cases have the similar conclusions. Then we choose to set fix $\rho=250$ and we will obtain a truncated $\ell_0$ model. From Table VI, we can observe that considerable improvements are achieved by the proposed SDSR algorithm with $\rho=250$ compared to $\rho=1$. Such a performance gain demonstrates that the significant improvement of SDSR algorithm can be achieved only owning to the truncation of the original $\ell_0$ model.

\subsection{Effect of the parameter $\rho$}
\begin{figure}[h]
  \centering
   \includegraphics[width=0.23\textwidth]{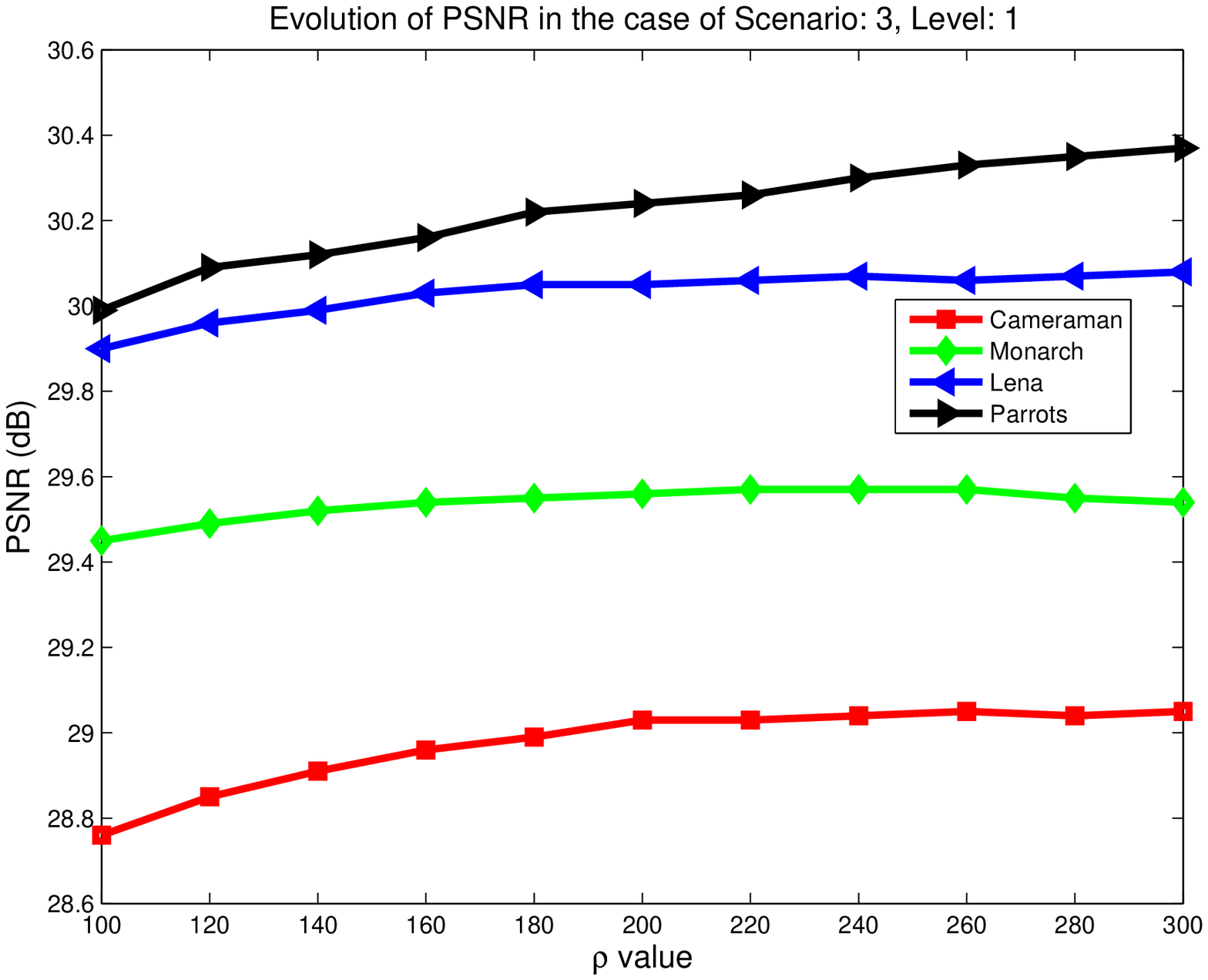}
   \includegraphics[width=0.23\textwidth]{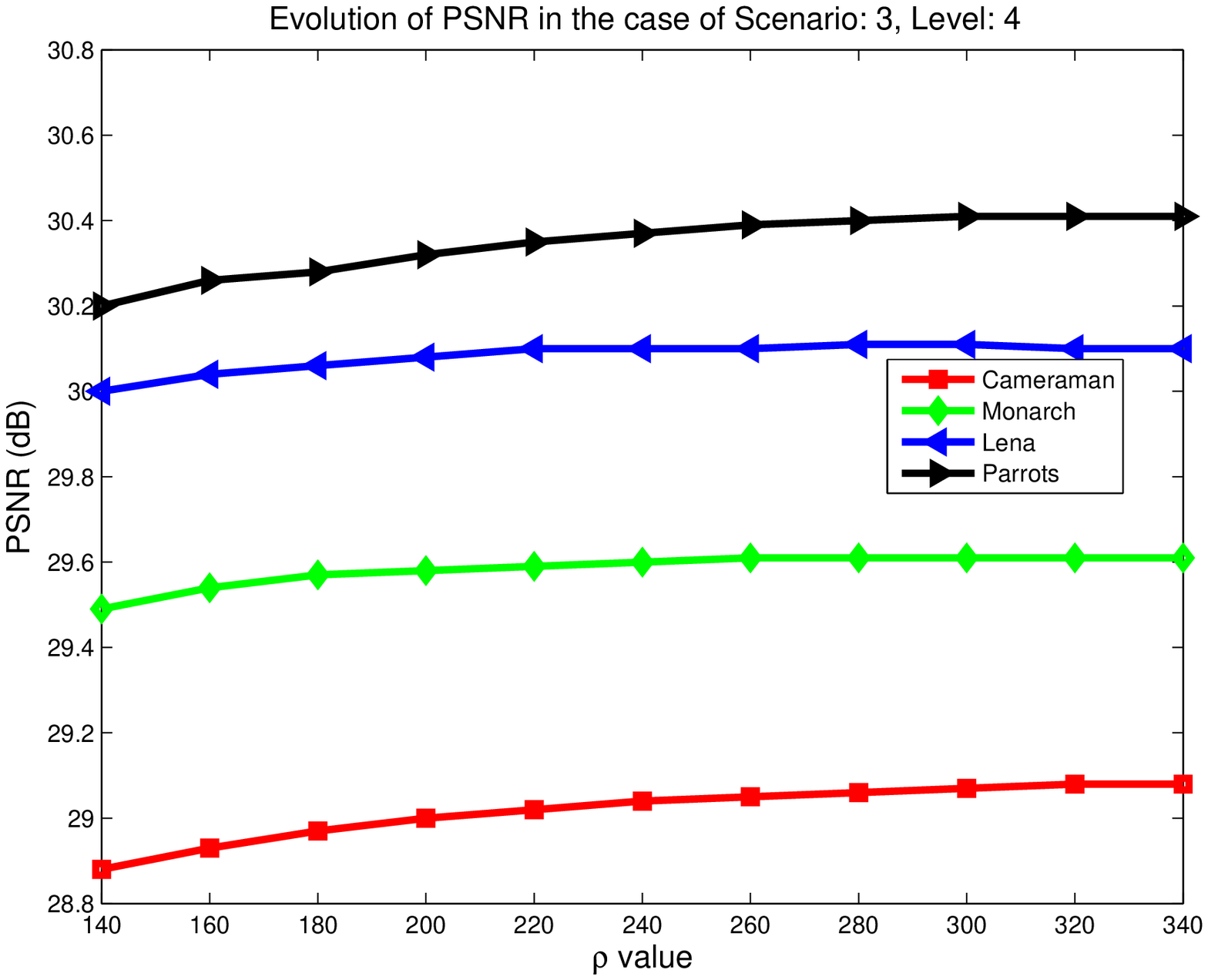} \\
   \includegraphics[width=0.23\textwidth]{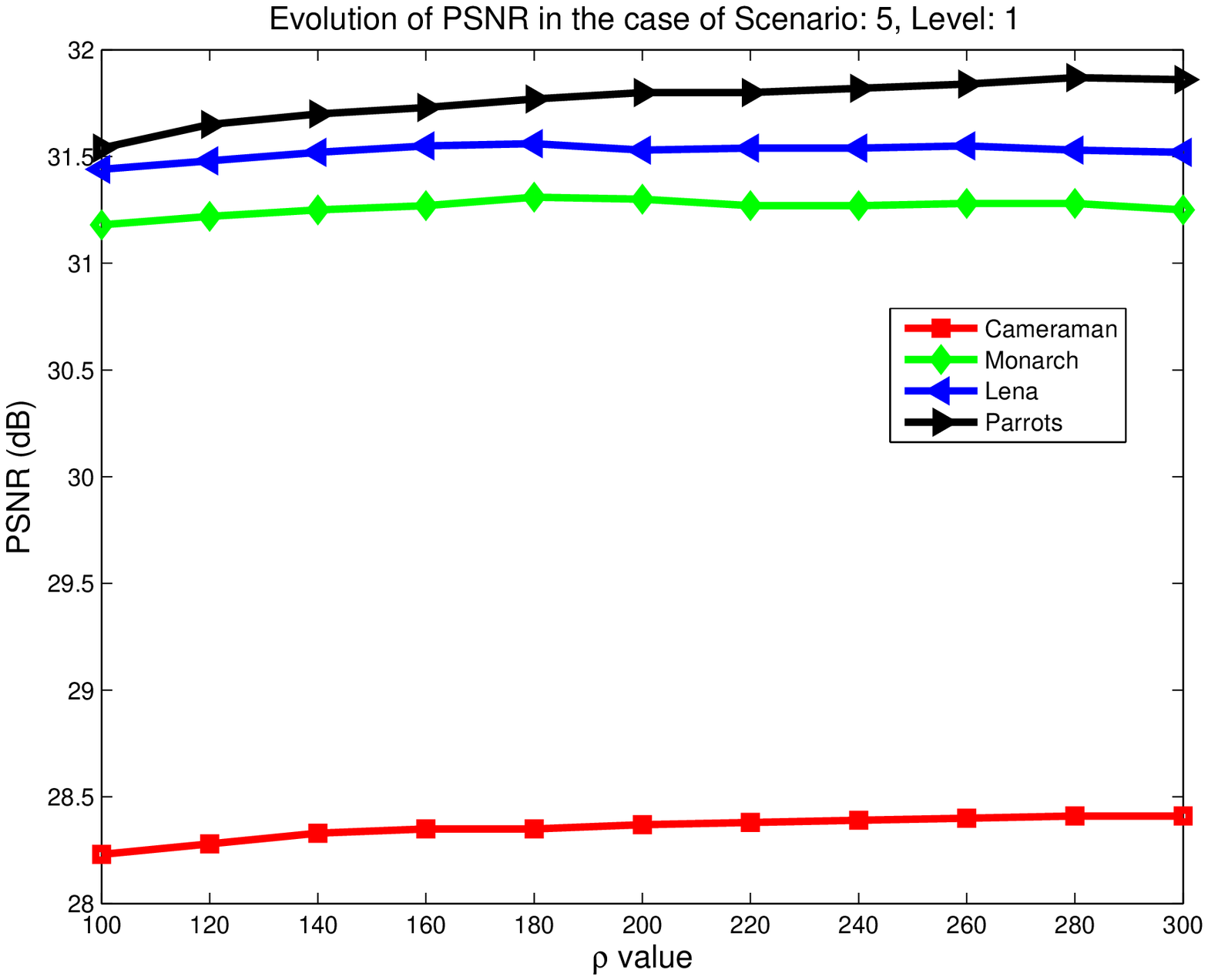}
   \includegraphics[width=0.23\textwidth]{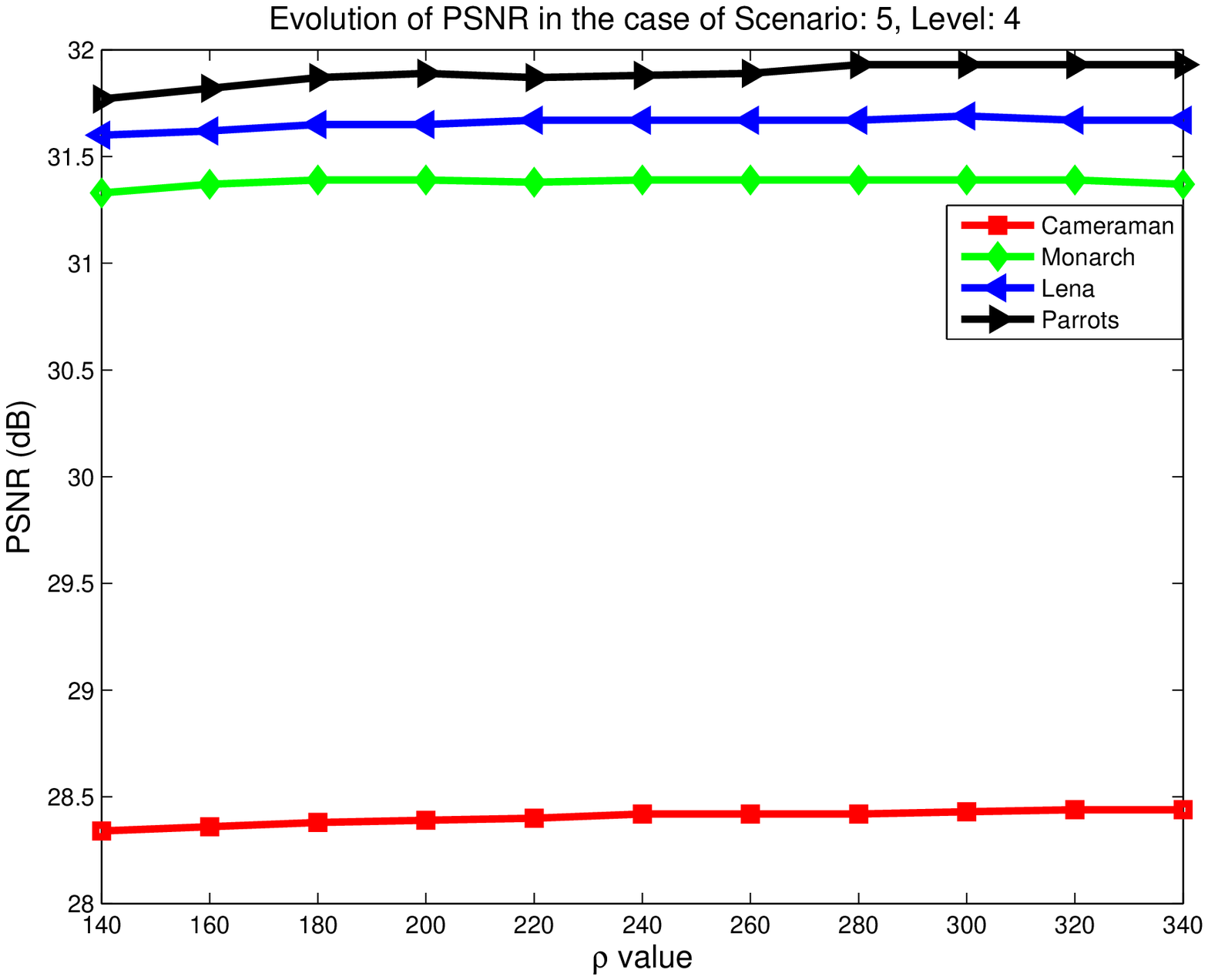} \\
  \caption{\small Curves of PSNR value v.s. $\rho$ choices for proposed IDD-BM3D+SDSR (1st stage). The scenarios are 3 and 5, respectively. The decomposition levels are $L=1$ and $L=4$, respectively. }
\label{fig: rho PSNR}
\end{figure}

\begin{figure}[h]
  \centering
   \includegraphics[width=0.23\textwidth]{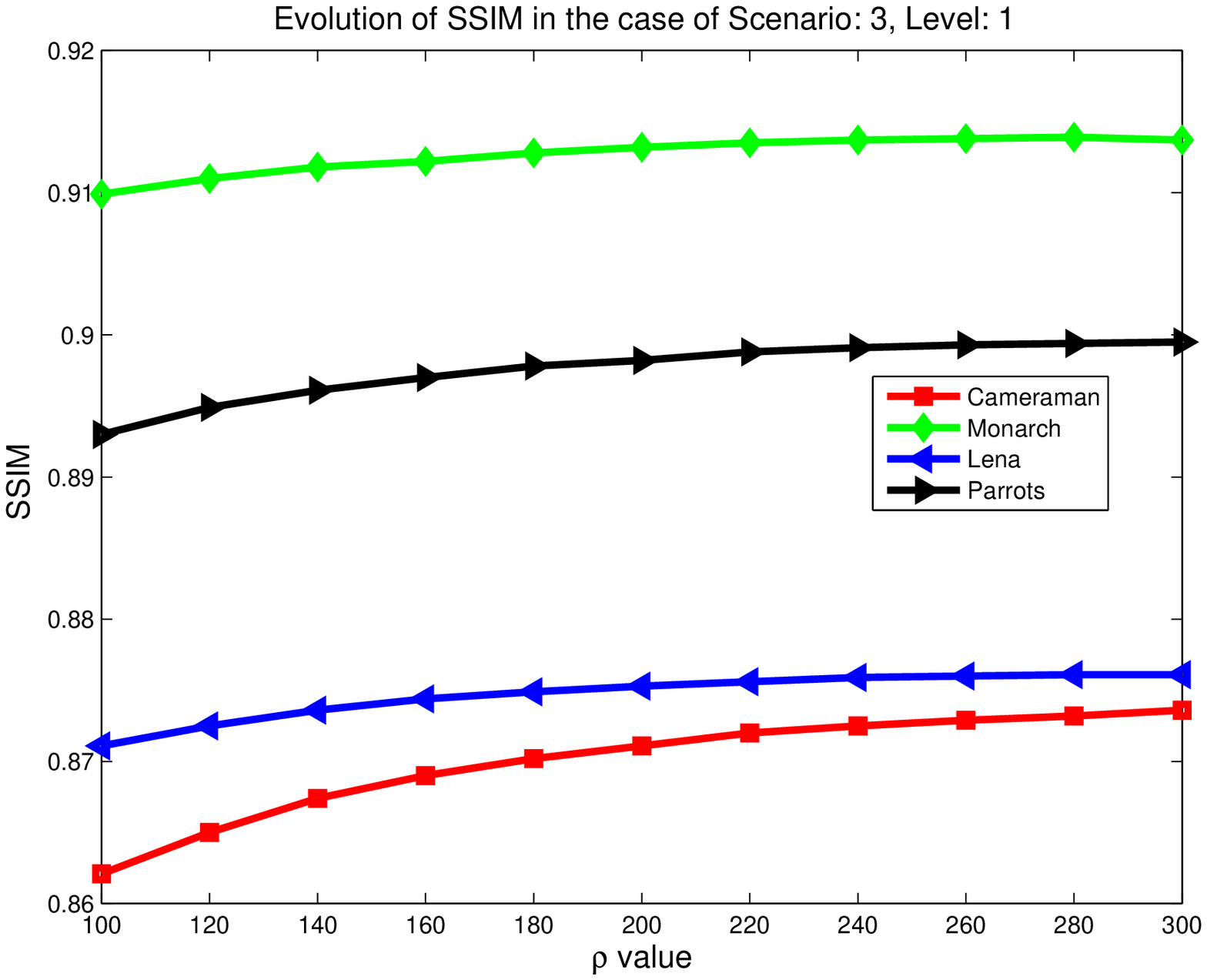}
   \includegraphics[width=0.23\textwidth]{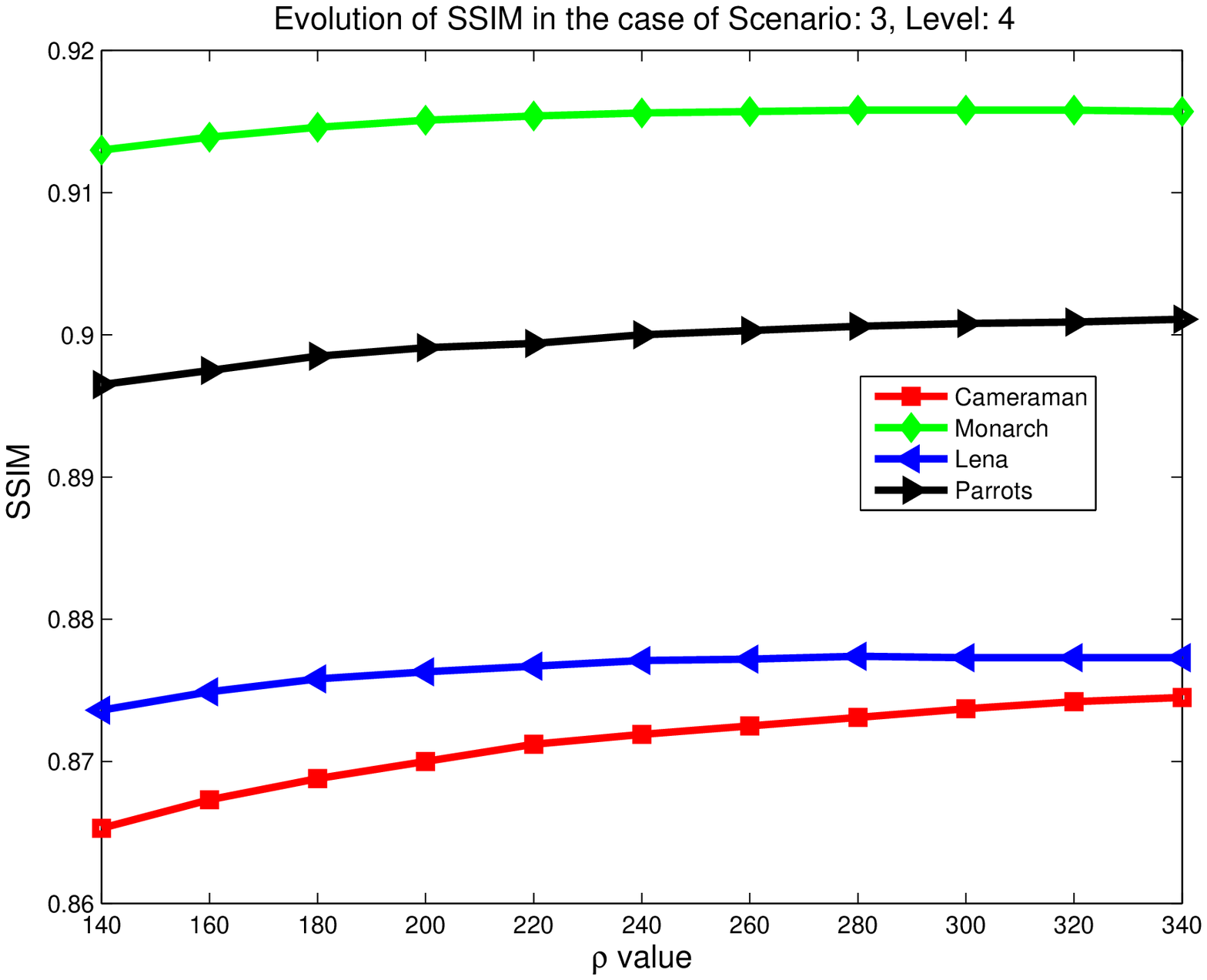} \\
   \includegraphics[width=0.23\textwidth]{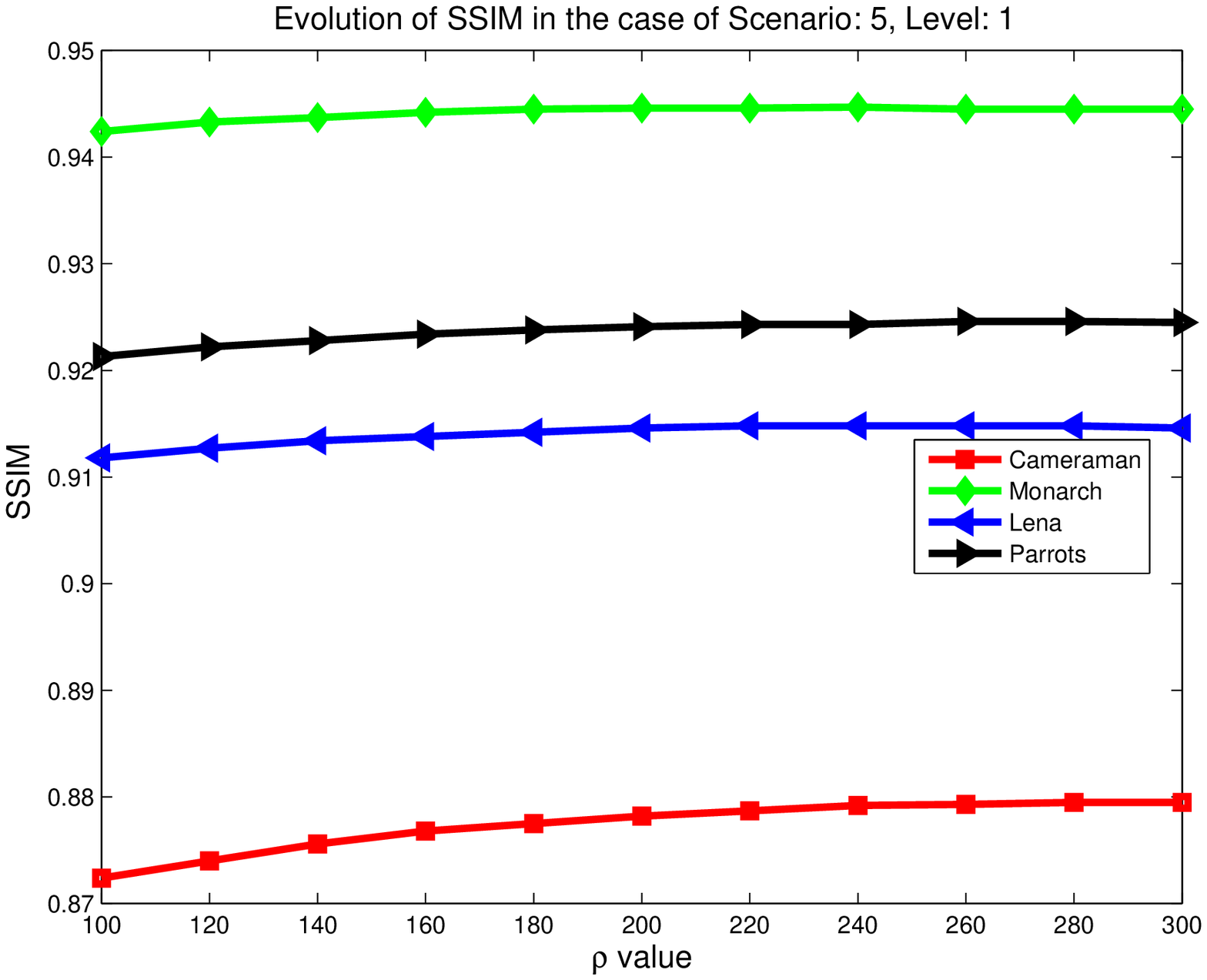}
    \includegraphics[width=0.23\textwidth]{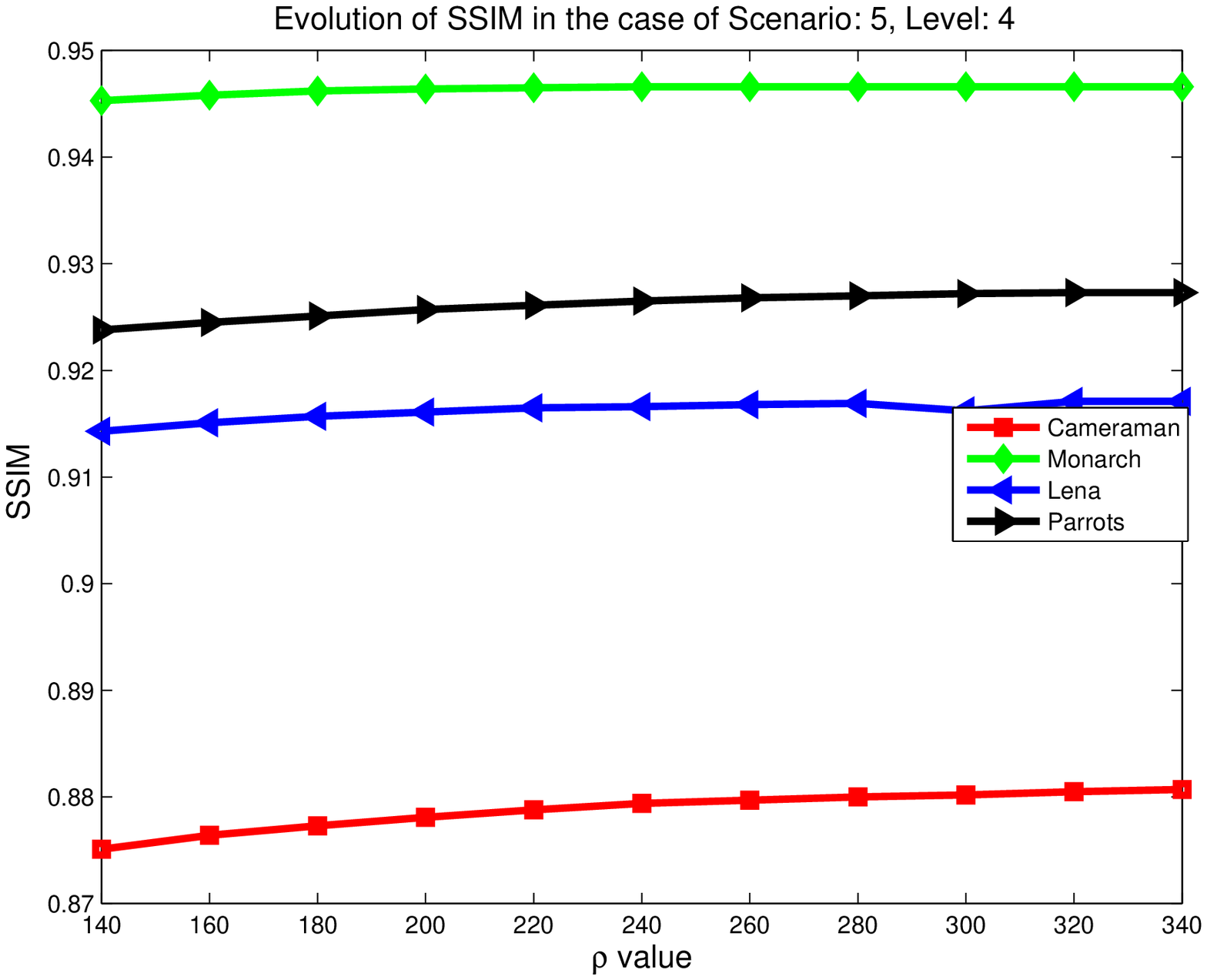} \\
  \caption{\small  Curves of SSIM value v.s. $\rho$ choices for proposed IDD-BM3D+SDSR (1st stage). The scenarios are 3 and 5, respectively. The decomposition levels are $L=1$ and $L=4$, respectively. }
\label{fig: rho SSIM}
\end{figure}

This section will give the detailed description about how sensitive the performance of the proposed algorithm is affected by $\rho$. In order to investigate the sensitivity of the parameter $\rho$ for the performance, the curves of PSNR and SSIM values versus the $\rho$ choices are presented in Figure \ref{fig: rho PSNR} and Figure \ref{fig: rho SSIM}, respectively. We can observe that the proposed SDSR algorithm is very robust to the parameter $\rho$.

\begin{table} \centering {
\begin{tabular}{|c|c|c|}
\hline 
Image name (Scenario) & $\rho=1$   & $\rho=250$ \\
 \hline
{Cameraman (3)} & 28.16/0.8552 & 29.07/0.8744 \\
\cline{1-3}
{Cameraman (5)} & 27.76/0.8692 & 28.42/0.8796  \\
\cline{1-3}
{Monarch (3)} & 28.11/0.8949  &  29.63/0.9159 \\
\cline{1-3}
{Monarch (5)} & 29.62/0.9336 & 31.37/0.9466  \\
 \cline{1-3}
{Lena (3)} & 28.91/0.8560  &  30.10/0.8771 \\
 \cline{1-3}
{Lena (5)} & 30.45/0.9025 & 31.67/0.9168  \\
\cline{1-3}
{Parrots (3)} & 29.53/0.8909   & 30.36/0.9001  \\
\cline{1-3}
{Parrots (5)} & 30.80/0.9197 &  31.93/0.9267 \\
 \hline

\end{tabular}}

\caption{ The PSNR value and SSIM value of IDD-BM3D+SDSR(L=4) when the parameter $\rho=1$ and $\rho=250$. The scenarios are 3 and 5, respectively. The decomposition level is $L=4$. }
\label{Table: reduced plain }
\end{table}

\subsection{Algorithm Stability}
\begin{figure}[h]
  \centering
   \includegraphics[width=0.23\textwidth]{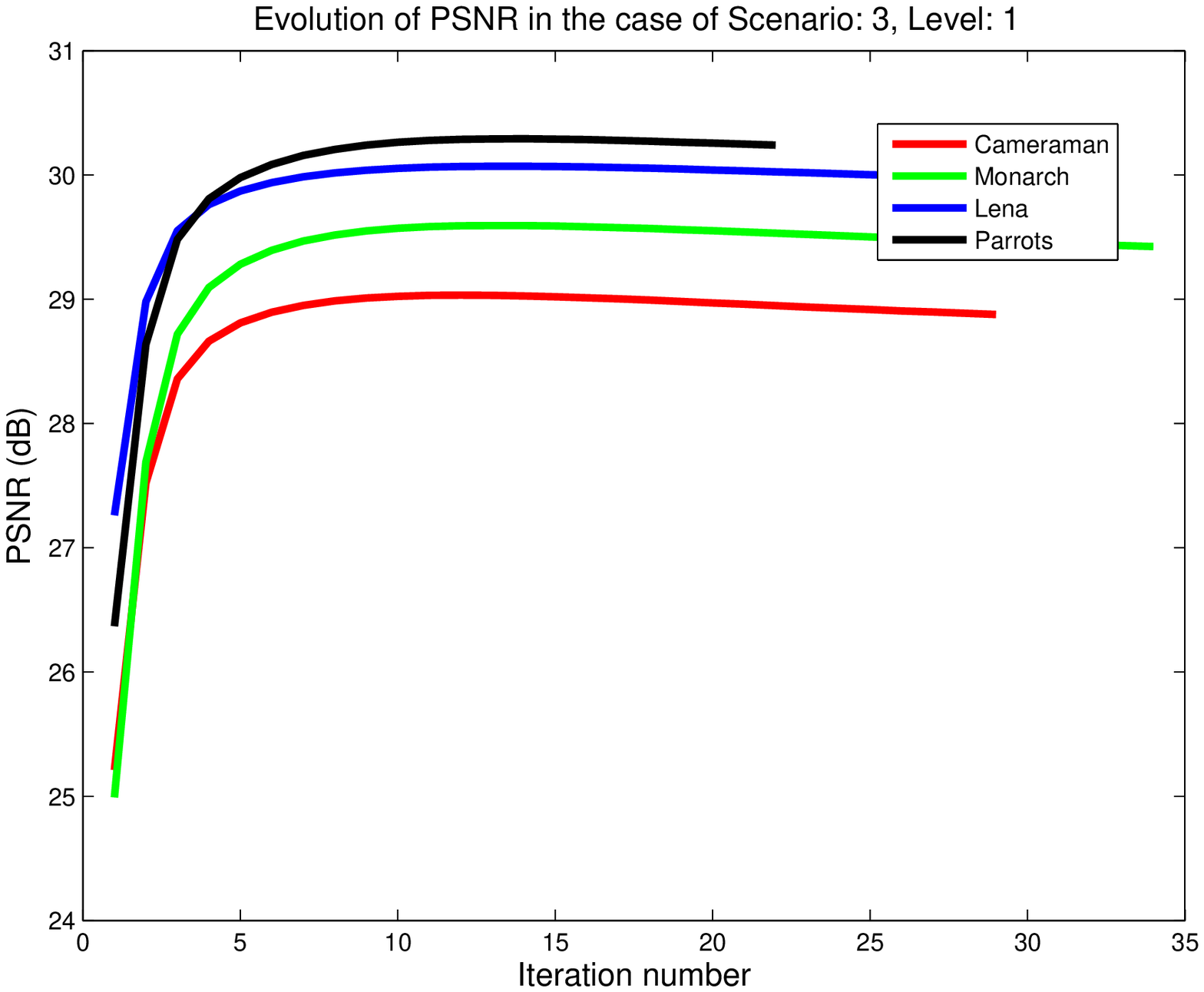}
   \includegraphics[width=0.23\textwidth]{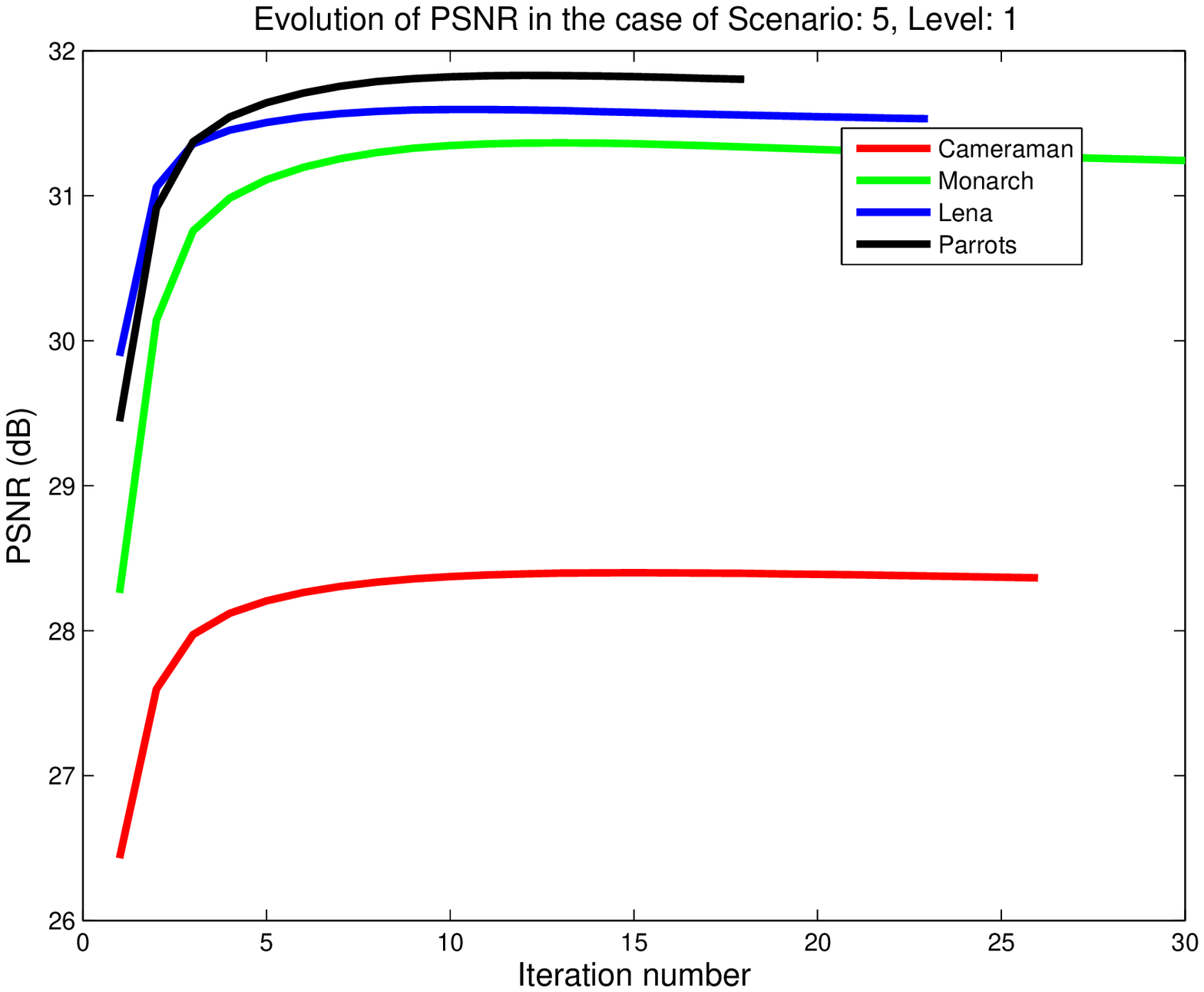} \\
   \includegraphics[width=0.23\textwidth]{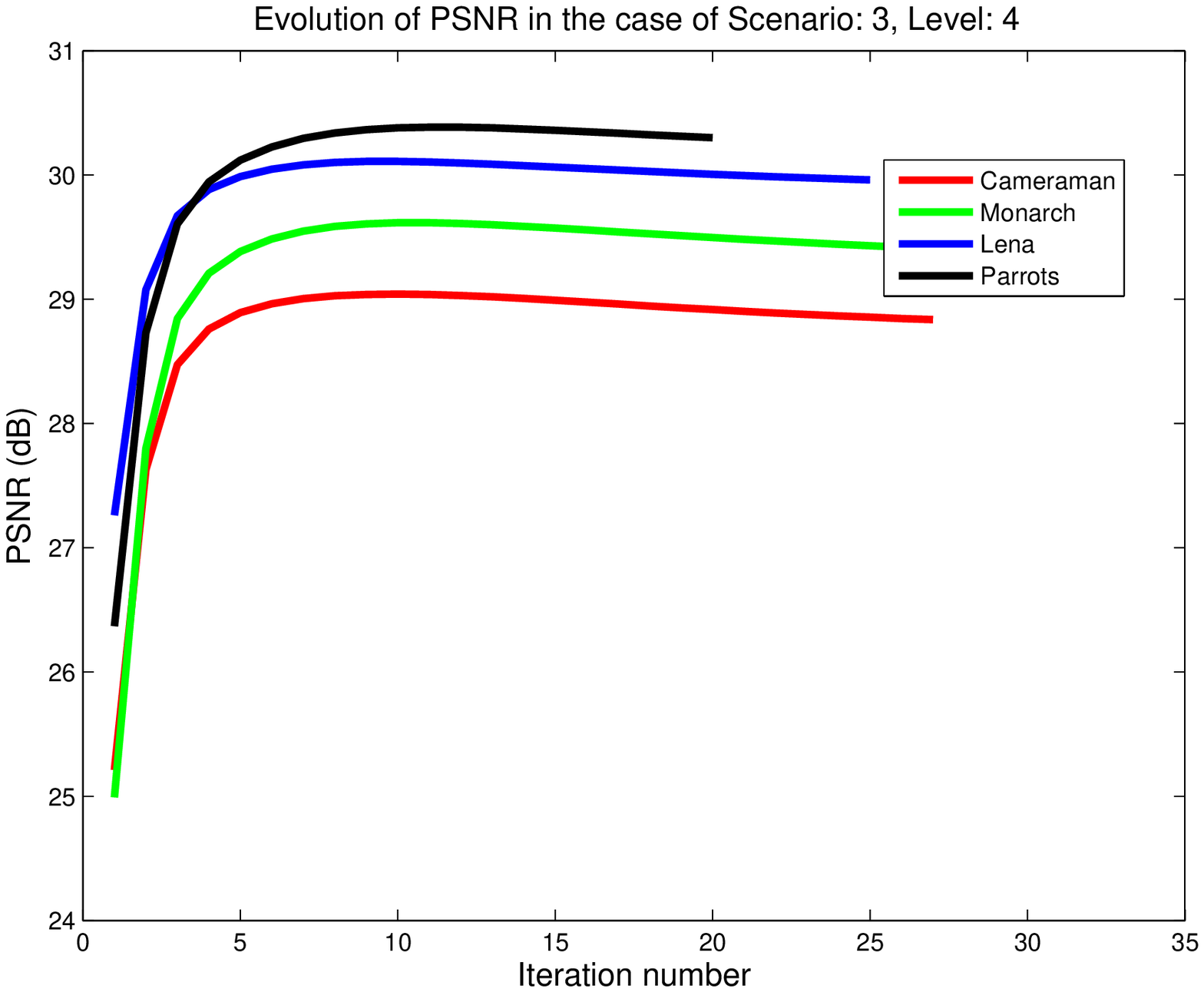}
  \includegraphics[width=0.23\textwidth]{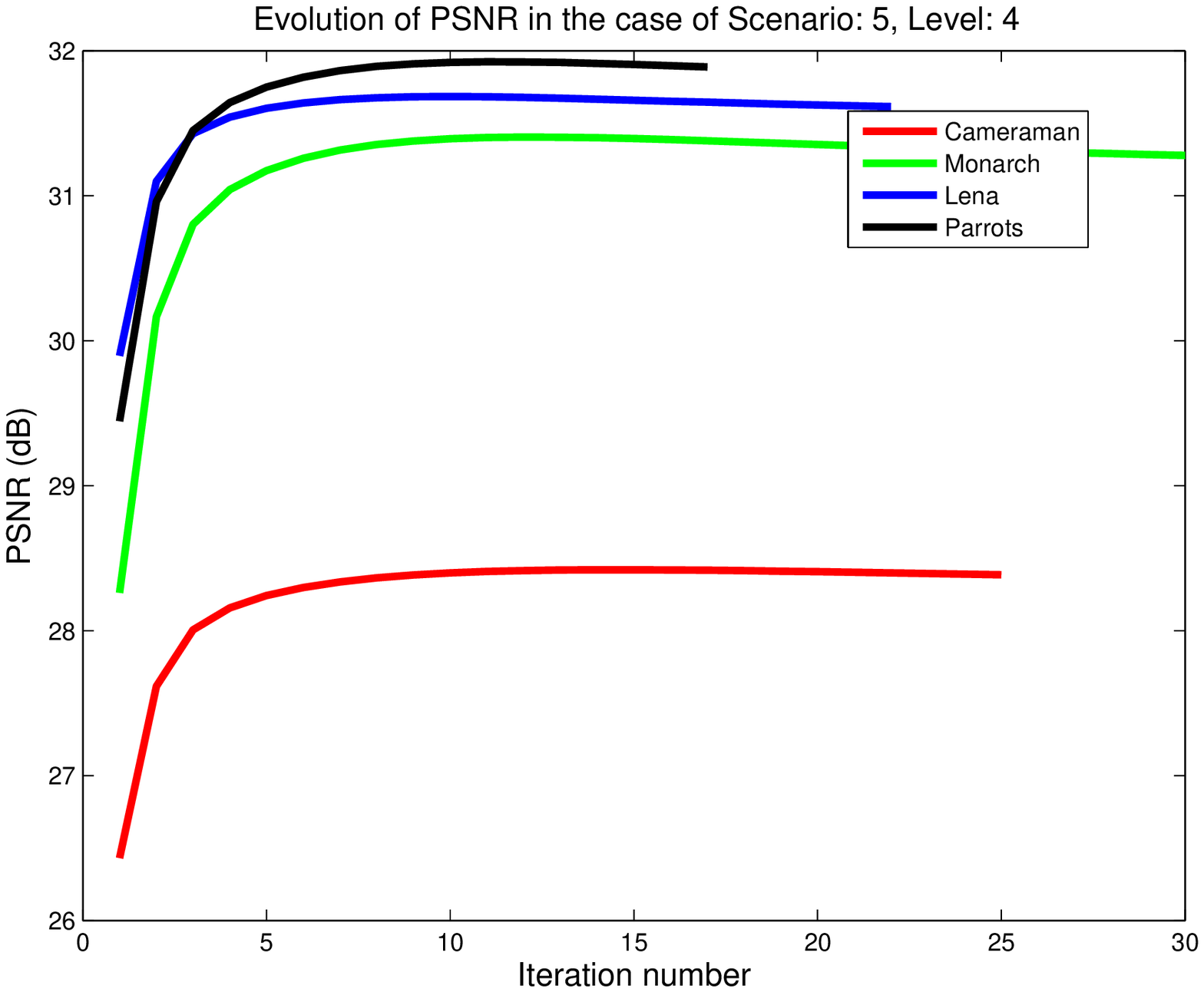} \\
  \caption{\small  Stability of the proposed SDSR algorithm.  Stable progress of the PSNR (dB) results is achieved by the proposed IDD-BM3D+SDSR (1st stage) for the 4 test images with respect to the iteration number in the cases of Scenario 3 and Scenario 5. The decomposition levels are $L=1$ and $L=4$, respectively. }
\label{fig: Iteration PSNR SSIM}
\end{figure}

Since the objective function (\ref{eq:new truncatedl0l2}) is non-convex with $T$ known, it is difficult to give its theoretical proof for global convergence. Here, we only provide the empirical evidence to illustrate the stability of the proposed SDSR algorithm. Figure \ref{fig: Iteration PSNR SSIM} plots the evolutions of PSNR versus iteration numbers. It is observed that with the growth of iteration number, all the PSNR curves increase monotonically and ultimately become flat and stable, exhibiting good stability of the proposed SDSR model.

\section{Conclusions and future work}
Image deblurring  is a fundamental topic in image processing and computer vision fields. In this paper, we propose the wavelet frame based support driven sparse regularization (SDSR) model. The partial support information of frame coefficients is self-learned and incorporated into the truncated $\ell_0$ quasi-norm frame-based model.  To attain reliable support set,  the results of the state-of-the-art image restoration methods are used as the initial reference image for support detection.  Experimental results demonstrated that the SDSR method outperforms the other state-of-the-art competing methods. The key component of the proposed SDSR model is the support estimation of frame coefficients.
The possible future work  along the same research line is to develop more effective support detection methods and extend SDSR to other image precessing tasks.
%
\section{Acknowledgements}
This work was partially supported by the Natural Science Foundation of China, Grant Nos. 11201054, 91330201, by the National Basic Research Program (973 Program), Grant No. 2015CB856000, and by the Fundamental Research Funds for the Central Universities ZYGX2013Z005.

\end{document}